# Autocorrelations Decay in Texts and Applicability Limits of Language Models

**Nikolay Mikhaylovskiy**
Higher IT School, Tomsk State University, Tomsk, Russia, 634050
NTR Labs, Moscow, Russia, 129594
nickm@ntr.ai

**Ilya Churilov**
NTR Labs, Moscow, Russia, 129594
ichurilov@ntr.ai

**Abstract**

We show that the laws of autocorrelations decay in texts are closely related to applicability limits of language models. Using distributional semantics we empirically demonstrate that autocorrelations of words in texts decay according to a power law. We show that distributional semantics provides coherent autocorrelations decay exponents for texts translated to multiple languages. The autocorrelations decay in generated texts is quantitatively and often qualitatively different from the literary texts. We conclude that language models exhibiting Markov behavior, including large autoregressive language models, may have limitations when applied to long texts, whether analysis or generation.

## 1 Introduction

In this work, we endeavor into outlining statistically the limits of applicability of popular contemporary language models. To avoid any terminological doubt, when we write "models of the language", we refer to any models that explain some linguistic phenomena, while "language models" refer to probabilistic language models as defined in Subsection 2.3 Probabilistic Language Models. While not long ago probabilistic language models were just models that assign probabilities to sequences of words [4], now they are the cornerstone of any task in computational linguistics through few-shot learning [6], prompt engineering [38] or fine-tuning [13]. On the other hand, current language models fail to catch long-range dependencies in the text consistently. For example, text generation with maximum likelihood target leads to rapid text degeneration, and consistent text generation requires probabilistic sampling and other tricks [22]. Large language models such as GPT-3 [6] push the boundary of "short text" rather far (specifically, to 2048 tokens), but do not remove the problem.

Our contributions in this work are the following:

- We explain how the laws of autocorrelations decay in texts are related to applicability of language models to long texts;
- We pioneer the use of pretrained word vectors for autocorrelation computations that allows us to study a widest range of autocorrelation distances;
- We show that the autocorrelations in literary texts decay according to power laws for all these distances;
- We show that distributional semantics typically provides coherent autocorrelations decay exponents for texts translated to multiple languages, unlike earlier flawed approaches;
- We show that the behavior of autocorrelations decay in generated texts is quantitatively and often qualitatively different from the literary texts.

| Grammar type (low → high) | Automaton | Memory |
|---|---|---|
| Regular (R) | Finite-state automaton (FSA) | Automaton state |
| Context-free (CF) | Push-down automaton (PDA) | + infinite stack (only top entry accessible) |
| Context-sensitive (CS) | Linear bounded automaton (LBA) | + bounded tape (all entries accessible) |
| Recursively enumerable (RE) | Turing machine (TM) | + infinite tape (all entries accessible) |

Table 1: Chomsky hierarchy of formal grammars (from [10])

## 2 Models of the Language

In this section, we briefly introduce models of the language that are important for the further considerations.

### 2.1 Formal Grammars

Formal grammars describe how to form strings from a language's alphabet that are valid according to the language's syntax. They were introduced by Chomsky in 1950s [7][8]. A formal grammar consists of a finite set of production rules in the form

$$left - hand\ side \rightarrow right - hand\ side, \tag{1}$$

where each side consists of a finite sequence of the following symbols:

- a finite set of nonterminal symbols (indicating that some production rule can yet be applied)
- a finite set of terminal symbols (indicating that no production rule can be applied)
- a start symbol (a distinguished nonterminal symbol)

Chomsky grammars constitute a hierarchy, see Table 1. While the original hierarchy implies strict inclusion of lower class grammars to higher ones, now there are several types of grammars known to fall between or partially overlap with the original classes (see, for example, [10]).

### 2.2 Distributional Semantics and Models

Distributional hypothesis assumes that linguistic items with similar distributions have similar meanings or function and was likely first introduced by Harris [20] in 1954 and was popularized in the form "a word is characterized by the company it keeps" by Firth [17]. The basic idea is to collect distributional information in, say, high-dimensional vectors, and then to define similarity in terms of some metric, say Euclidean distance or the angle between the vectors.

Early distributional approaches from 60s relied on hand-crafted features of the words [35], while more recent – on statistics of varied sorts. The first generation of statistical distributional semantics approaches included Latent Semantic Analysis (LSA) [11][12], Hyperspace Analogue to Language (HAL) [24][25], and many others, see [15] for a review. The second generation primarily consists of word2vec [31][32] and GloVe [37] models, the first, implicitly, and the second, explicitly adding the word analogy task into the training objective, so that similar relationships between words should be described by similar difference vectors between embeddings. The third generation of statistical distributional semantics models was started by emergence of BERT contextual word embeddings [13]. BERT have combined the word and its current context into a single vector embedding and used Masked Language Modelling training objective. A lot of recent work sprouted from BERT.

### 2.3 Probabilistic Language Models

Probabilistic language models consider sequences

$$t_{1:m} = \{t_1, t_2, \dots, t_m\} \tag{2}$$

of tokens from the lexicon $\mathcal{L}$. An autoregressive language model estimates the probability of such a sequence

$$P(t_{1:m}) = P(t_1)P(t_2|t_1)P(t_3|t_{1:2}) \dots P(t_m|t_{1:m-1}) = \prod_{k=1}^{m} P(t_k|t_{1:k-1}) \tag{3}$$

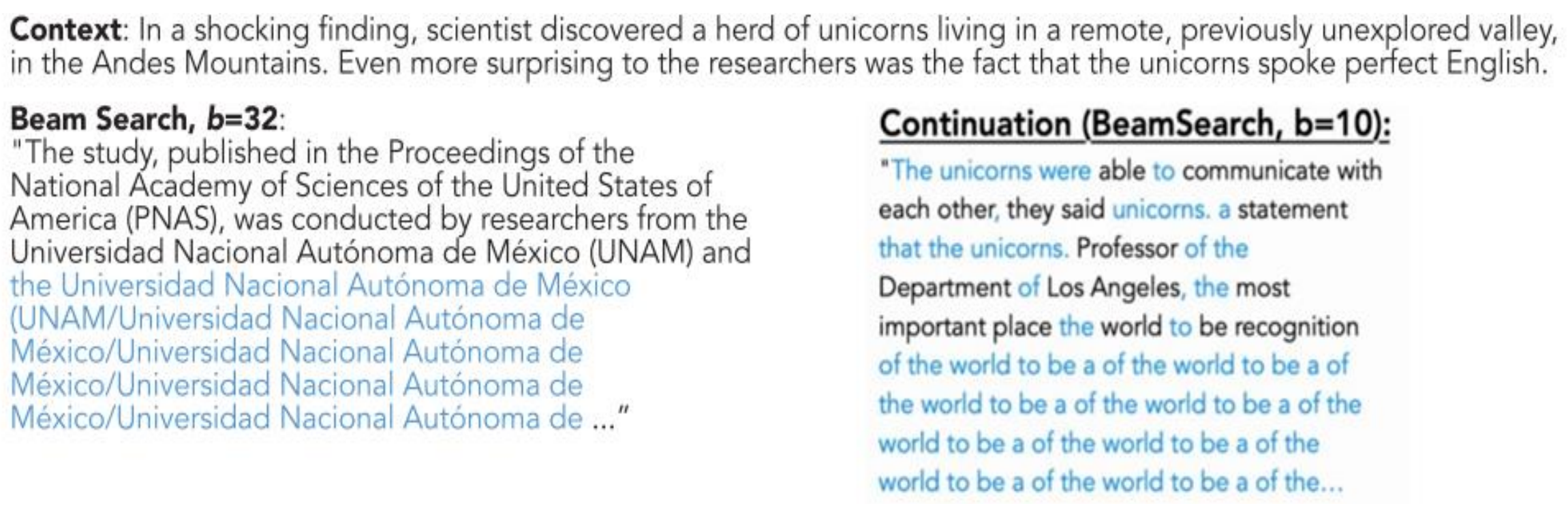

Figure 1: Beam search produces degenerate text (from [22])

using the chain rule. Most models introduce the Markov [30] assumption that the probability of a token depends on the previous $n-1$ tokens only, thus approximating (3) with a truncated version

$$P(t_{1:m}) \approx \prod_{k=1}^{m} P(t_k|t_{k-n+1:k-1}) \tag{4}$$

While the language models based on recurrent [33], and specifically, LSTM [41] neural networks do not introduce the Markov assumption explicitly, we will shortly see that in practice they do exhibit Markovian behavior. On the other hand, it is long known that Markov models describe stochastic regular grammars [42].

## 3 Why Autocorrelations Decay Laws Matter?

In this section we explain why autocorrelation decay laws matter a lot to computational linguistics' near-future.

### 3.1 Computing Autocorrelations Using Distributional Semantics

Suppose we have a sequence of $N$ vectors $V_i \in R^d, i \in [1, N]$. Autocorrelation function $C(\tau)$ is the average similarity between the vectors as a function of the lag $\tau = i - j$ between them. The simplest metric of vector similarity is the cosine distance $d(V_i, V_j) = \cos\angle(V_i, V_j) = \frac{V_i \cdot V_j}{\|V_i\|\|V_j\|}$, where $\cdot$ is a dot product between two vectors and $\| \quad \|$ is an Euclidean norm of a vector. Thus,

$$C(\tau) = \frac{1}{N-\tau} \sum_{i=1}^{N-\tau} \frac{V_i \cdot V_{i+\tau}}{\|V_i\|\|V_{i+\tau}\|}. \tag{5}$$

$C(\tau)$ ranges from $-1$ for perfectly anticorrelated sequence (for $\tau = 1$ and $d = 1$ that would be $1, -1, 1, -1$ etc.) to 1 for a perfectly correlated one (for $\tau = 1$ and $d = 1$ that would be $1, 1, 1, 1$ etc.).

A distributional semantic assigns a vector to each word or context in a text. Thus, a text is transformed into a sequence of vectors, and we can calculate an autocorrelation function for the text.

### 3.2 Transformer Language Models Exhibit Markovian Behavior

In this paper, by Markovian behavior, we mean that large language models actually use only a limited context, often significantly less than the maximum context possible. Thus they implicitly or explicitly use the Markov assumption. Two separate phenomena classes that prove that transformer language models exhibit Markovian behavior are known, and in Section 5.5 we introduce the third one.

One such phenomenon is the rapid text degeneration when a transformer language model is used to generate text with maximum likelihood target [21][28]. Maximization-based decoding methods such as beam search lead to output text that is bland, incoherent, or gets stuck in repetitive loops [22] that are extremely reminiscent of positively recurrent Markov chains (see Figure 1).

| Level | Architecture | Description |
|---|---|---|
| R- | Transformer | The encoder with stacked multi-head attention layers and dense layers. |
| R | RNN | A classical RNN with ReLU activations. |
| R+ | LSTM | A classical LSTM. |
| DCF+ | Stack-RNN | An RNN with an external stack, with PUSH, POP, and NO-OP actions. |
| NDCF | NDStack-RNN | An RNN with a nondeterministic stack, simulated using dynamic programming |
| CS | Tape-RNN | An RNN with a finite tape, as in a Turing machine (similar to Baby-NTM ). |

Table 2: Alignment of neural network architectures with Chomsky hierarchy (from [10])

The other phenomenon is studied in detail in [10]. The authors have found that the networks roughly match the computational models associated with the Chomsky hierarchy: RNNs can solve tasks up to the regular level, Stack-RNNs up to the DCF level, and Tape-RNNs up to the CS level. Finally, they observed that Transformers and LSTMs are less aligned with the Chomsky hierarchy: Transformers fail on regular tasks, while LSTMs can solve tasks more difficult than regular. The results of [10] are summarized in Table 2. As transformers can at most generalize to regular languages and Markov models describe stochastic regular grammars [42], we can safely say that transformers exhibit behavior no richer than regular.

### 3.3 Markovian Implies Exponential Correlations Decay, Probabilistic Context-Free Grammars Can Generate Power Laws

Assume that the sequence (2) is an output of a random source that takes values in $\mathcal{L}$. If the source is Markovian, it can be shown [23] that the autocorrelations (or, equivalently, mutual information between chunks of the text) decay exponentially. Namely, the following theorem holds:

**Theorem 1 ([23]).** Let $M$ be a Markov matrix that generates a Markov process. If $M$ is irreducible and aperiodic, then the asymptotic behavior of the mutual information $I(t_1, t_2)$ is exponential decay toward zero for $|t_2 - t_1| \gg 1$ with decay timescale $\log \frac{1}{|\lambda_2|}$, where $\lambda_2$ is the second largest eigenvalue of $M$. If $M$ is reducible or periodic, $I$ can instead decay to a constant; no Markov process whatsoever can produce power law decay.

One the other hand, the following theorem holds:

**Theorem 3 ([23]).** There exist a probabilistic context-free grammar such that the mutual information $I(A, B)$ between two symbols $A$ and $B$ in the terminal strings of the language decay like $d^{-k}$, where $d$ is the number of symbols between A and B.

### 3.4 If the Natural Language Exhibits Power Law Correlations Decay, We Can Do Better Than Autoregressive Language Models

Summarizing the above, if texts in the natural languages exhibit exponential autocorrelations decay, autoregressive language models are good to analyze or generate texts of any length. On the other hand, if texts in the natural languages exhibit power law autocorrelations decay, building language models that exhibit at least hierarchical, context-free-grammar-ish, slow-correlation-decay behavior may be beneficial for a variety of downstream tasks. This may be not enough to model long texts successfully because natural languages cannot be completely described by context-free grammars (see, for example, [40]), but may be a meaningful step.

## 4 Studying Autocorrelations Decay Laws in Texts

### 4.1 Prior Research

Schenkel, Zhang, and Zhang [39] were likely the first to empirically find the power law autocorrelations decay in texts using a random walk model with an arbitrary mapping of characters to fixed length, 5 bit sequences. They studied 10 texts in English. The obvious drawback of their approach is dependency on encoding. Amit et al. [3] explored this problem in various translations of the Bible and have shown that the power law exponent depends on both the language and the codification. Testing multiple random mappings would provide a more reliable estimate of power law exponents, but such a research is a matter of future. Random walk models have later been used to find the power law in text by several researchers, including Ebeling and Neiman [14], Kokol et al. [26] (who, by the way, in our opinion have not found power-law autocorrelations in literary writing on distances studied, but found power-law autocorrelations in computer programs, in a perfect agreement with the fact that computer programs are described by context-free grammars), Pavlov et al. [36], who find multifractal structures in the text, and Manin [29], who attribute long-range correlations to slow variations in lexical composition within the text.

Alvarez-Lacalle et al. [2] used a version of first-generation distributional semantic model to study autocorrelations in 12 literary texts in English to find power law autocorrelations decay. Altmann, Cristadoro, and Degli [1] analyze 41 binary functions on words separately on ten English versions of international novels. They separate the effects of burstiness and long-range correlations in the power spectrum and find a power law correlations decay. Lin and Tegmark [23] in a short empirical part of their study use three text corpora: 100 MB from Wikipedia, the first 114 MB of a French corpus and 27 MB of English articles from slate.com. They observe the power law decay of mutual information, but note that the large portion of the long-range mutual information appears to be dominated by poems in the French sample and by the html-like syntax in the Wikipedia sample. They have also shown similar power decay laws for autocorrelations in natural music and exponential laws in generated music, the result reproduced by different means by Yamshchikov and Tikhonov [43]. Corral et al. [10] study intervals between consecutive appearances of specific words in literary texts in 4 languages, including Finnish (a rare study of highly agglutinative language) to find that most words have a universal dimensionless probability density function described by gamma distribution. Gillet and Ausloos [18] and Montemurro and Pury [34] study sequences built from word frequencies and word lengths to find the power law autocorrelations decay.

### 4.2 Research Questions

Given the prior art, many research question remain unanswered. The ones we address in this work are:
**Q1.** How accurately can we say that autocorrelations in texts decay according to a power law?
**Q2.** Can we reject the hypothesis of exponential decay of correlations?
**Q3.** Does the law of decay depend on the language of the text?
**Q4.** Over what range of distances does the decay in autocorrelations follow a power law?
**Q5.** Are autocorrelations in LM-generated texts any different from literary texts?

### 4.3 Methods

In this work we use two distributional semantic models to estimate autocorrelations in long texts. One is a bag-of-words (BOW) embedding model of Alvarez-Lacalle et al. [2]. The other distributional semantic model we use is GloVe [37]. We use pretrained multilingual GloVe vector embeddings from [16]. We filter out both frequent and rare words filtering similarly to [2] when using BOW.

BOW assigns a vector of dimension $d$ to each word first, and then averages these vectors over a window of the size $a$. This averaged vector is then assigned to a word in the center of averaging window. The exact procedure for BOW is described in detail in [2]. GloVe naturally maps each word to a vector; we then center the vector system by subtracting the average of vectors over the whole text, and, similarly, average over a window of the size $a$ when we need direct comparison to BOW. After that in both cases we can compute the autocorrelation function following Section 3.1.

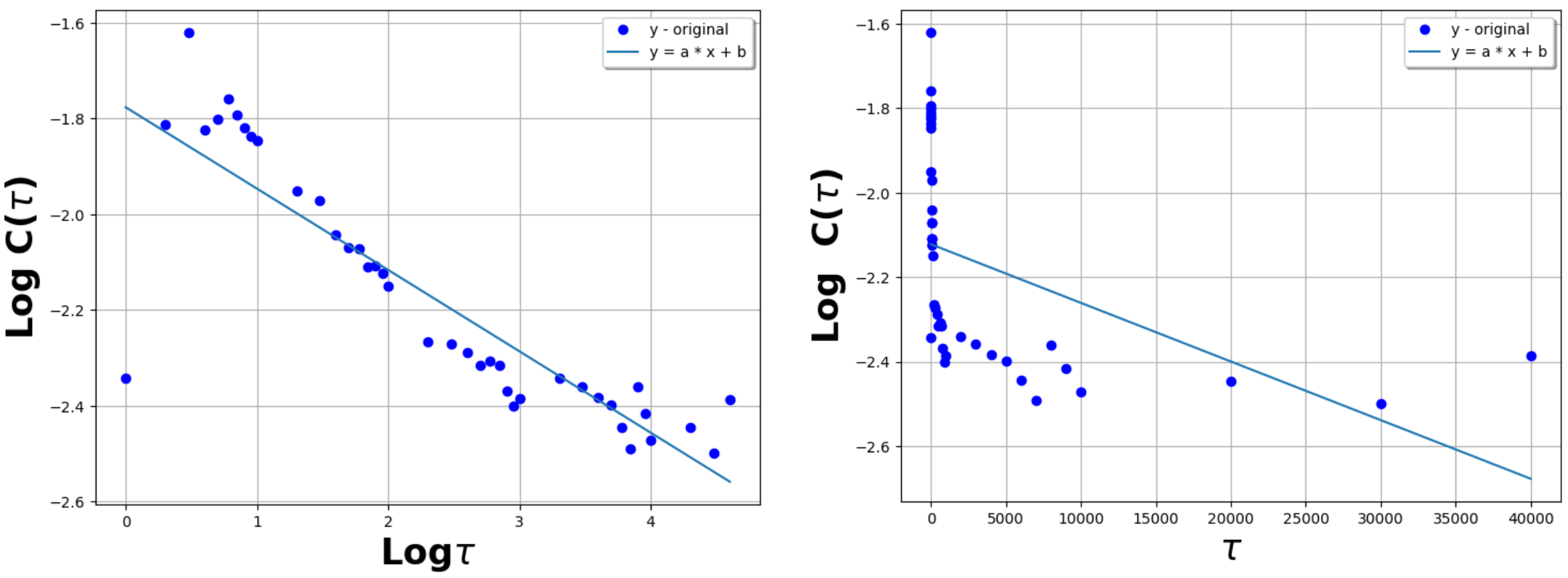

Figure 2: Autocorrelations in Don Quixote (English) computed using GloVe, a = 1, d = 300, $\tau \in [1, 40000]$ Left: log-log coordinates. Right: log-linear coordinates.

# 5 Experiments

## 5.1 The Dataset

For our studies we have collected a dataset of long literary and philosophical works in English, Spanish, French, German and Russian[i] each: Critique of Pure Reason, Don Quixote de la Mancha, Moby-Dick or, The Whale, The Adventures of Tom Sawyer, The Iliad, The Republic and War and Peace. The only translation absent is Moby-Dick in German, which happened to be substantially abridged. The texts have been obtained from Project Gutenberg, Wikisource, Royallib and lib.ru and preprocessed so as to fit our research purposes:

- copyright texts were removed from the files;
- author and translator notes were removed;
- table of contents and any indices were removed, except for the table of contents from Don Quixote;
- any links to illustrations have been removed;
- in the Russian version of War and Peace any non-Russian text have been replaced with Russian translations;
- etymology section was removed from Moby-Dick or, The Whale, where encountered, as some languages missed it.

## 5.2 Choosing Between Hypotheses of Power Law and Exponential Decay of Correlations

To address **Q1.** "How accurately can we say that autocorrelations in texts decay according to a power law?" and **Q2.** "Can we reject the hypothesis of exponential decay of correlations?" for each text, we have computed autocorrelations for a series of distances $\tau = n * 10^k, n \in [1, 9]$ , and approximated the points produced by a straight line in log-log and log-linear coordinates using the least squares regression. We have evaluated the goodness of fit of each model by MAPE (Mean Absolute Percentage Error). The range of $\tau$ for Glove was chosen from the first non-negative autocorrelation value ε (autocorrelations on small distances $\tau = [1, 2]$ happened to be sometimes negative).

The results for the English translation of Don Quixote are presented in the Figure 2. It can be seen that in log-log coordinates the regressed straight line approximates data well enough, unlike log-linear coordinates.

Table 3 lists the MAPE metrics of goodness of fit of autocorrelation by power and exponential laws (the smaller the better). It can be easily seen that for all the texts but one the hypothesis of exponential decay of correlations can be rejected. The peculiarity of the French translation of The Iliad is that the autocorrelation with $\tau = 1$ is very small but still positive, thus both producing significantly larger MAPE. Additional graphs are presented in the Appendix A.

| | Power Law | | | | | Exponential Law | | | | |
|---|---|---|---|---|---|---|---|---|---|---|
| | BOW en | fr | es | ru | en | BOW en | fr | es | ru | en |
| The Adventures of Tom Sawyer | **0,16** | **0,11** | **0,16** | **0,14** | **0,21** | 0,52 | 0,32 | 0,33 | 0,33 | 0,55 |
| The Republic | **0,21** | **0,15** | **0,09** | **0,10** | **0,13** | 0,58 | 0,28 | 0,25 | 0,31 | 0,38 |
| Don Quixote | **0,20** | **0,11** | **0,12** | **0,09** | **0,20** | 0,66 | 0,24 | 0,22 | 0,23 | 0,44 |
| War and Peace | **0,20** | **0,13** | **0,11** | **0,08** | **0,09** | 0,54 | 0,24 | 0,24 | 0,28 | 0,42 |
| Critique of Pure Reason | **0,09** | **0,07** | **0,15** | **0,10** | **0,14** | 0,27 | 0,17 | 0,20 | 0,21 | 0,25 |
| The Iliad | **0,24** | 2,37 | **0,16** | **0,10** | **0,19** | 0,63 | **2,33** | 0,17 | 0,19 | 0,54 |
| | | | | | | | | | | |
| Moby-Dick or, The Whale | **0,14** | **0,12** | **0,11** | **0,09** | **0,15** | 0,40 | 0,22 | 0,22 | 0,22 | 0,47 |

Table 3: Goodness of fit of autocorrelation by power and exponential laws in terms of MAPE.

| | BOW | | | GloVe | | |
|---|---|---|---|---|---|---|
| | $\alpha$ | $\beta$ | MAPE | $\alpha$ | $\beta$ | MAPE |
| en | -0.7718 | 0.9545 | 0.1054 | -0.7246 | 1.1582 | 0.1044 |
| fr | -0.8836 | 1.1407 | 0.2154 | -0.7749 | 1.1051 | 0.2150 |
| es | -0.7601 | 0.9332 | 0.1057 | -0.7083 | 0.9947 | 0.1271 |
| ru | -0.7412 | 0.7874 | 0.0787 | -0.6431 | 0.9173 | 0.0548 |
| de | -0.8072 | 0.9542 | 0.1411 | -0.8326 | 1.3478 | 0.1657 |

Table 4: Dependence of the autocorrelations power decay law in Don Quixote on the language and embedding. $\tau$ ranges from 200 to 4000 words, d=300, a = 200

### 5.3 Determining the Dependency of the Autocorrelations Decay Law on the Language of the Text

To study the dependency of the autocorrelations decay law on the language of the text, we have measured $C(\tau)$ for the same multilingual dataset as in Section 5.1 and fitted with power law $C(\tau) = \beta \cdot \tau^{\alpha}$. Table 4 presents results for Don Quixote. It can be easily seen that the parameters of power law, as well as the accuracy of the approximation are extremely consistent among languages for both embeddings, with standard deviation of exponent being 7% for BOW and 10% for GloVe. Moreover, the exponents for BOW and GloVe are also consistent within 15%, which we consider a very good agreement. This is in a stark contrast with the results from [3] that critically depend on the codification and language.

### 5.4 Determining the Range of Distances Where the Decay in Autocorrelations Can Be Considered Subject to a Power Law

As the BOW approach requires a sufficiently large window size $a$, we have studied the dependence of autocorrelations on distance ranges using GloVe embeddings with a window size $a = 1$. For each text we explored all the ranges of $\tau$ spanning at least a decimal order of magnitude, and fitted the autocorrelations with the best fitting log, power and exponential functions. We then mapped the differences between MAPE of power and other approximations, as well as the ranges where each function fits the data the best. The results for the Critique of Pure Reason in English and The Adventures of Tom Sawyer in Spanish are presented on Figure 3. Each small square on these images correspond to a range of $\tau$ determined by its vertical (start) and horizontal (end) coordinates, for example, the full range of $\tau \in [1, 40000]$ corresponds to the top right corner. Additional graphs are presented in Appendix B.

It can be seen that for the shorter spans of $\tau$ the best approximations are sometimes logarithmic or exponential but their advantage is not significant, while for the longer ranges the best approximations

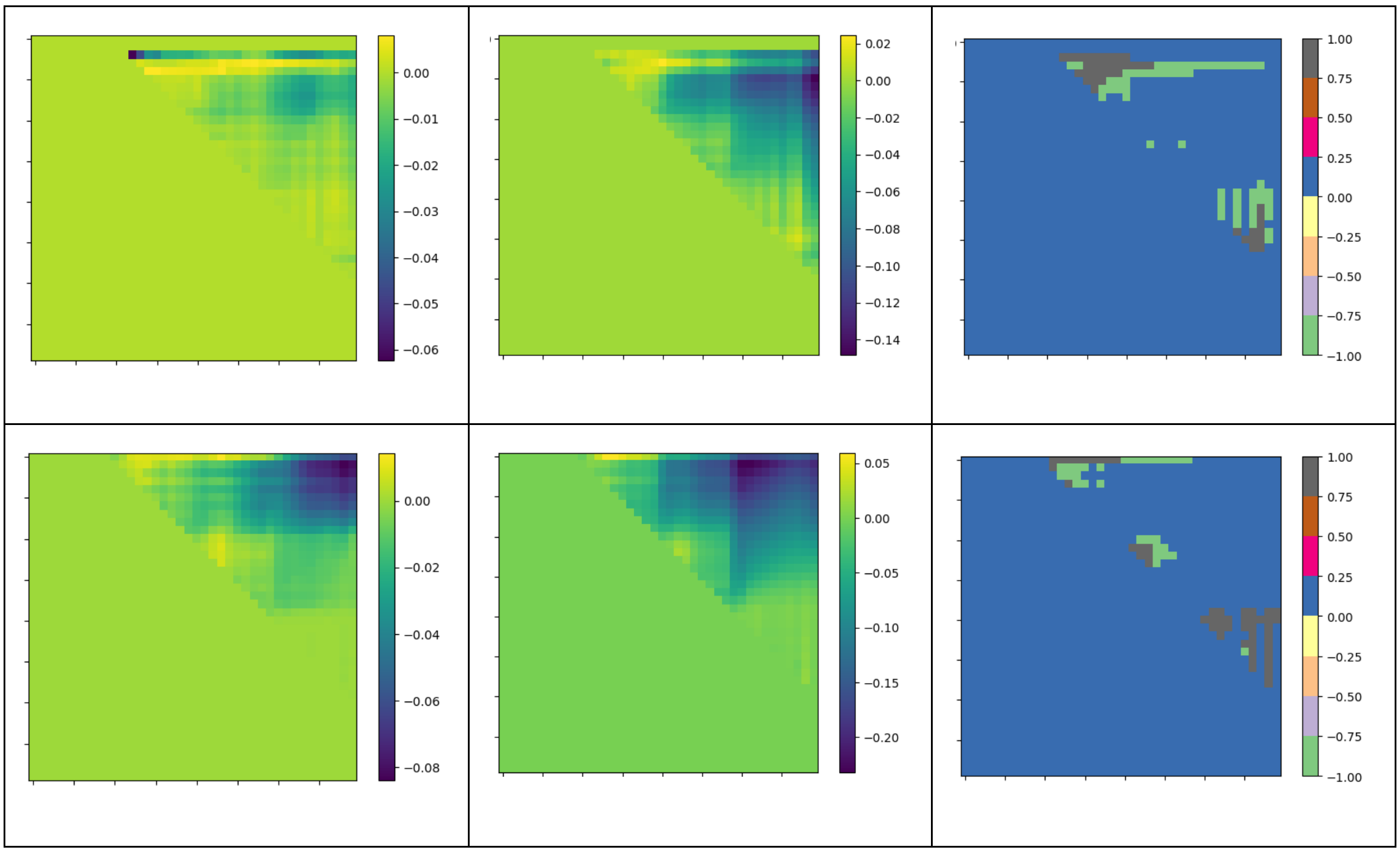

Figure 3: Autocorrelations in Critique of Pure Reason in English (top) and The Adventures of Tom Sawyer in Spanish (bottom) computed using GloVe, $a = 1, d = 300$. Vertical axis: start of $\tau$ range. Horizontal axis: end of $\tau$ range. Left: difference between power and log approximation MAPE. Middle: difference between power and exp approximation MAPE. Right: ranges where power (blue), exp (gray), and log (green) approximations are the best.

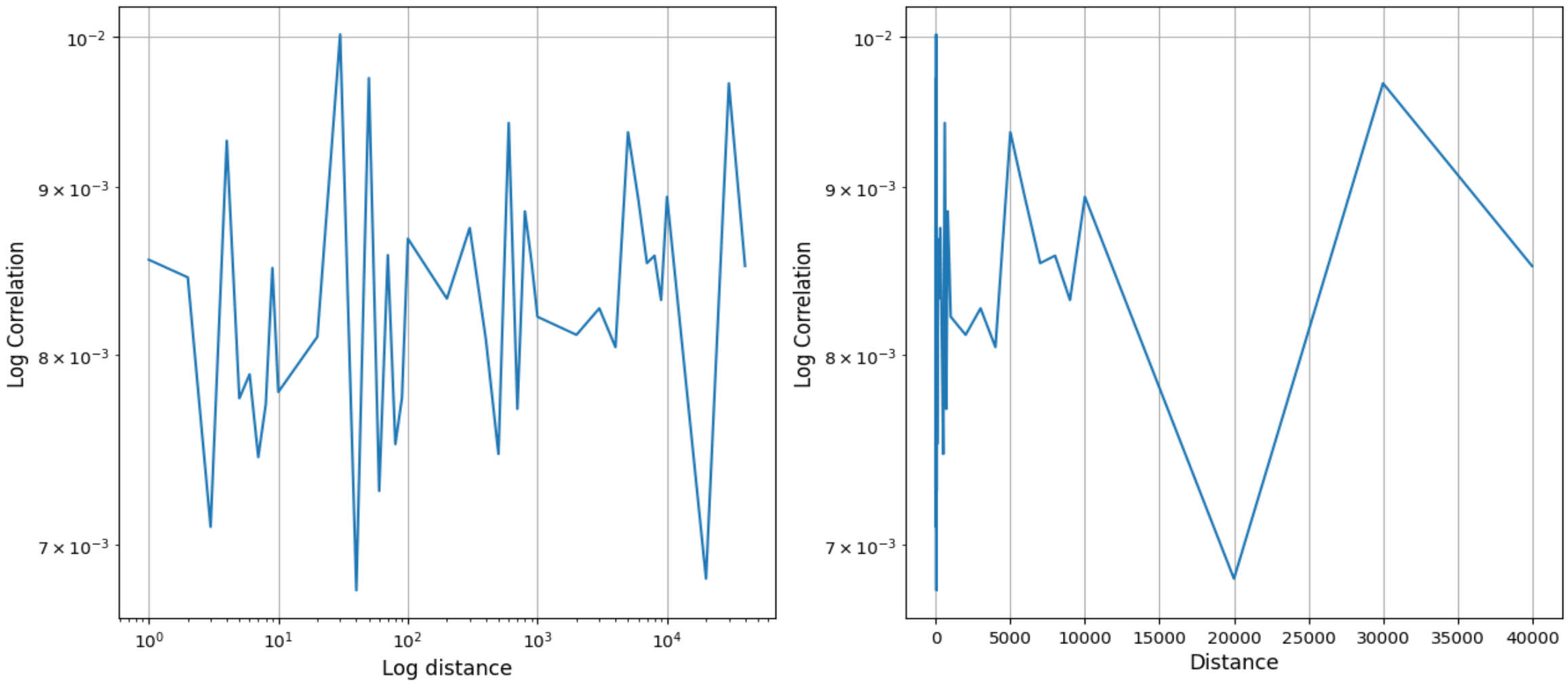

Figure 4: Autocorrelations in a randomly shuffled The Adventures of Tom Sawyer in Spanish computed using GloVe, a=1, d=300. Left: log-log, to right: log-linear coordinates

are always power law. Additionally, the location of such ranges is hectic. We conclude that the cases where exponential or logarithmic approximation is better than the power law approximation represent natural short-range variability and cannot be considered a regularity.

### 5.5 Autocorrelations in Generated Texts

The behavior of autocorrelations is qualitatively different when the text is generated. The simplest way to generate an incoherent text is to shuffle words in a text. Figure 4 demonstrates that there is no specific autocorrelations decay law for an incoherent text.

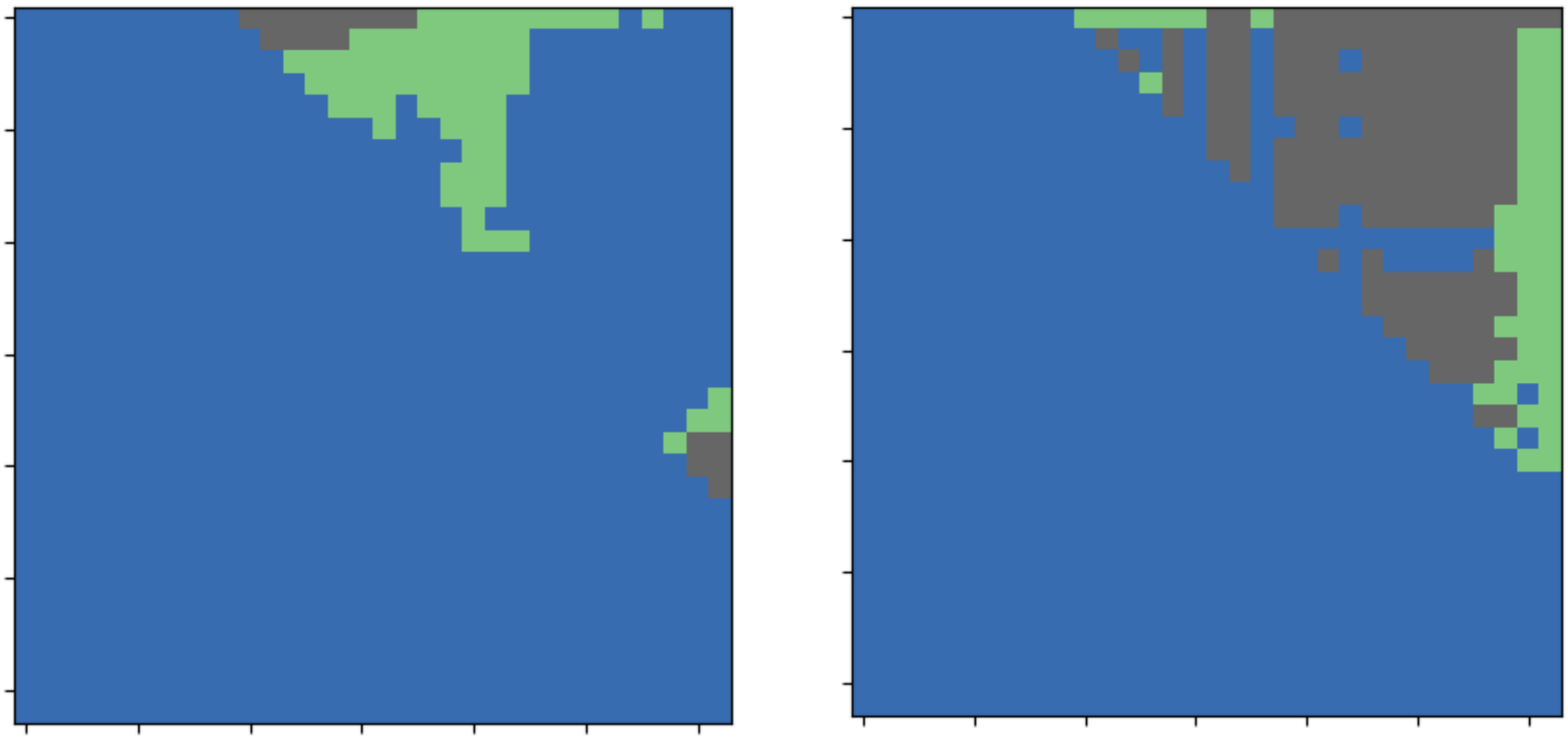

Figure 5: Autocorrelations in texts generated by GPT-2 (left) and S4 (right) models computed using GloVe, $a = 1, d = 300$, ranges where power (blue), exp (gray), and log (green) approximations are the best depicted. Vertical axis: start of $\tau$ range. Horizontal axis: end of $\tau$ range.

To study autocorrelations in texts generated by large language models, we have used GPT-2 base [6] with the default generation parameters, and Structured State Space model S4 base [19] with the default generation parameters, and generated some 10K word continuous text with each model. The generated texts are listed in Appendix C and Appendix D, respectively. We then performed the same procedure as in Section 5.4, mapping ranges where each decay law provides the best approximation. The results are presented on Figure 5.

The autocorrelations decay in an exponential manner in the text generated by S4 model, while according to a power law on long distances and to power or log law – on short distances in the text generated by GPT-2. The autocorrelations in generated texts are significantly larger and decay much slower than the ones in the natural texts. In our S4 and GPT-2 generated examples, the power law coefficients are $a = -0.045, b = -0.71$ and $a = -0.027, b = -0.77$, respectively. At the same time we have not seen the coefficient a less than 0.1 for any natural text in English we have studied, and the average is closer to 0.2, indicating almost 10-fold gap between the power law decay rates in natural and generated texts. Typical values of coefficient b for natural texts are between -1.5 and -2, indicating at least 2-fold gap between natural and generated texts.

Thus we can say that the autocorrelations decay in generated texts are quantitatively and often qualitatively different from the literary texts. The conditions that influence the autocorrelations decay laws in generated texts may include sampling approach, temperature and other hyperparameters. This is a matter of future research.

## 6 Conclusions

We have shown empirically that autocorrelations in literary texts are decaying following the power law with the only upper limit being the length of the work itself and the hypothesis of exponential decay can be rejected for these distances. We have also shown empirically that the laws of autocorrelation decay, if measured using distributional semantics models are typically the same for the literary work translated to different languages. This contrasts previous findings that used flawed technique based on encoding-dependent random walks. Thus, we believe that distributional semantics models are a robust enough tool to measure autocorrelations in long texts.

The autocorrelations decay in generated texts is quantitatively and often qualitatively different from the literary texts. Based on the above, we can conclude that for long text processing one may need architectures different from the autoregressive ones, and many questions remain unanswered.

## Acknowledgements

The authors are grateful to their colleagues at NTR Labs ML division for the discussions and support. Early versions of this work were discussed with Anton Kolonin, Dmitry Manin and Alexey Tikhonov. These discussions have improved our approach and research design, for which we are very grateful. We are also extremely grateful to Tatiana Sherstinova who discussed early versions of this work, suggested numerous improvements and provided a webinar venue at HSE to discuss this work publicly.

# Appendix A

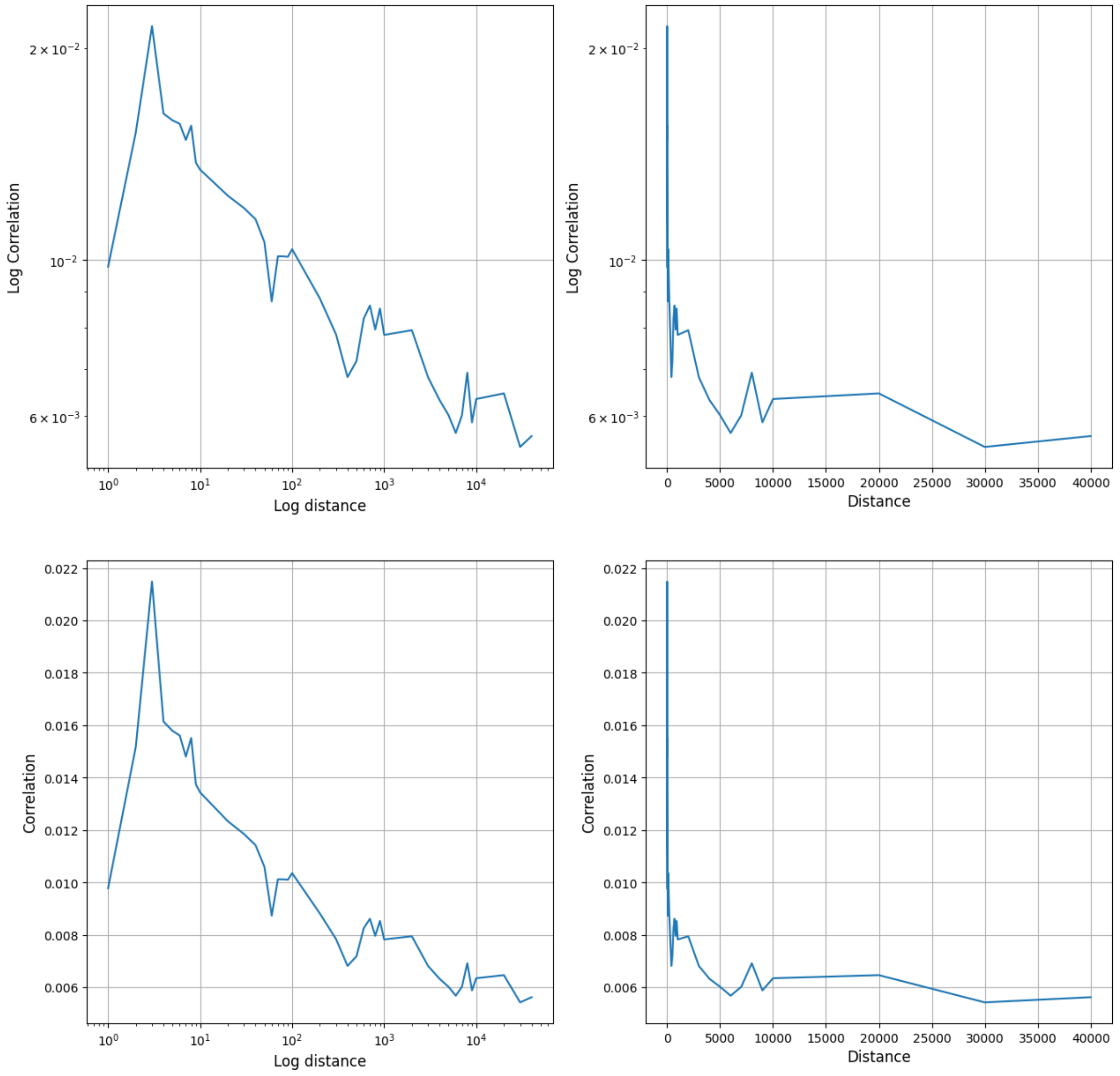

Figure 6: GloVe computation of autocorrelations in The Republic in Spanish in different coordinates

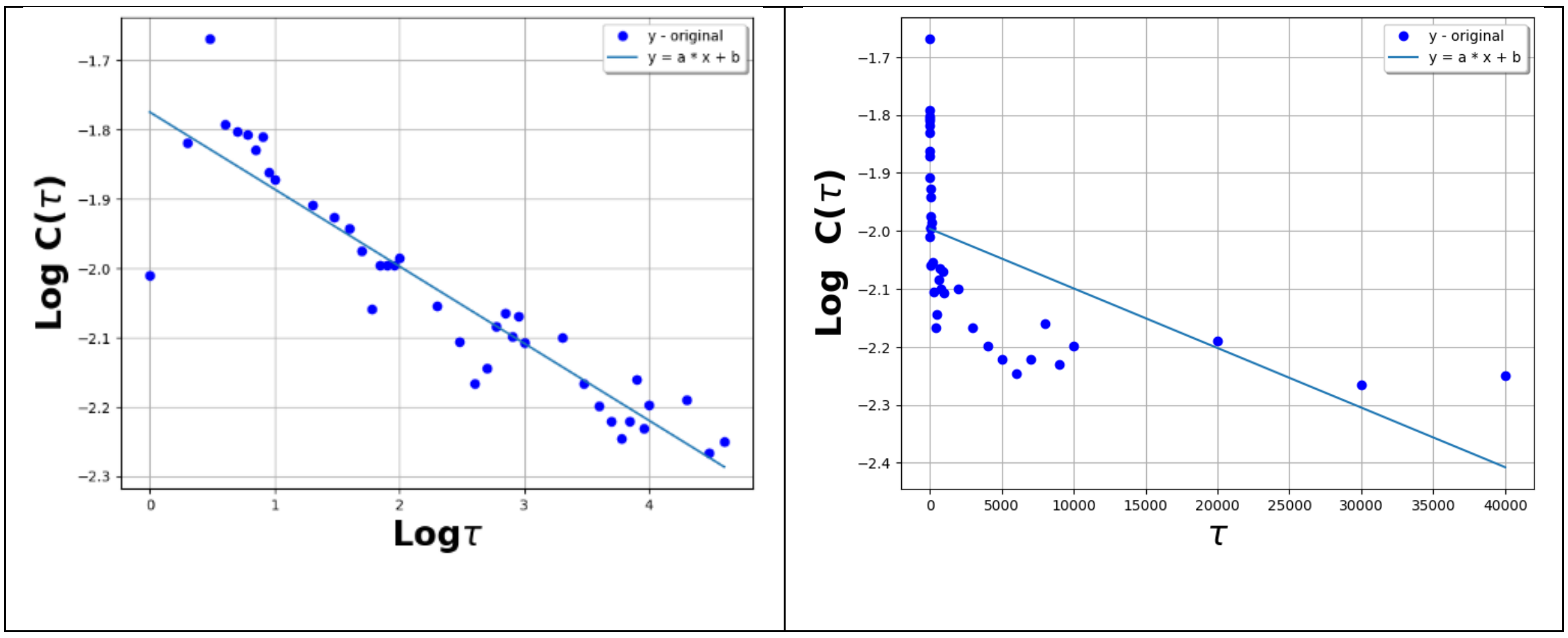

Figure 7: Best fit of GloVe computation of autocorrelations in The Republic in Spanish by power (left) and exp (right) functions

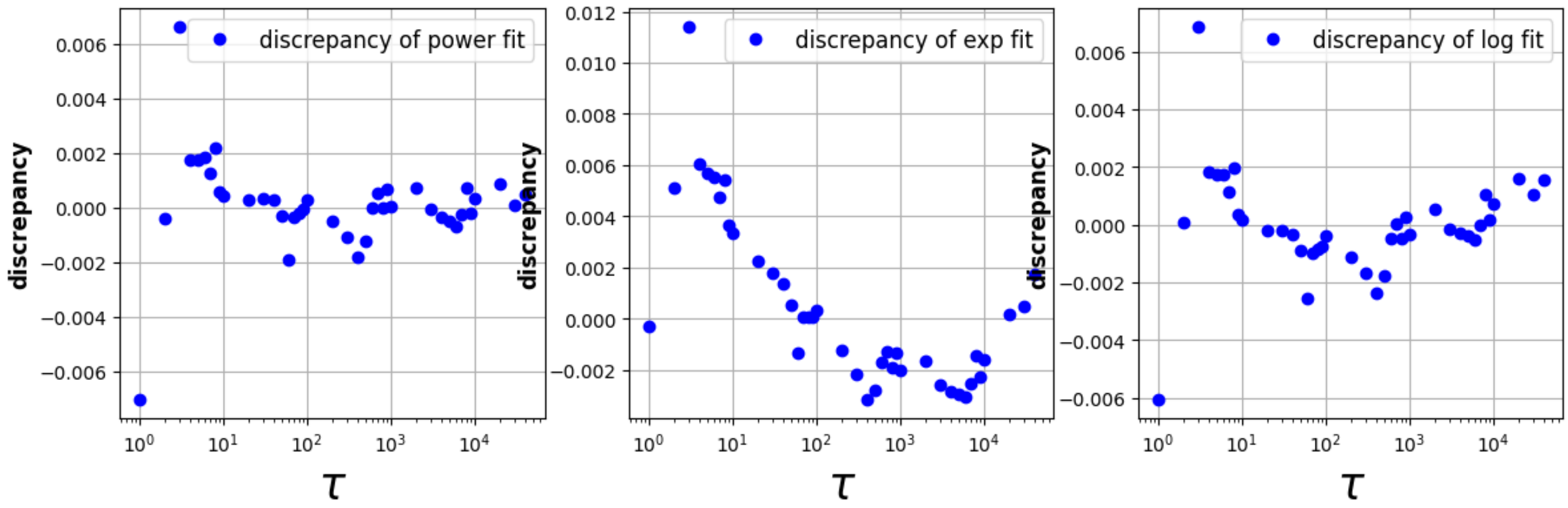

Figure 8: Residual graphs of the best fit of GloVe computation of autocorrelations in The Republic in Spanish

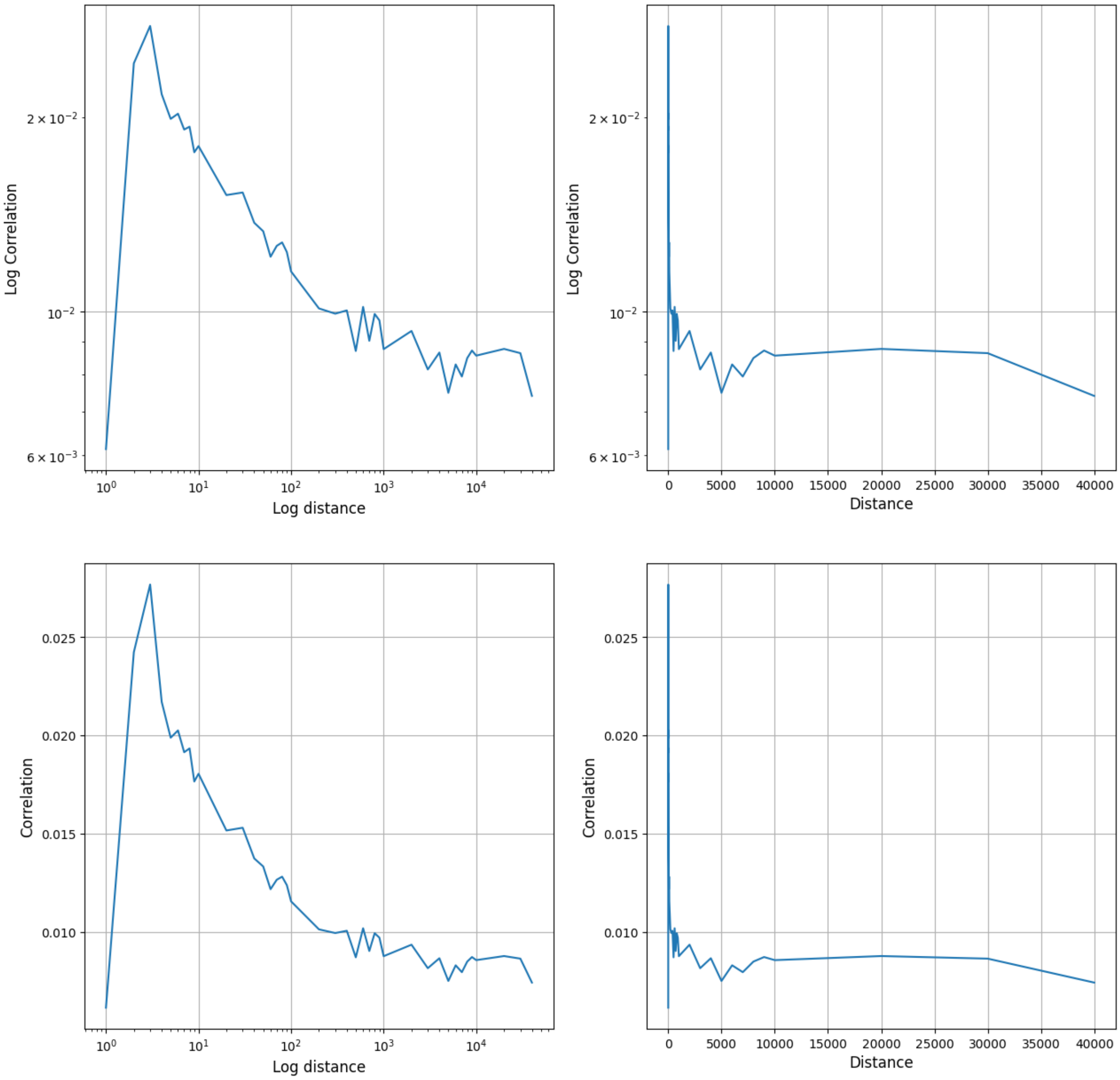

Figure 9: GloVe computation of autocorrelations in The Republic in French in different coordinates

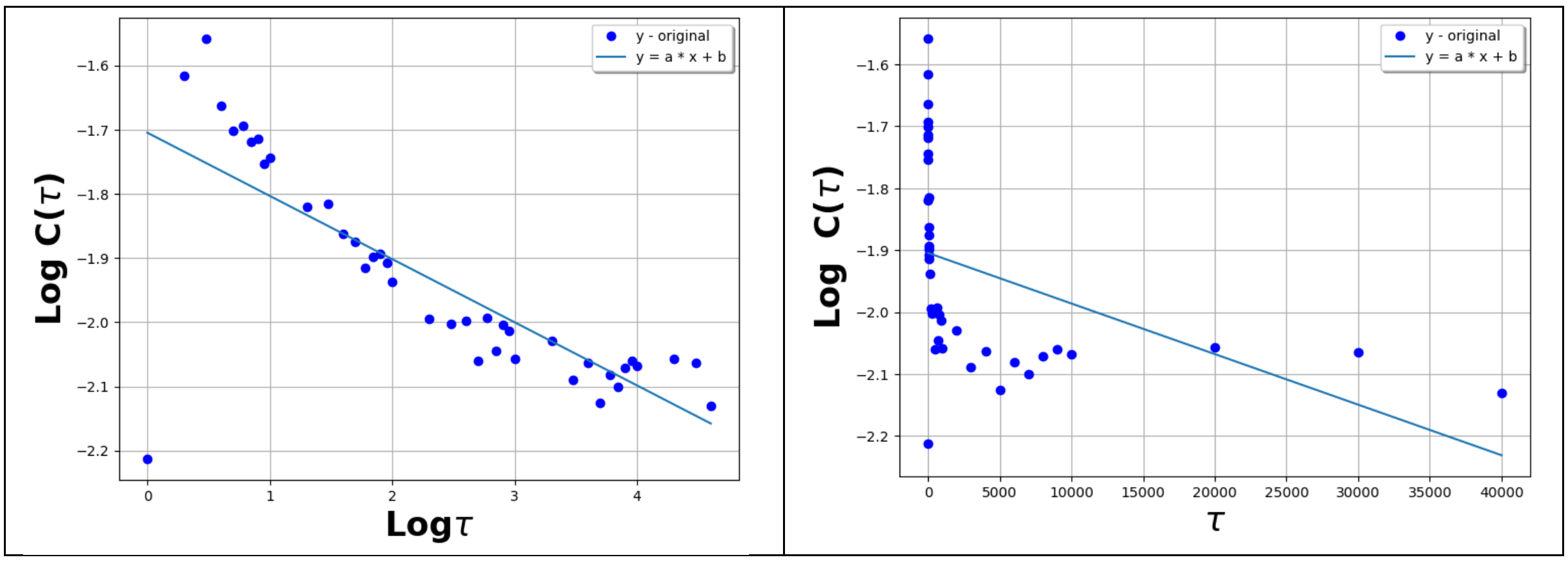

Figure 10: Best fit of GloVe computation of autocorrelations in The Republic in French by power (left) and exp (right) functions

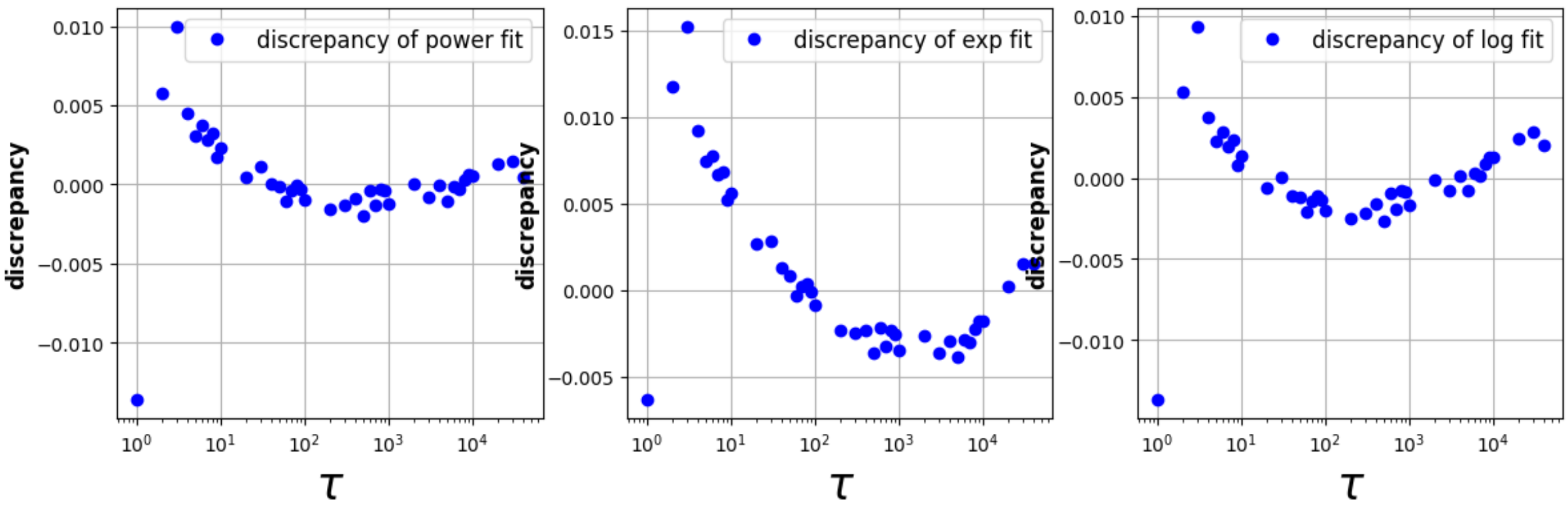

Figure 11: Residual graphs of the best fit of GloVe computation of autocorrelations in The Republic in French

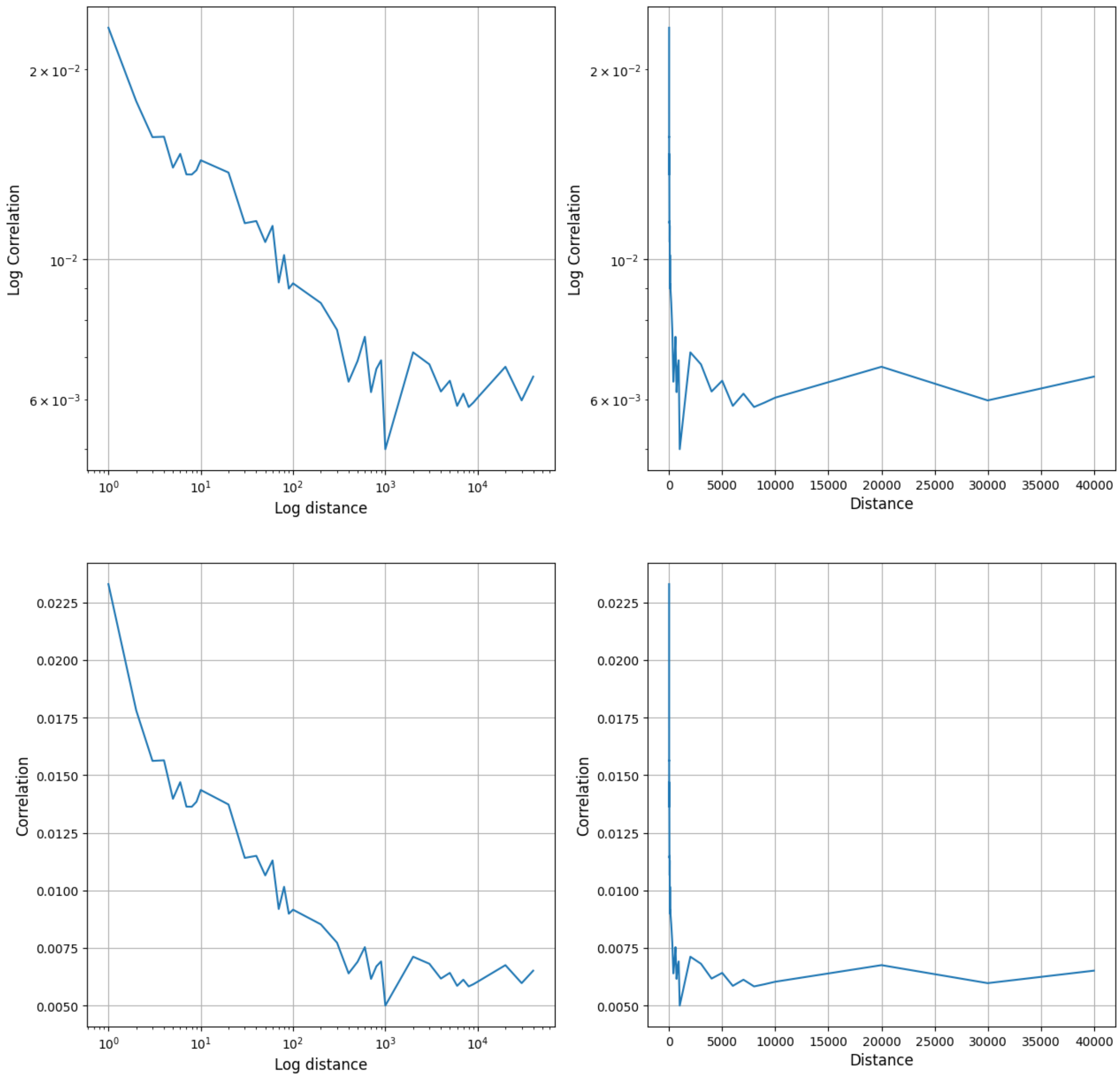

Figure 12: GloVe computation of autocorrelations in The Republic in Russian in different coordinates

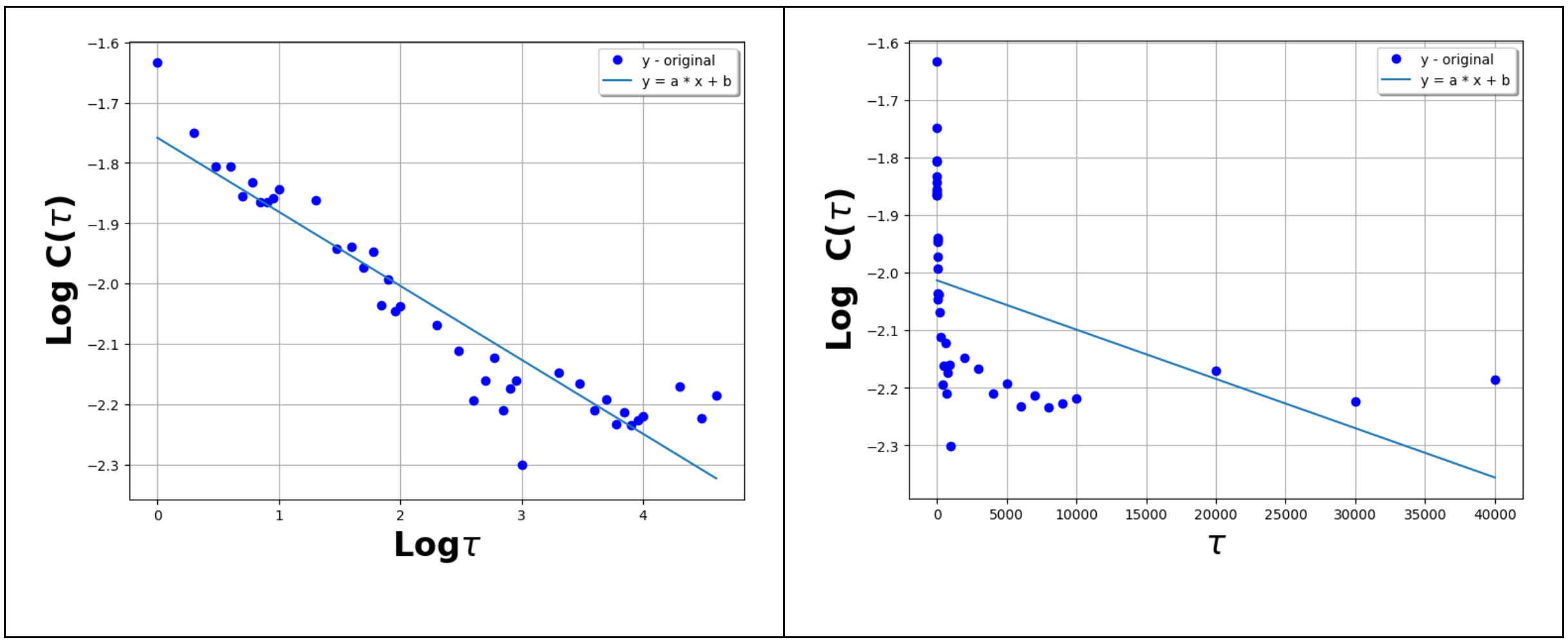

Figure 13: Best fit of GloVe computation of autocorrelations in The Republic in Russian by power (left) and exp (right) functions

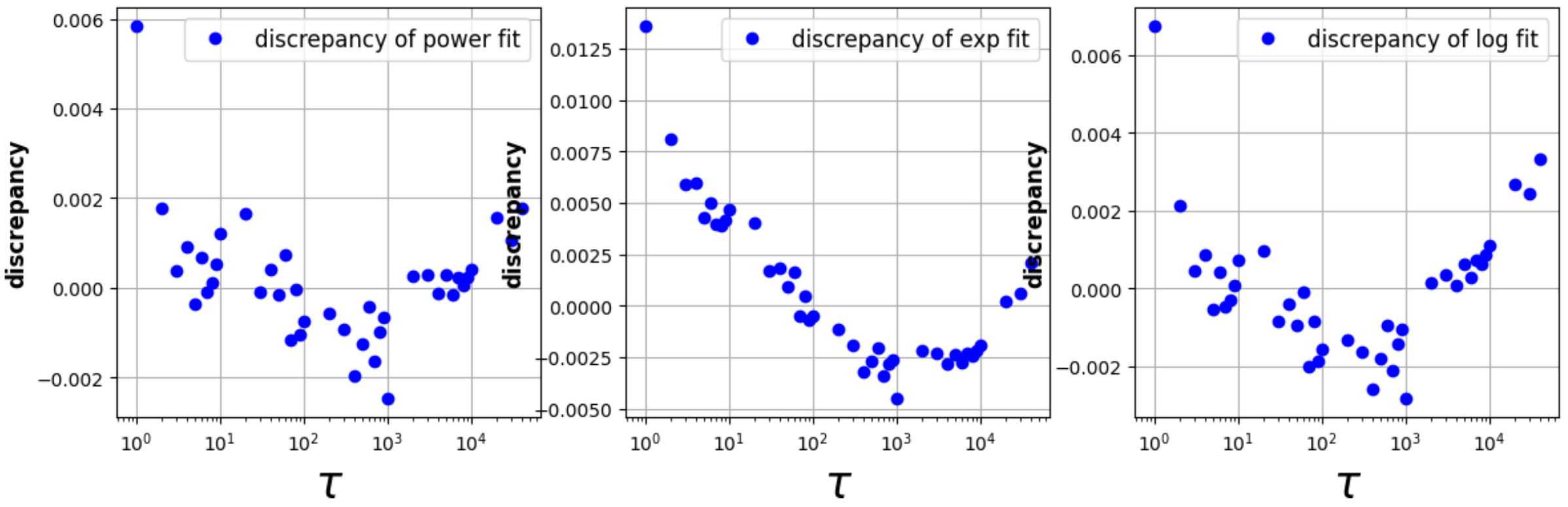

Figure 14: Residual graphs of the best fit of GloVe computation of autocorrelations in The Republic in Russian

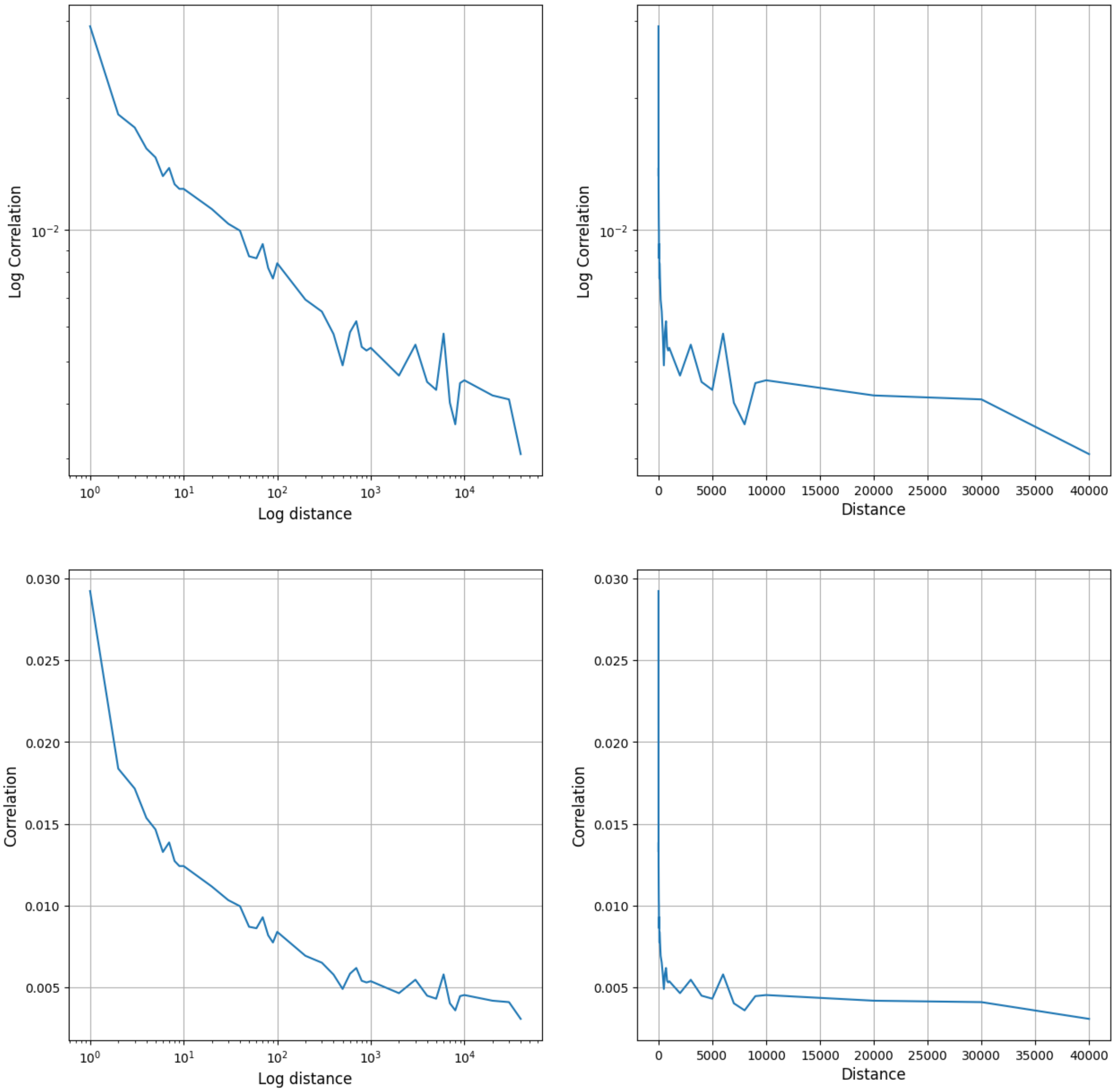

Figure 15: GloVe computation of autocorrelations in The Republic in German in different coordinates

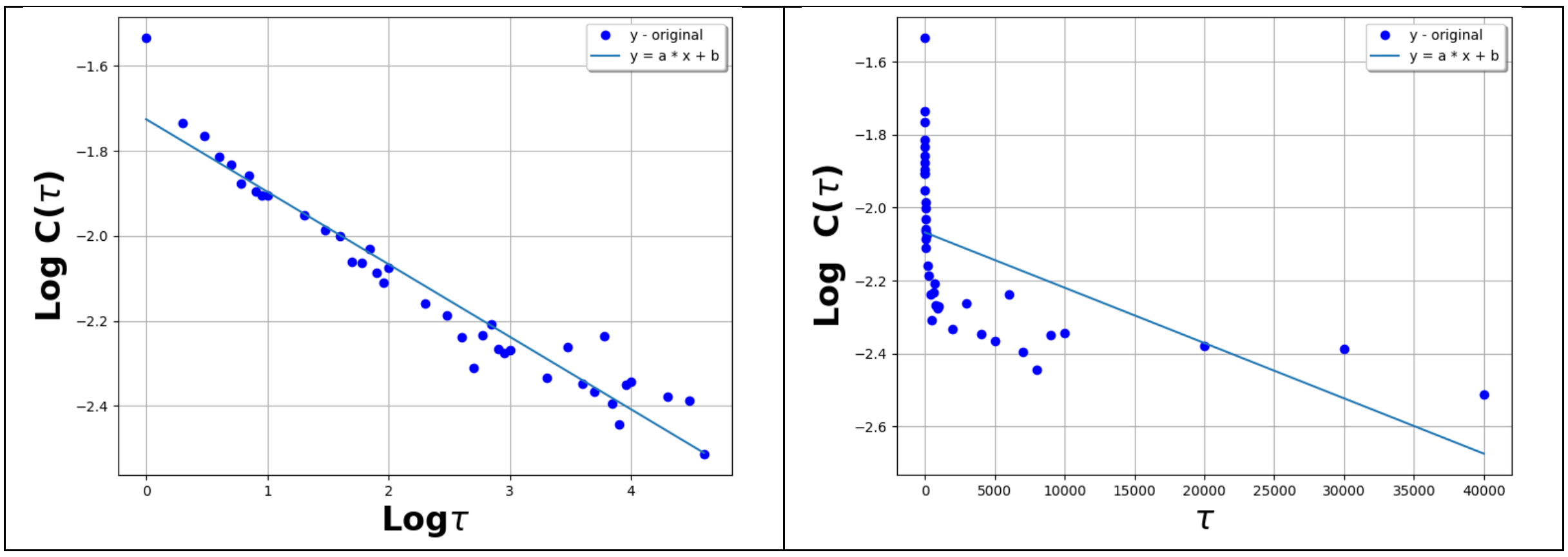

Figure 16: Best fit of GloVe computation of autocorrelations in The Republic in German by power (left) and exp (right) functions

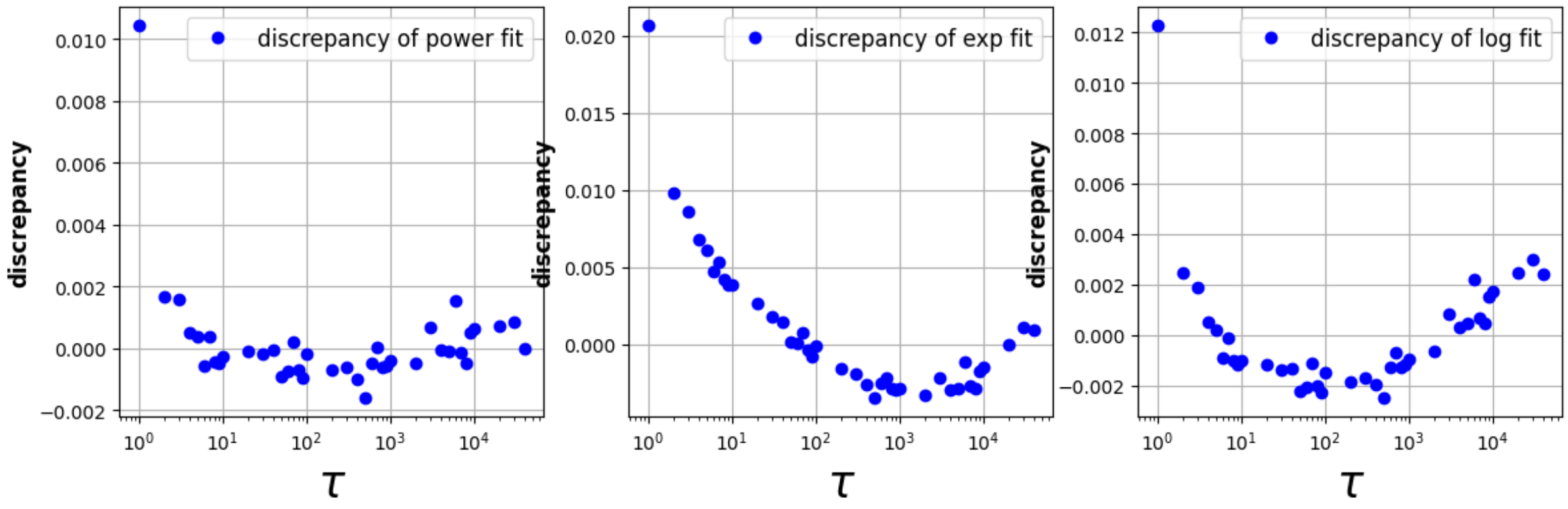

Figure 17: Residual graphs of the best fit of GloVe computation of autocorrelations in The Republic in German

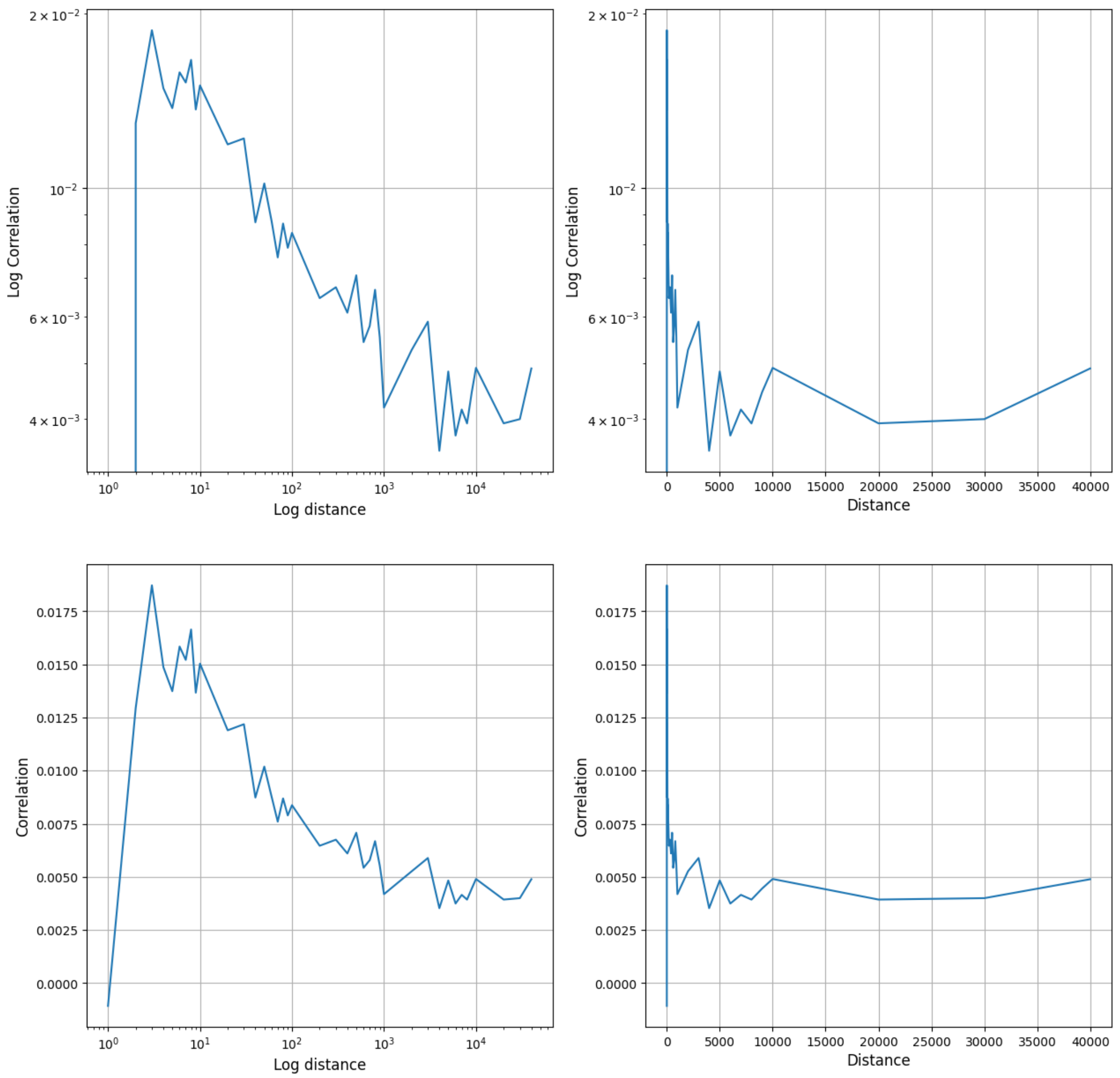

Figure 18: GloVe computation of autocorrelations in The Republic in English in different coordinates

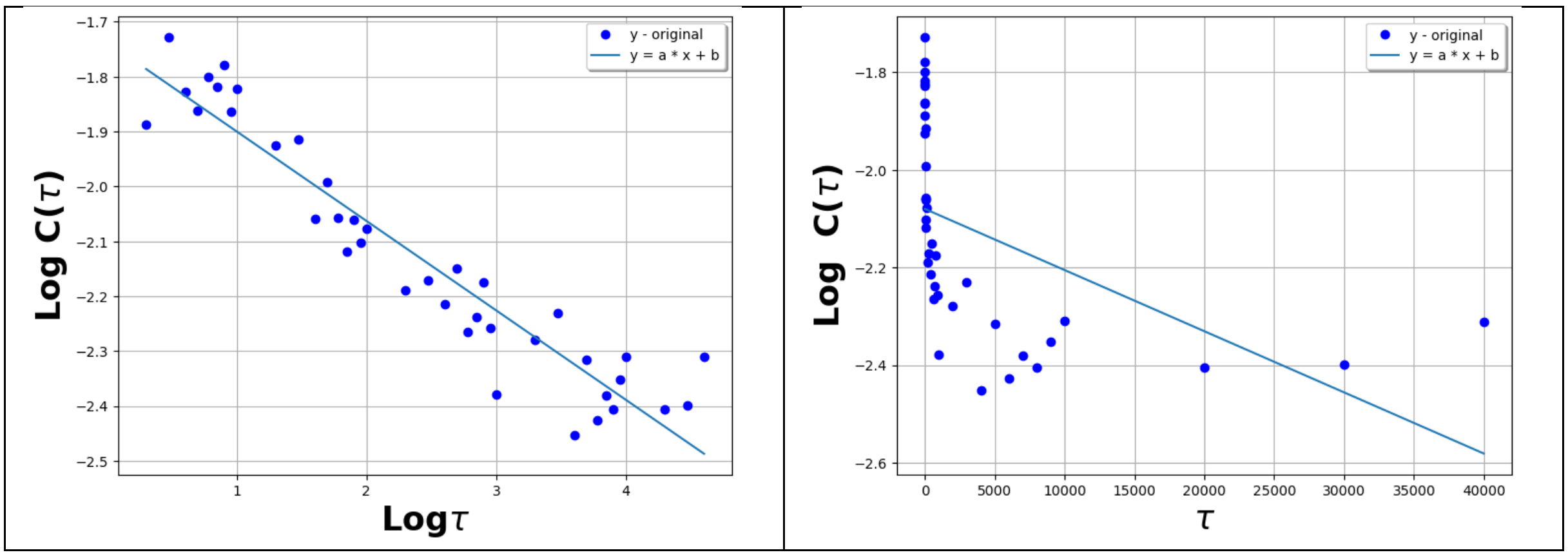

Figure 19: Best fit of GloVe computation of autocorrelations in The Republic in Engilsh by power (left) and exp (right)

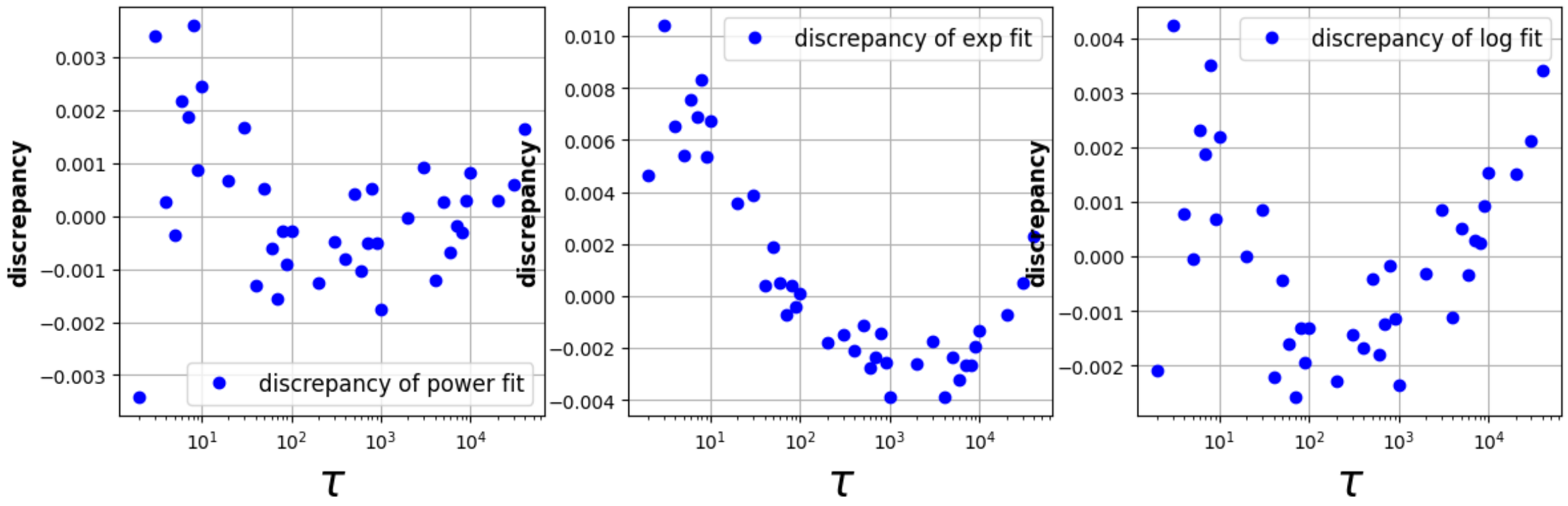

Figure 20: Residual graphs of the best fit of GloVe computation of autocorrelations in The Republic in German

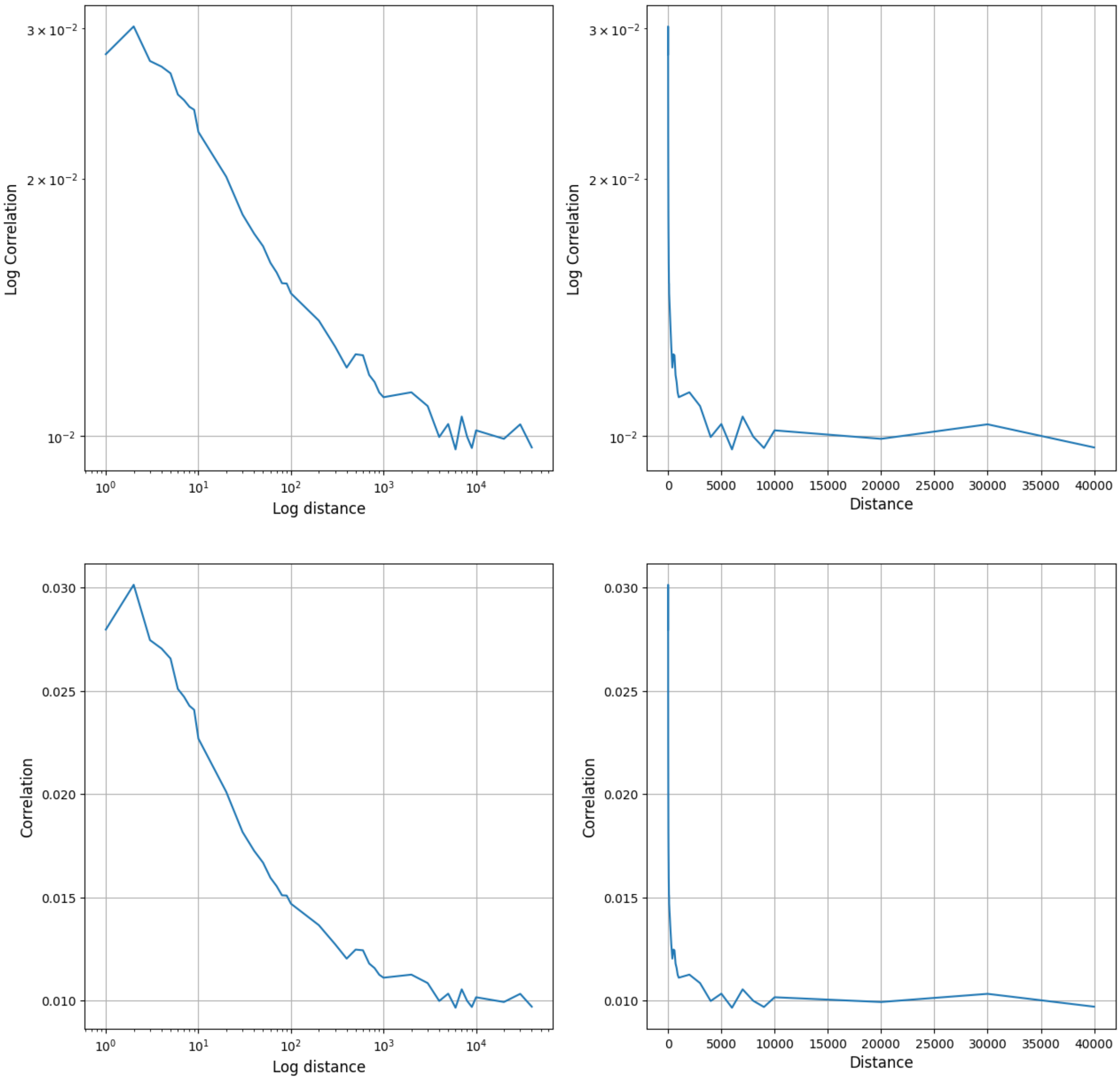

Figure 21: GloVe computation of autocorrelations in War and Pease in Russian in different coordinates

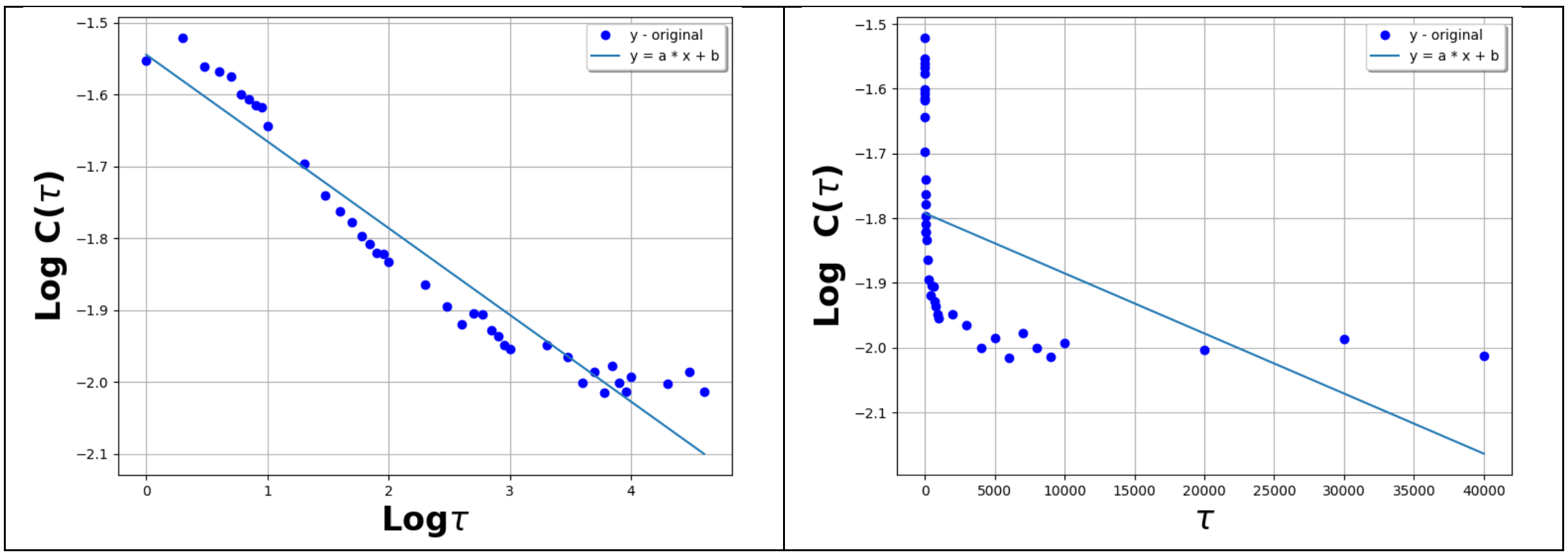

Figure 22: Best fit of GloVe computation of autocorrelations in War and Pease in Russian by power (left) and exp (right)

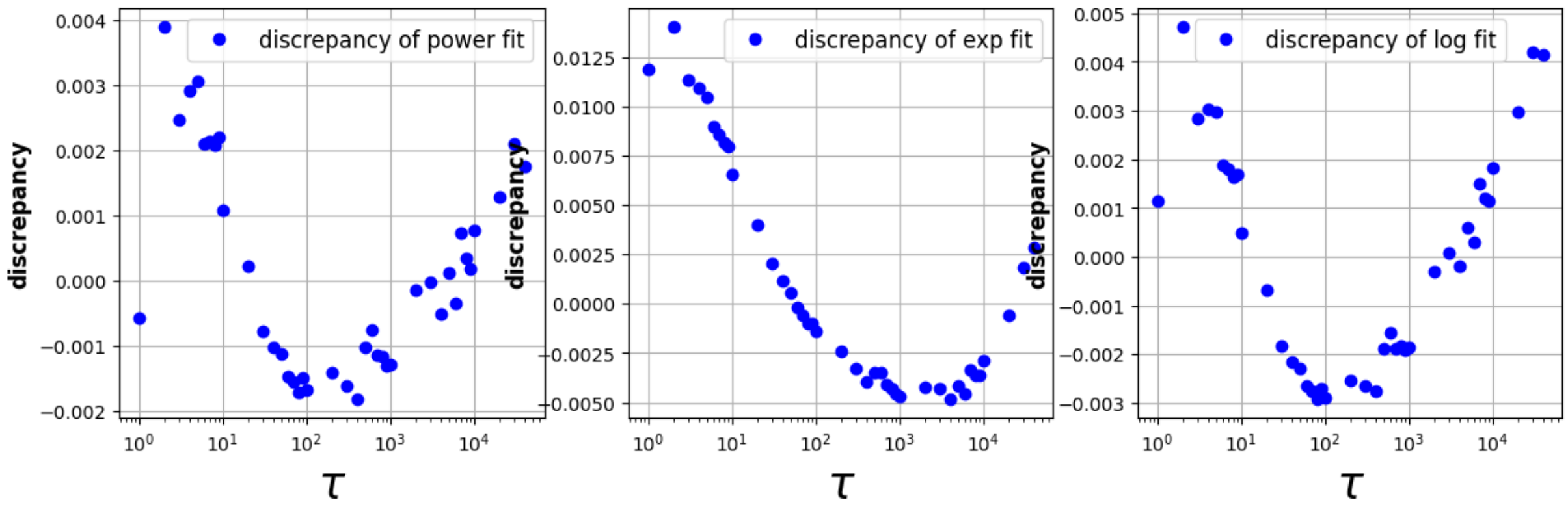

Figure 23: Residual graphs of the best fit of GloVe computation of autocorrelations in in War and Pease in Russian

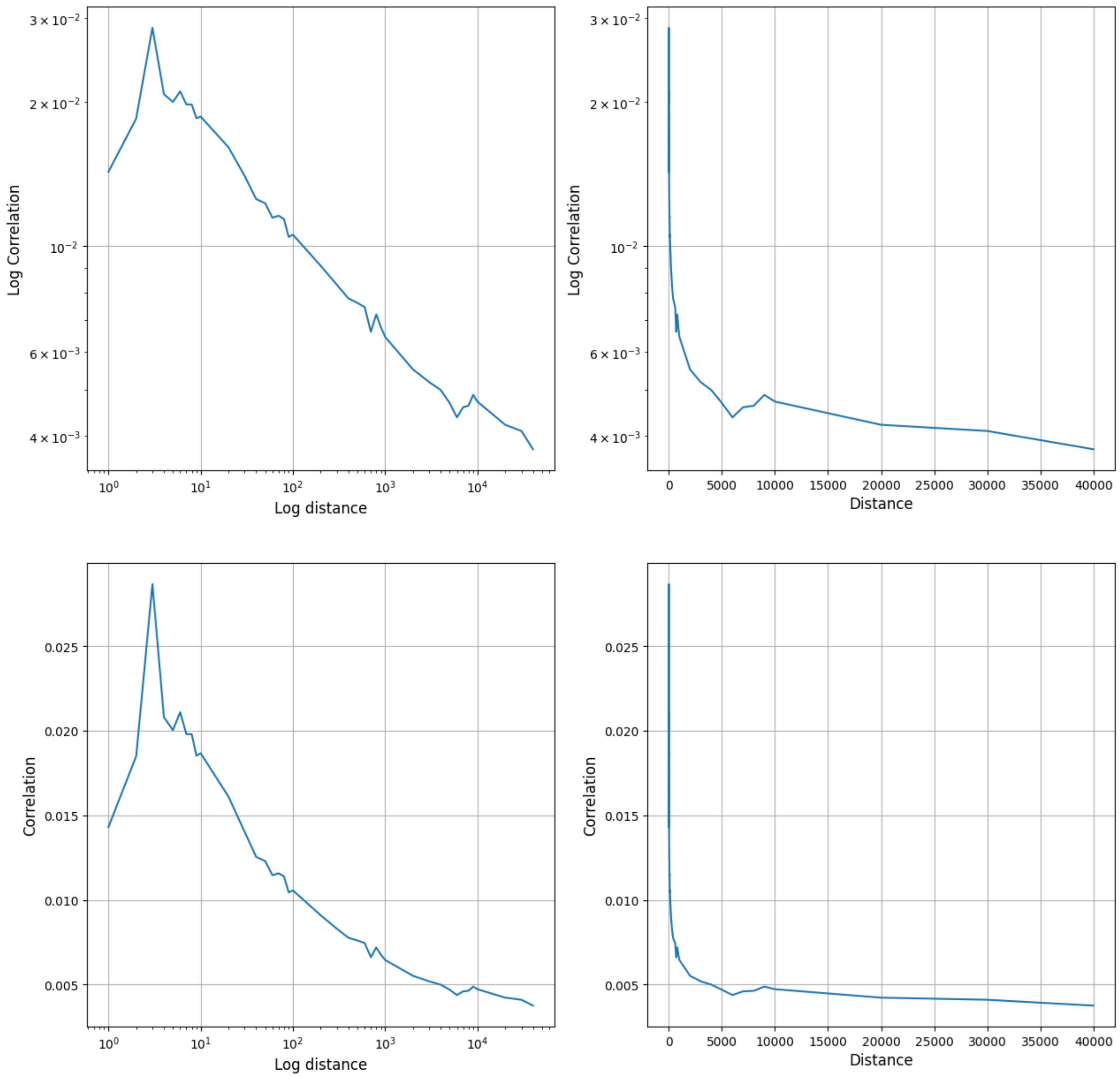

Figure 24: GloVe computation of autocorrelations in War and Pease in English in different coordinates

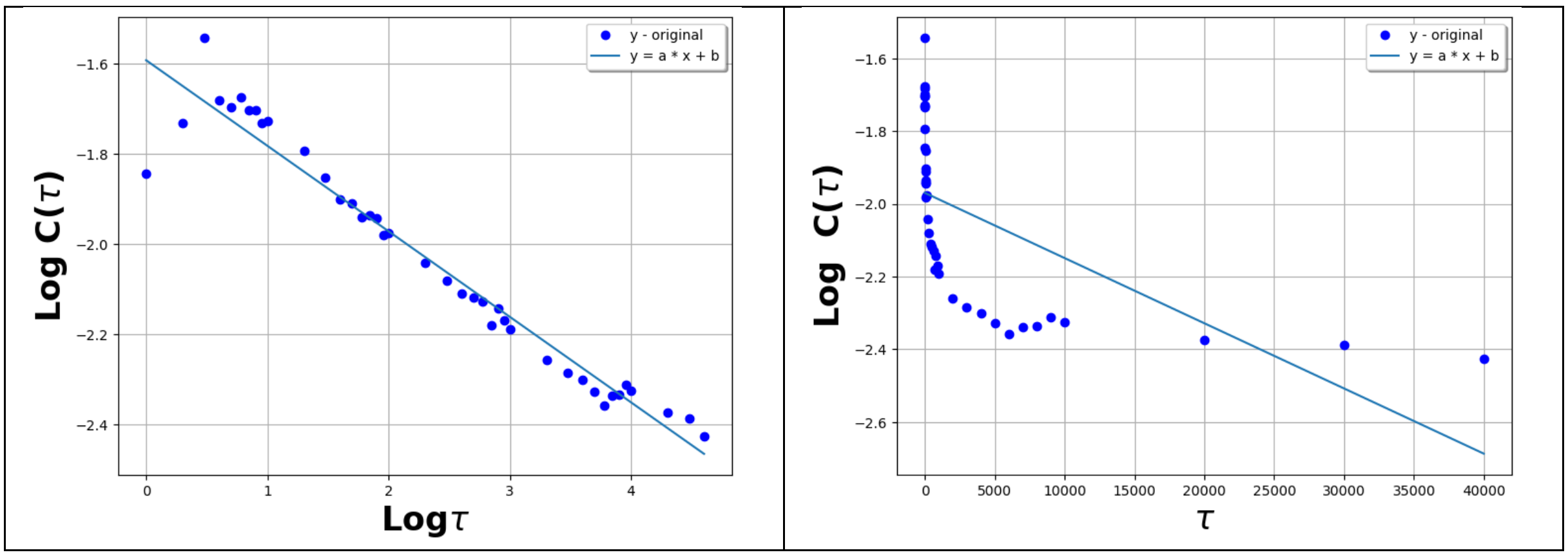

Figure 25: Best fit of GloVe computation of autocorrelations in War and Pease in English by power (left) and exp (right)

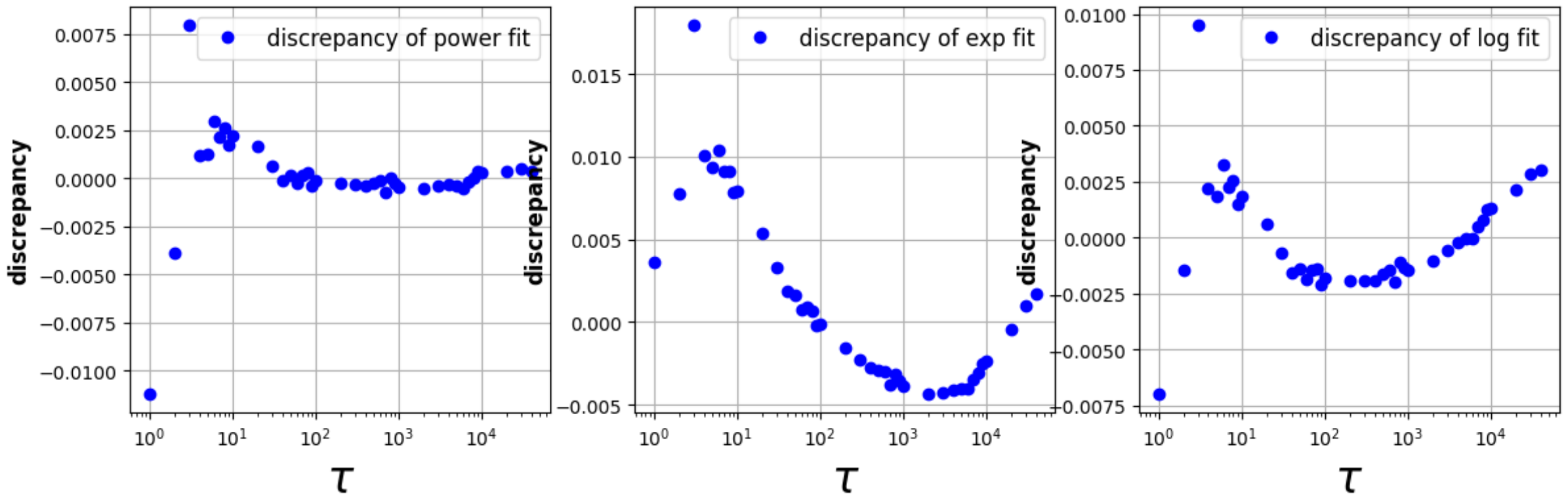

Figure 26: Residual graphs of the best fit of GloVe computation of autocorrelations in in War and Pease in English

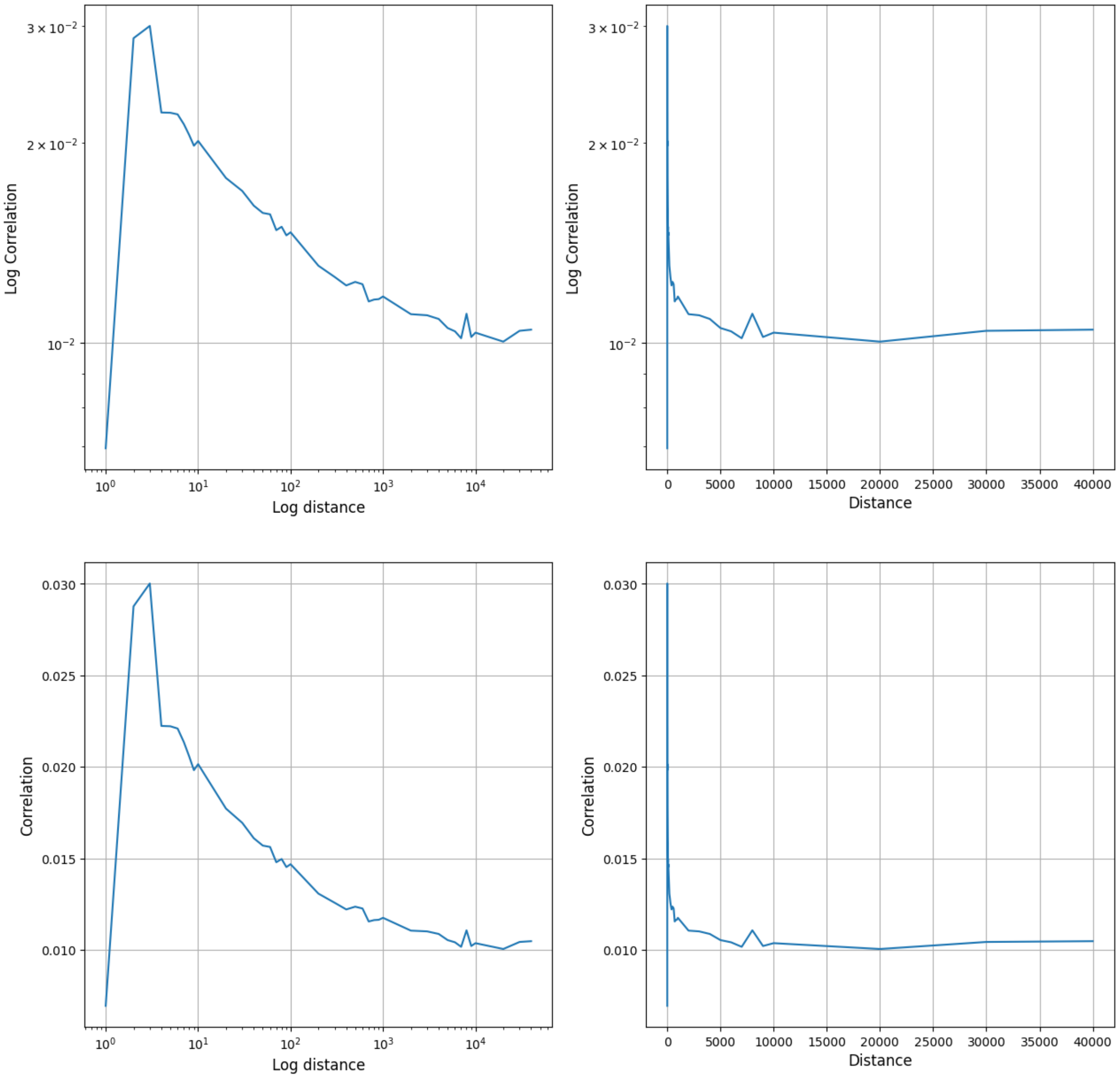

Figure 27: GloVe computation of autocorrelations in War and Pease in French in different coordinates

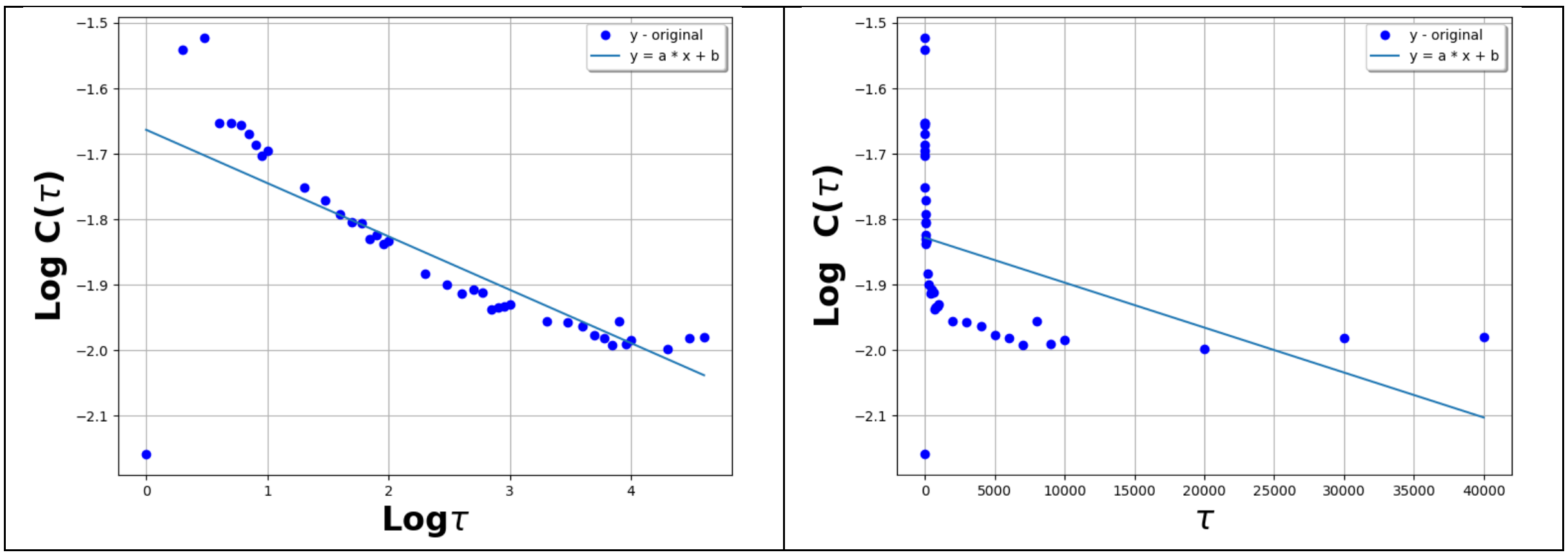

Figure 28: Best fit of GloVe computation of autocorrelations in War and Pease in French by power (left) and exp (right)

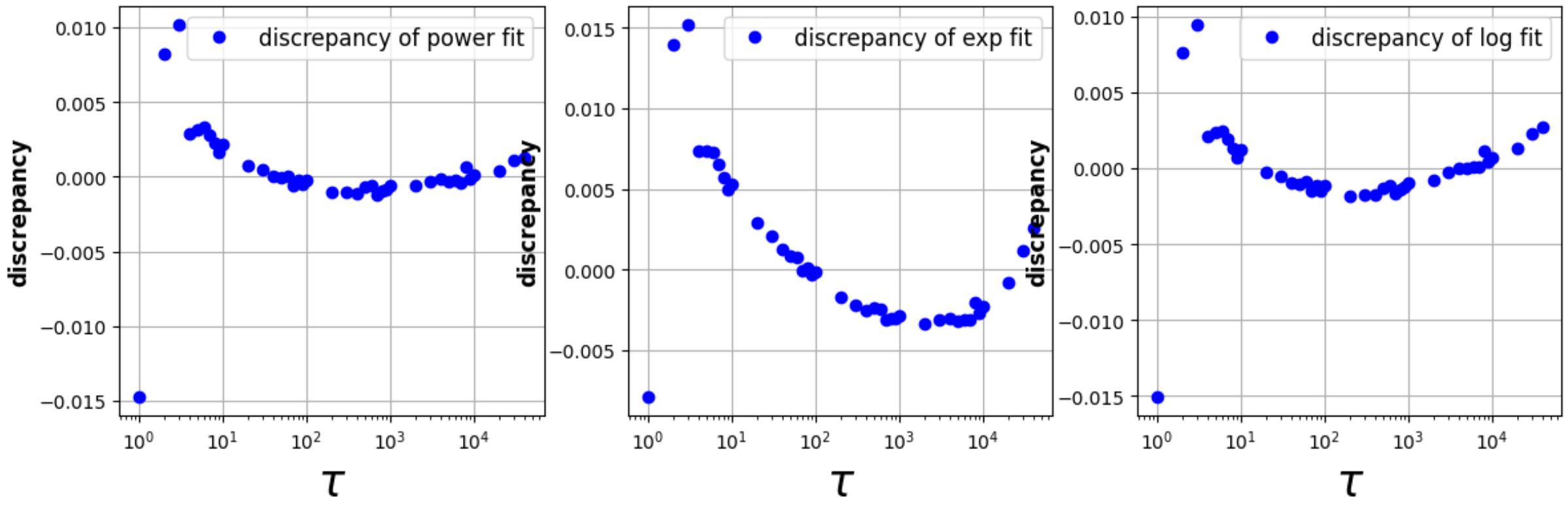

Figure 29: Residual graphs of the best fit of GloVe computation of autocorrelations in in War and Pease in French

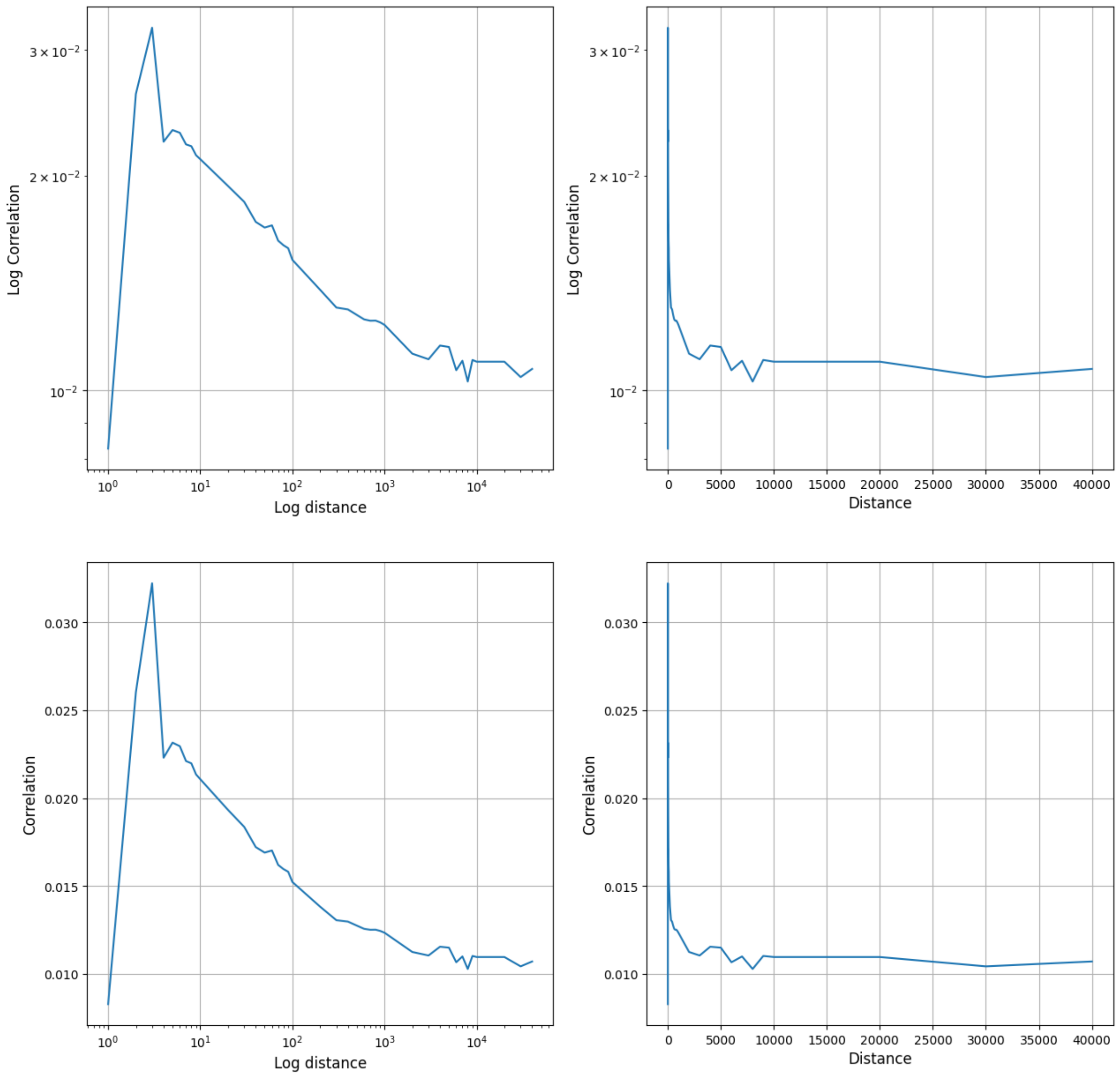

Figure 30: GloVe computation of autocorrelations in War and Pease in Spanish in different coordinates

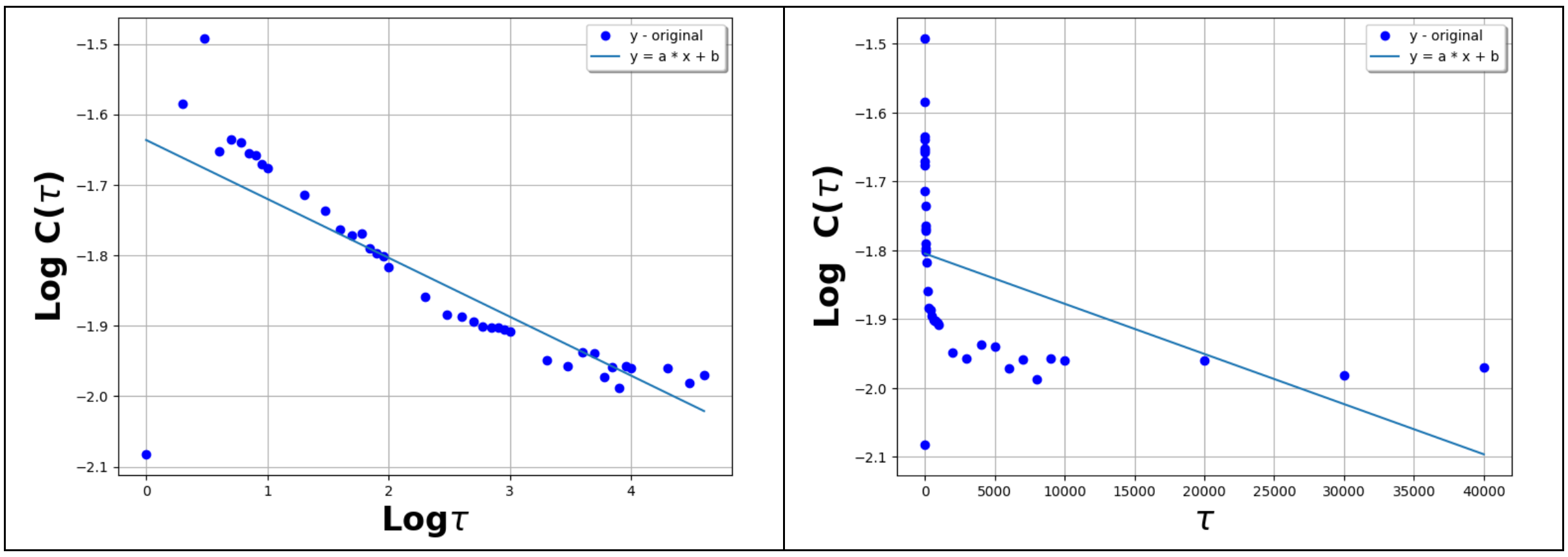

Figure 31: Best fit of GloVe computation of autocorrelations in War and Pease in Spanish by power (left) and exp (right)

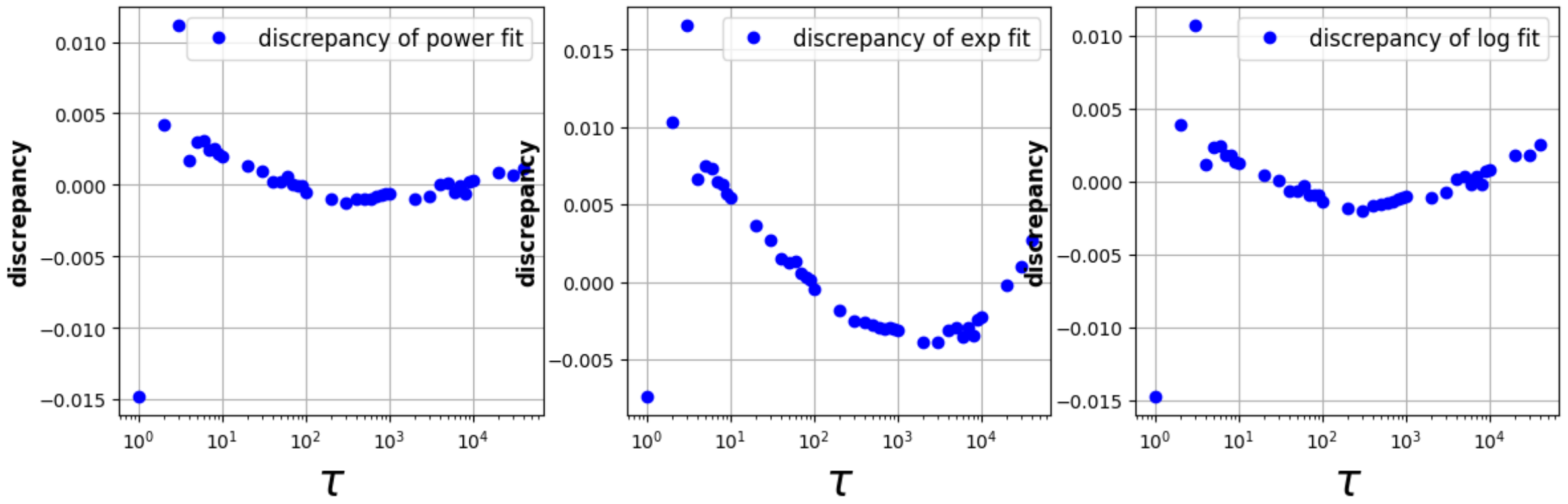

Figure 32: Residual graphs of the best fit of GloVe computation of autocorrelations in in War and Pease in Spanish

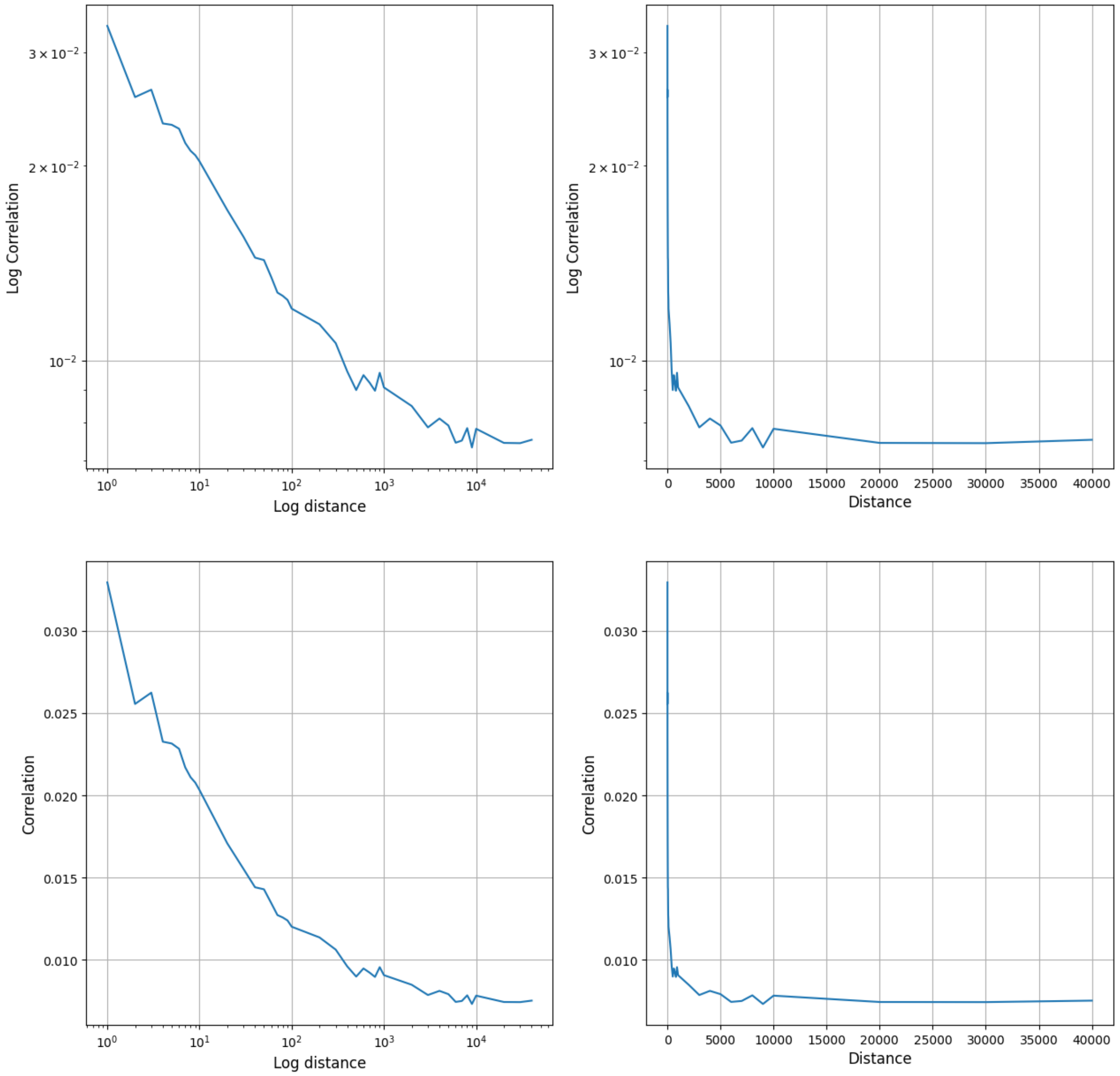

Figure 33: GloVe computation of autocorrelations in War and Pease in German in different coordinates

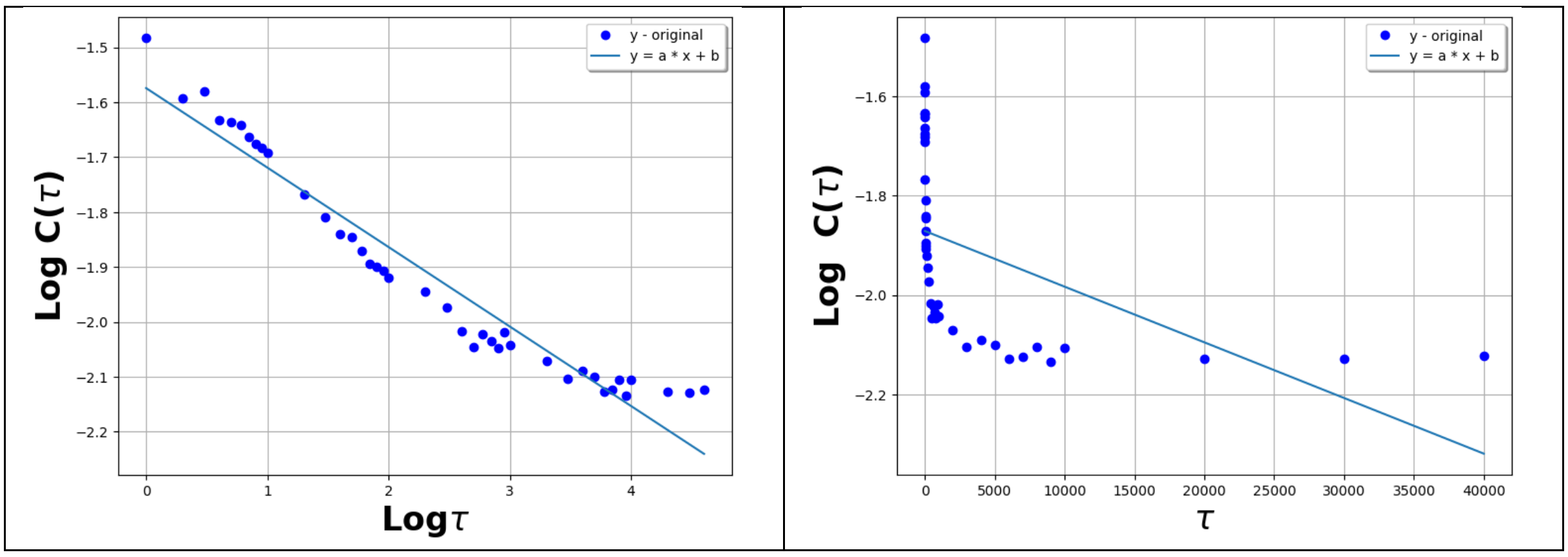

Figure 34: Best fit of GloVe computation of autocorrelations in War and Pease in German by power (left) and exp (right)

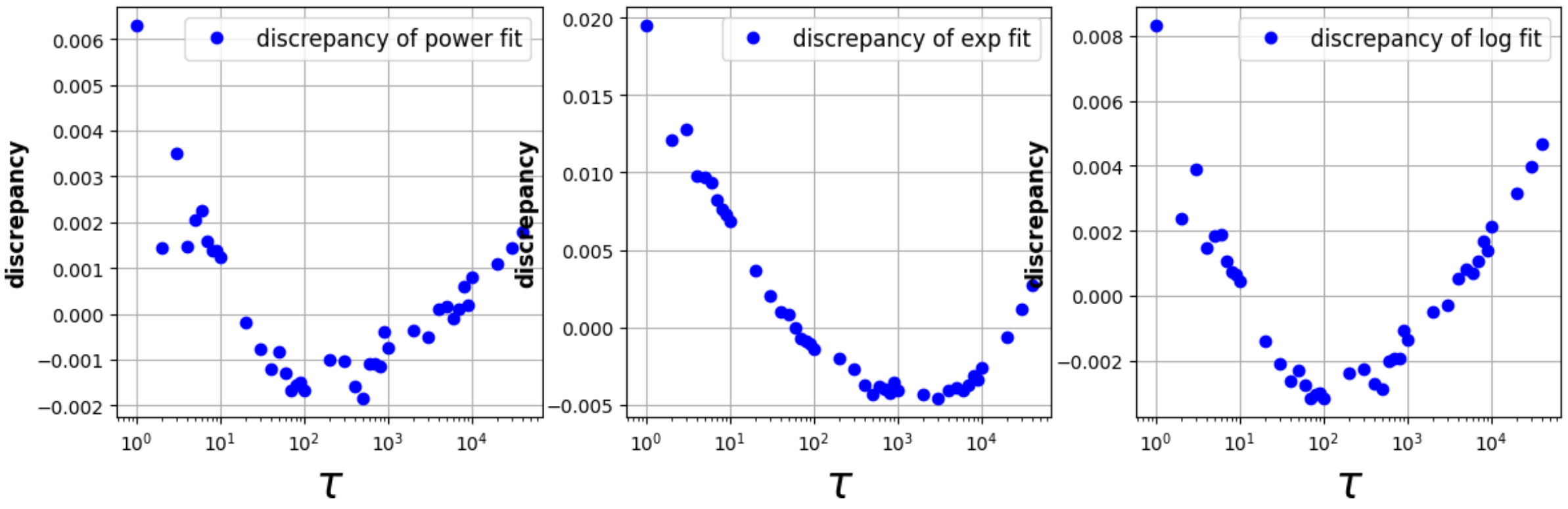

Figure 35: Residual graphs of the best fit of GloVe computation of autocorrelations in War and Pease in German

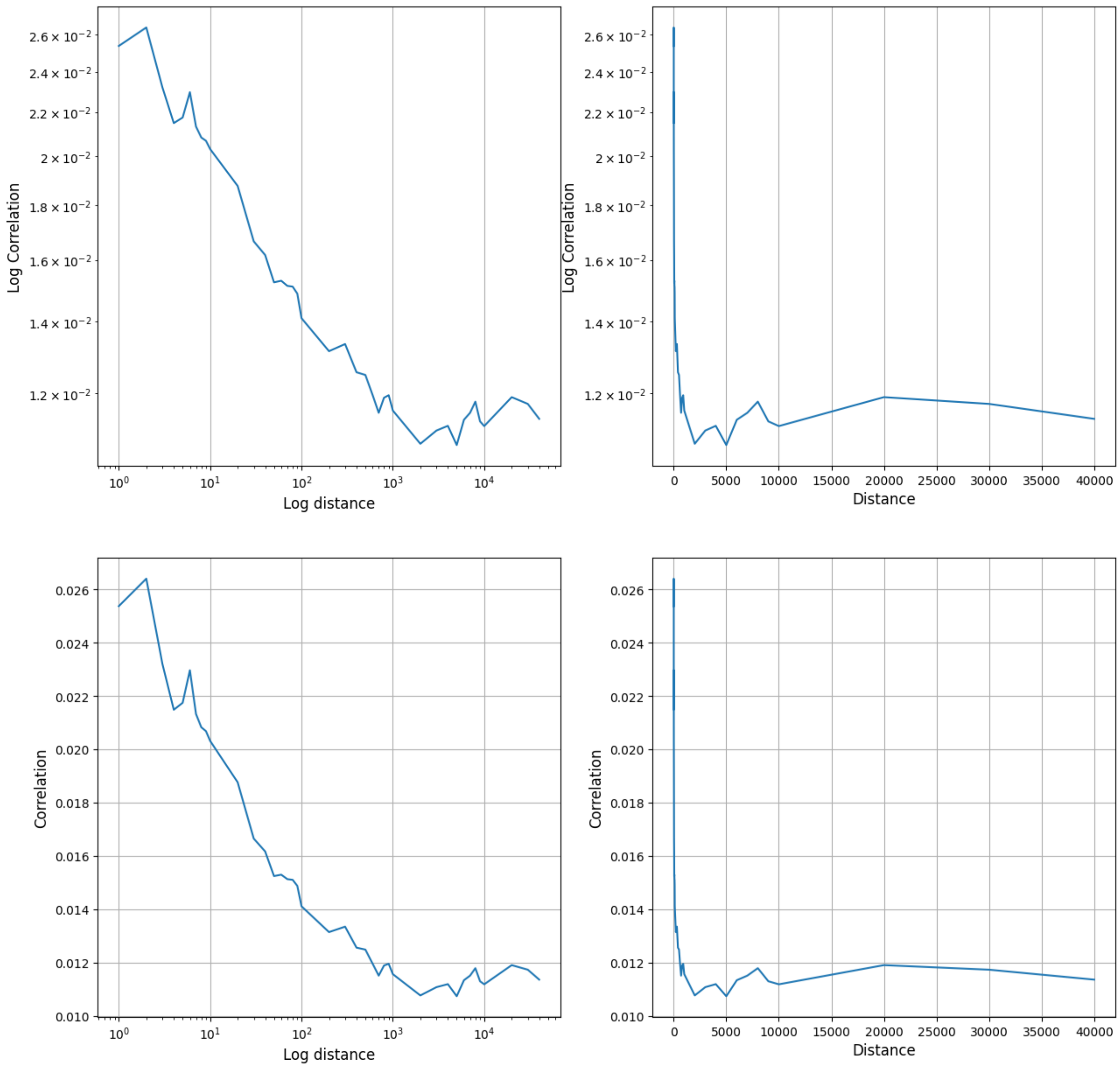

Figure 36: GloVe computation of autocorrelations in Moby-Dick or, The Whale in Russian in different coordinates

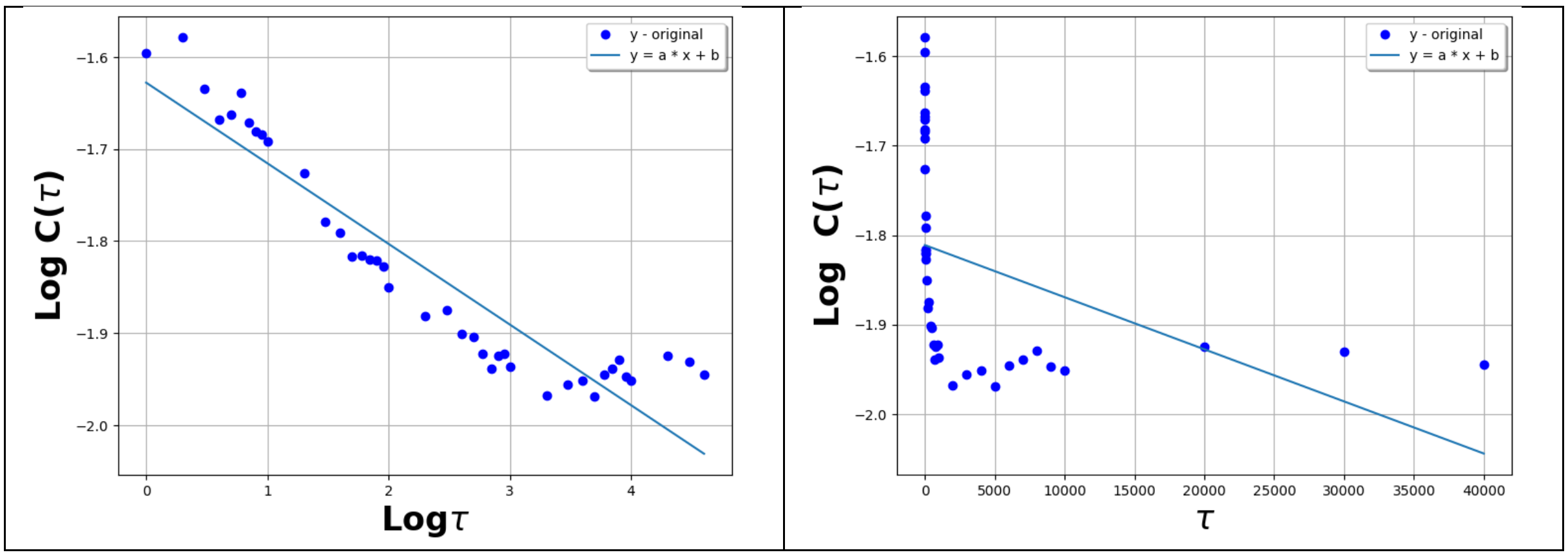

Figure 37: Best fit of GloVe computation of autocorrelations in Moby-Dick or, The Whale in Russian by power (left) and exp (right)

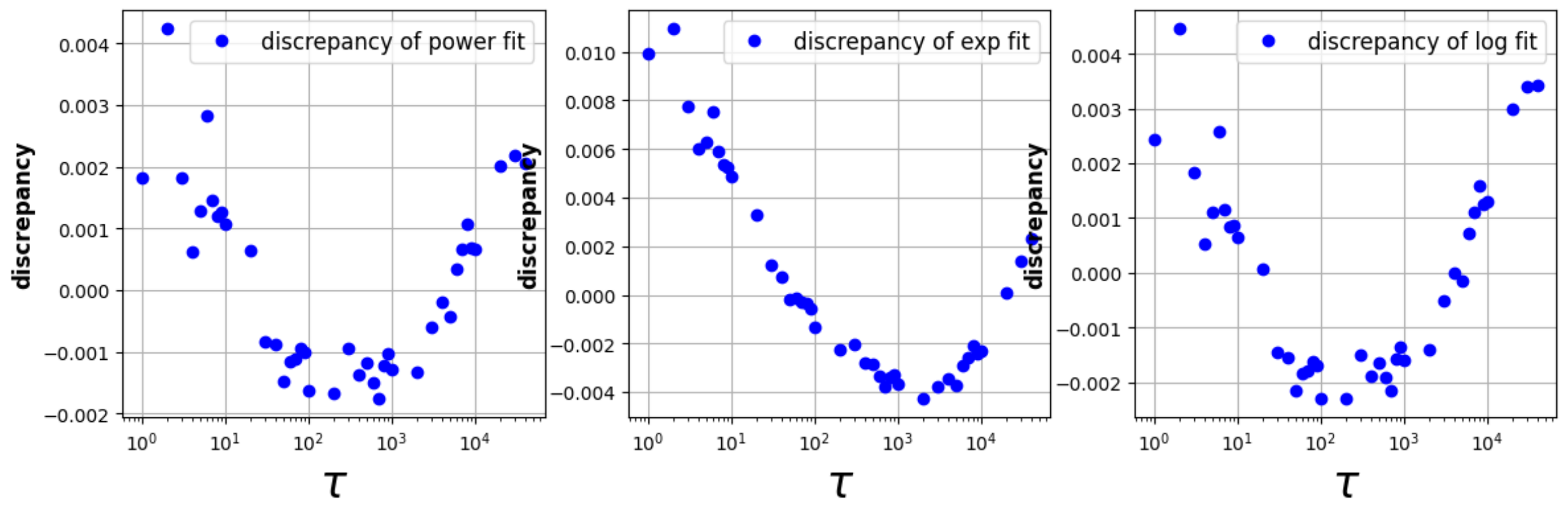

Figure 38: Residual graphs of the best fit of GloVe computation of autocorrelations in Moby-Dick or, The Whale in Russian

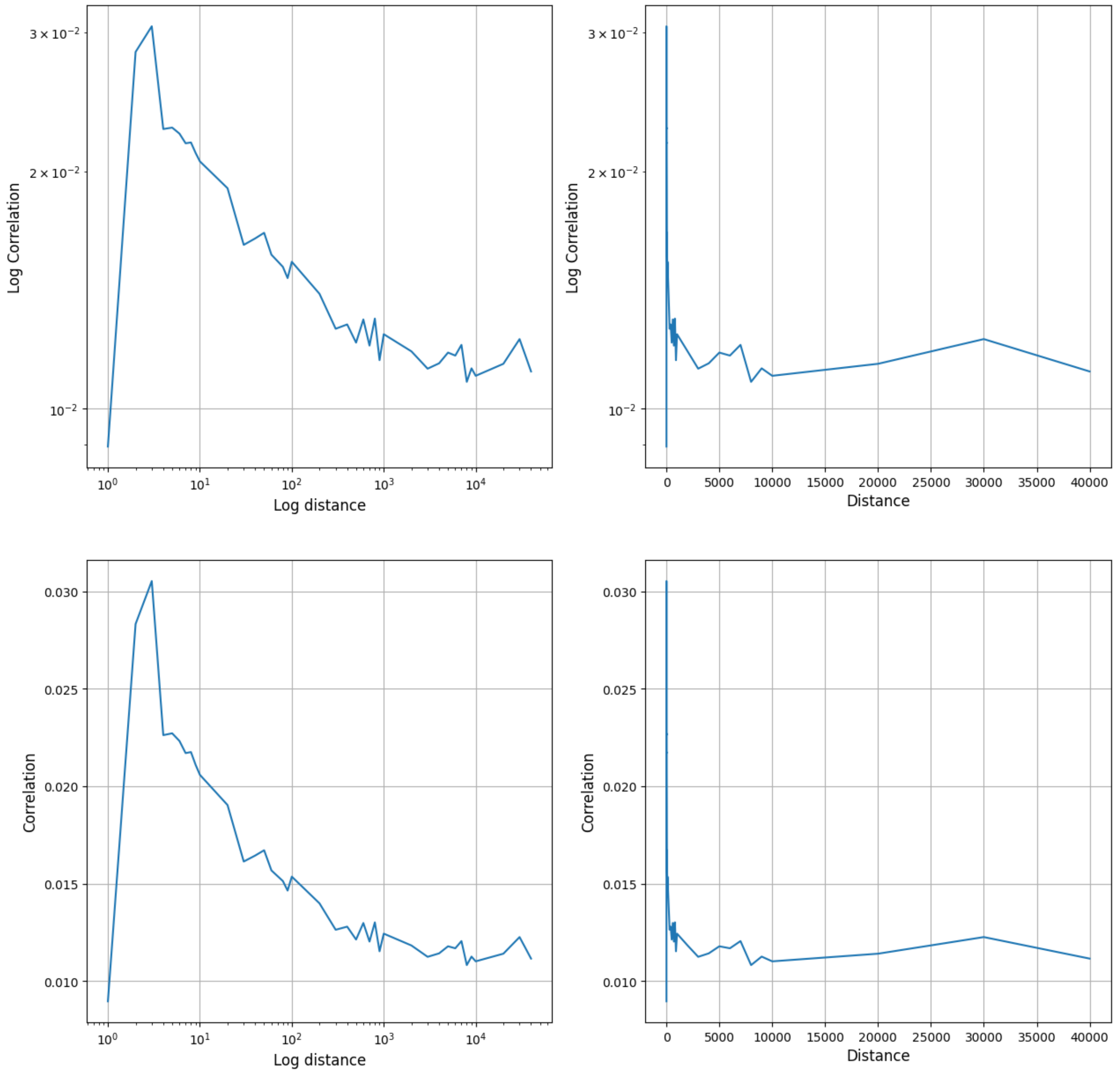

Figure 39: GloVe computation of autocorrelations in Moby-Dick or, The Whale in French in different coordinates

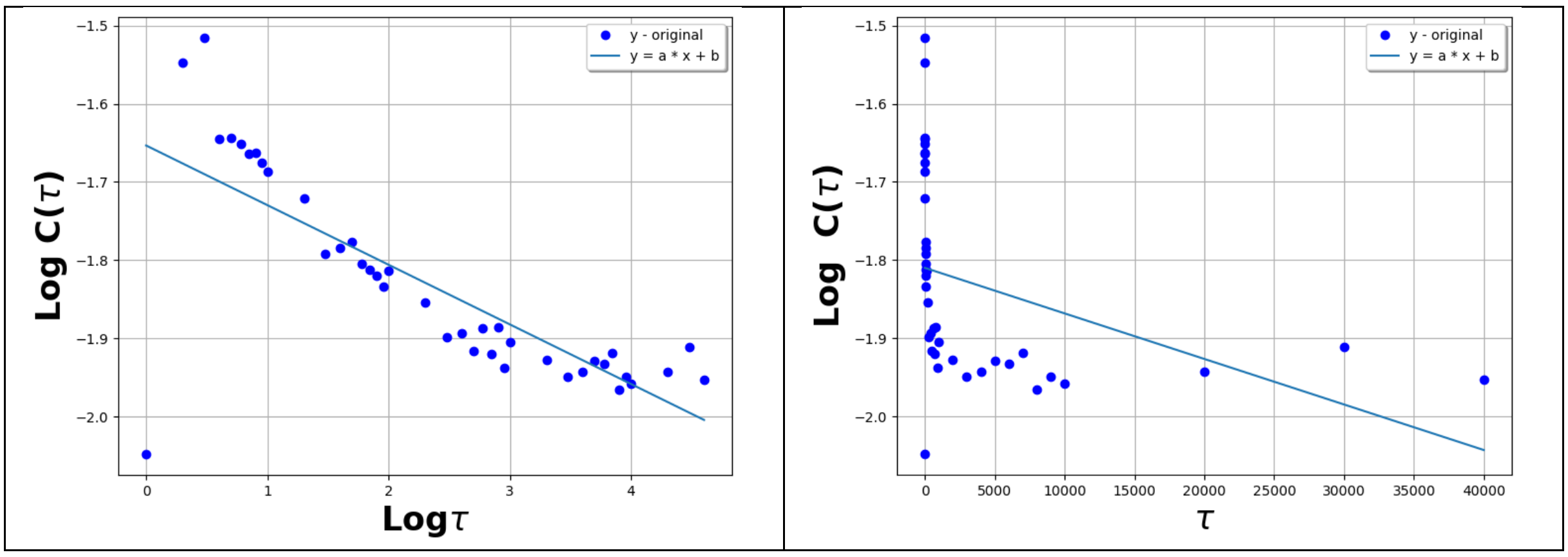

Figure 40: Best fit of GloVe computation of autocorrelations in Moby-Dick or, The Whale in French by power (left) and exp (right)

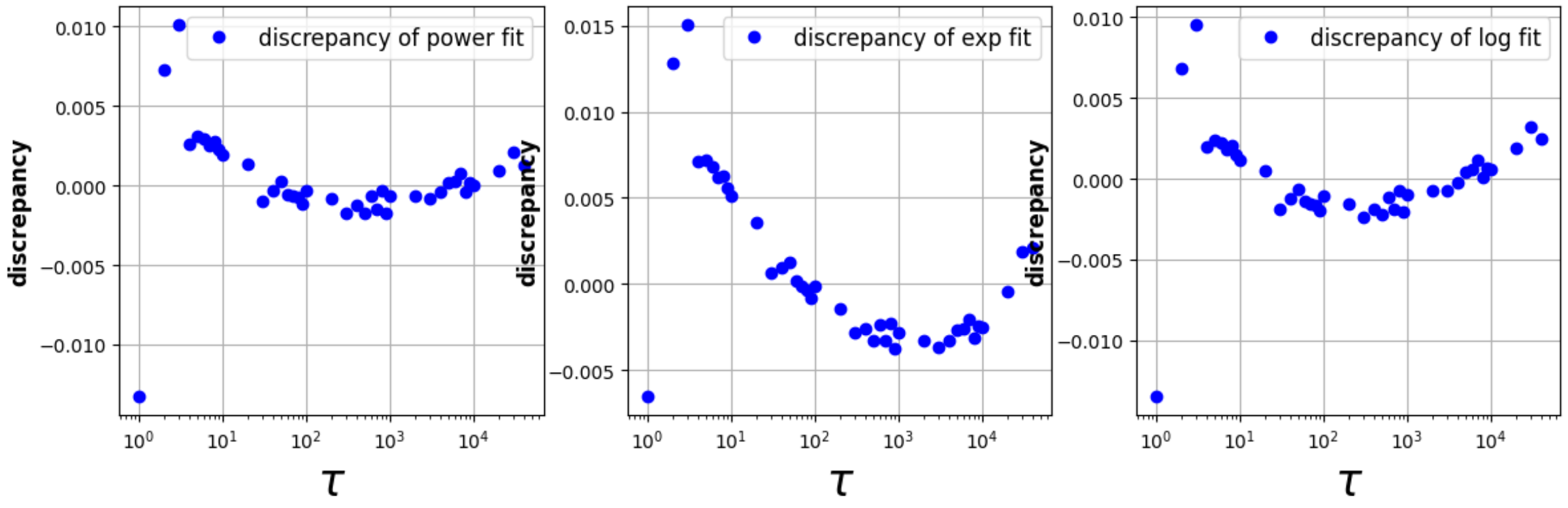

Figure 41: Residual graphs of the best fit of GloVe computation of autocorrelations in Moby-Dick or, The Whale in French

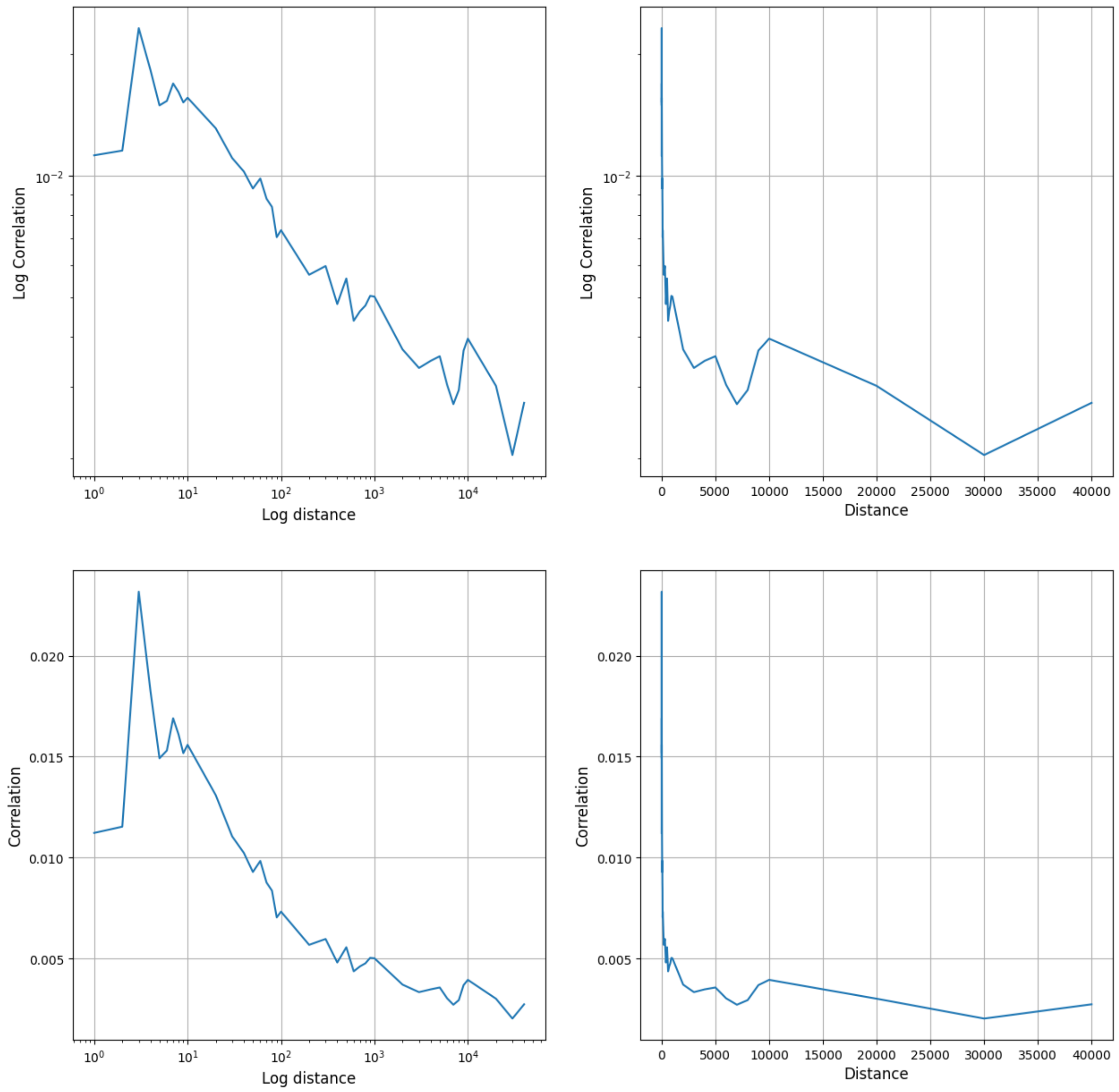

Figure 42: GloVe computation of autocorrelations in Moby-Dick or, The Whale in English in different coordinates

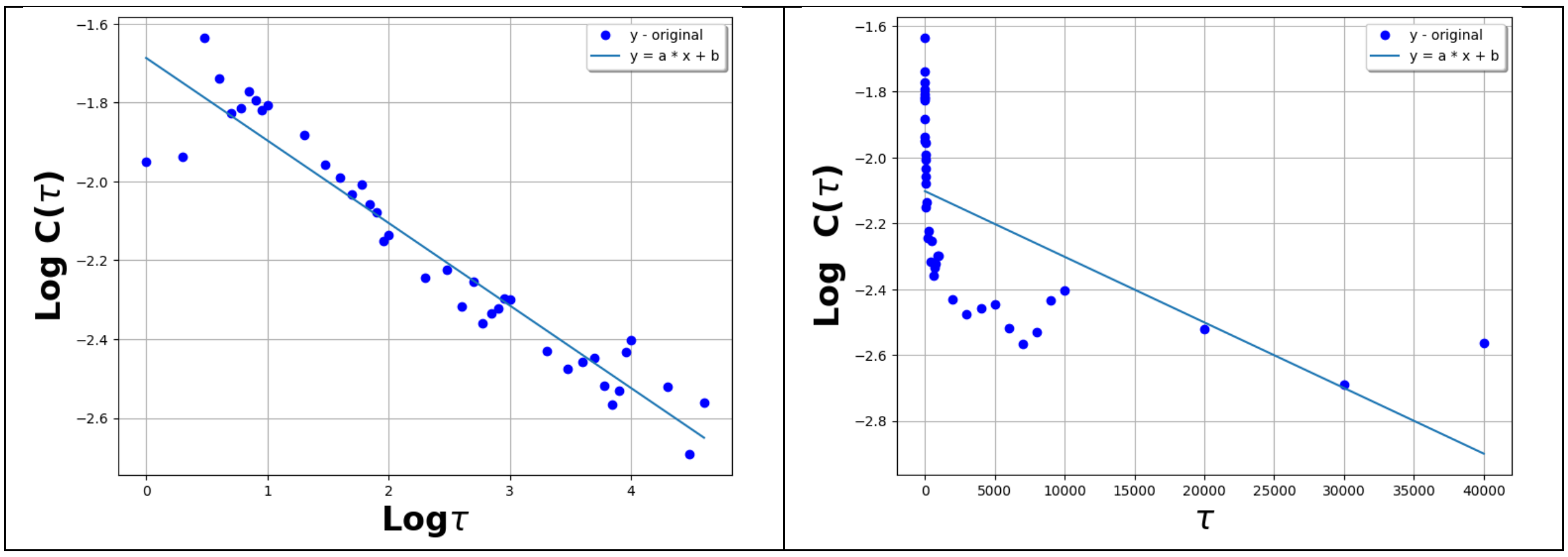

Figure 43: Best fit of GloVe computation of autocorrelations in Moby-Dick or, The Whale in English by power (left) and exp (right)

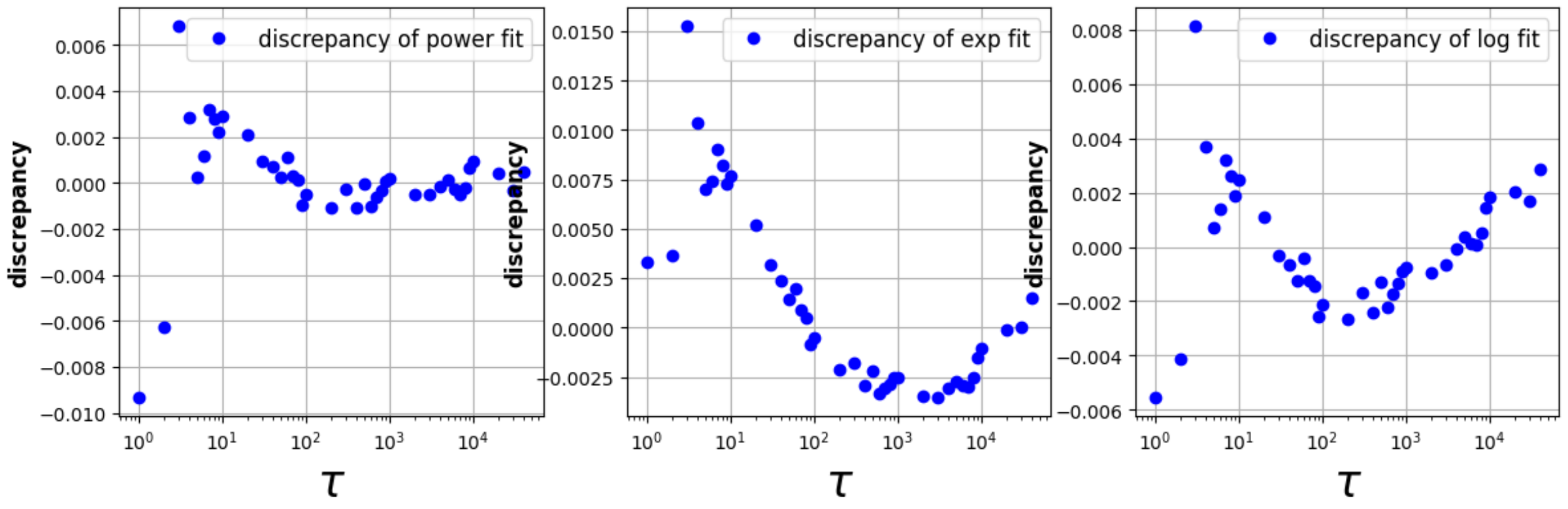

Figure 44: Residual graphs of the best fit of GloVe computation of autocorrelations in Moby-Dick or, The Whale in English

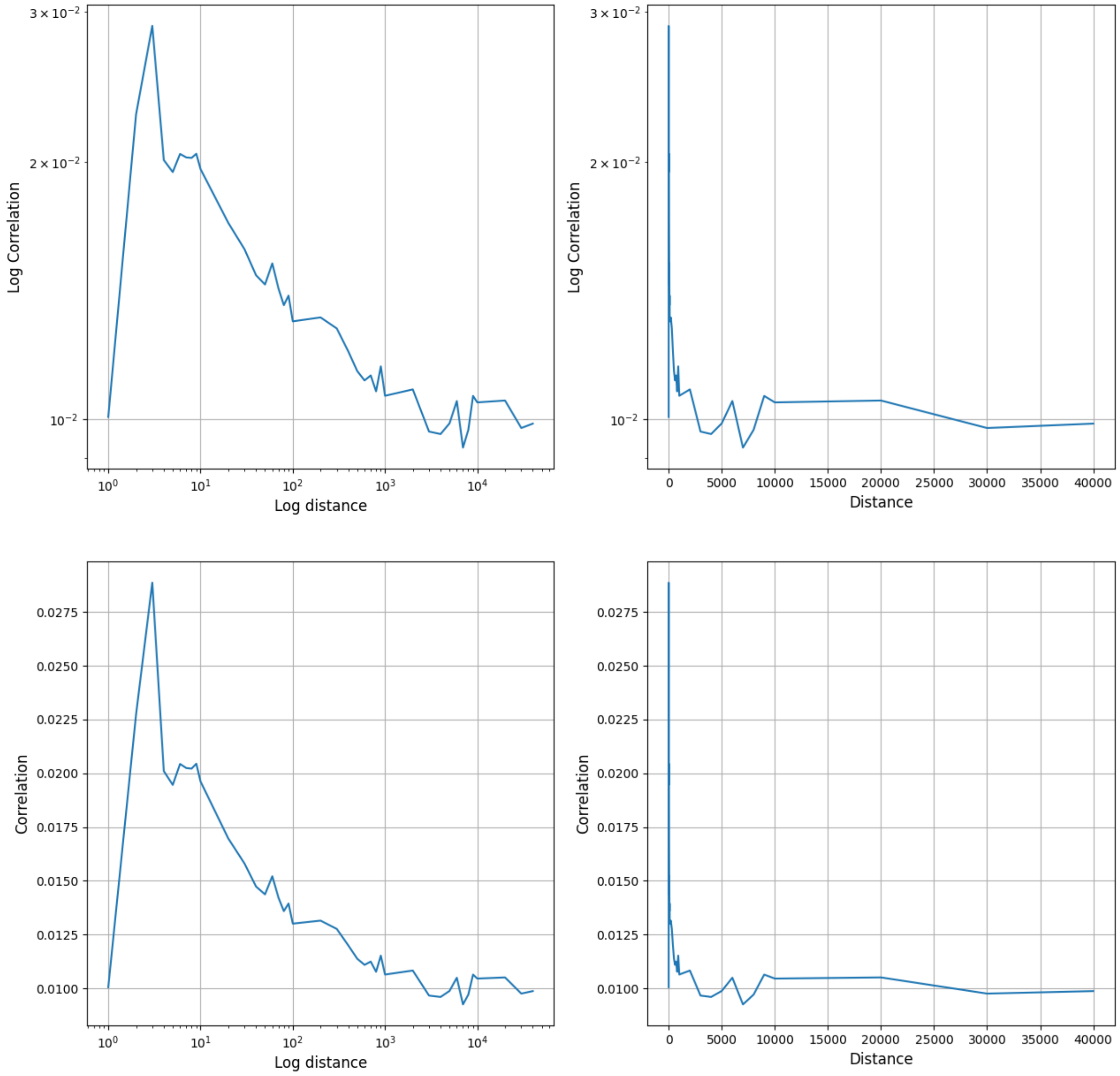

Figure 45: GloVe computation of autocorrelations in Moby-Dick or, The Whale in Spanish in different coordinates

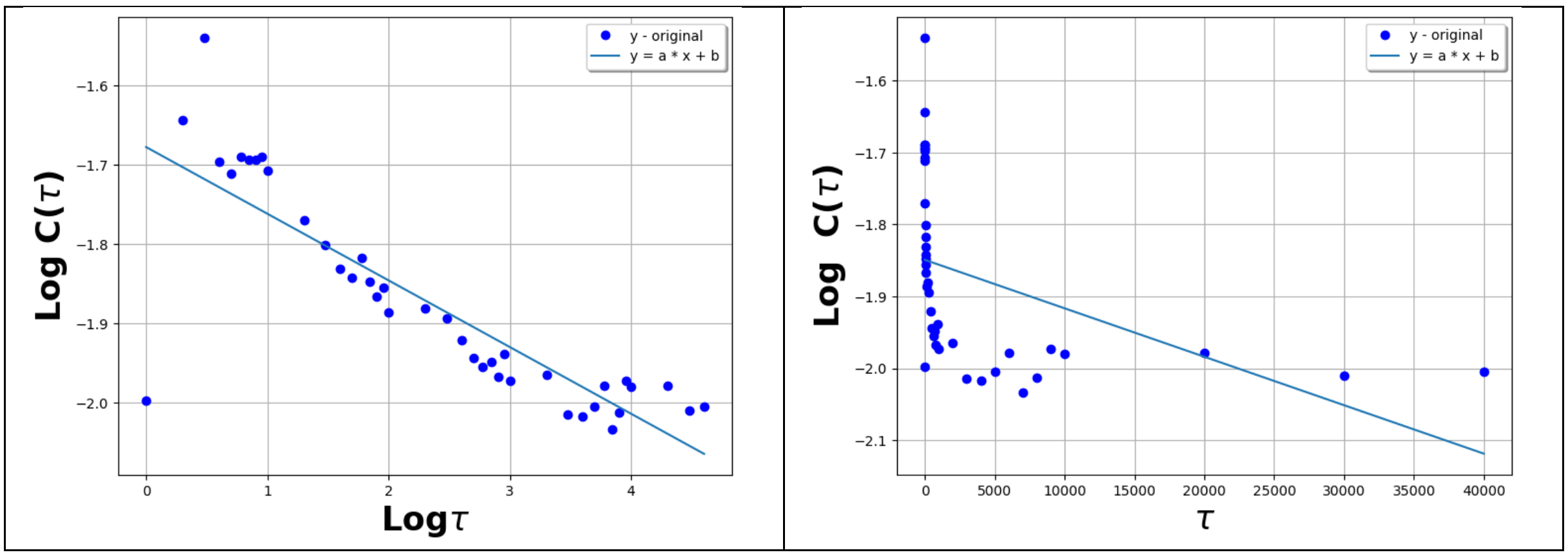

Figure 46: Best fit of GloVe computation of autocorrelations in Moby-Dick or, The Whale in Spanish by power (left) and exp (right)

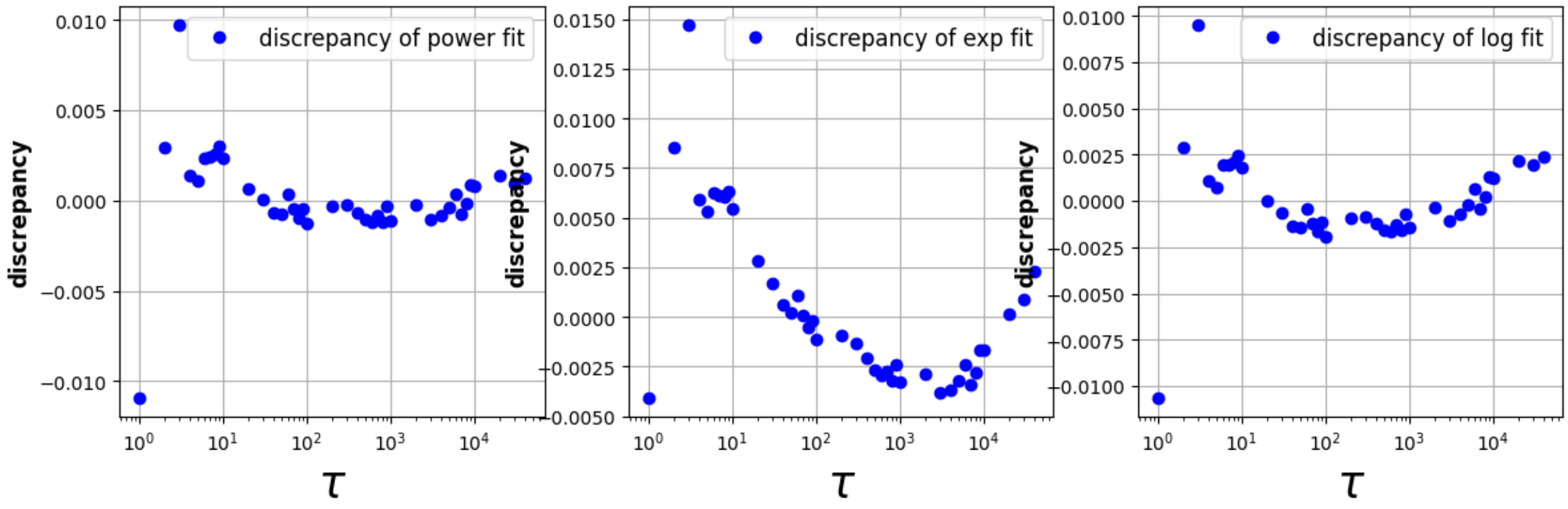

Figure 47: Residual graphs of the best fit of GloVe computation of autocorrelations in Moby-Dick or, The Whale in Spanish

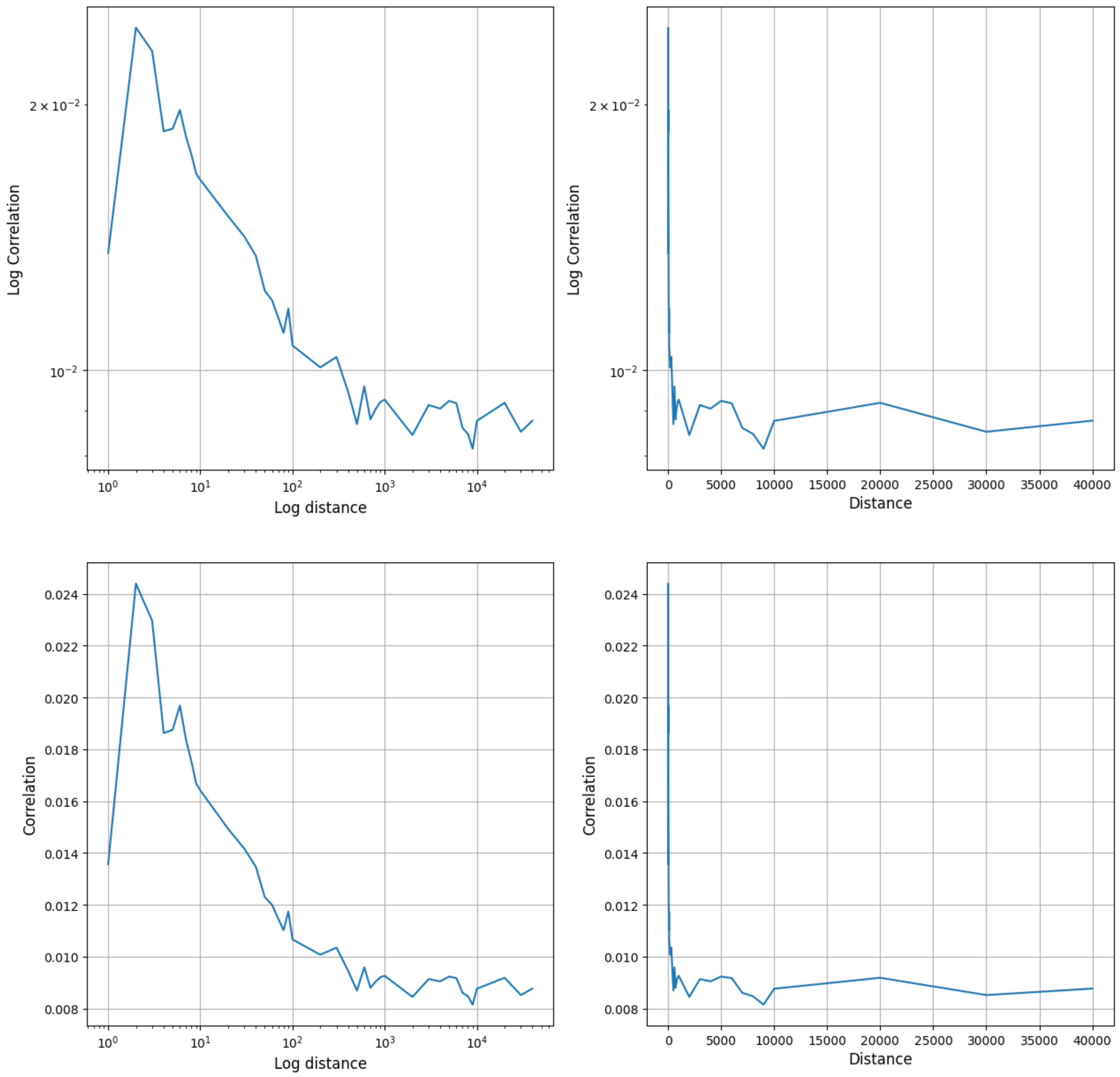

Figure 48: GloVe computation of autocorrelations in Don Quixote de la Mancha in French in different coordinates

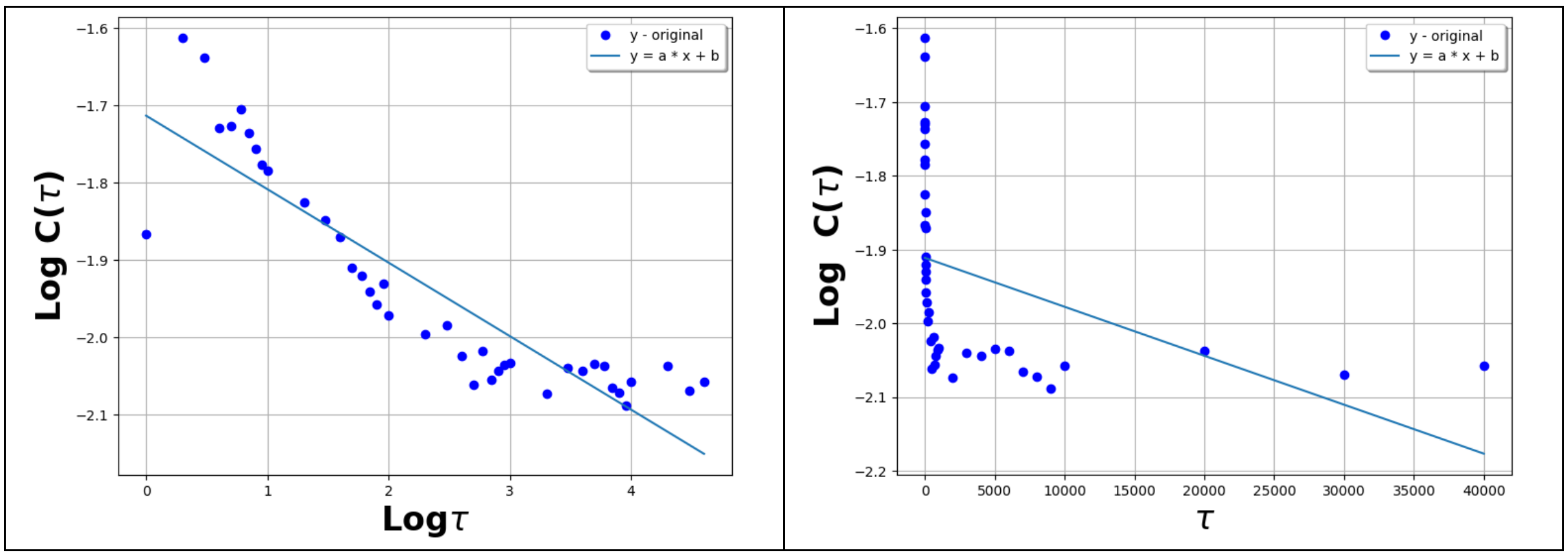

Figure 49: Best fit of GloVe computation of autocorrelations in Don Quixote de la Mancha in French by power (left) and exp (right)

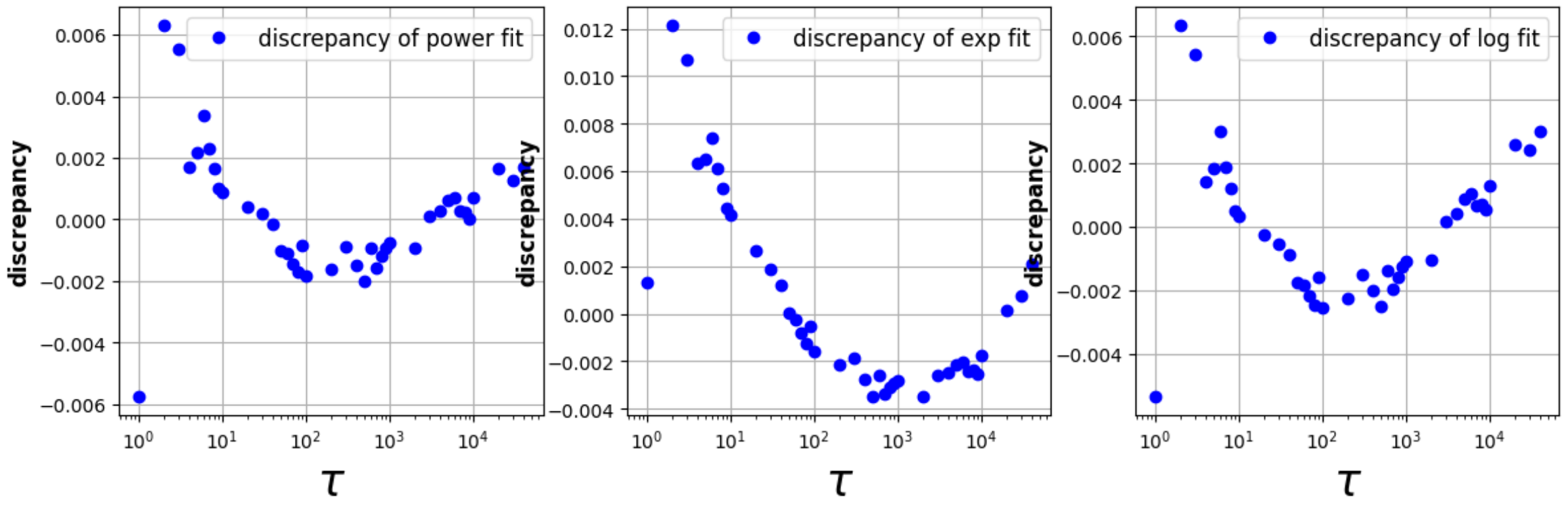

Figure 50: Residual graphs of the best fit of GloVe computation of autocorrelations in Don Quixote de la Mancha in French

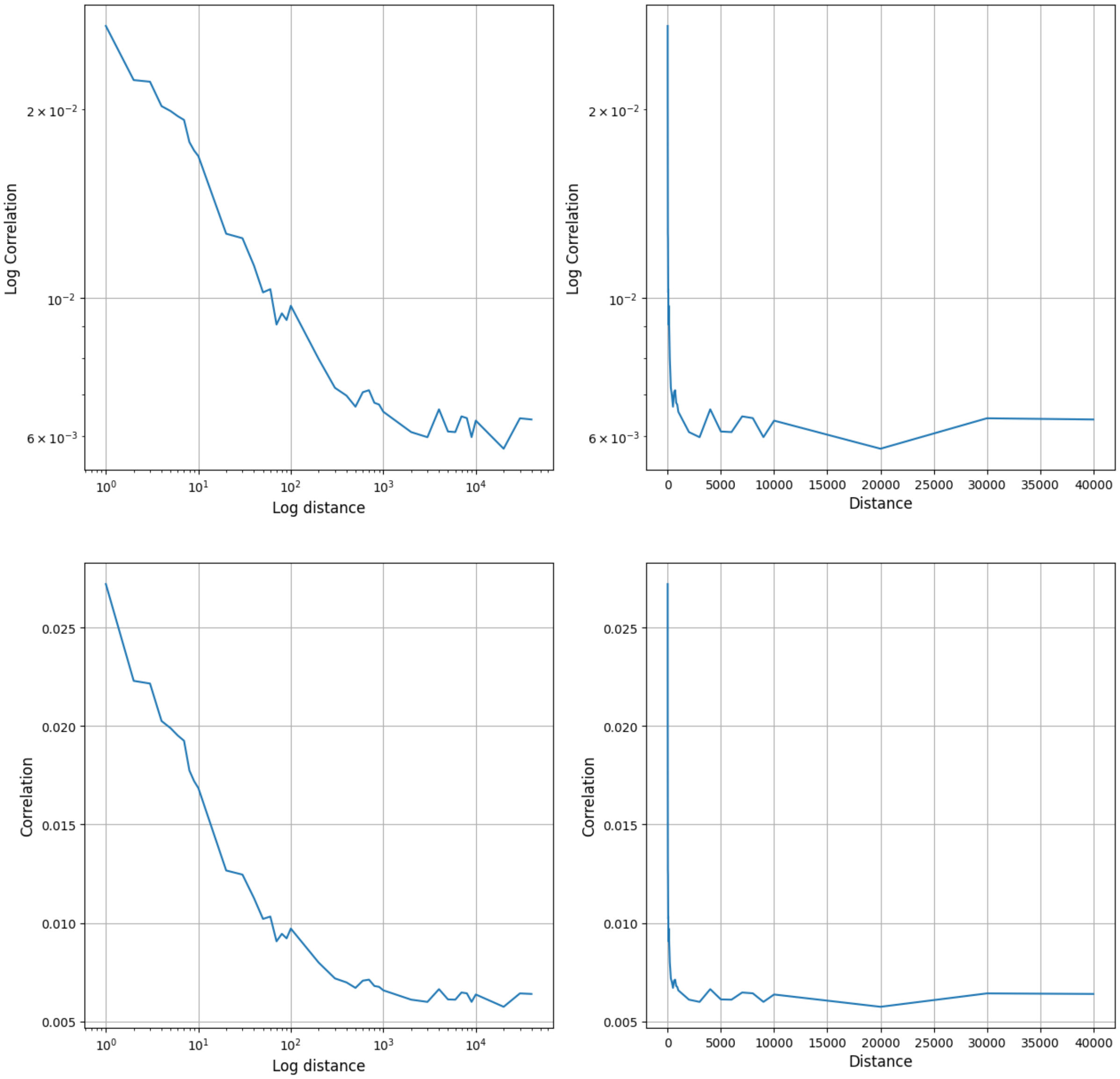

Figure 51: GloVe computation of autocorrelations in Don Quixote de la Mancha in German in different coordinates

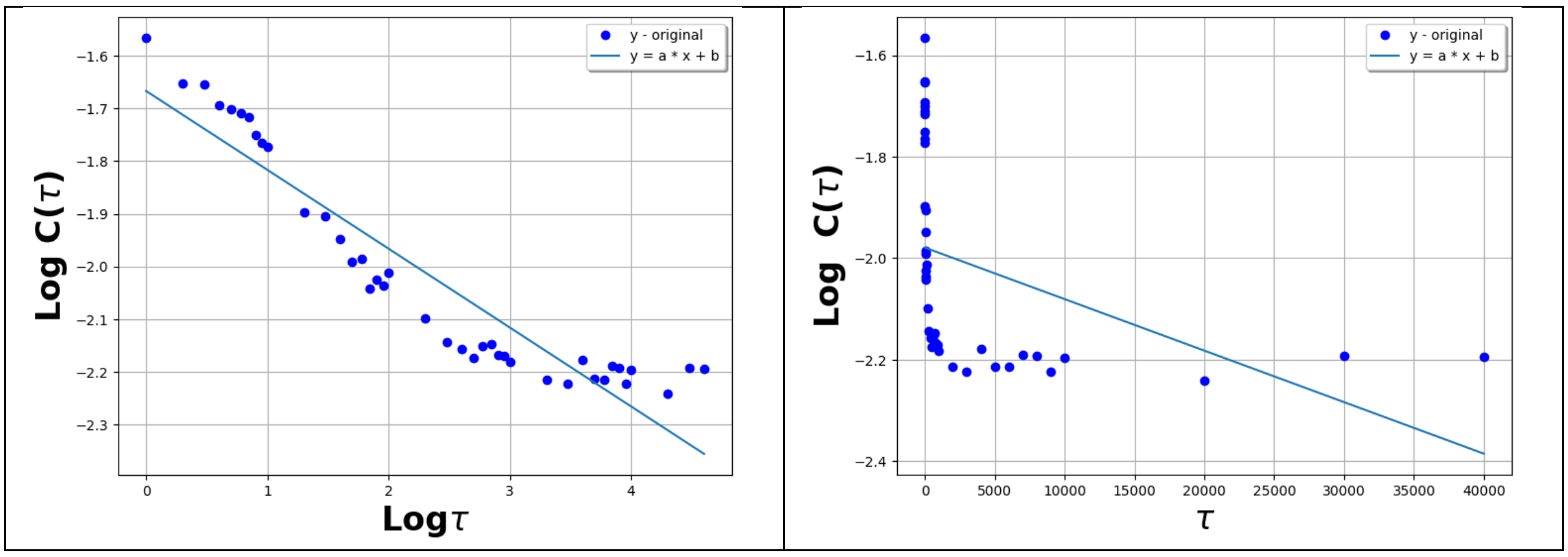

Figure 52: Best fit of GloVe computation of autocorrelations in Don Quixote de la Mancha in German by power (left) and exp (right)

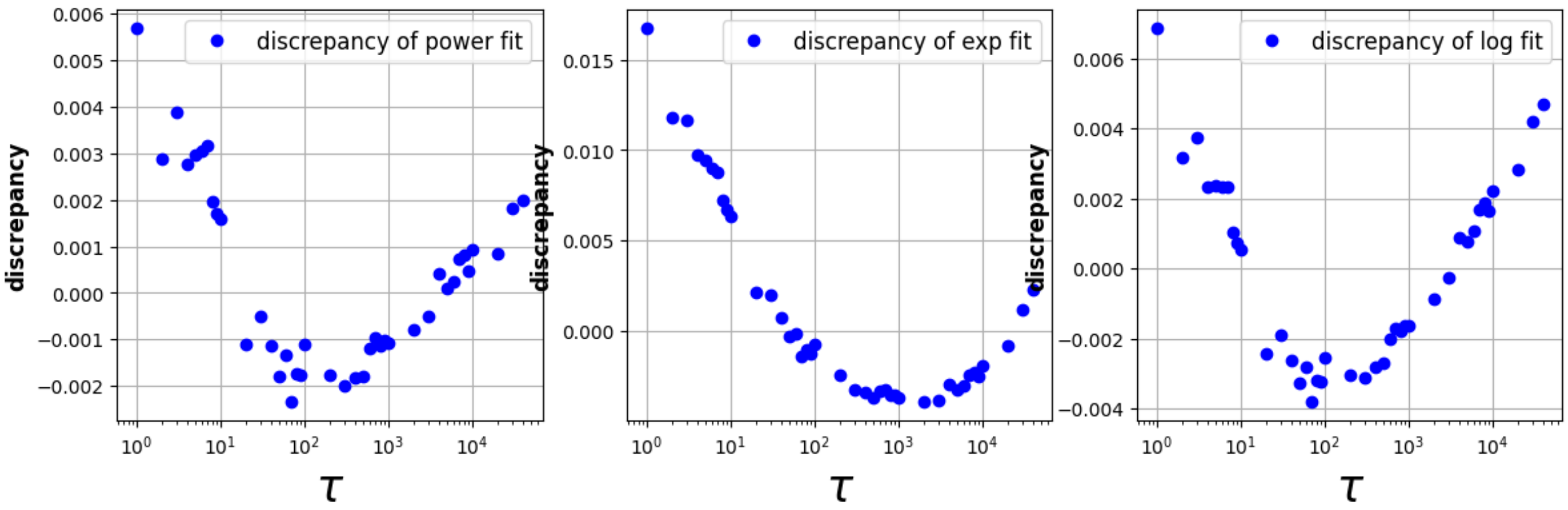

Figure 53: Residual graphs of the best fit of GloVe computation of autocorrelations in Don Quixote de la Mancha in German

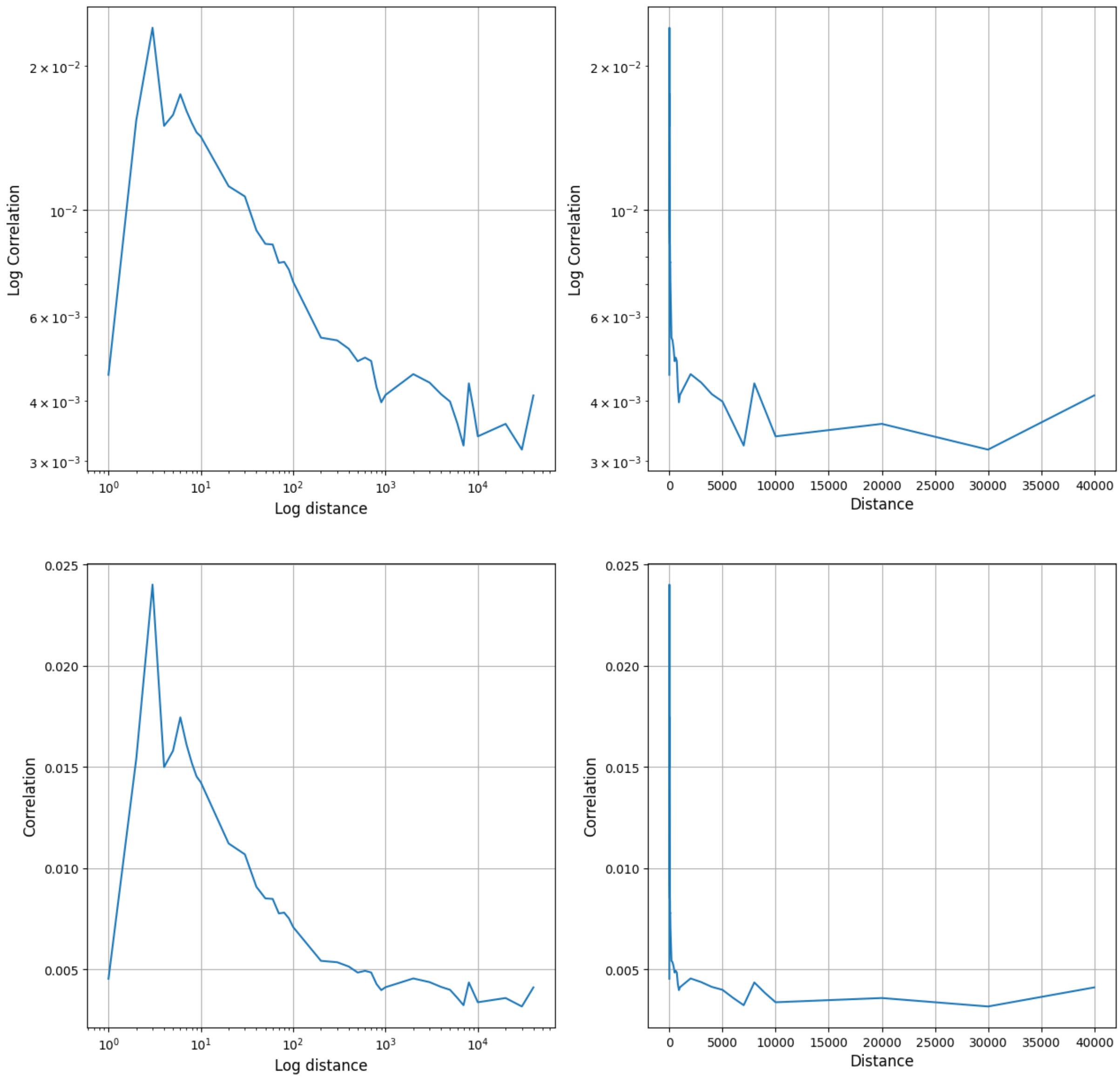

Figure 54: GloVe computation of autocorrelations in Don Quixote de la Mancha in English in different coordinates

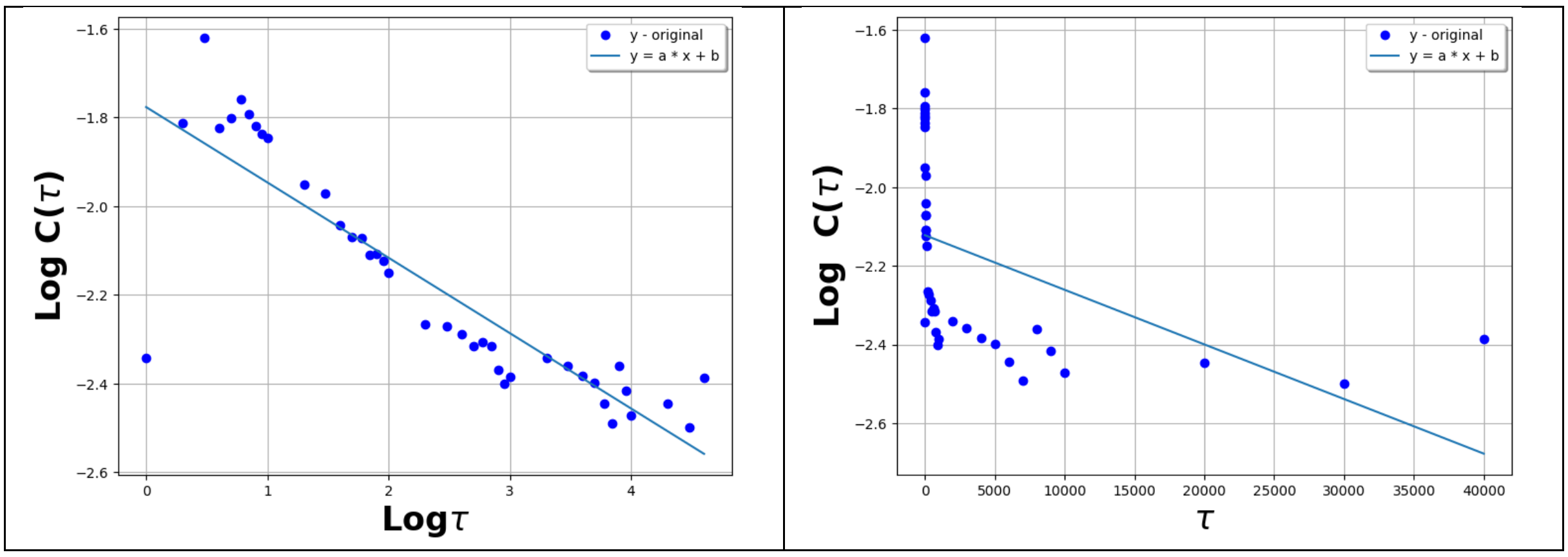

Figure 55: Best fit of GloVe computation of autocorrelations in Don Quixote de la Mancha in English by power (left) and exp (right)

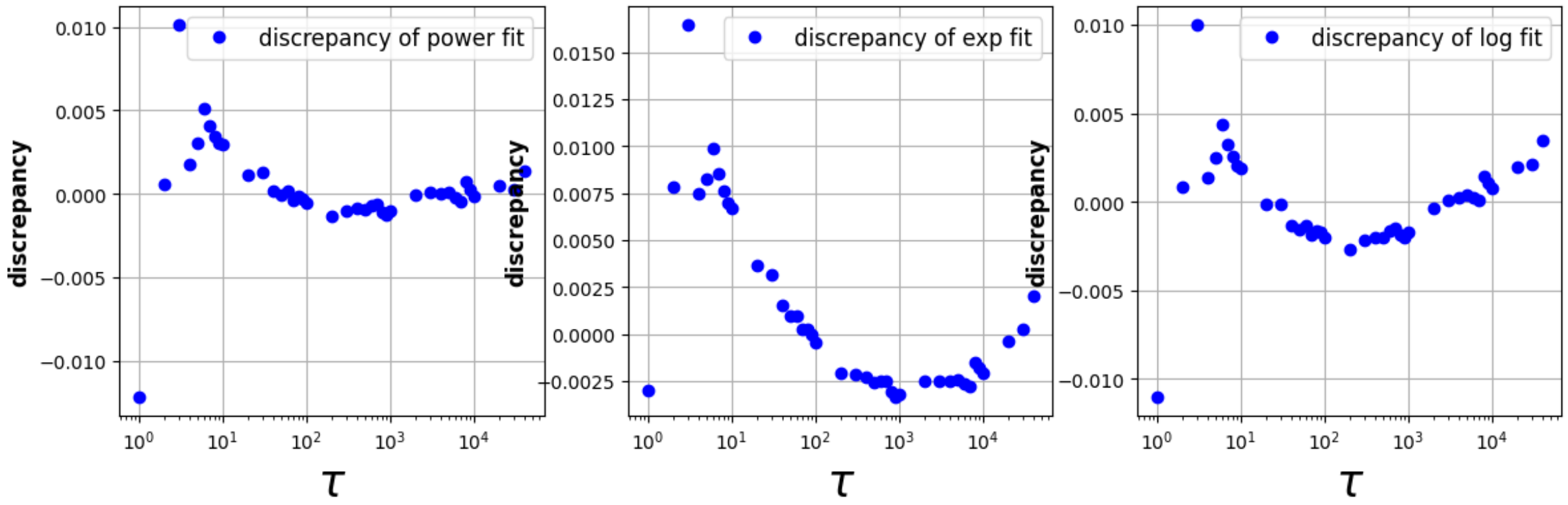

Figure 56: Residual graphs of the best fit of GloVe computation of autocorrelations in Don Quixote de la Mancha in English

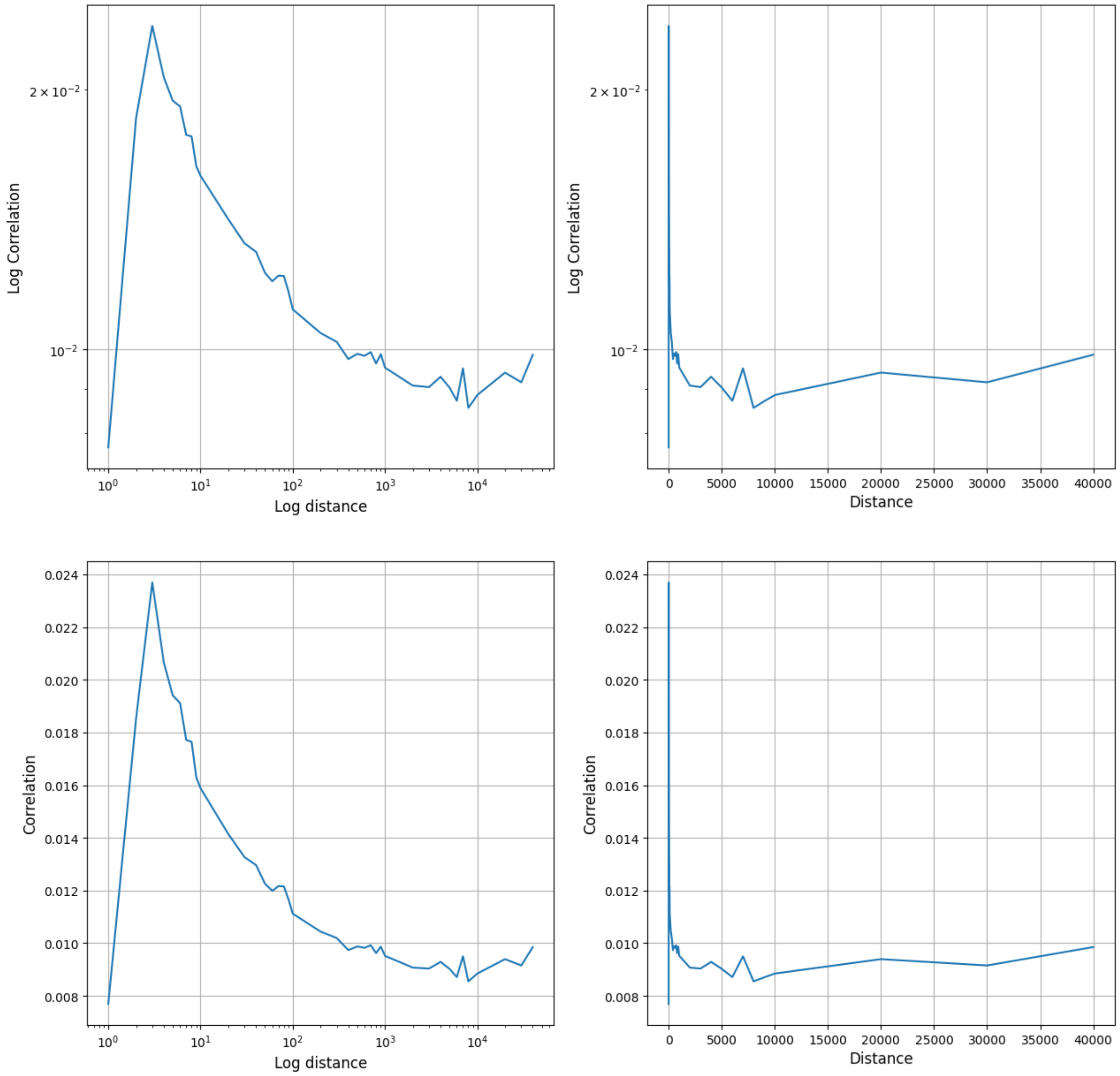

Figure 57: GloVe computation of autocorrelations in Don Quixote de la Mancha in Spanish in different coordinates

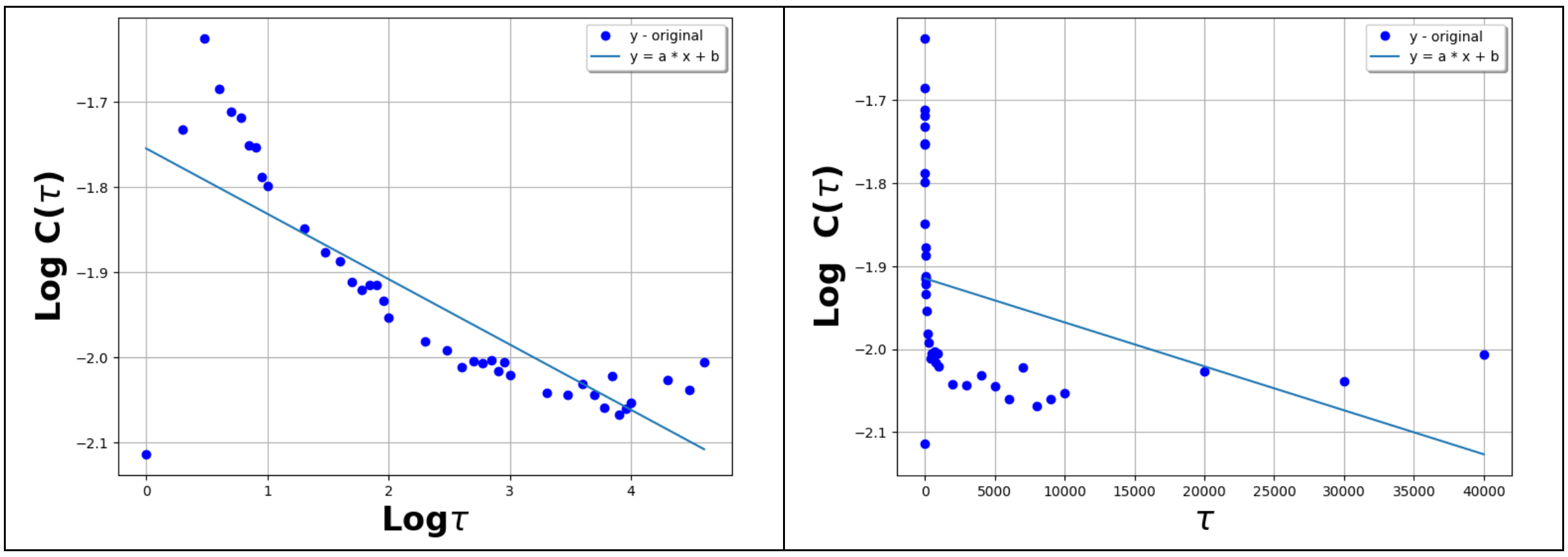

Figure 58: Best fit of GloVe computation of autocorrelations in Don Quixote de la Mancha in Spanish by power (left) and exp (right)

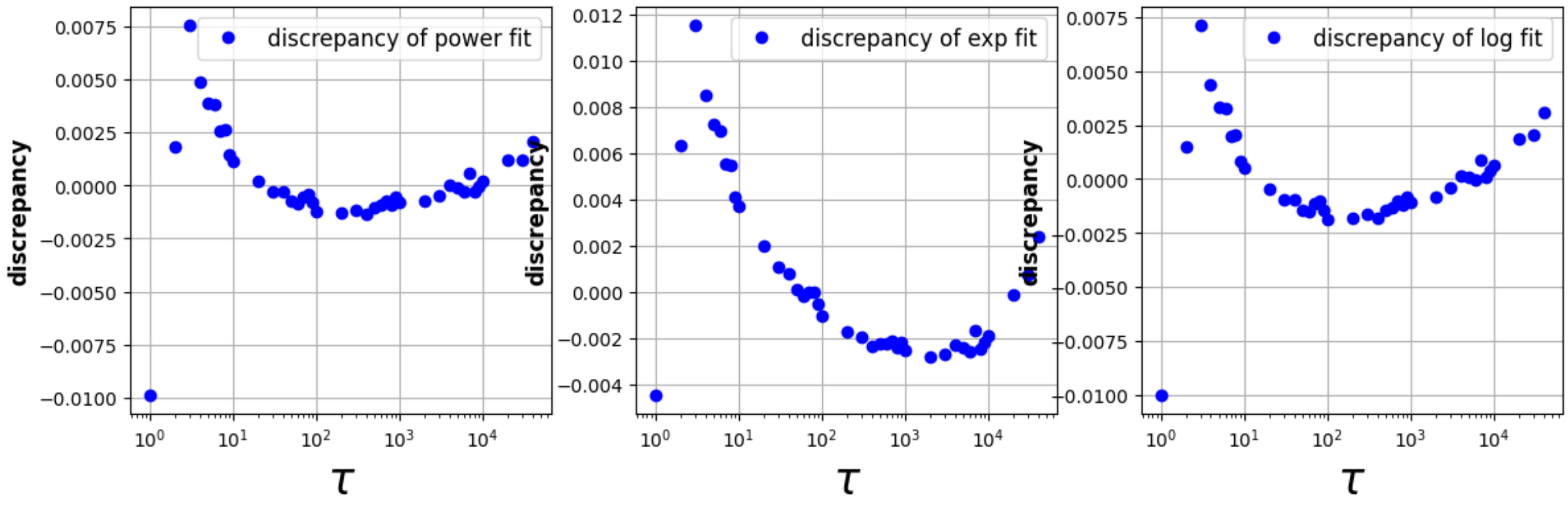

Figure 59: Residual graphs of the best fit of GloVe computation of autocorrelations in Don Quixote de la Mancha in Spanish

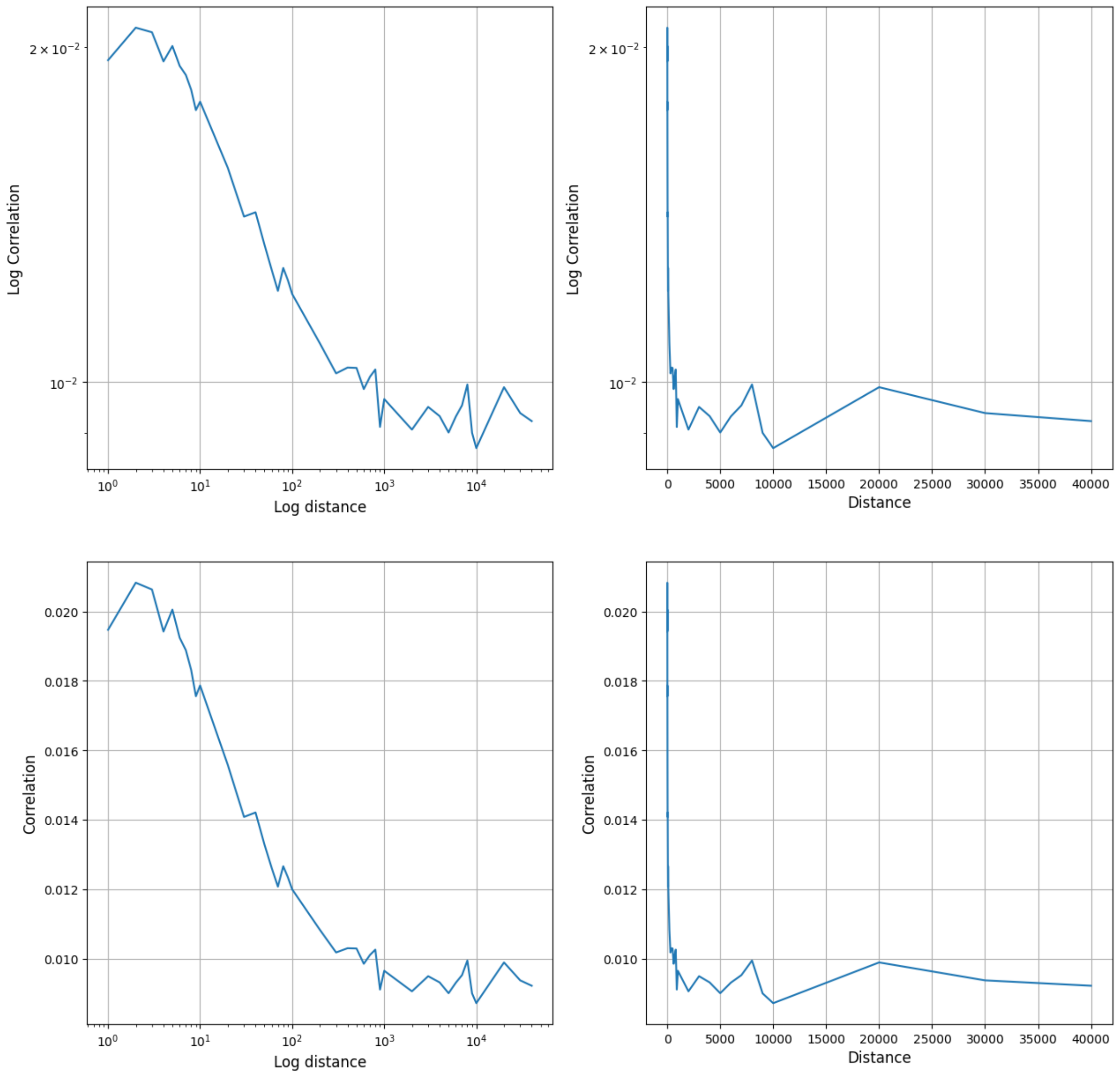

Figure 60: GloVe computation of autocorrelations in Don Quixote de la Mancha in Russian in different coordinates

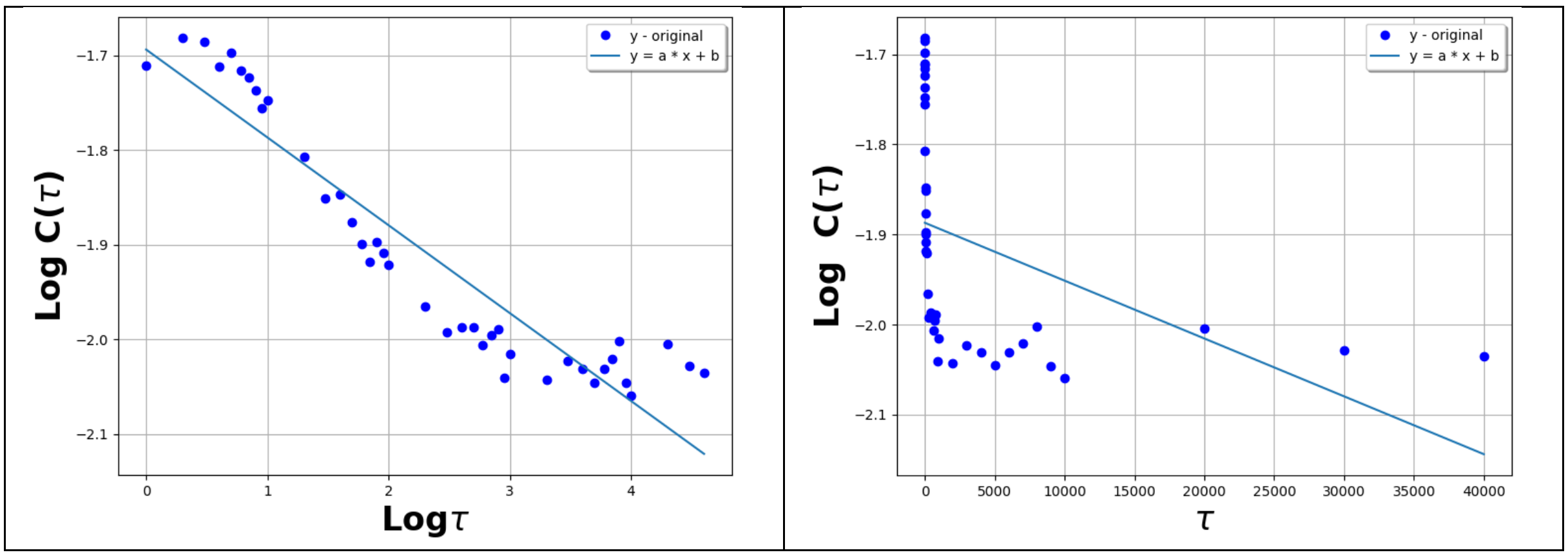

Figure 61: Best fit of GloVe computation of autocorrelations in Don Quixote de la Mancha in Russian by power (left) and exp (right)

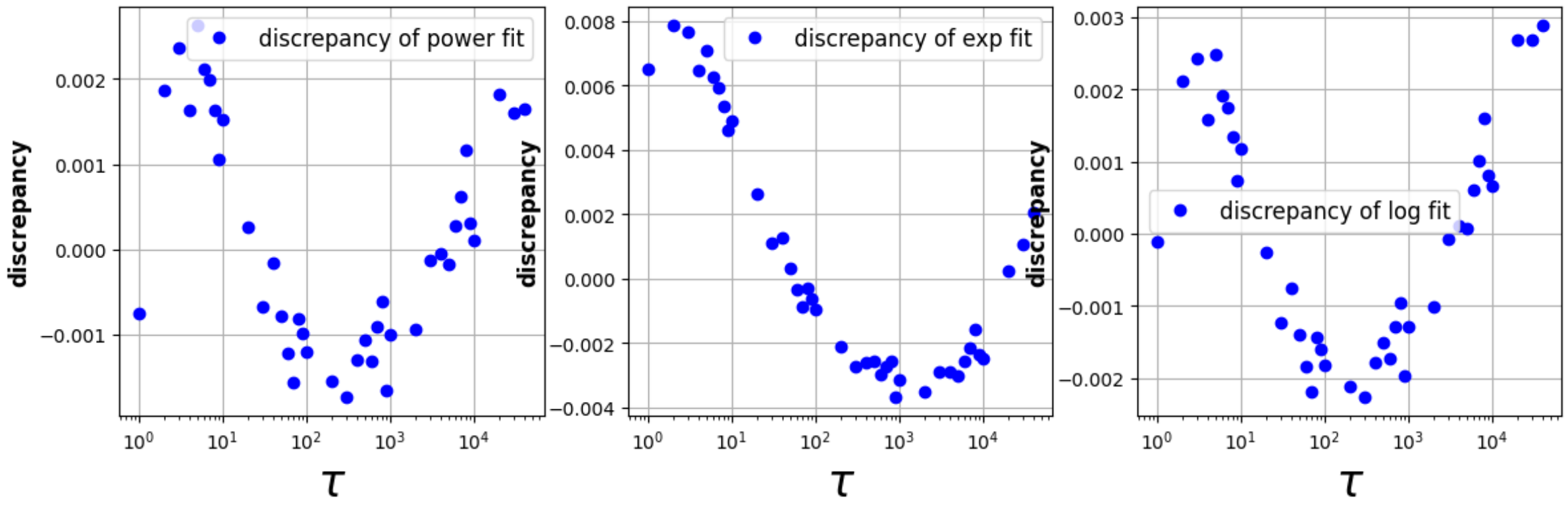

Figure 62: Residual graphs of the best fit of GloVe computation of autocorrelations in Don Quixote de la Mancha in Russian

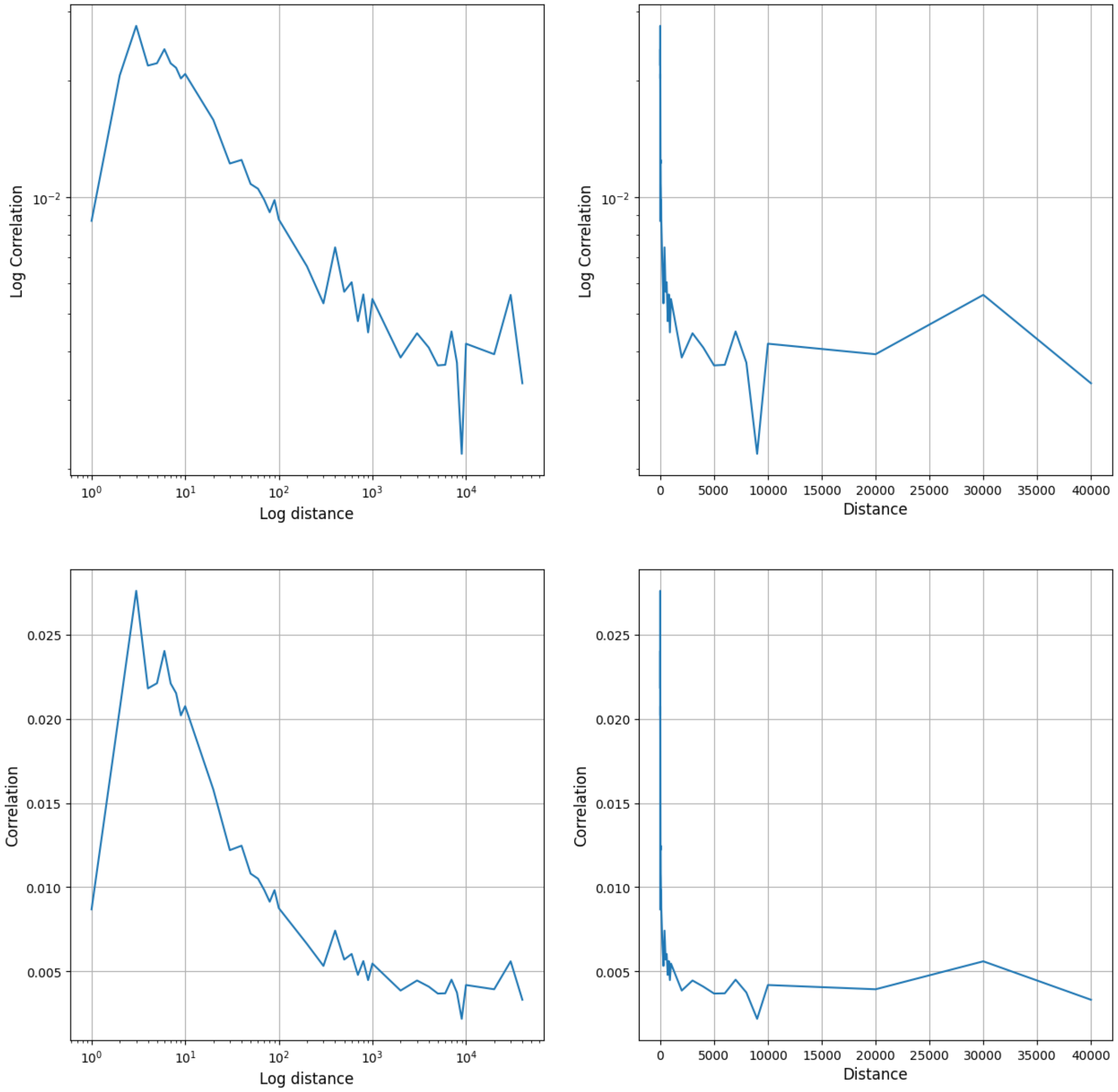

Figure 63: GloVe computation of autocorrelations in The Adventures of Tom Sawyer in English in different coordinates

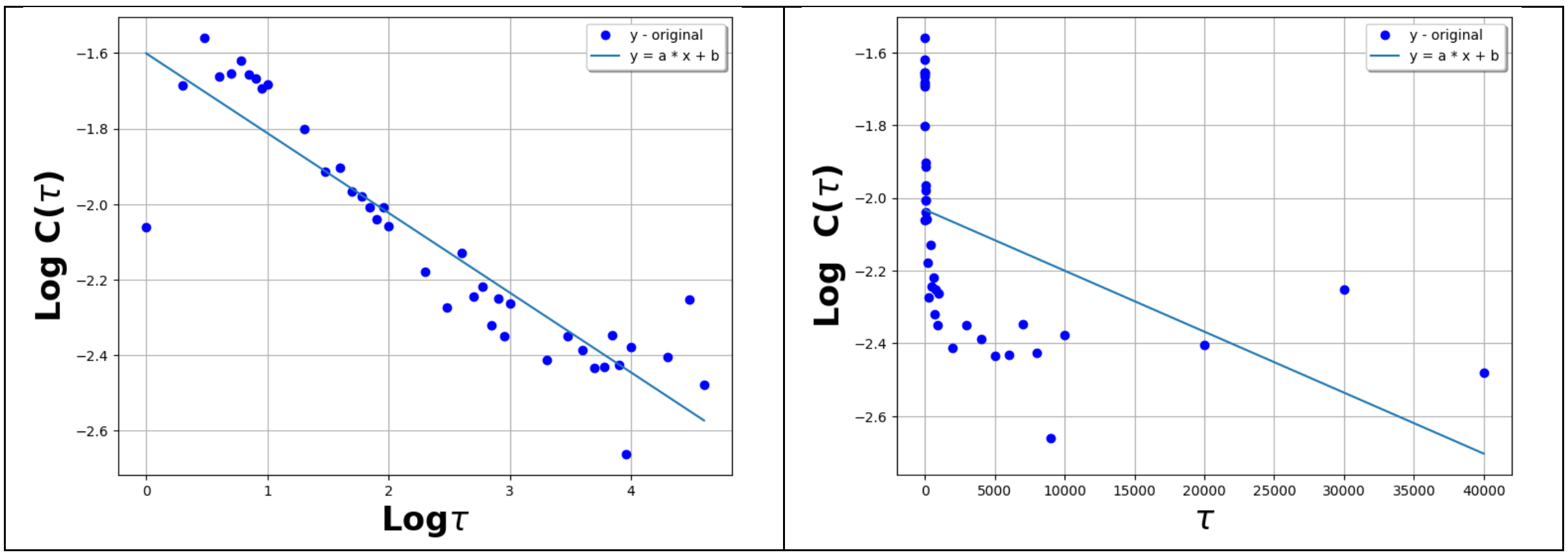

Figure 64: Best fit of GloVe computation of autocorrelations in The Adventures of Tom Sawyer in English by power (left) and exp (right)

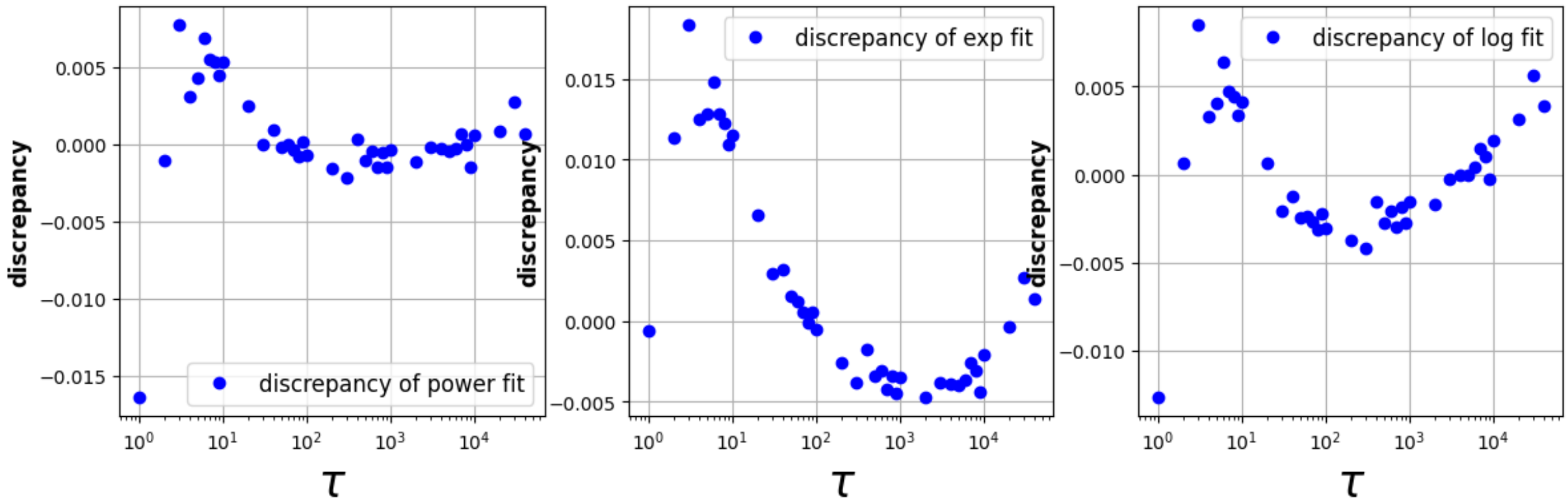

Figure 65: Residual graphs of the best fit of GloVe computation of autocorrelations in The Adventures of Tom Sawyer in English

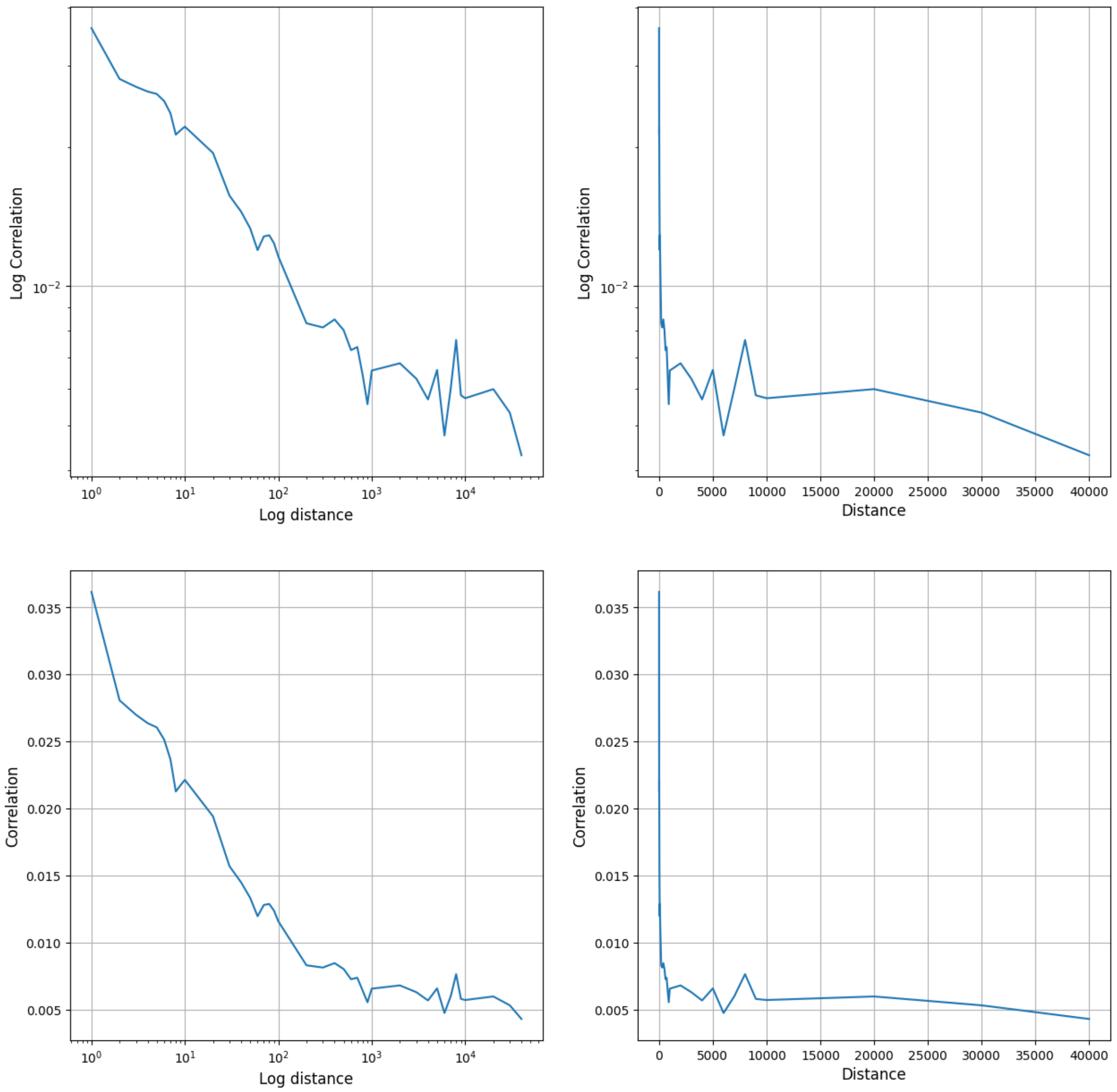

Figure 66: GloVe computation of autocorrelations in The Adventures of Tom Sawyer in German in different coordinates

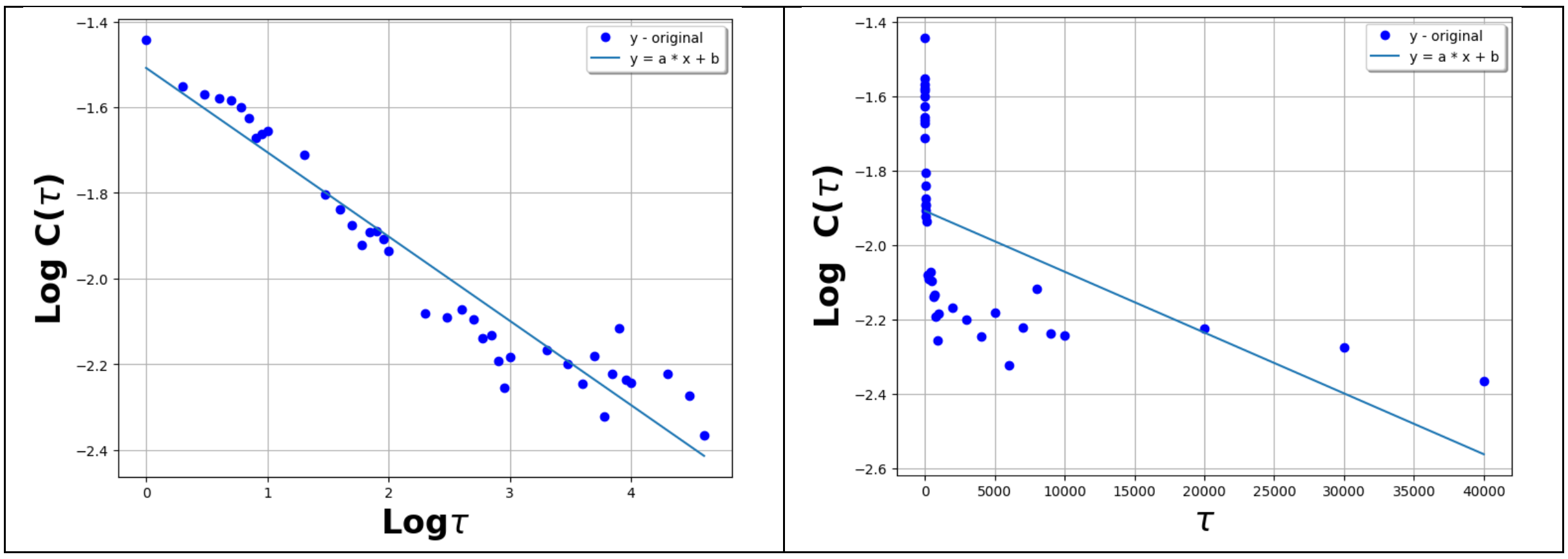

Figure 67: Best fit of GloVe computation of autocorrelations in The Adventures of Tom Sawyer in German by power (left) and exp (right)

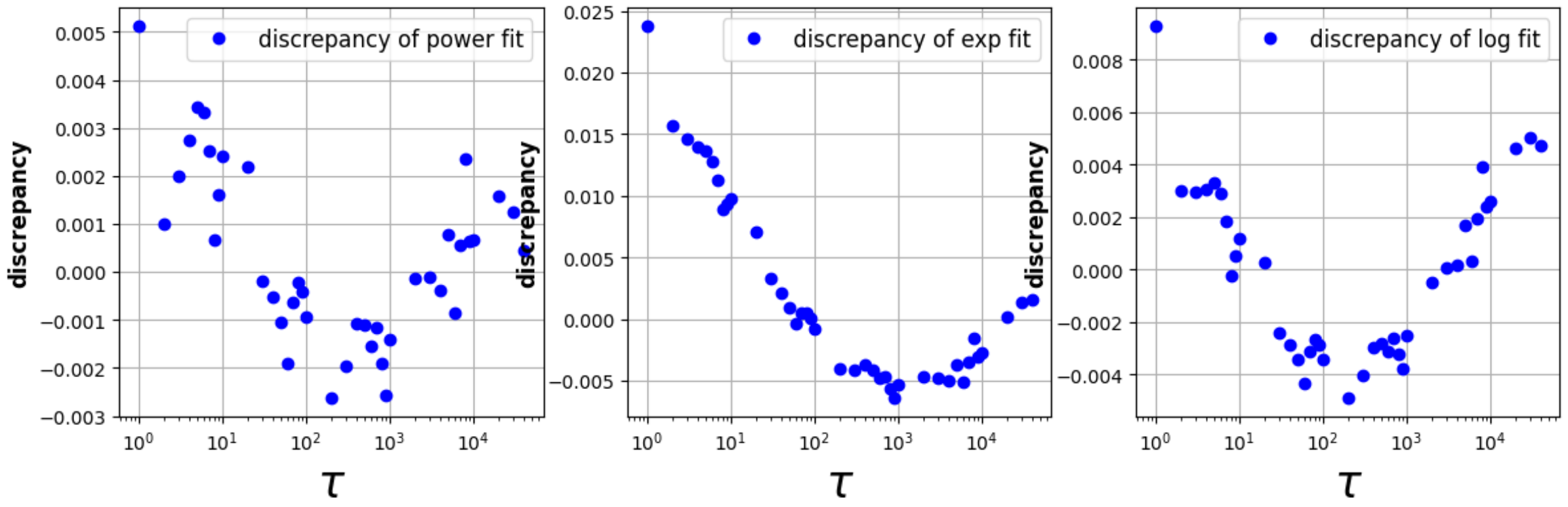

Figure 68: Residual graphs of the best fit of GloVe computation of autocorrelations in The Adventures of Tom Sawyer in German

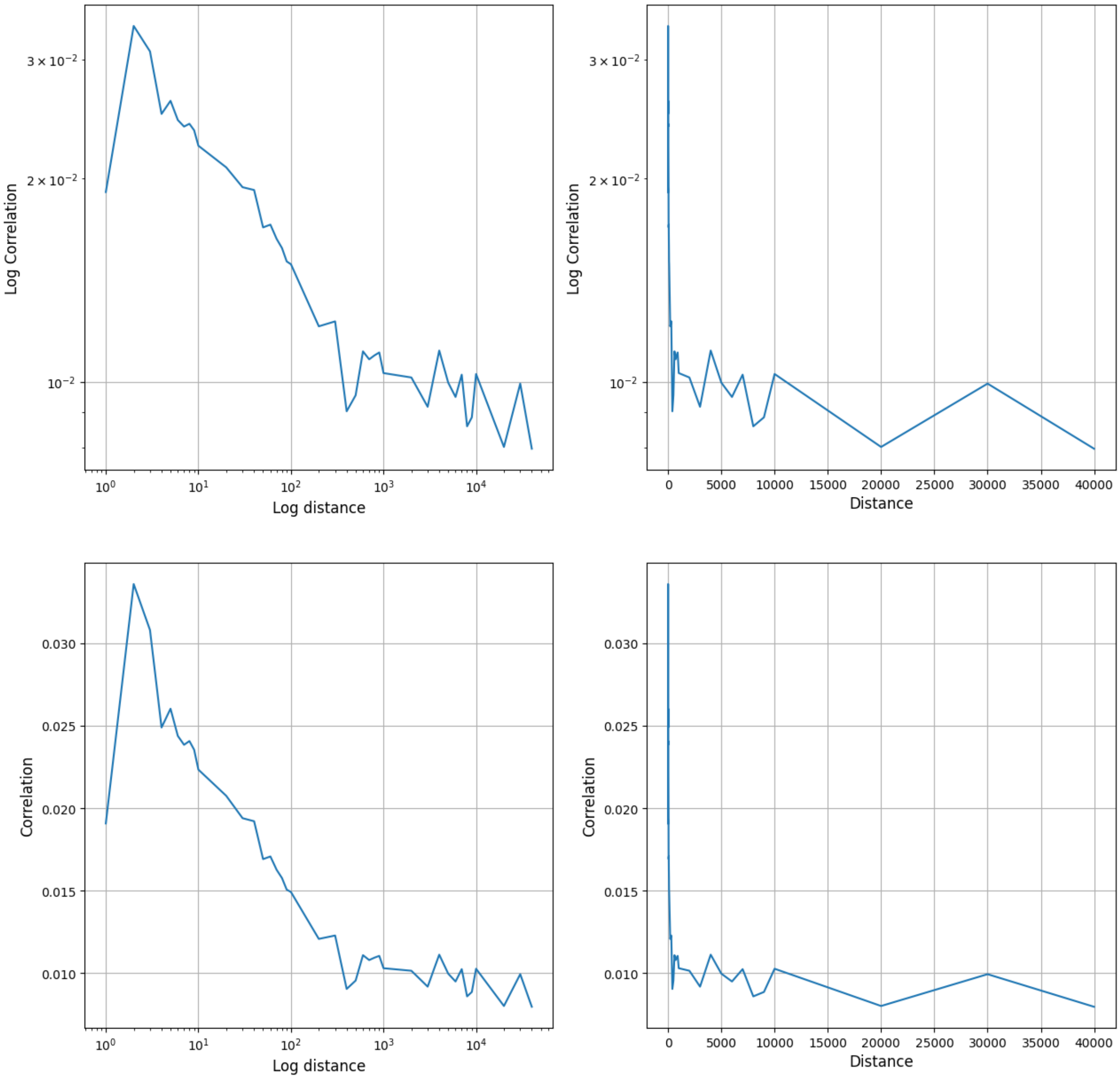

Figure 69: GloVe computation of autocorrelations in The Adventures of Tom Sawyer in French in different coordinates

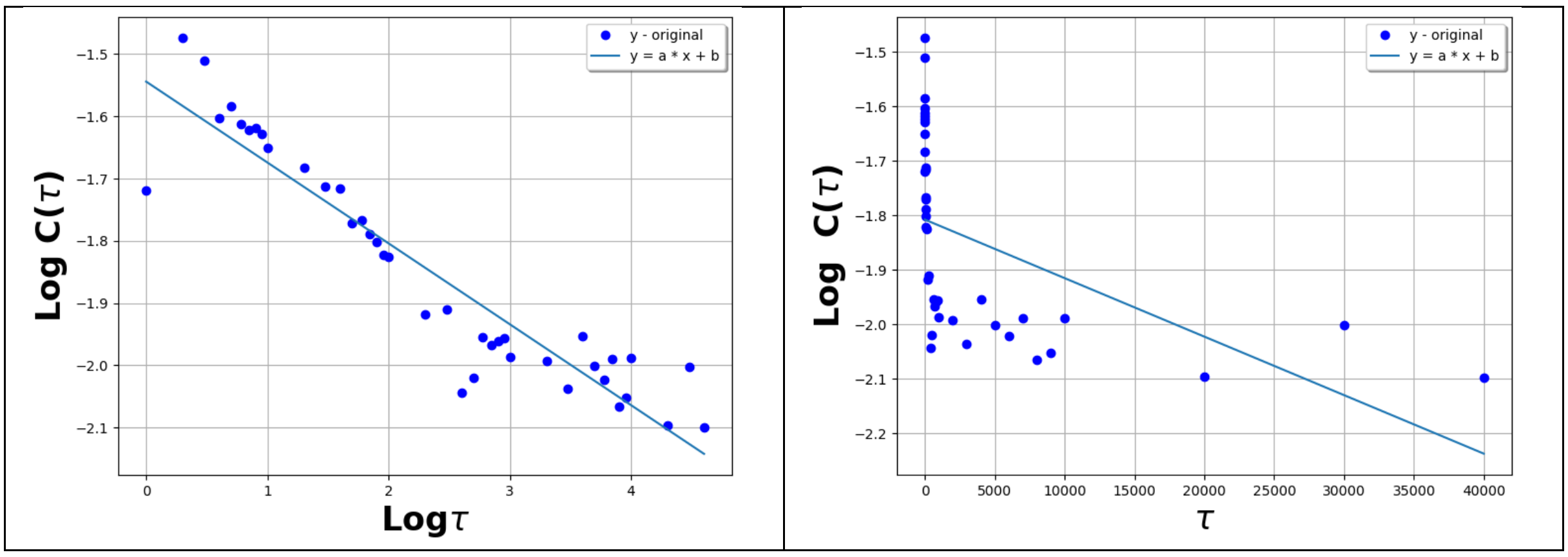

Figure 70: Best fit of GloVe computation of autocorrelations in The Adventures of Tom Sawyer in French by power (left) and exp (right)

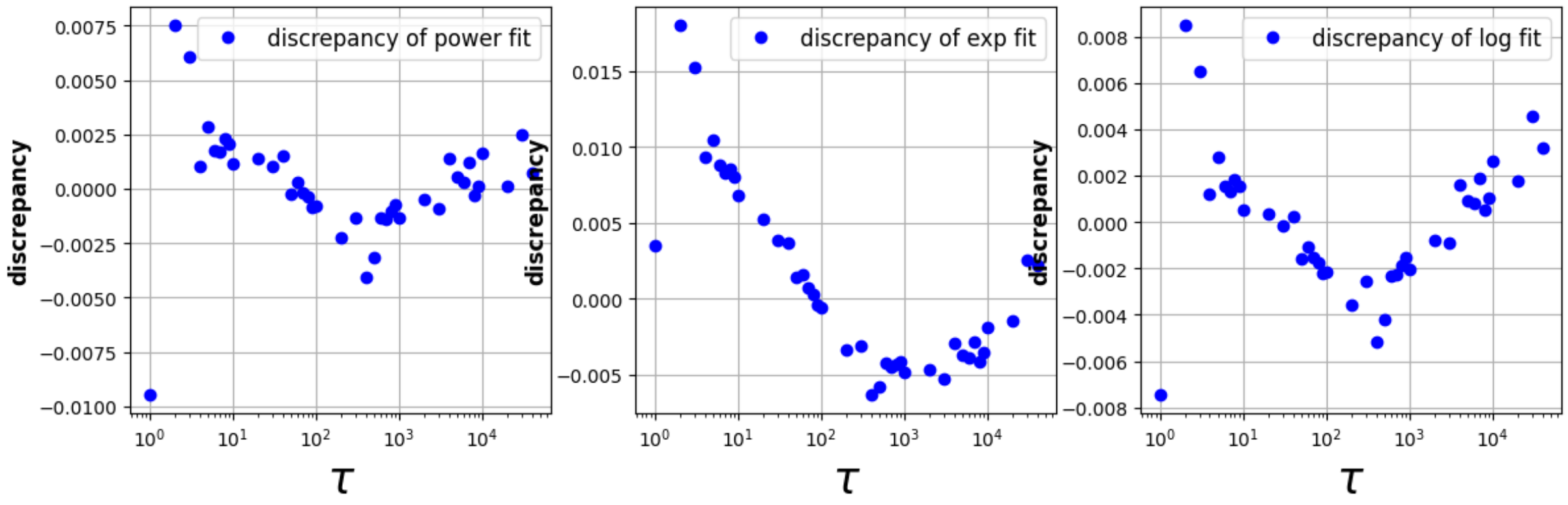

Figure 71: Residual graphs of the best fit of GloVe computation of autocorrelations in The Adventures of Tom Sawyer in French

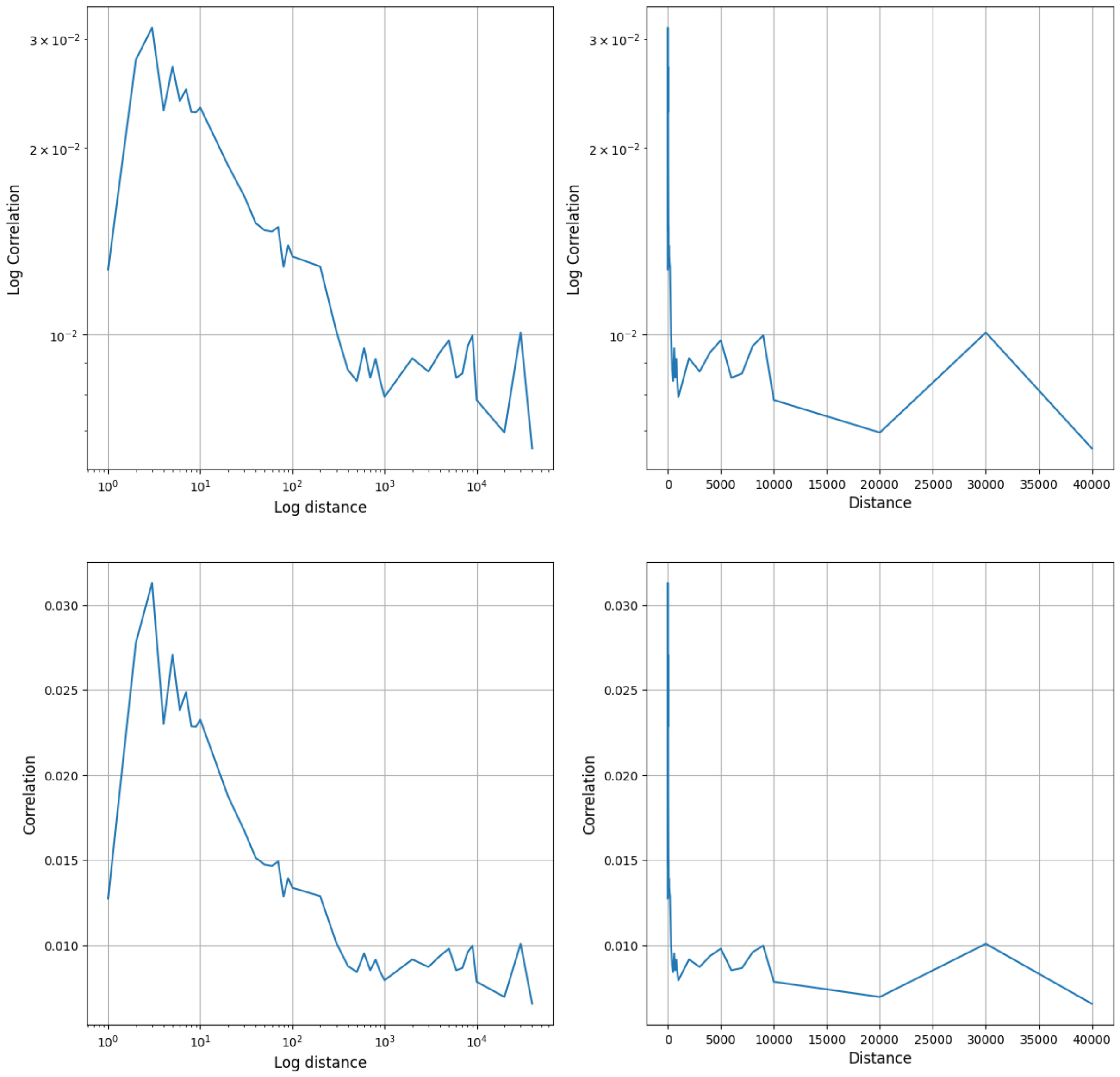

Figure 72: GloVe computation of autocorrelations in The Adventures of Tom Sawyer in Spanish in different coordinates

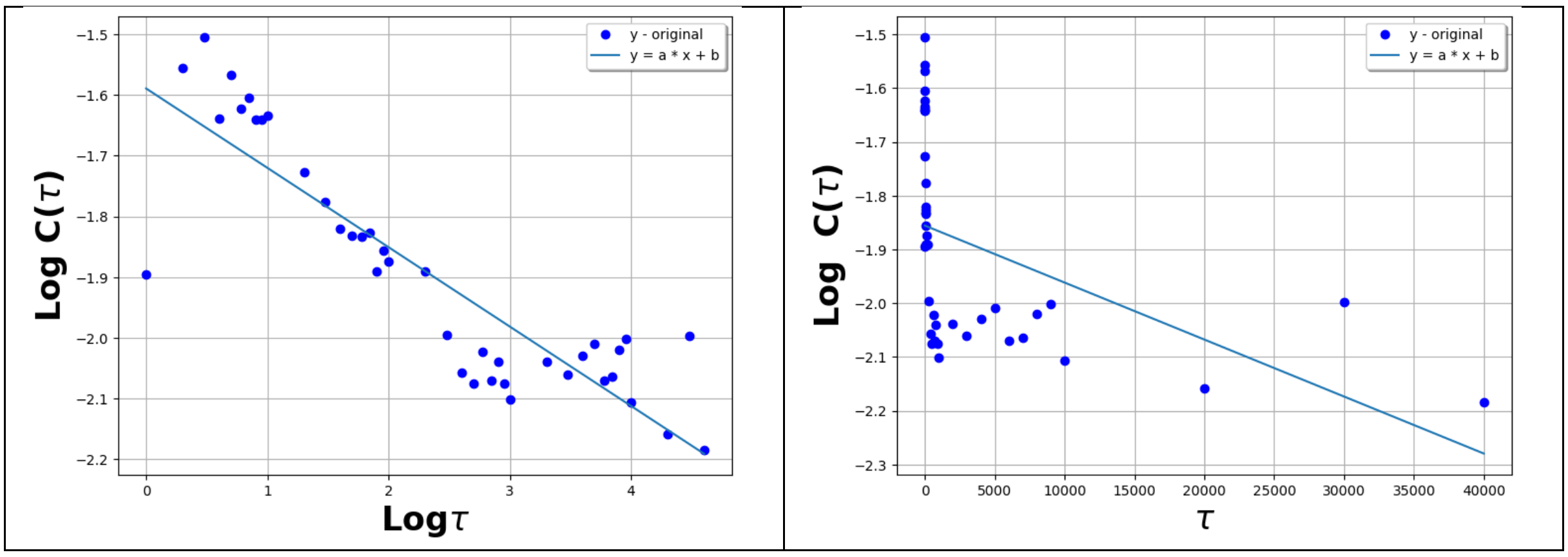

Figure 73: Best fit of GloVe computation of autocorrelations in The Adventures of Tom Sawyer in Spanish by power (left) and exp (right)

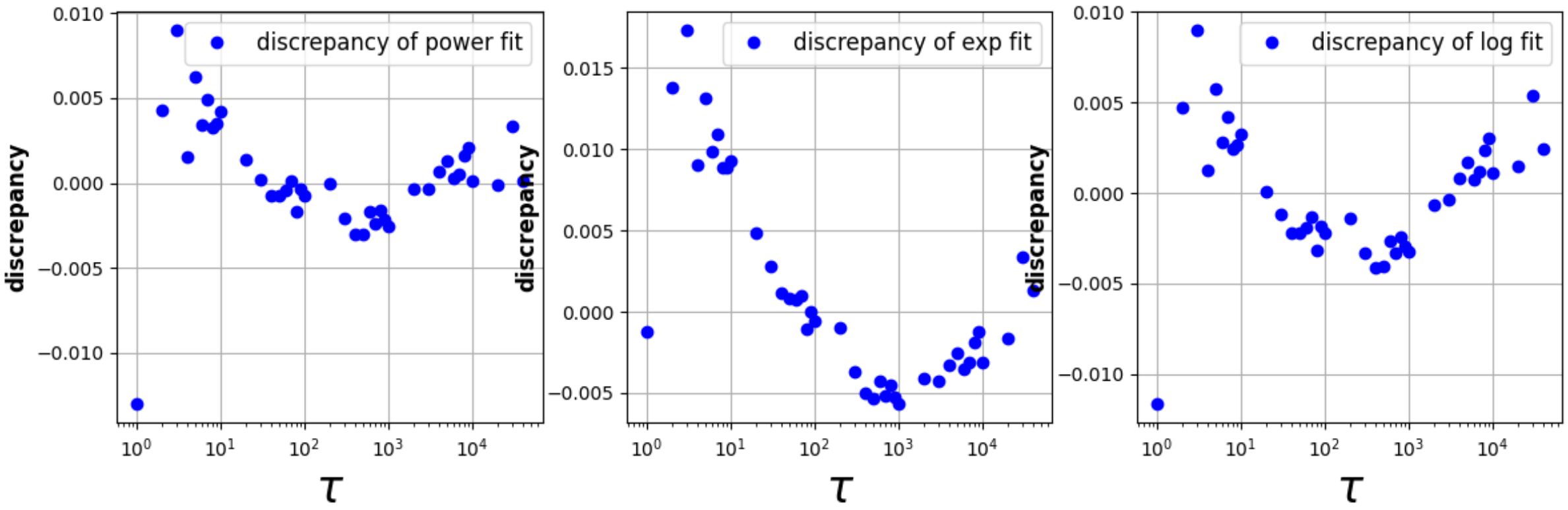

Figure 74: Residual graphs of the best fit of GloVe computation of autocorrelations in The Adventures of Tom Sawyer in Spanish

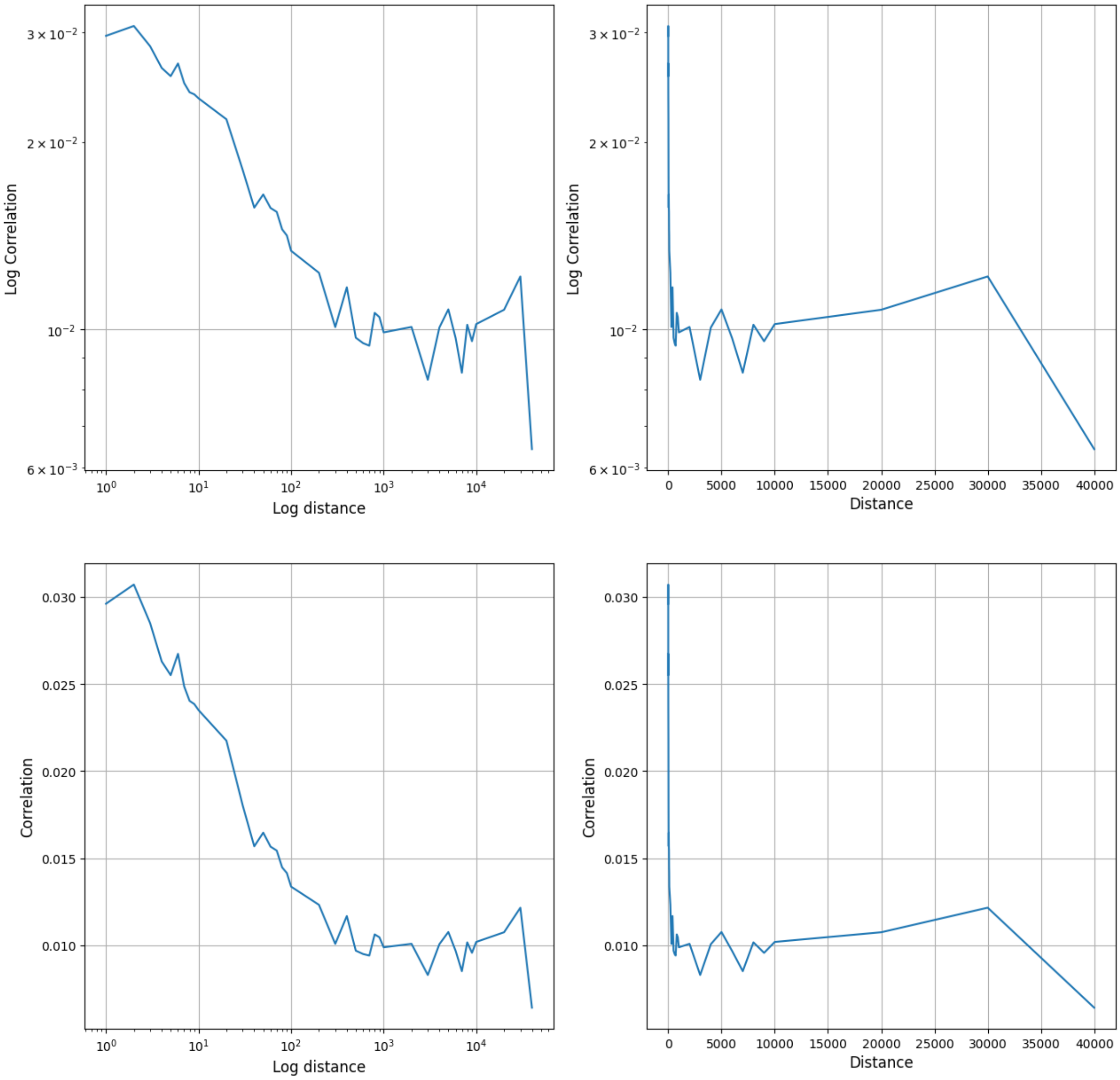

Figure 75: GloVe computation of autocorrelations in The Adventures of Tom Sawyer in Russian in different coordinates

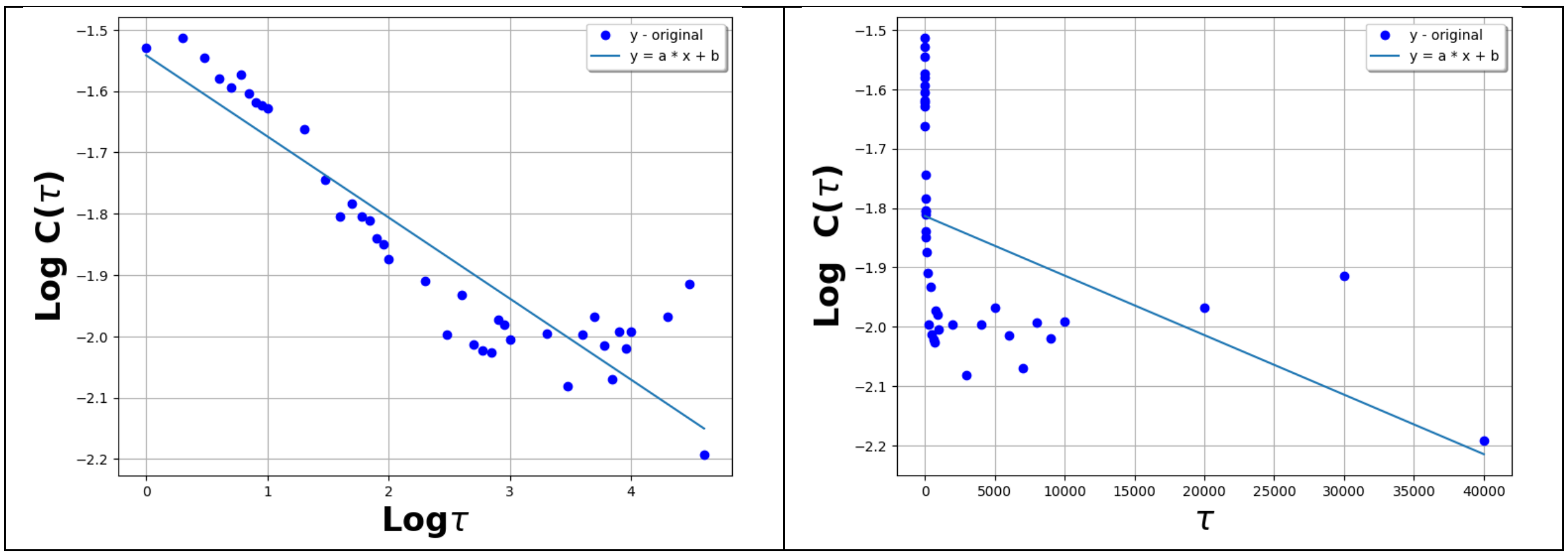

Figure 76: Best fit of GloVe computation of autocorrelations in The Adventures of Tom Sawyer in Russian by power (left) and exp (right)

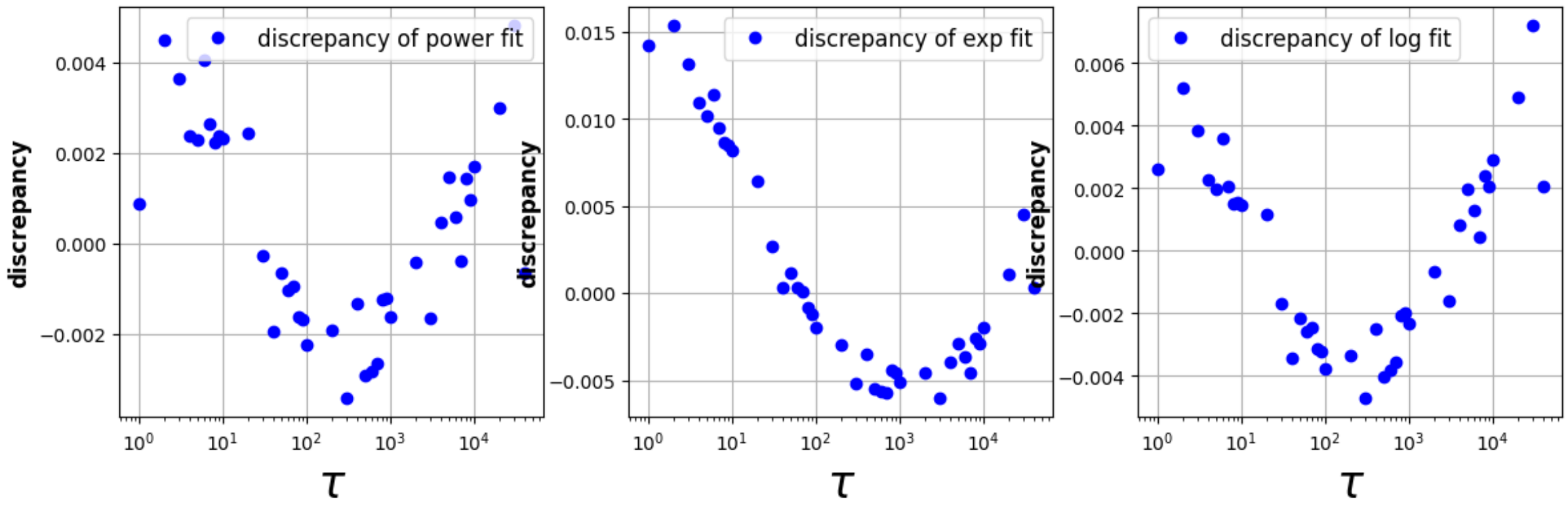

Figure 77: Residual graphs of the best fit of GloVe computation of autocorrelations in The Adventures of Tom Sawyer in Russian

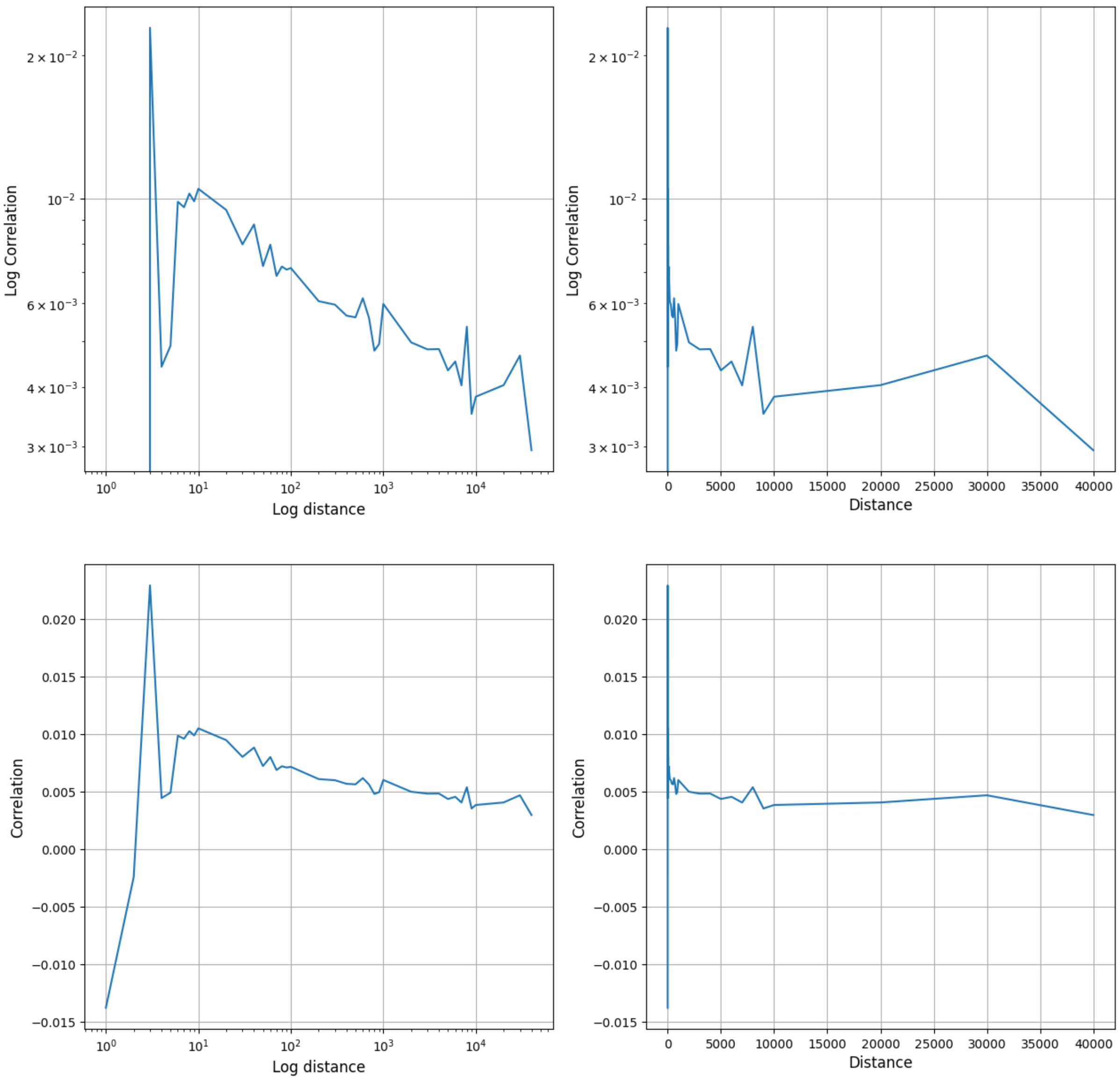

Figure 78: GloVe computation of autocorrelations in Critique of Pure Reason in English in different coordinates

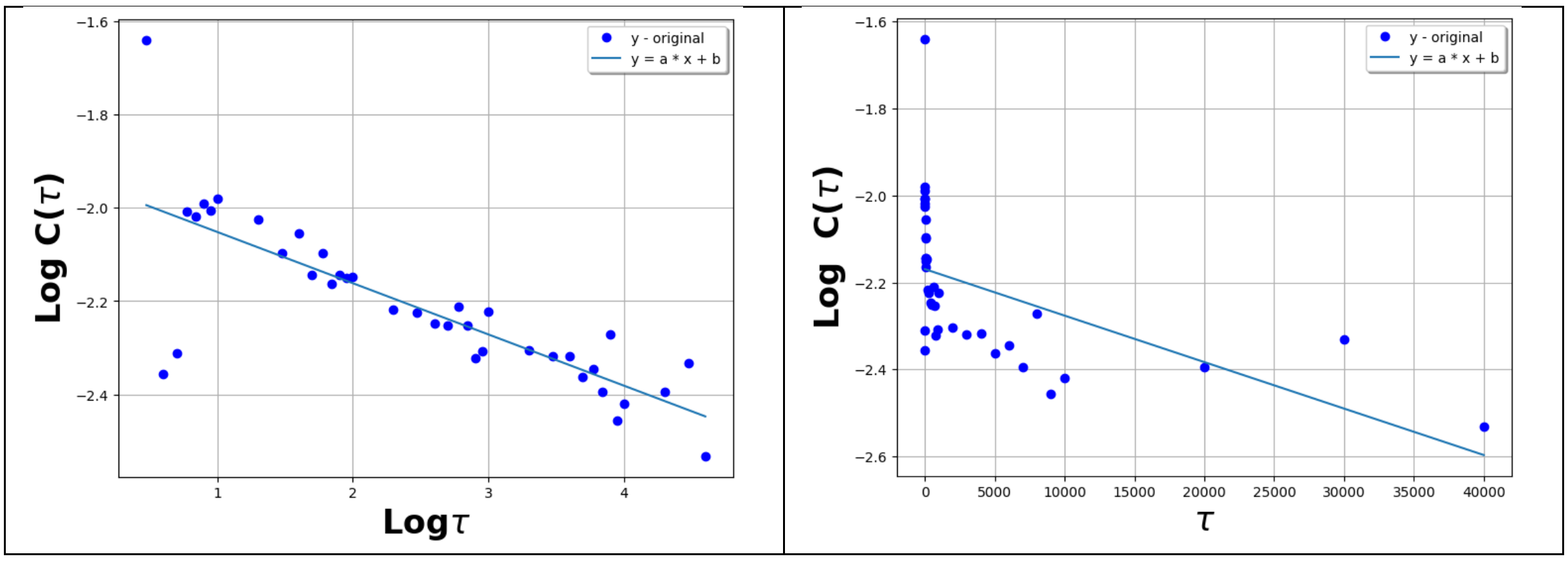

Figure 79: Best fit of GloVe computation of autocorrelations in Critique of Pure Reason in English by power (left) and exp (right)

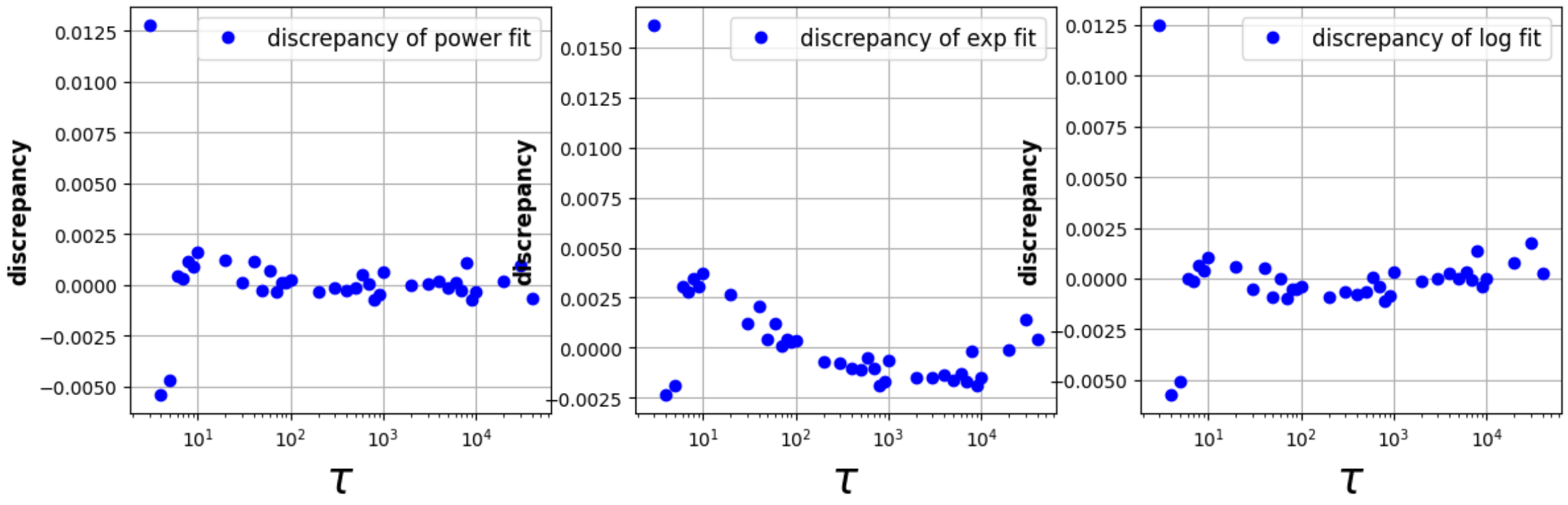

Figure 80: Residual graphs of the best fit of GloVe computation of autocorrelations in Critique of Pure Reason in English

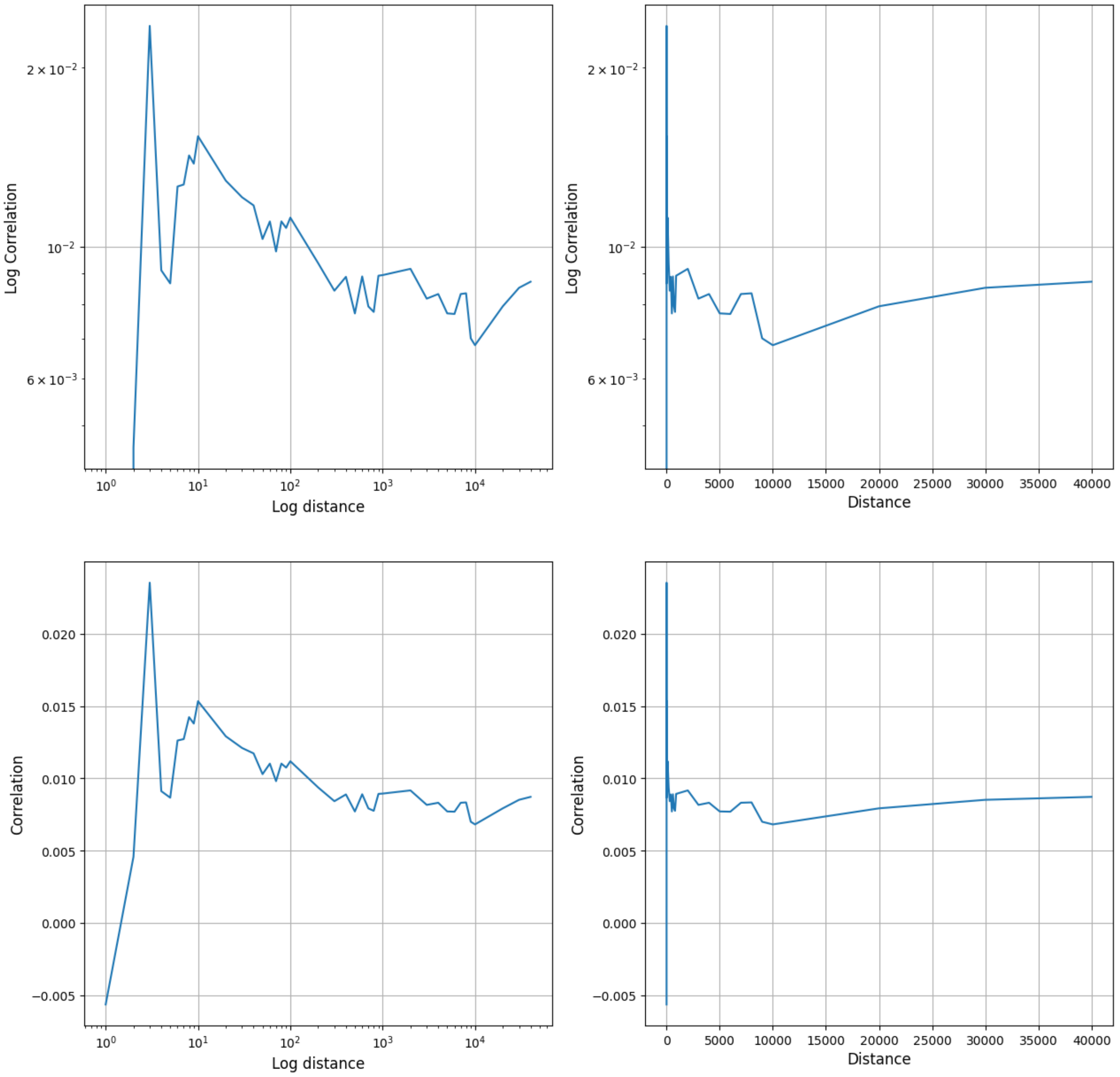

Figure 81: GloVe computation of autocorrelations in Critique of Pure Reason in Spanish in different coordinates

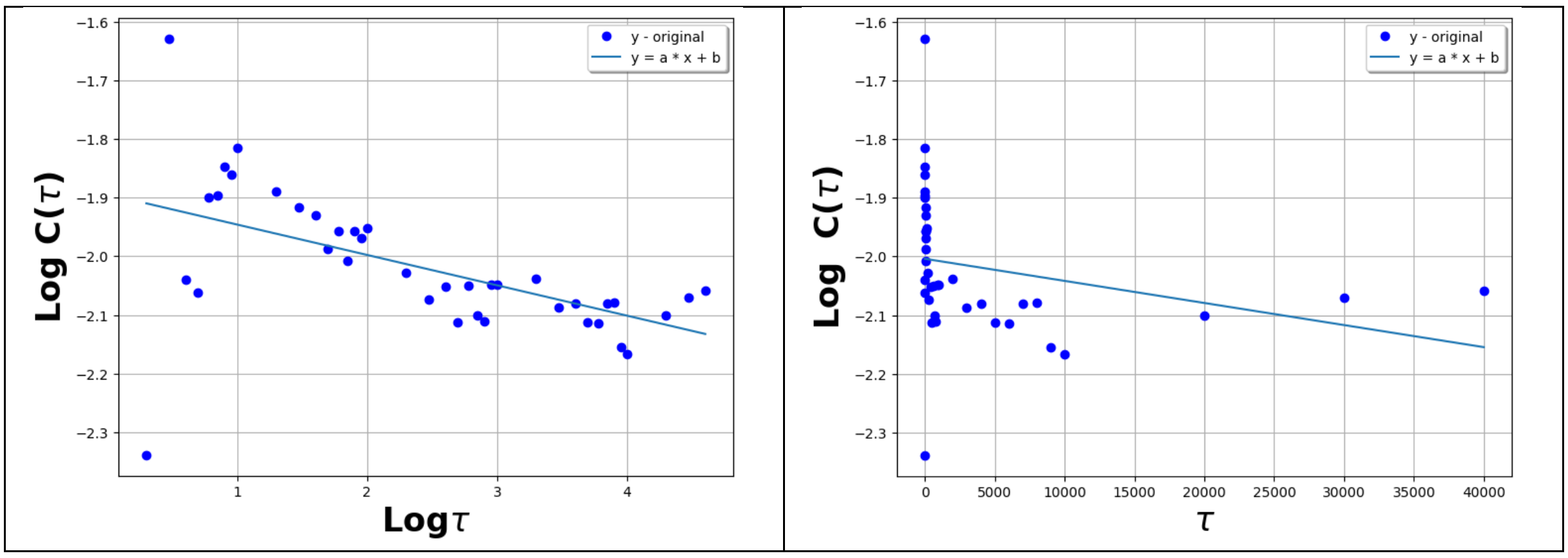

Figure 82: Best fit of GloVe computation of autocorrelations in Critique of Pure Reason in Spanish by power (left) and exp (right)

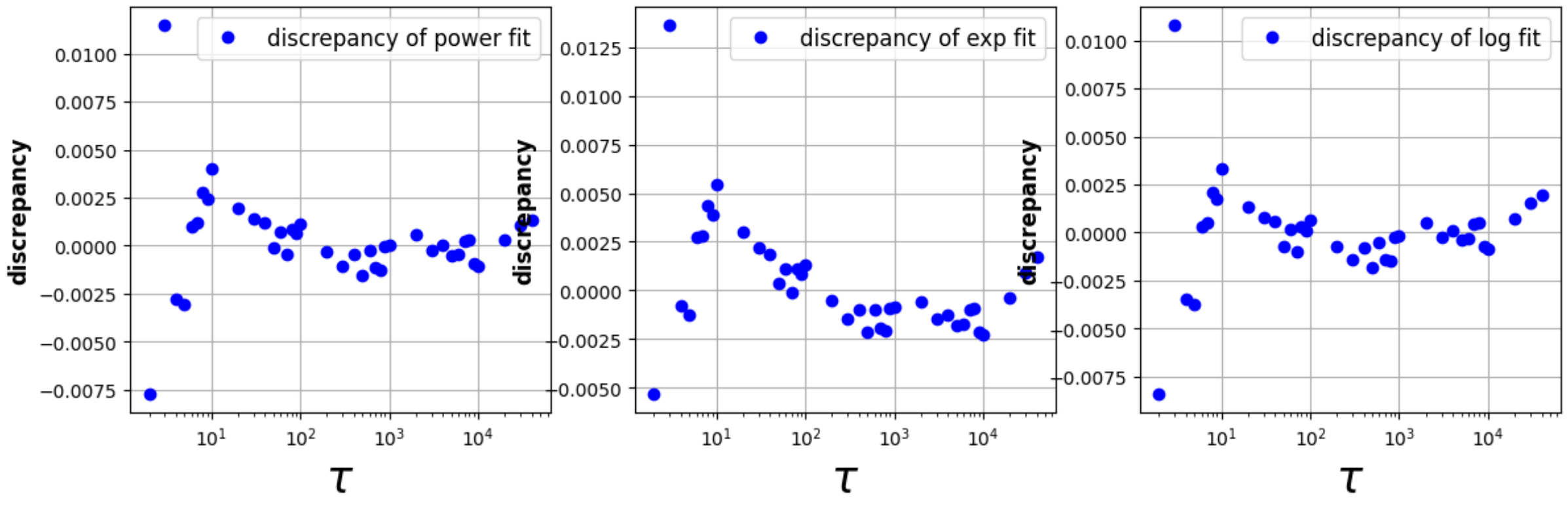

Figure 83: Residual graphs of the best fit of GloVe computation of autocorrelations in Critique of Pure Reason in Spanish

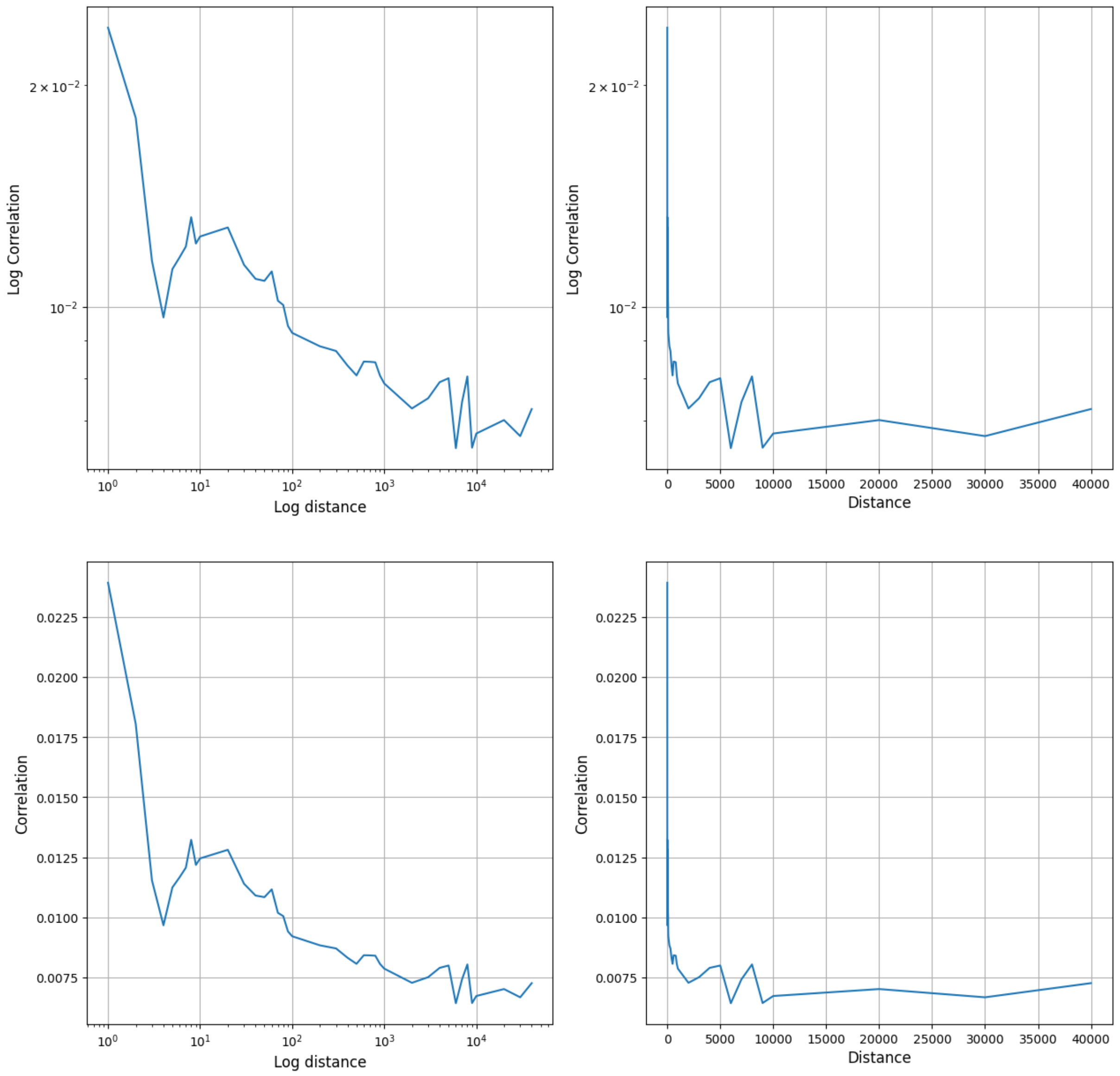

Figure 84: GloVe computation of autocorrelations in Critique of Pure Reason in German in different coordinates

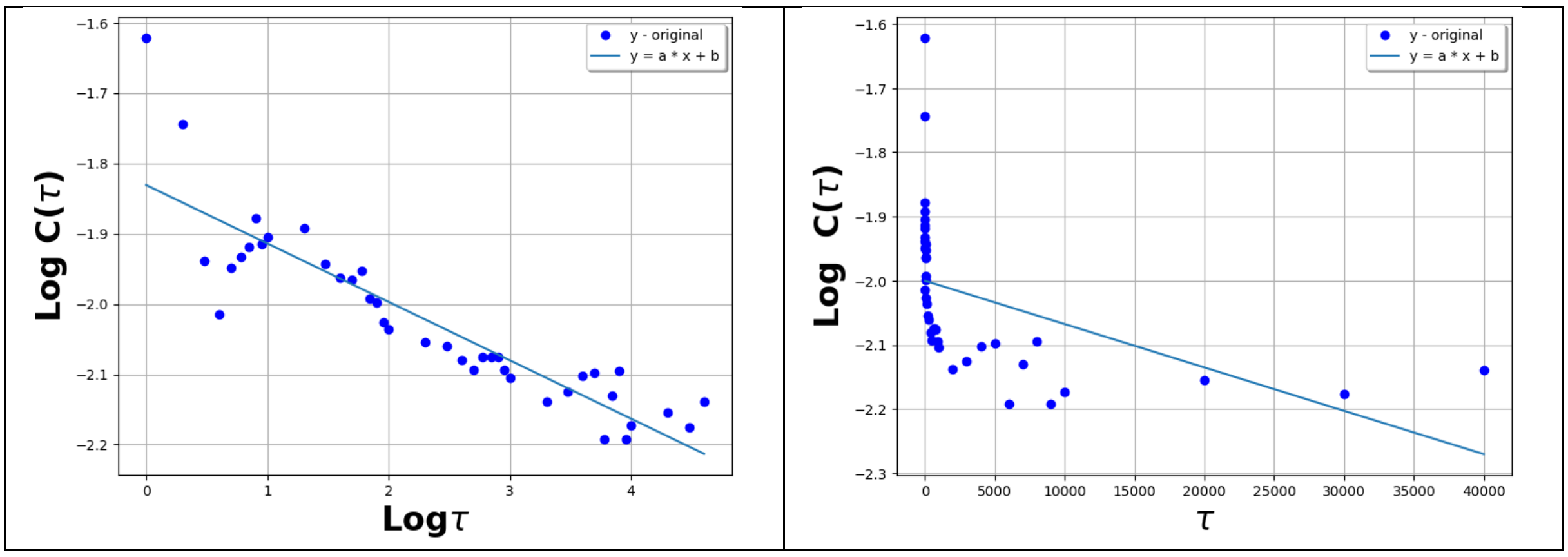

Figure 85: Best fit of GloVe computation of autocorrelations in Critique of Pure Reason in German by power (left) and exp (right)

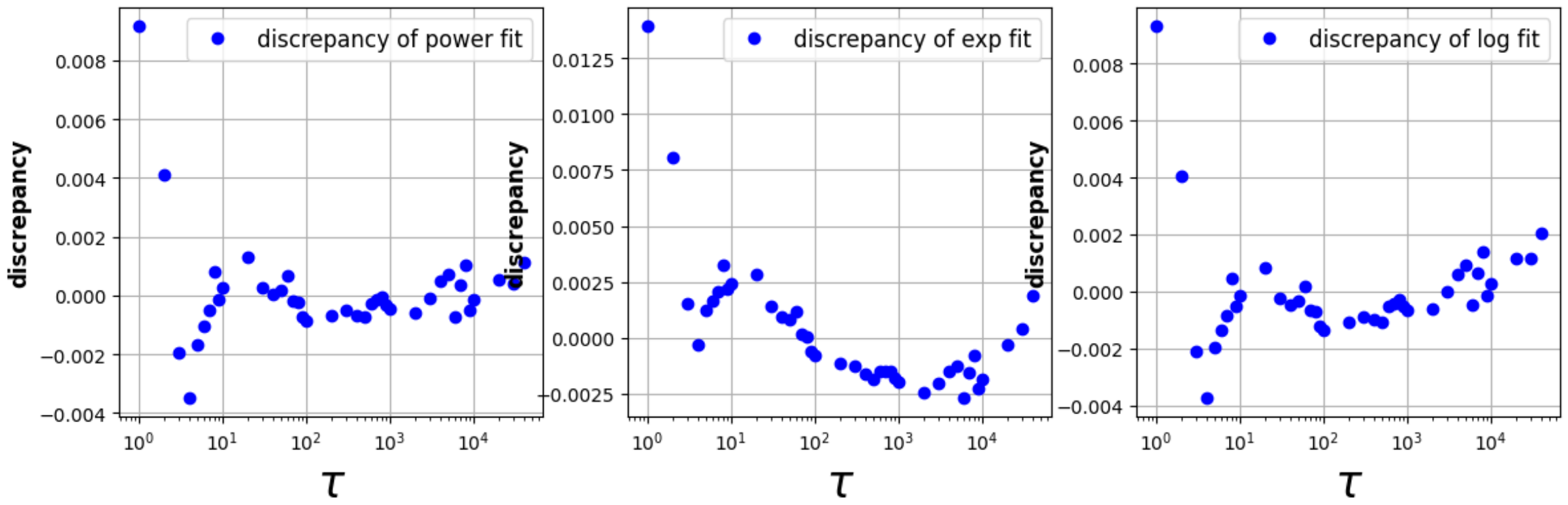

Figure 86: Residual graphs of the best fit of GloVe computation of autocorrelations in Critique of Pure Reason in German

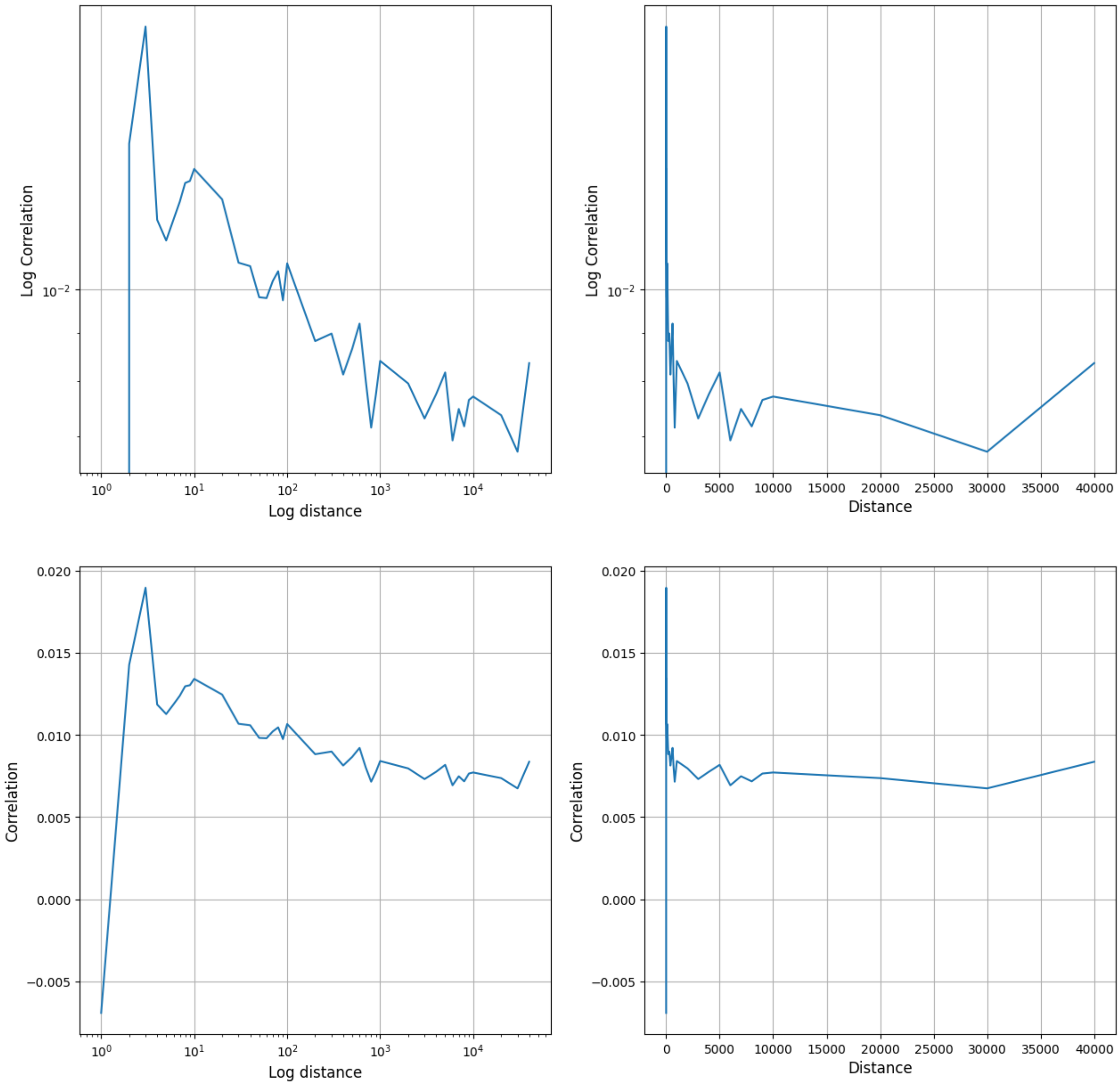

Figure 87: GloVe computation of autocorrelations in Critique of Pure Reason in German in different coordinates

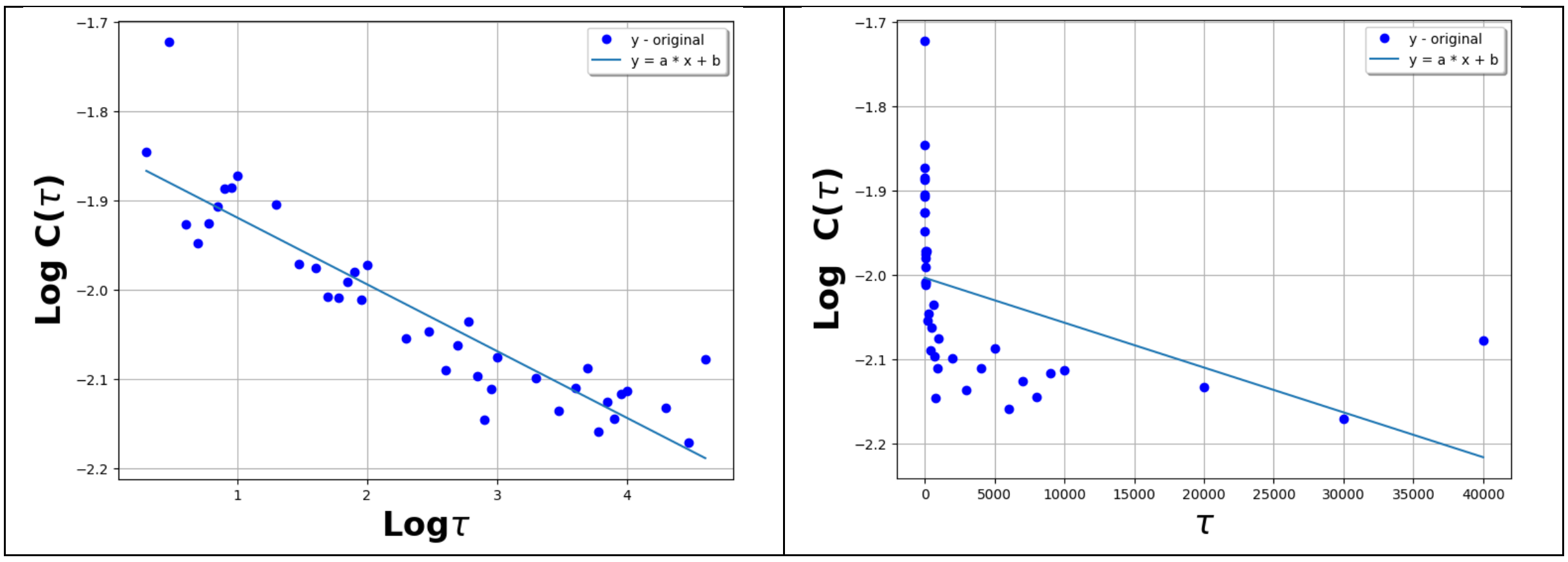

Figure 88: Best fit of GloVe computation of autocorrelations in Critique of Pure Reason in French by power (left) and exp (right)

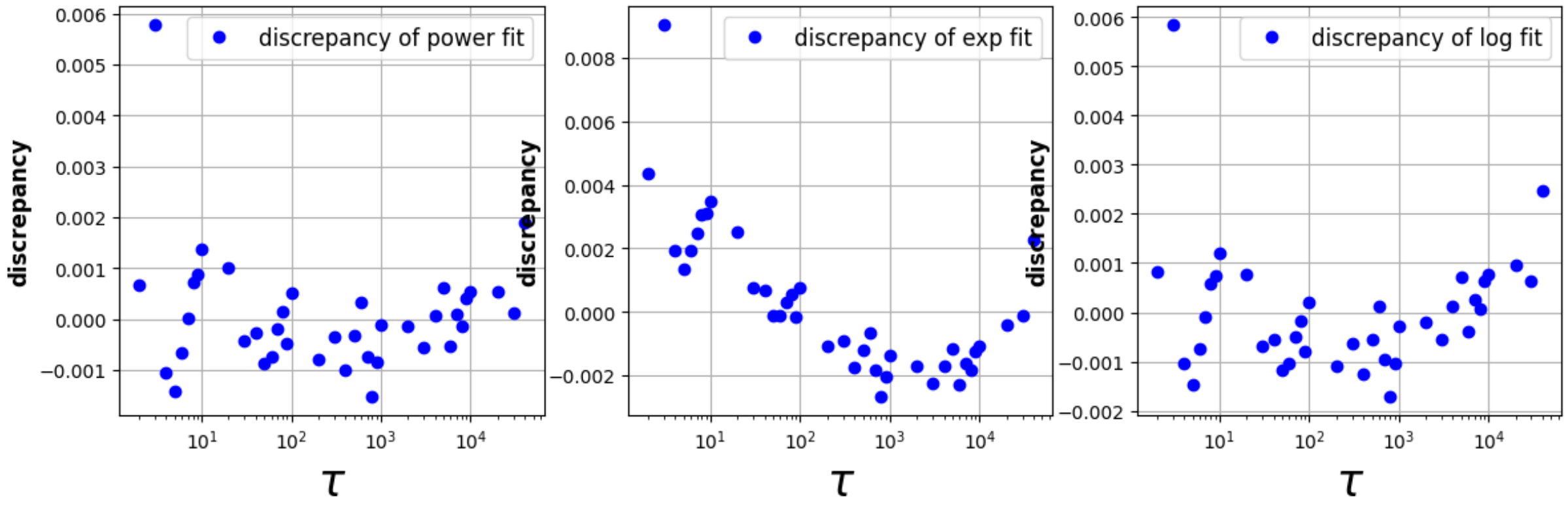

Figure 89: Residual graphs of the best fit of GloVe computation of autocorrelations in Critique of Pure Reason in French

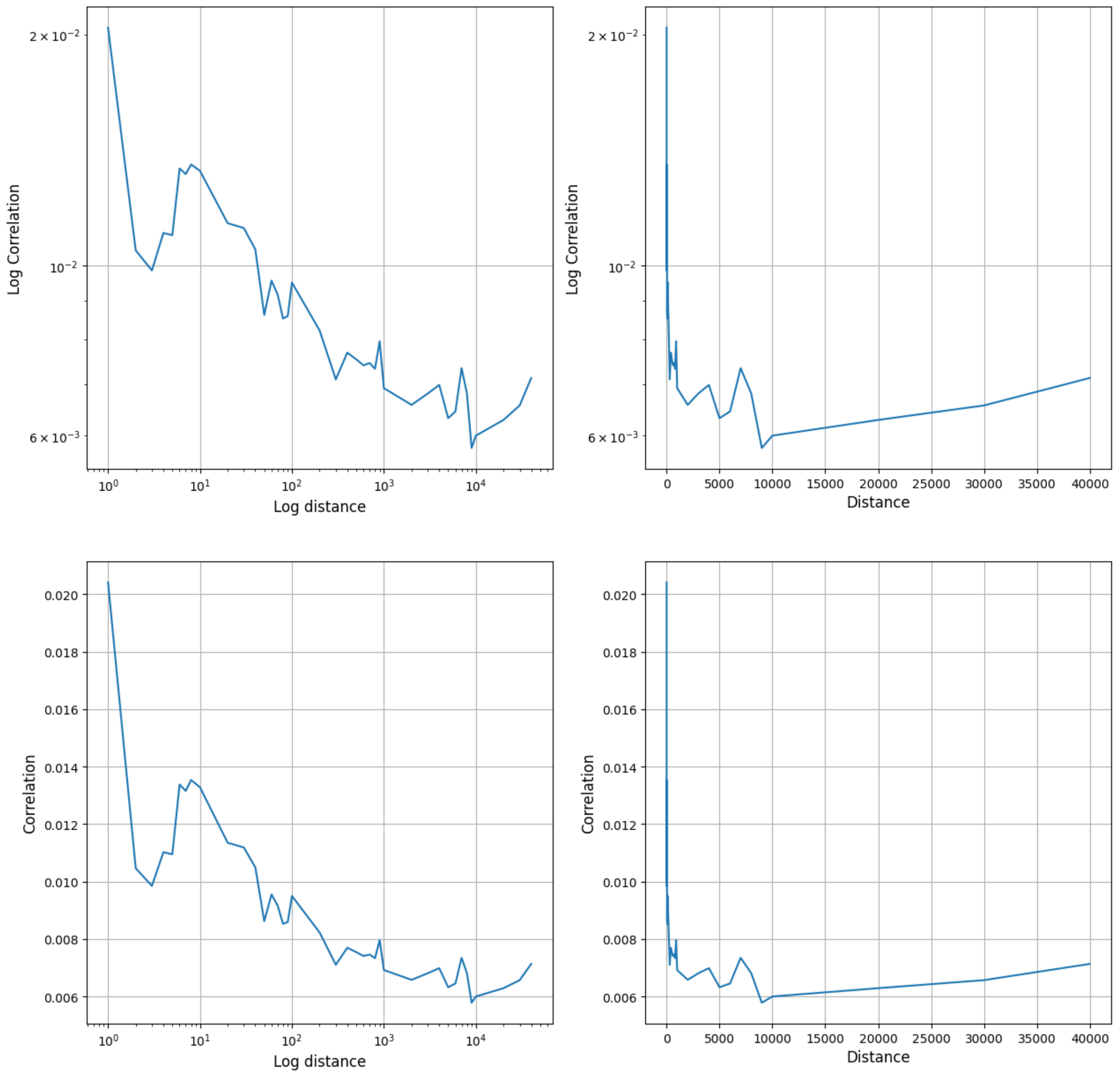

Figure 90: GloVe computation of autocorrelations in Critique of Pure Reason in Russian in different coordinates

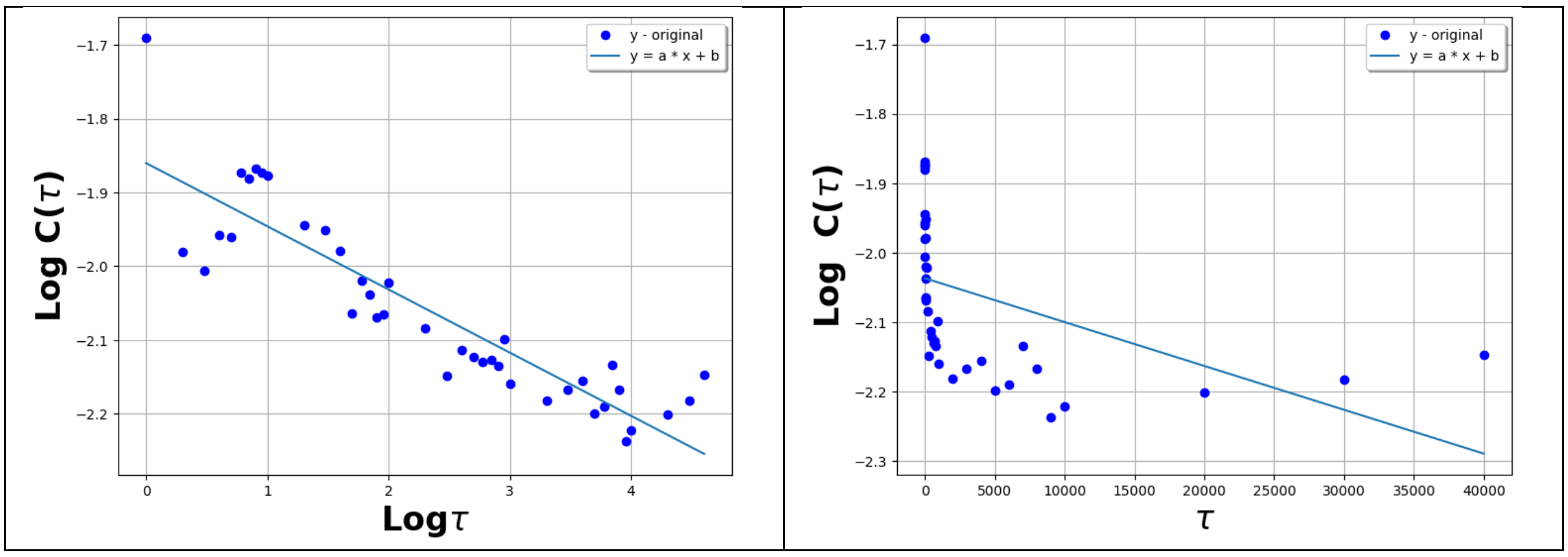

Figure 91: Best fit of GloVe computation of autocorrelations in Critique of Pure Reason in Russian by power (left) and exp (right)

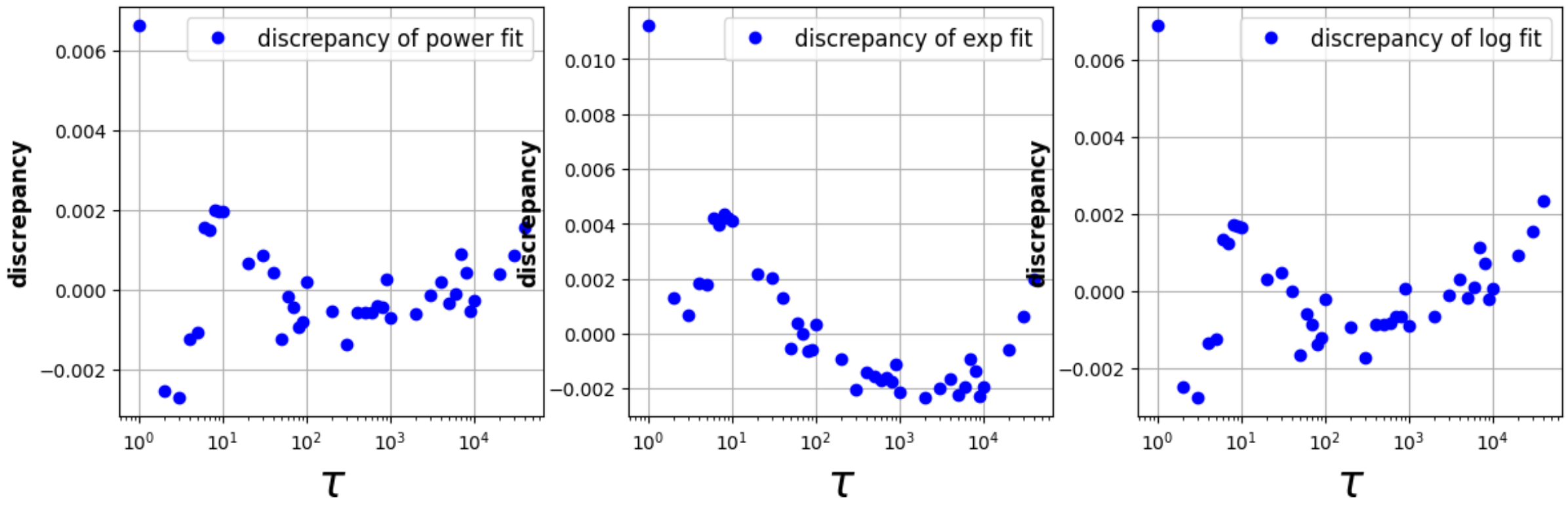

Figure 92: Residual graphs of the best fit of GloVe computation of autocorrelations in Critique of Pure Reason in Russian

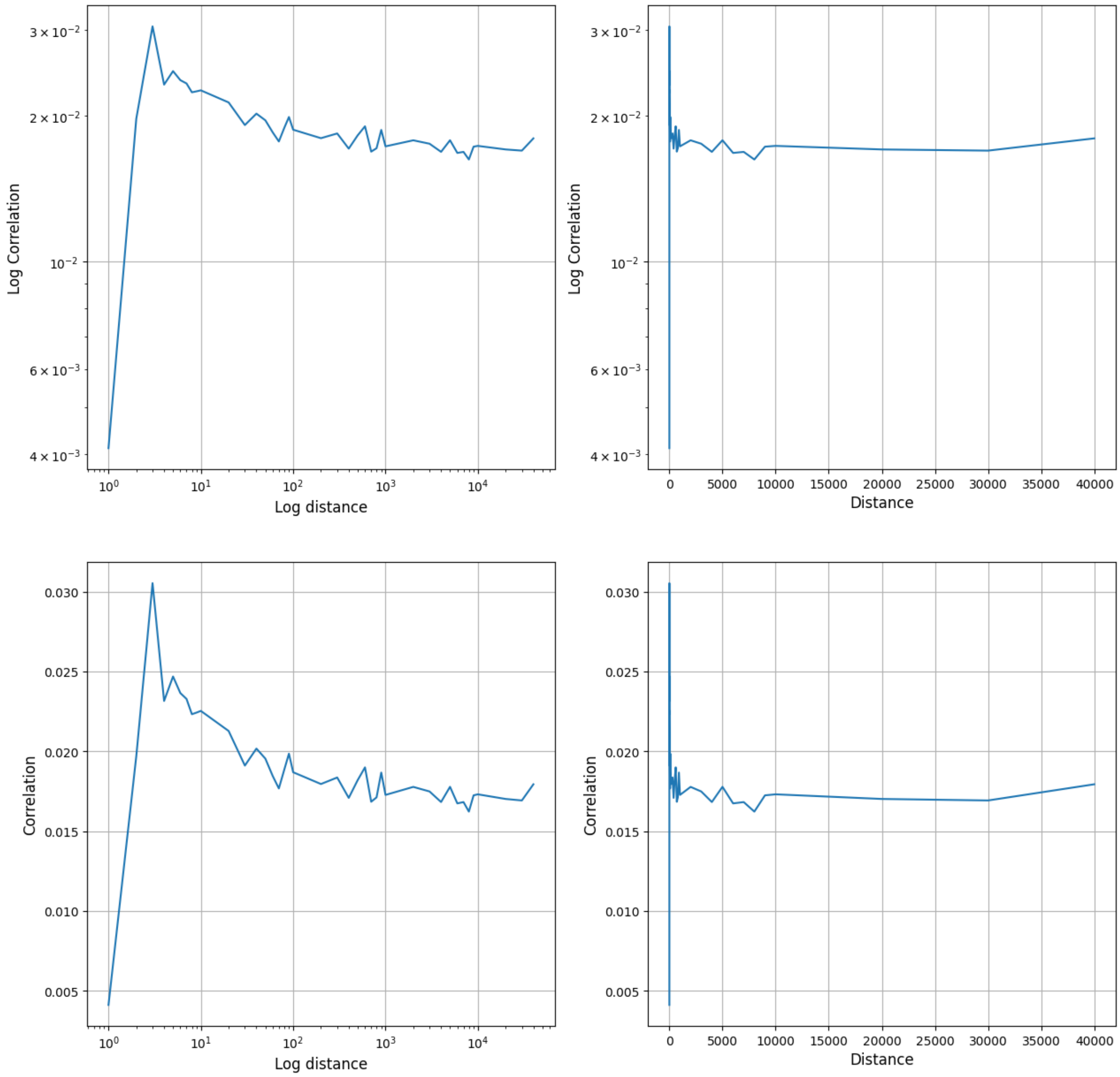

Figure 93: GloVe computation of autocorrelations in The Iliad in Spanish in different coordinates

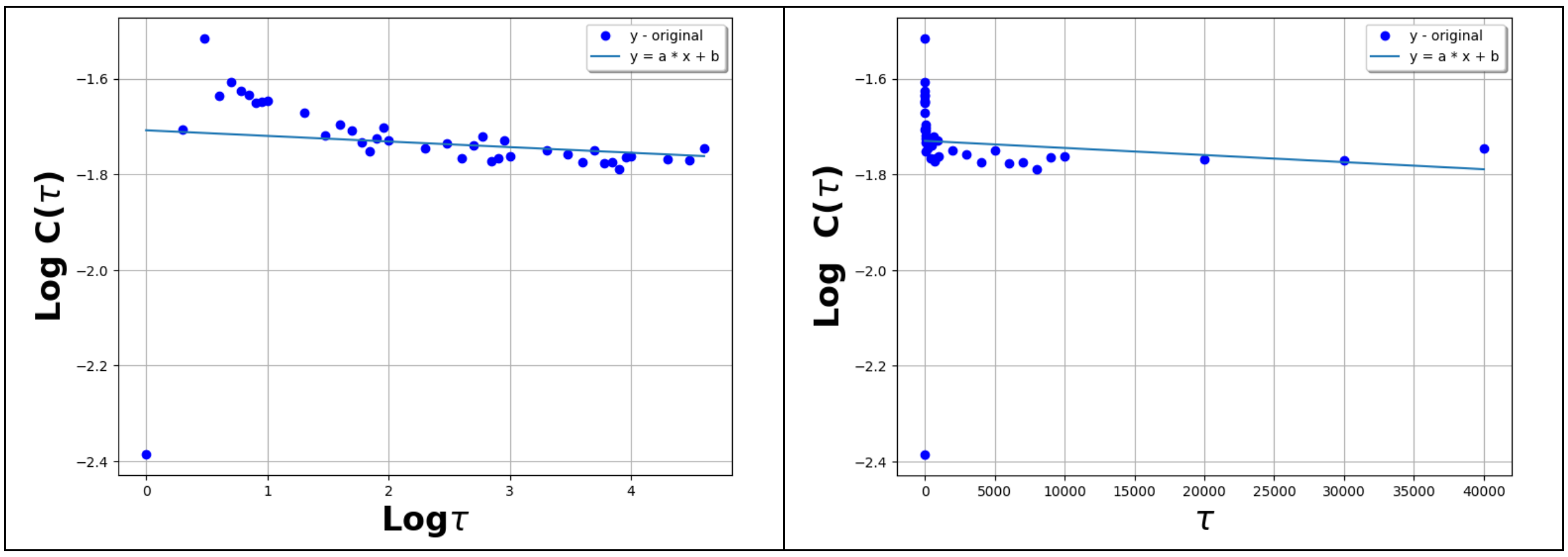

Figure 94: Best fit of GloVe computation of autocorrelations The Iliad in Spanish by power (left) and exp (right)

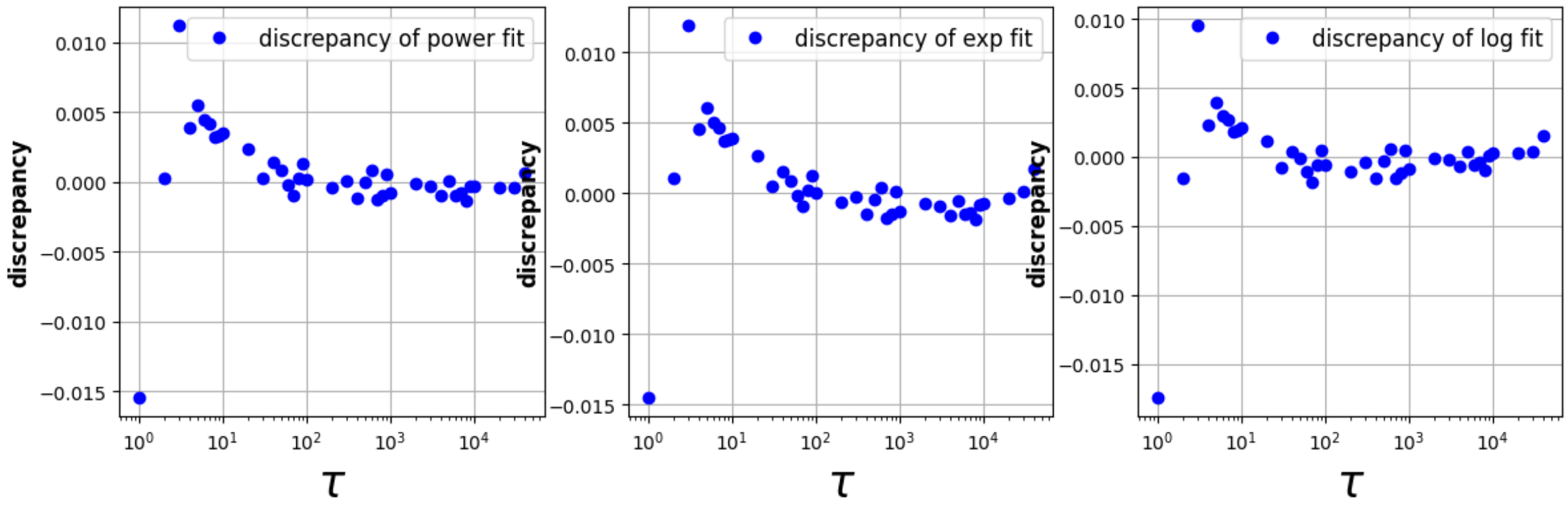

Figure 95: Residual graphs of the best fit of GloVe computation of autocorrelations in The Iliad in Spanish

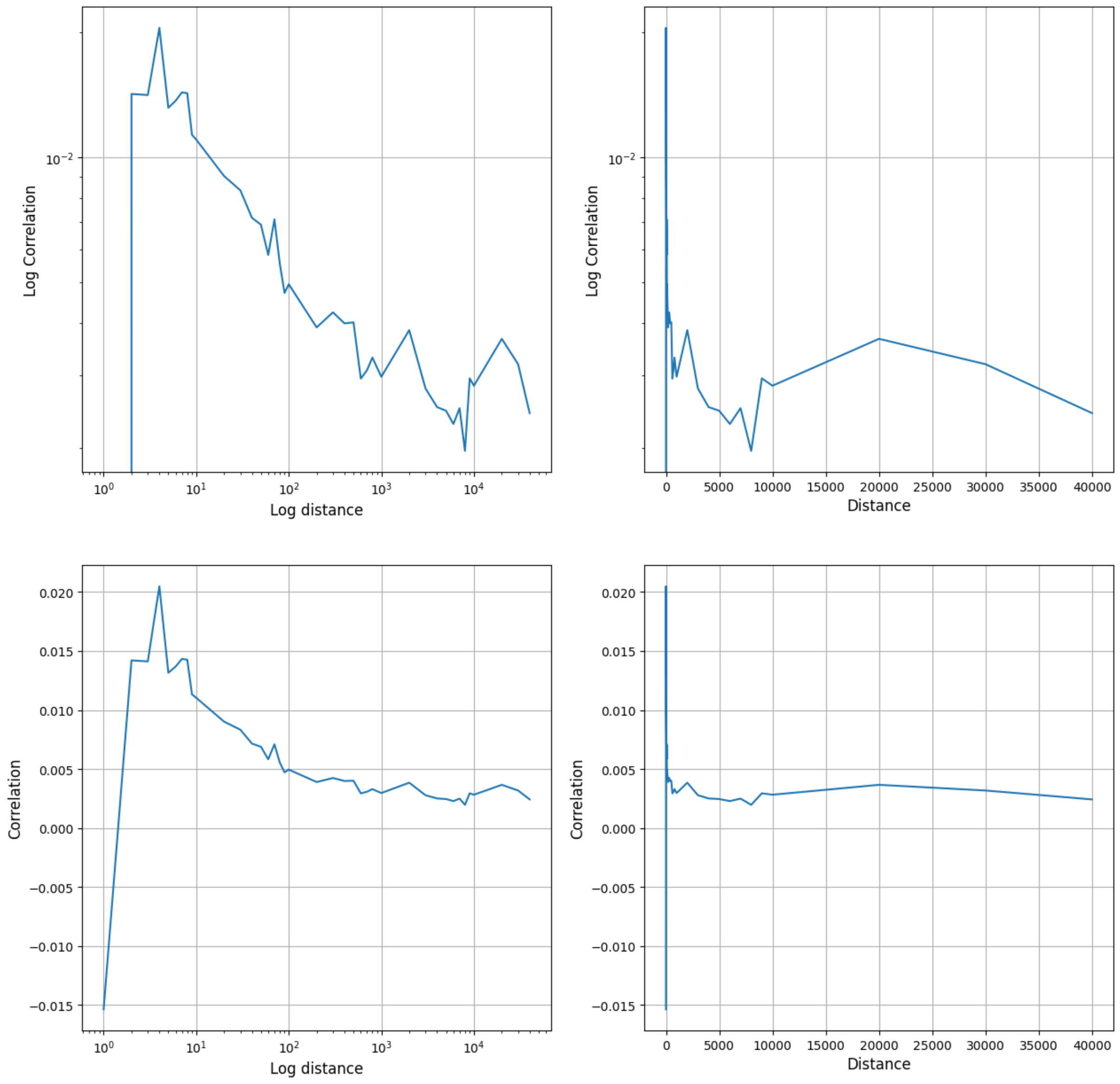

Figure 96: GloVe computation of autocorrelations in The Iliad in English in different coordinates

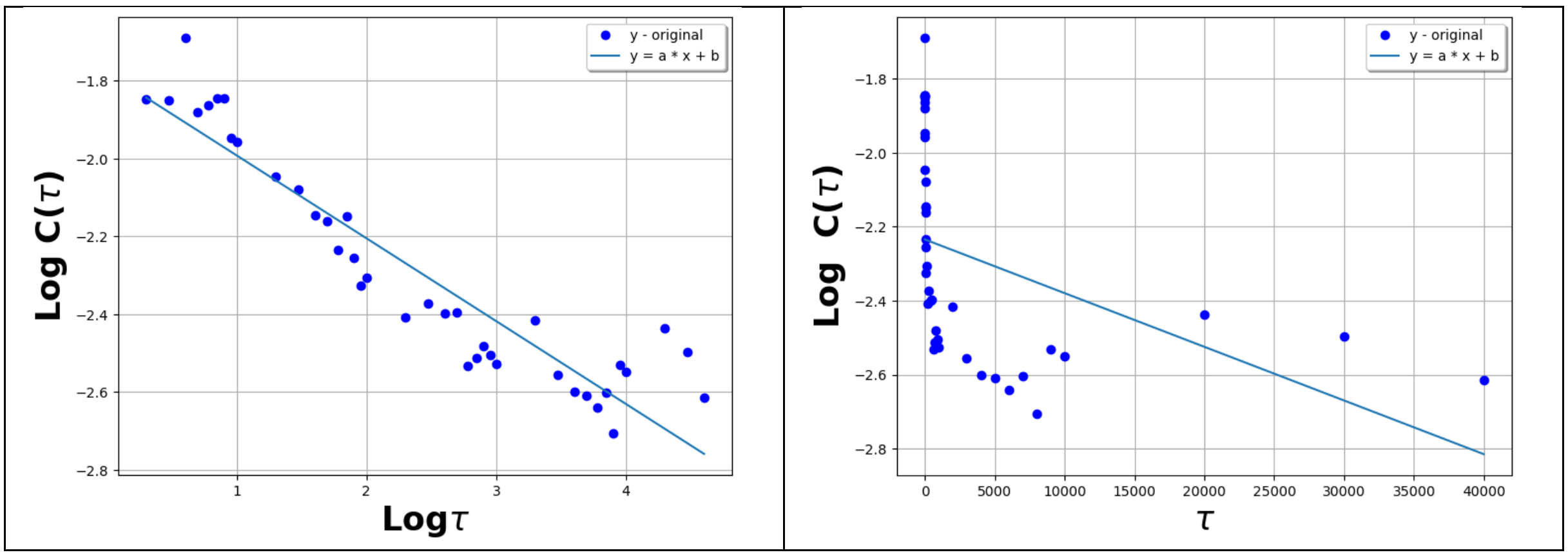

Figure 97: Best fit of GloVe computation of autocorrelations The Iliad in English by power (left) and exp (right)

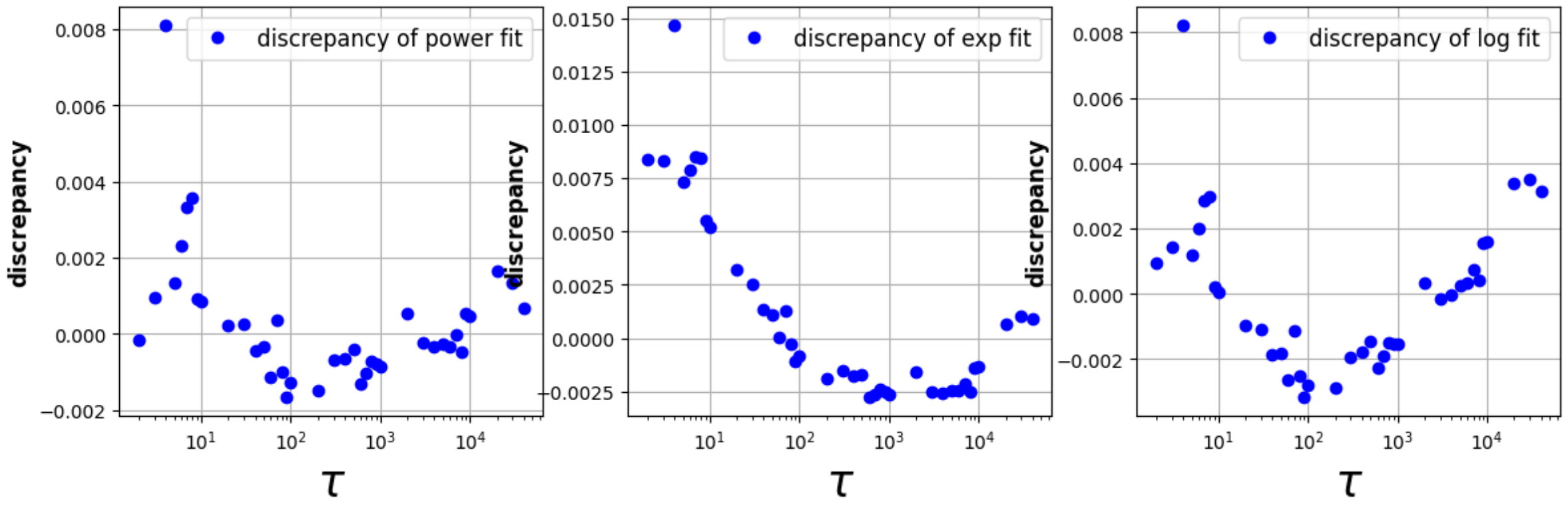

Figure 98 Residual graphs of the best fit of GloVe computation of autocorrelations in The Iliad in English

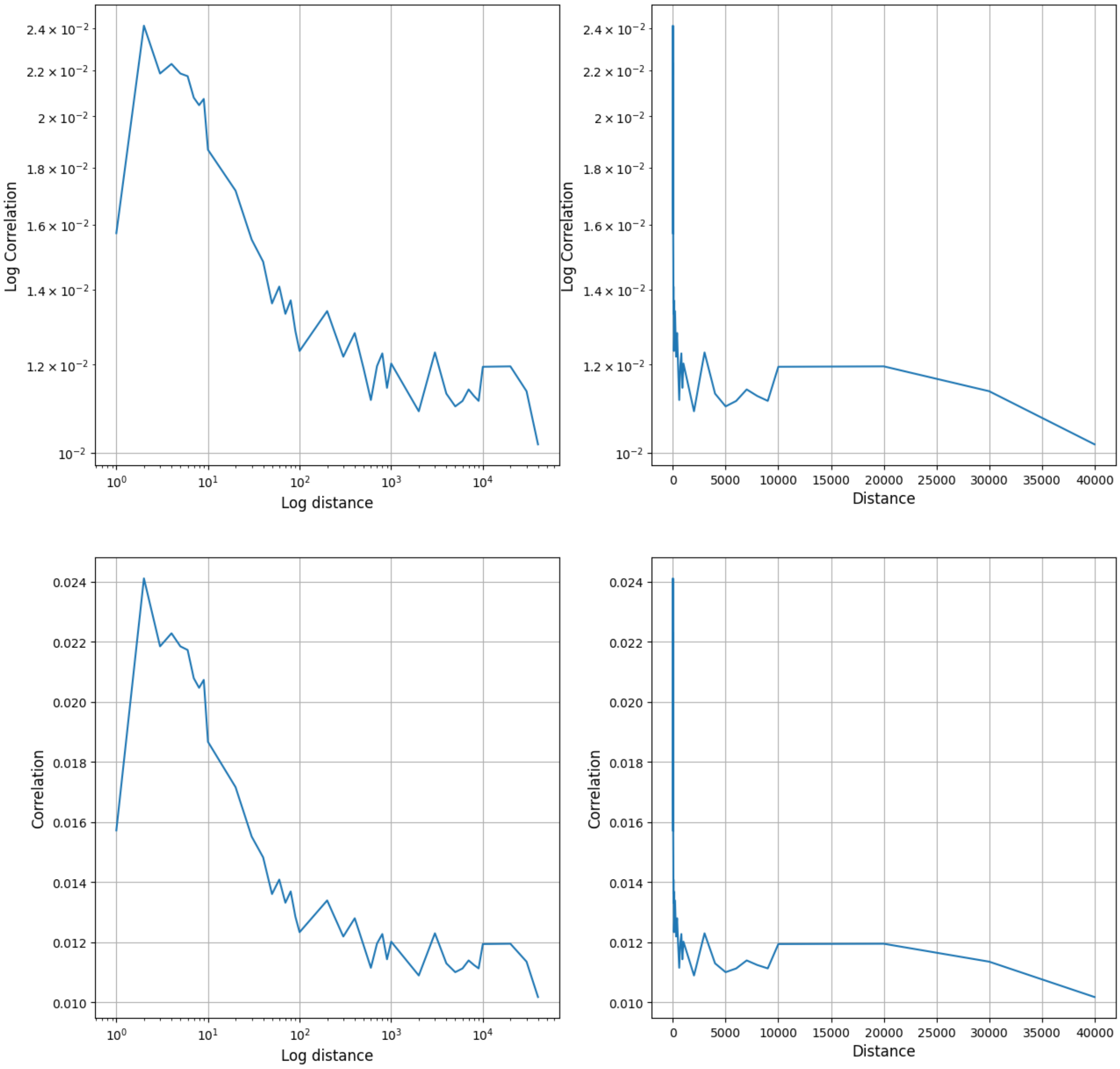

Figure 99: GloVe computation of autocorrelations in The Iliad in Russian in different coordinates

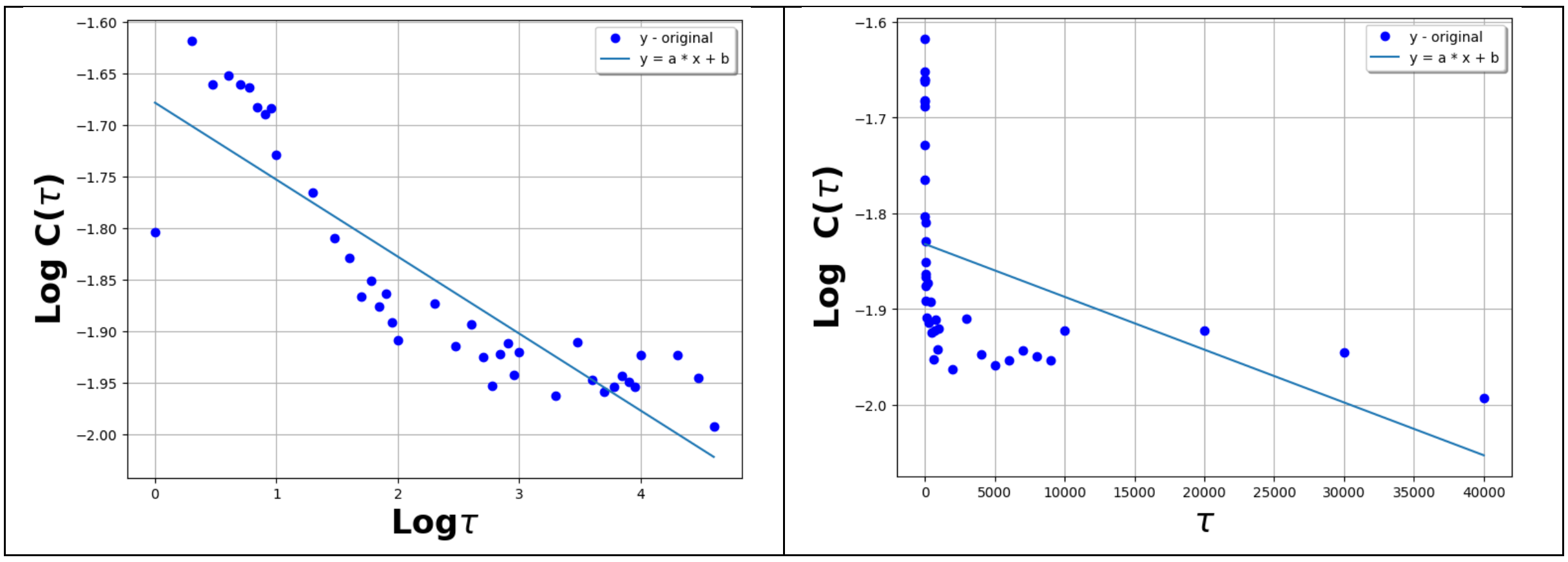

Figure 100: Best fit of GloVe computation of autocorrelations The Iliad in Russian by power (left) and exp (right)

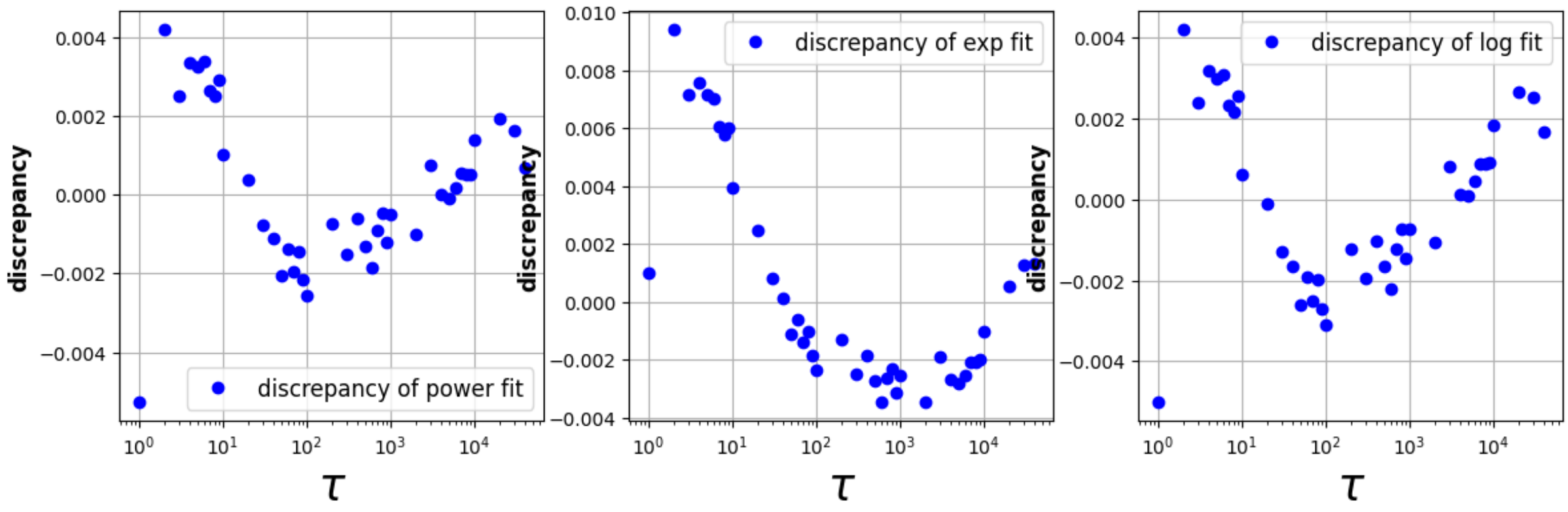

Figure 101: Residual graphs of the best fit of GloVe computation of autocorrelations in The Iliad in Russian

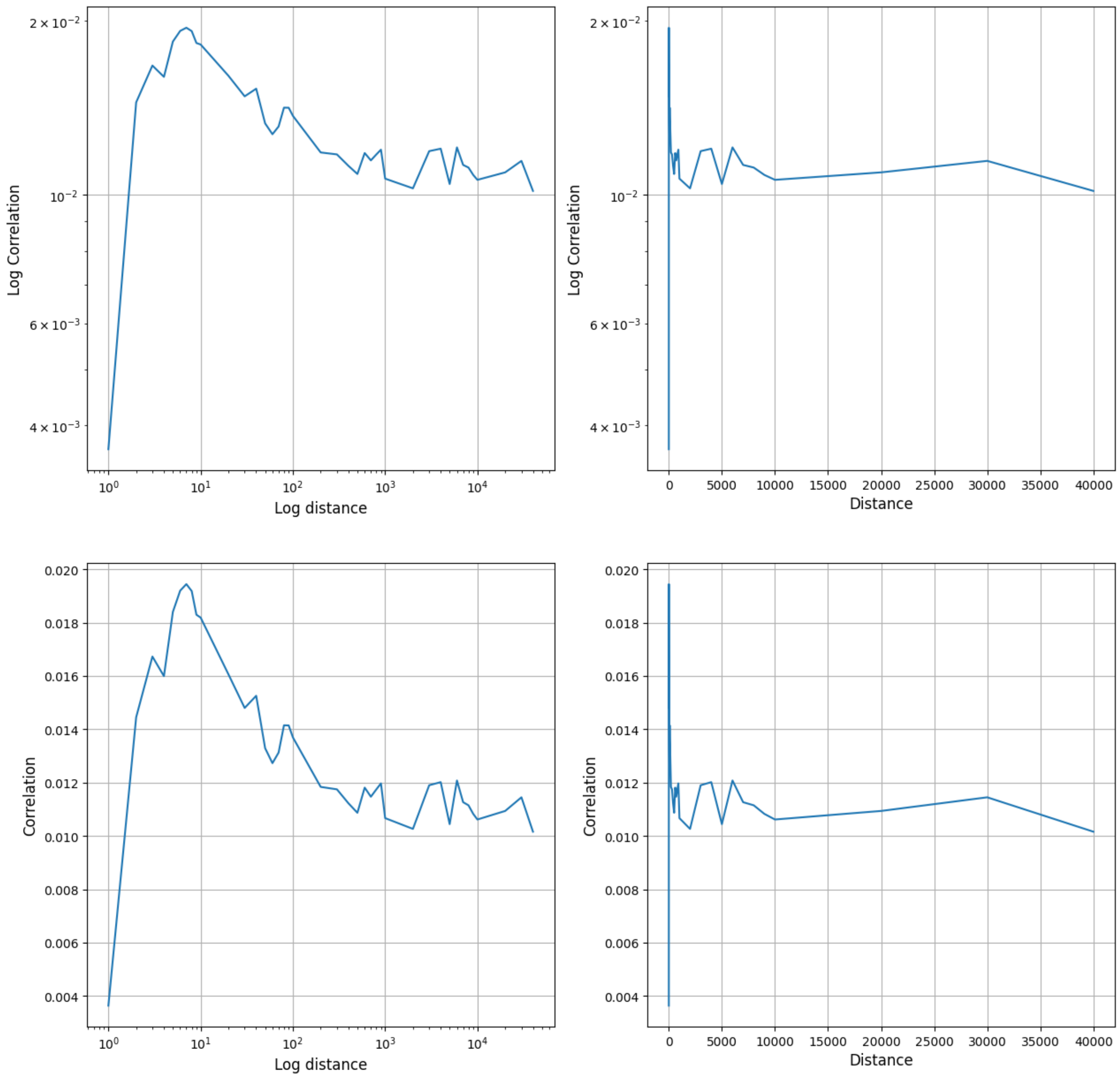

Figure 102: GloVe computation of autocorrelations in The Iliad in German in different coordinates

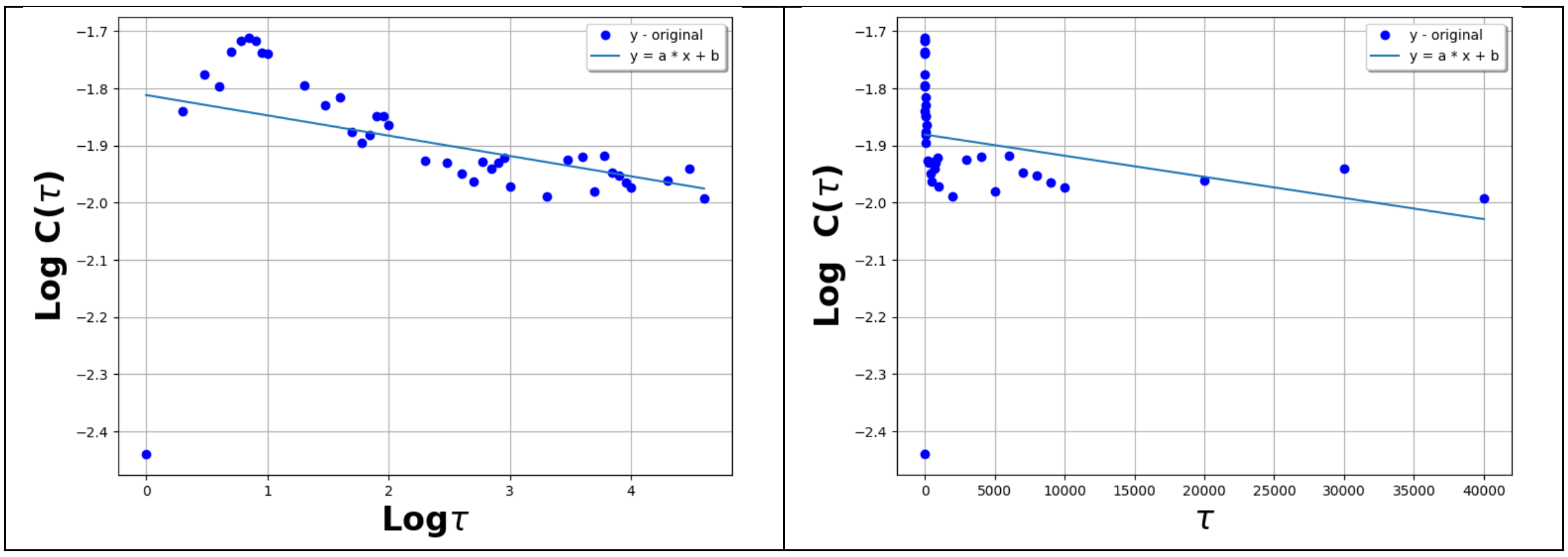

Figure 103: Best fit of GloVe computation of autocorrelations The Iliad in German by power (left) and exp (right)

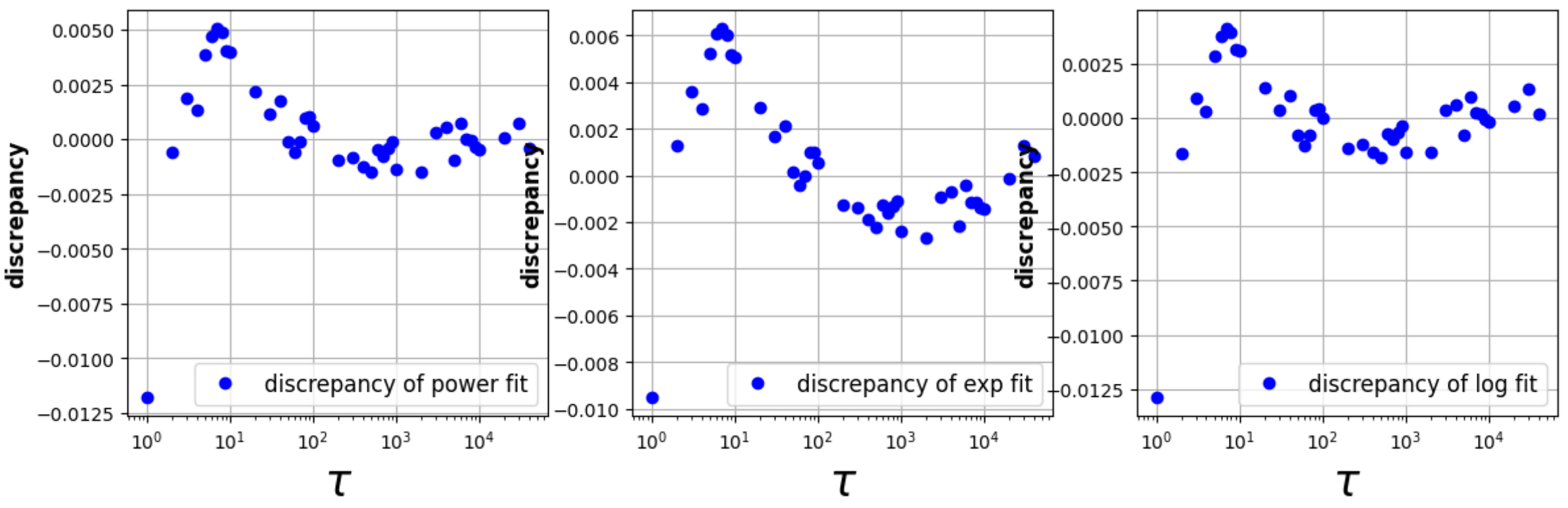

Figure 104: Residual graphs of the best fit of GloVe computation of autocorrelations in The Iliad in German

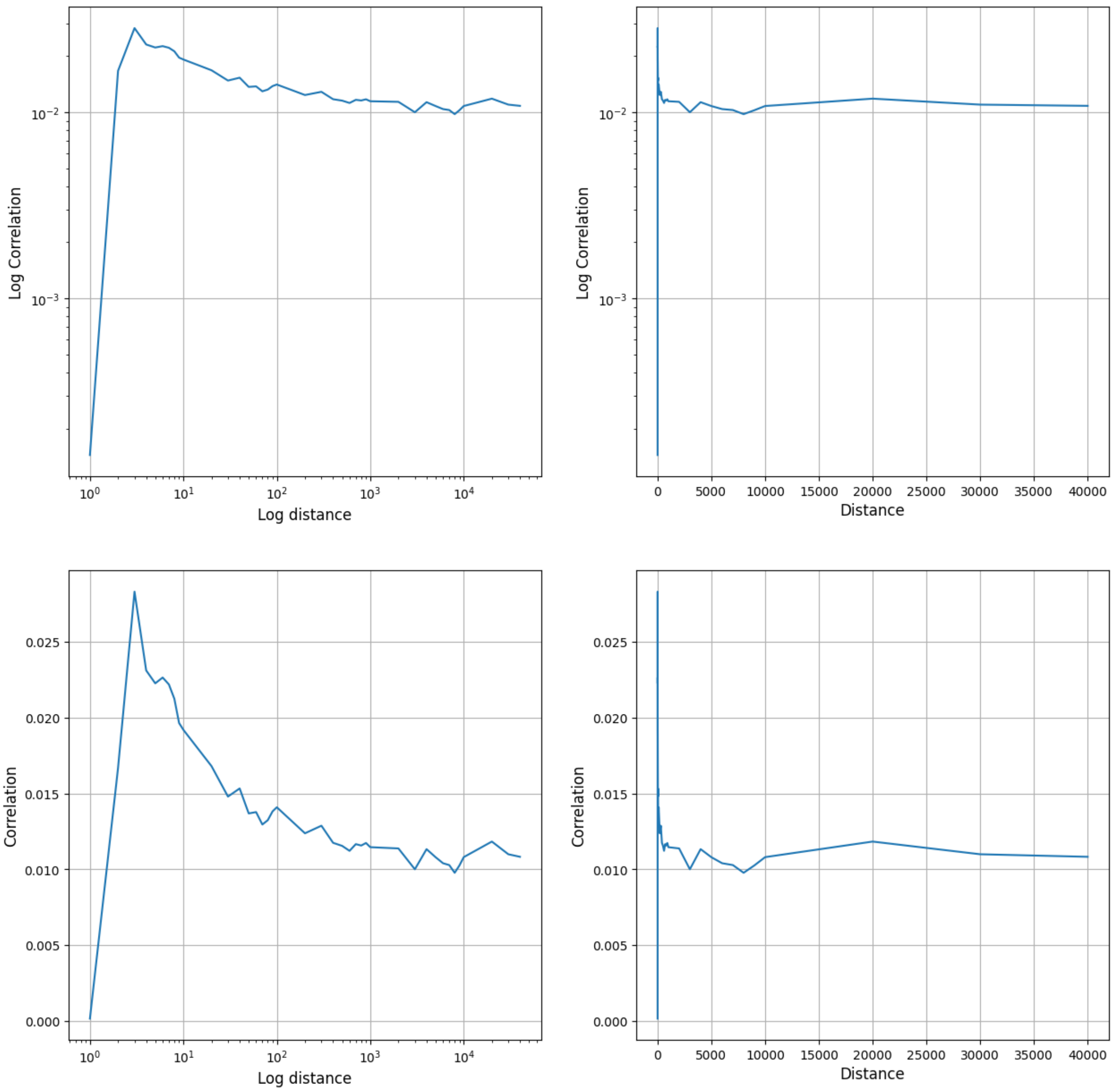

Figure 105: GloVe computation of autocorrelations in The Iliad in French in different coordinates

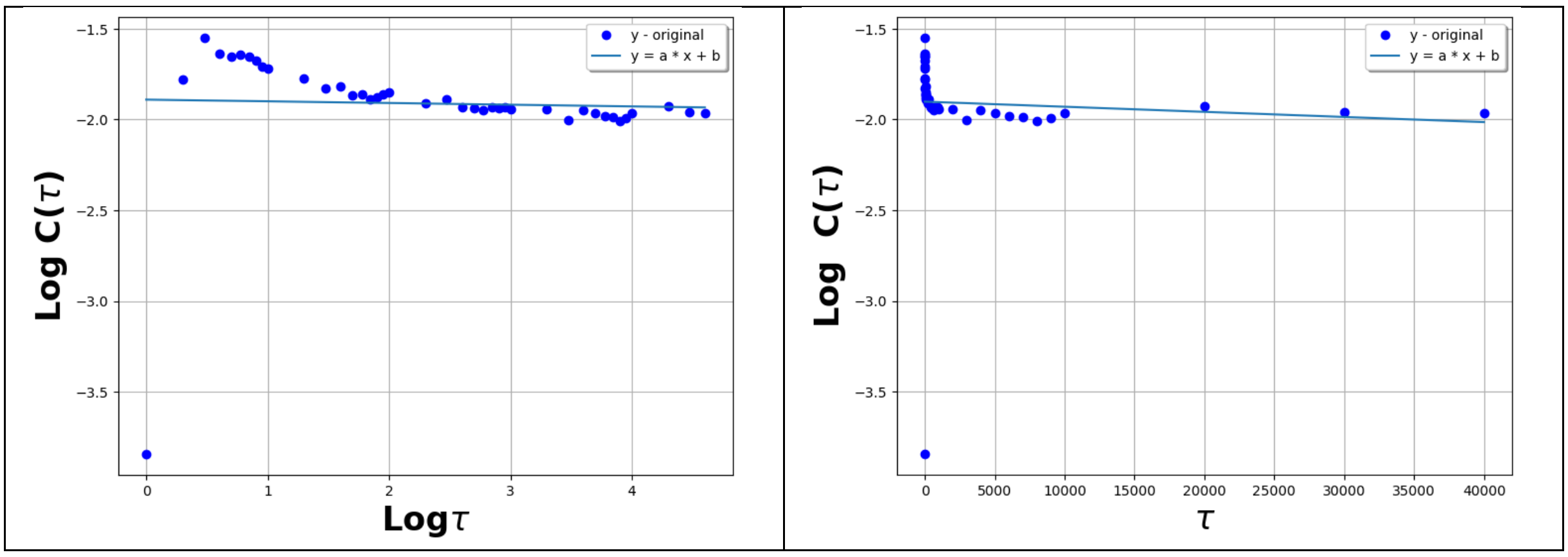

Figure 106: Best fit of GloVe computation of autocorrelations The Iliad in French by power (left) and exp (right)

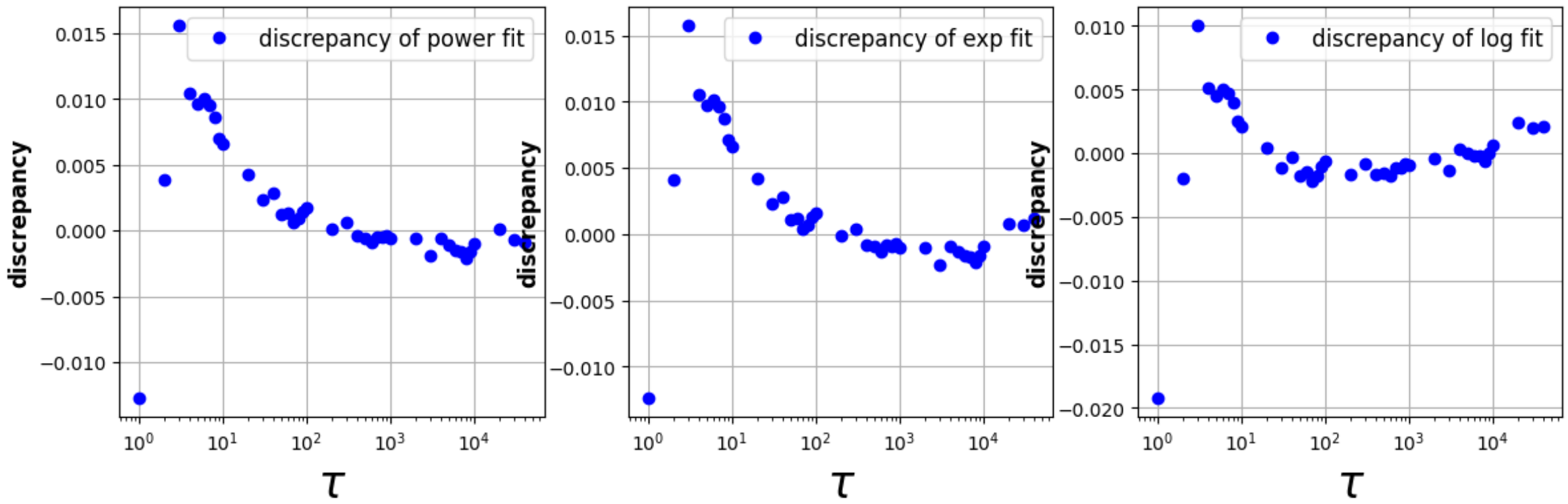

Figure 107: Residual graphs of the best fit of GloVe computation of autocorrelations in The Iliad in French

**Appendix B**

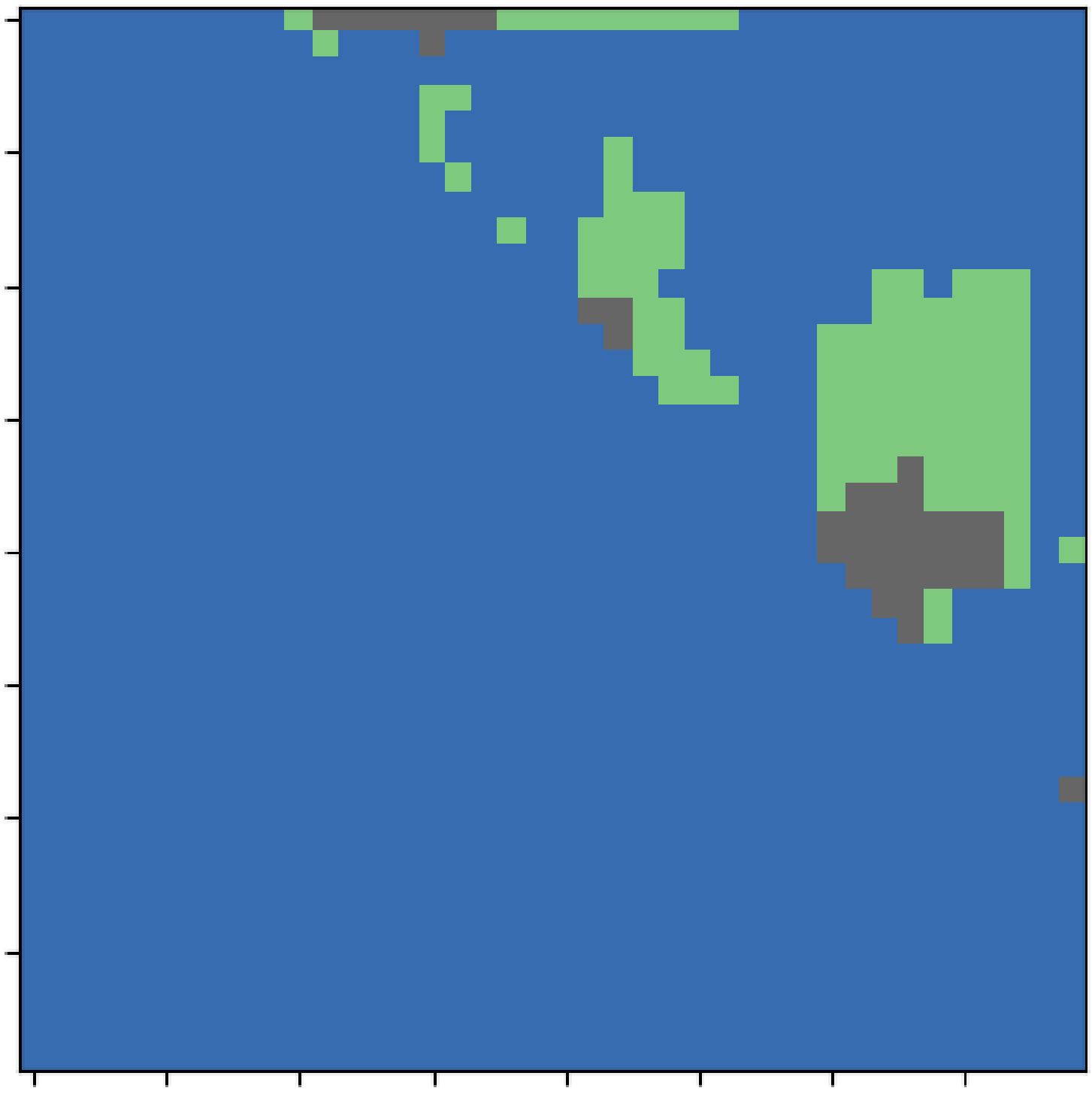

Figure 108: Ranges where power (blue), exp (gray), and log (green) functions are the best approximations to decay law of the autocorrelations in Republic in Spanish computed using GloVe, $a = 1, d = 300$. Vertical axis: start of $\tau$ range. Horizontal axis: end of $\tau$ range.

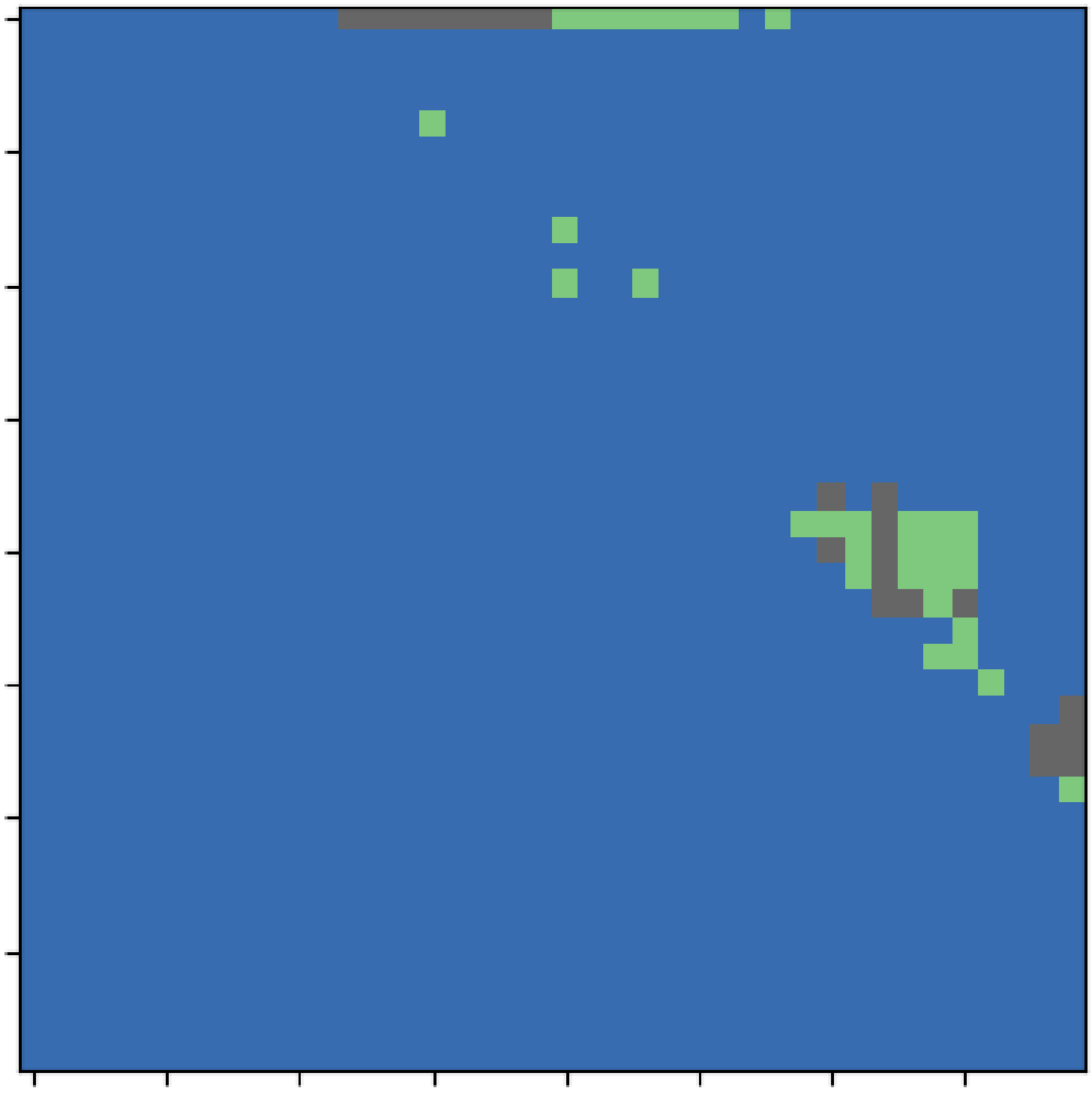

Figure 109: Ranges where power (blue), exp (gray), and log (green) functions are the best approximations to decay law of the autocorrelations in Republic in French computed using GloVe, $a = 1, d = 300$. Vertical axis: start of $\tau$ range. Horizontal axis: end of $\tau$ range.

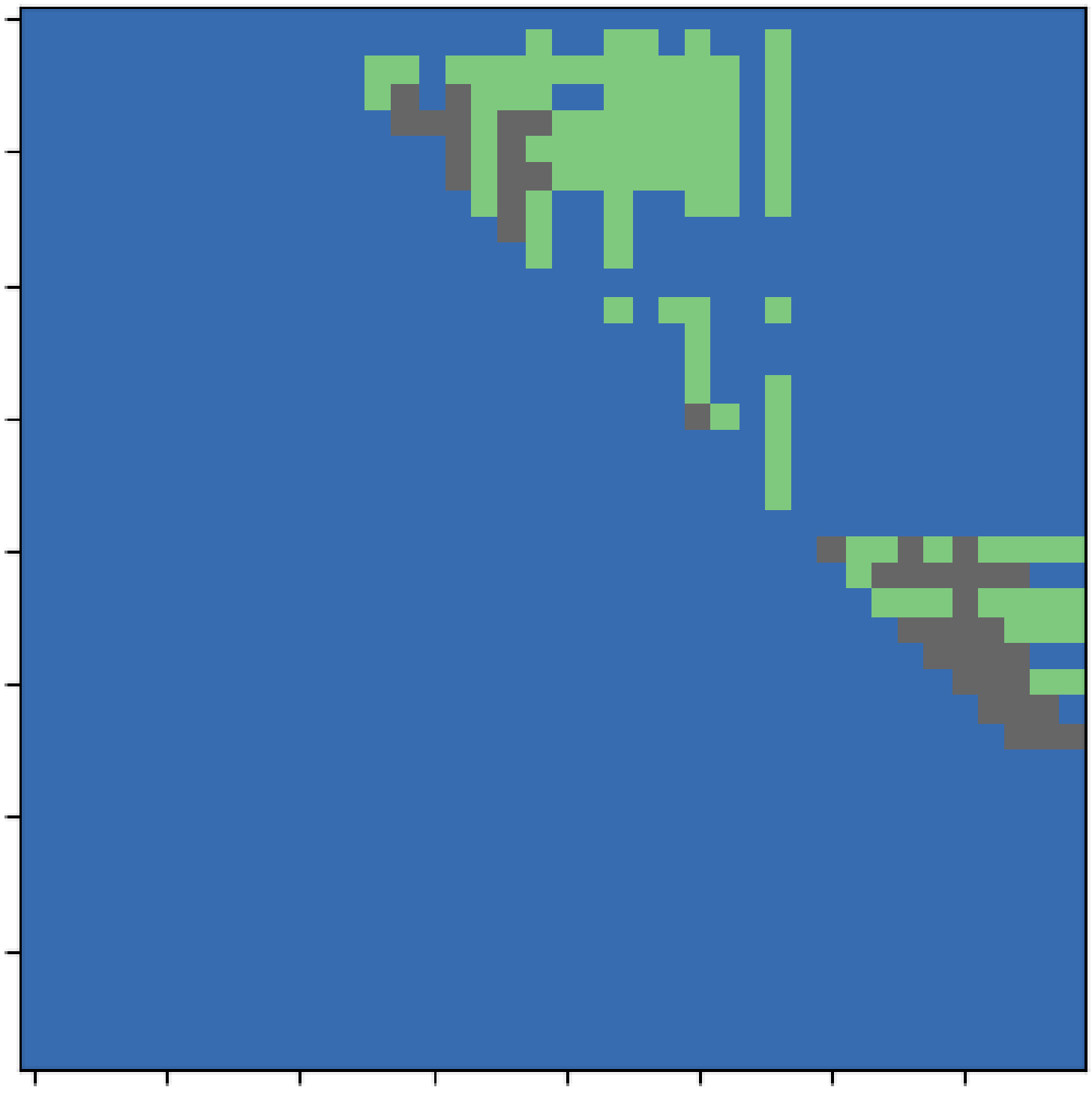

Figure 110: Ranges where power (blue), exp (gray), and log (green) functions are the best approximations to decay law of the autocorrelations in Republic in Russian computed using GloVe, $a = 1, d = 300$. Vertical axis: start of $\tau$ range. Horizontal axis: end of $\tau$ range.

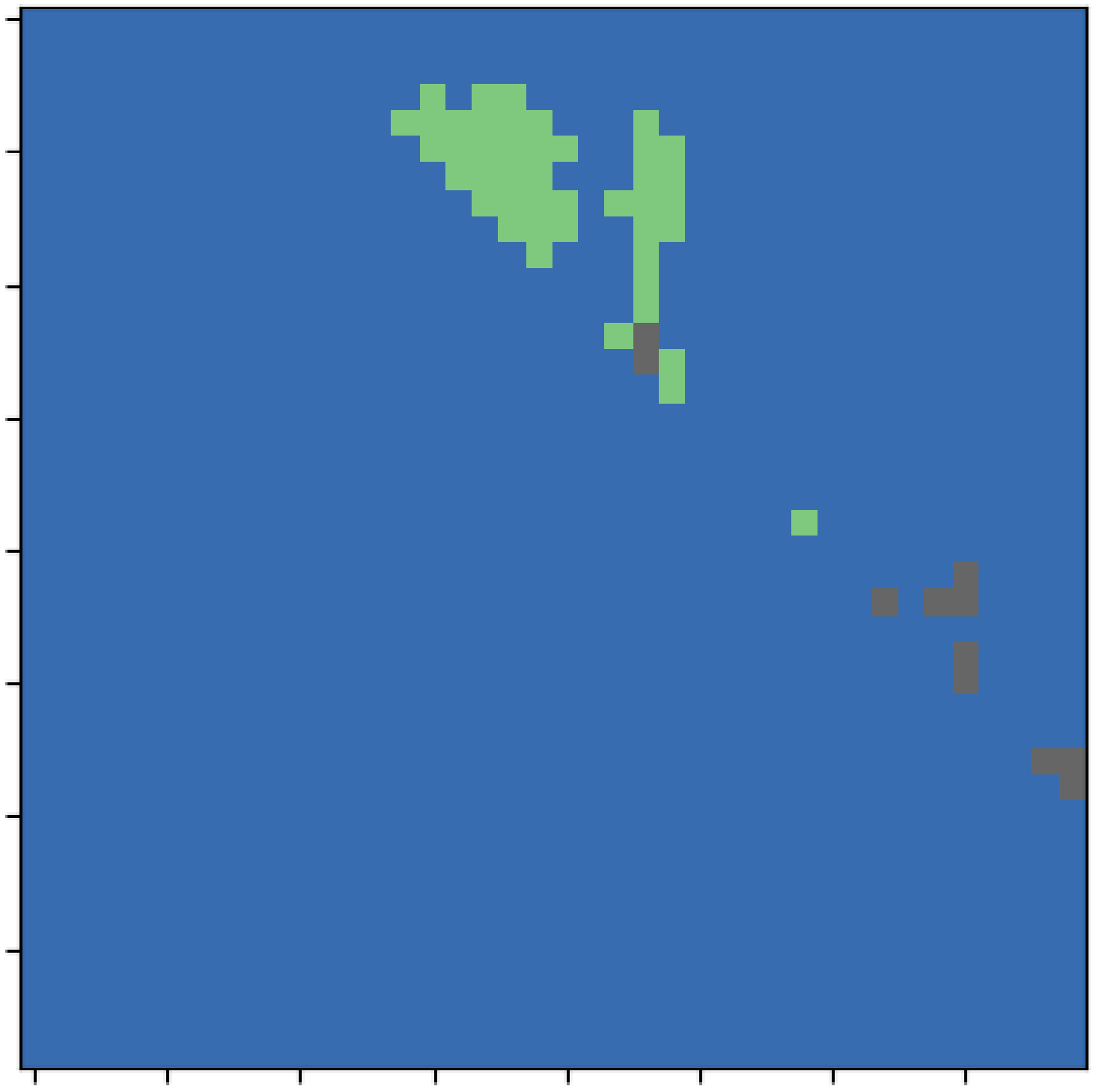

Figure 111: Ranges where power (blue), exp (gray), and log (green) functions are the best approximations to decay law of the autocorrelations in Republic in German computed using GloVe, $a = 1, d = 300$. Vertical axis: start of $\tau$ range. Horizontal axis: end of $\tau$ range.

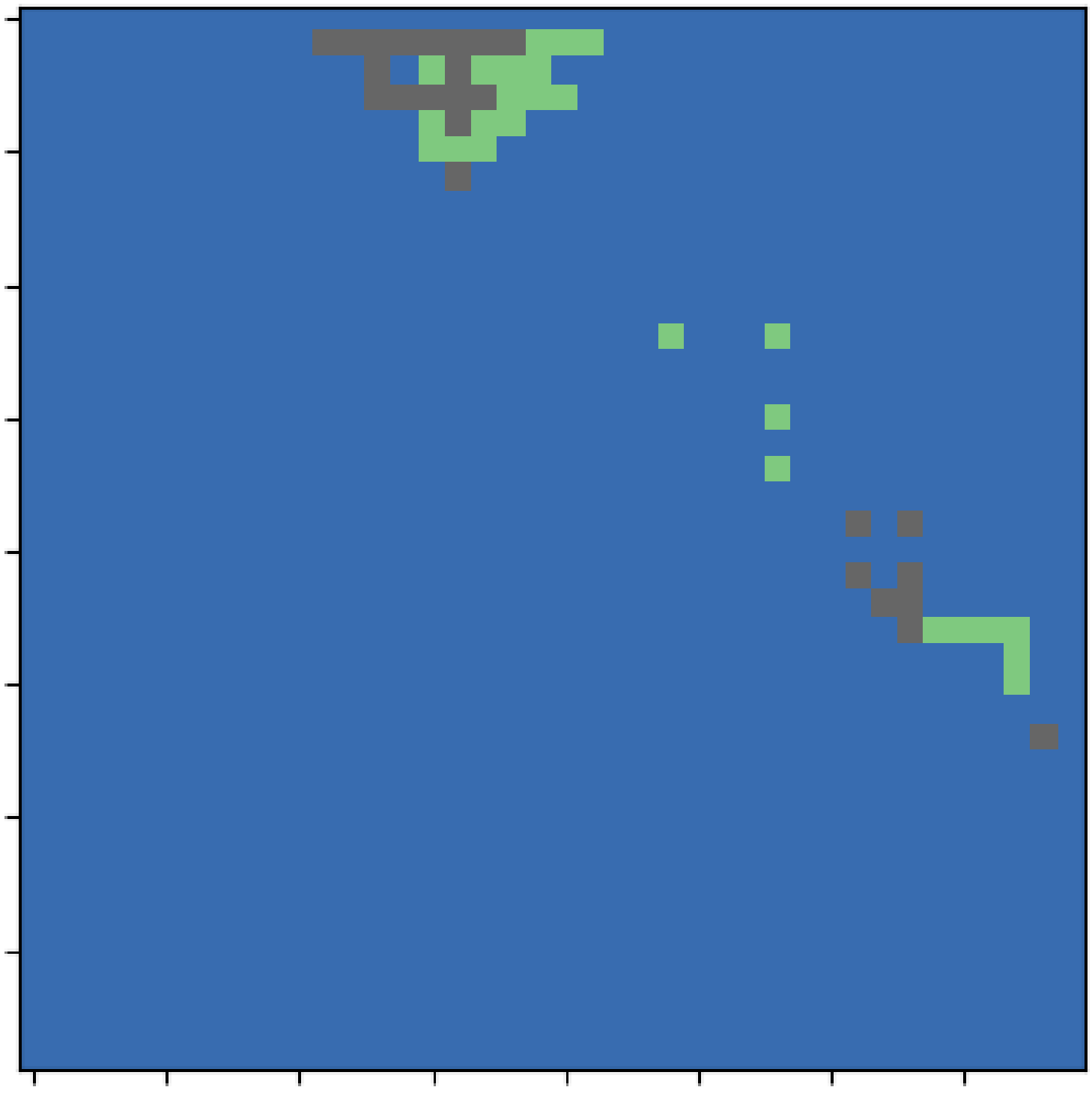

Figure 112: Ranges where power (blue), exp (gray), and log (green) functions are the best approximations to decay law of the autocorrelations in Republic in English computed using GloVe, $a = 1, d = 300$. Vertical axis: start of $\tau$ range. Horizontal axis: end of $\tau$ range.

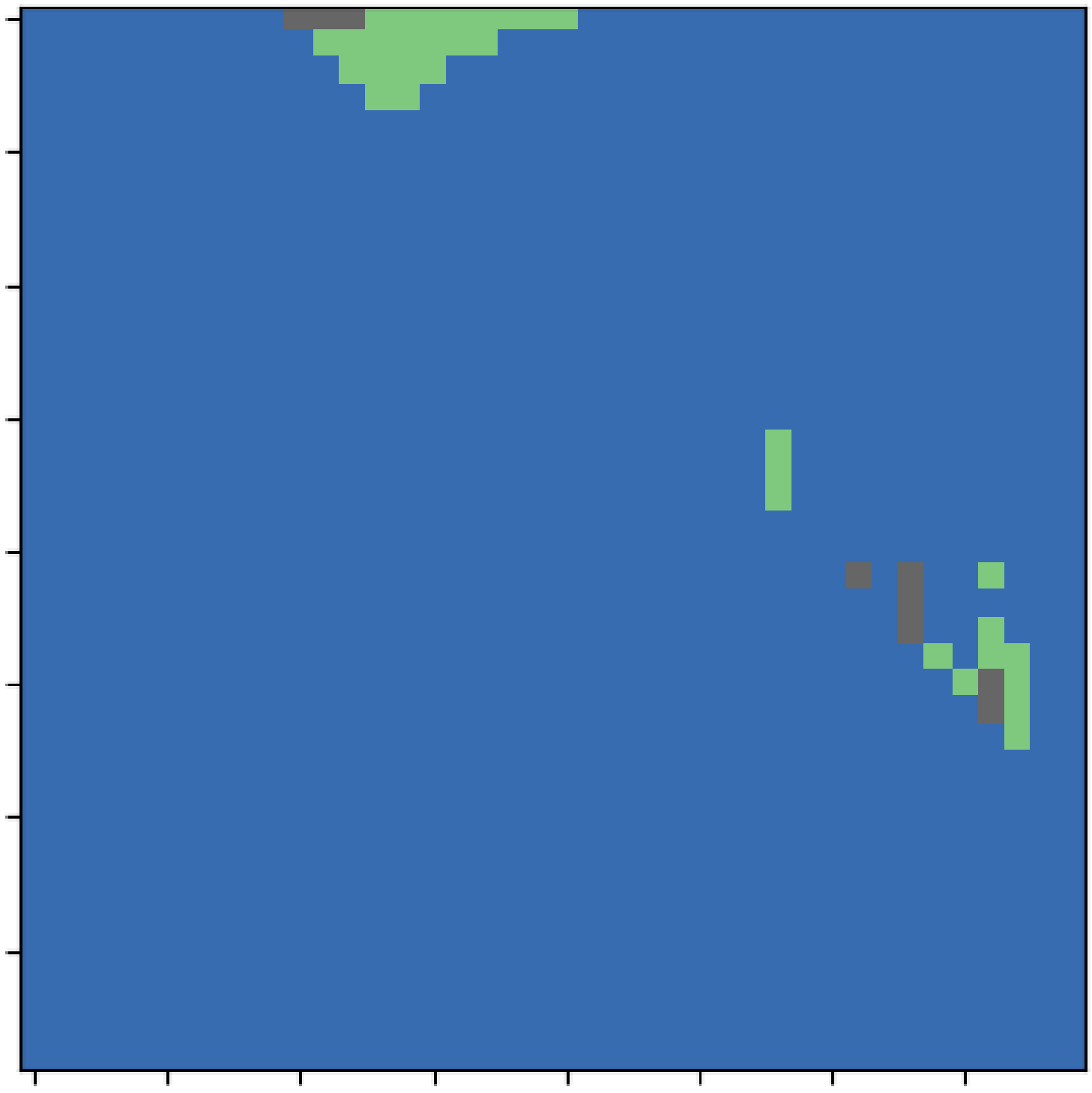

Figure 113: Ranges where power (blue), exp (gray), and log (green) functions are the best approximations to decay law of the autocorrelations in War and Peace in Russian computed using GloVe, $a = 1, d = 300$. Vertical axis: start of $\tau$ range. Horizontal axis: end of $\tau$ range.

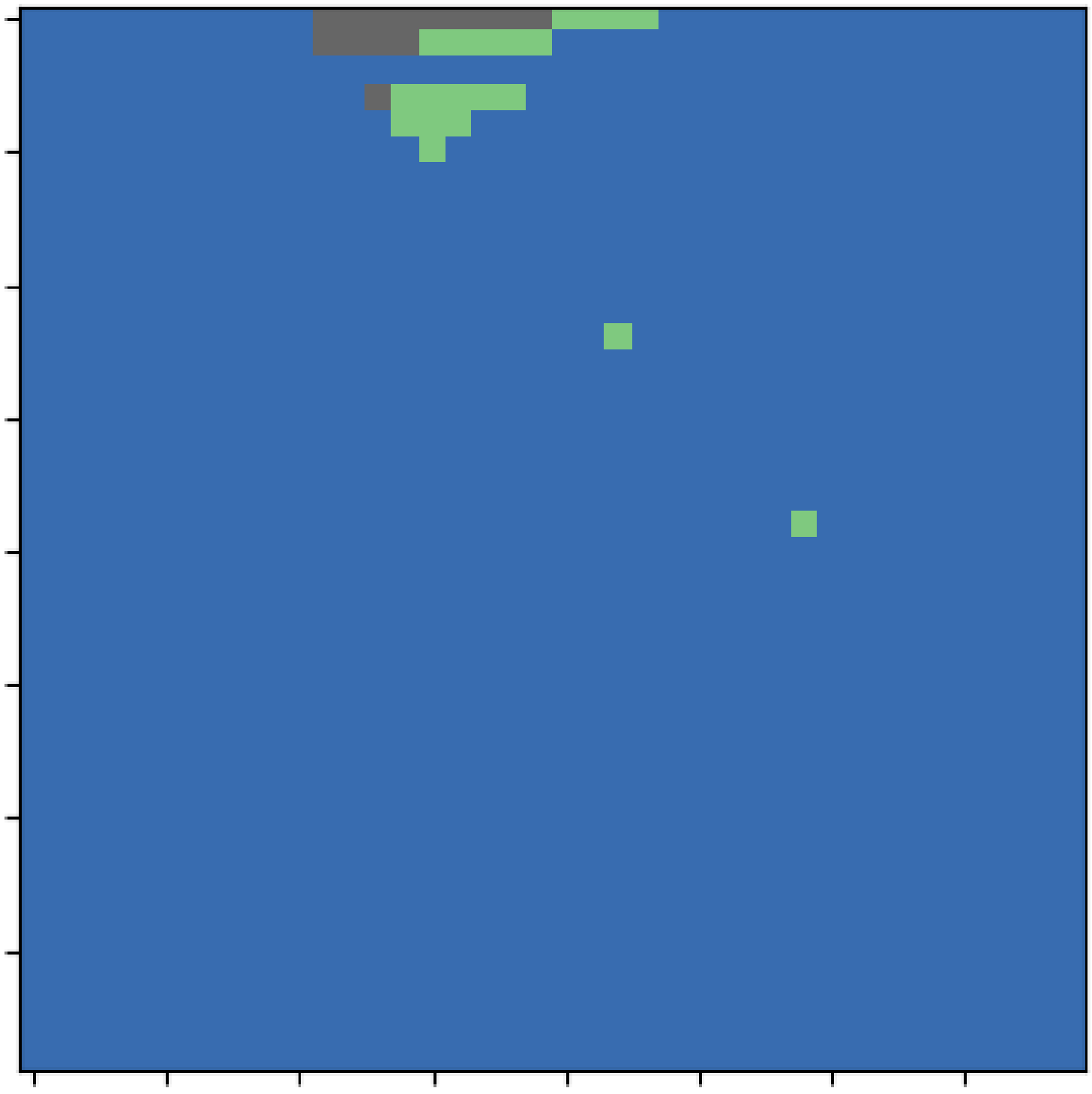

Figure 114: Ranges where power (blue), exp (gray), and log (green) functions are the best approximations to decay law of the autocorrelations in War and Peace in English computed using GloVe, $a = 1, d = 300$. Vertical axis: start of $\tau$ range. Horizontal axis: end of $\tau$ range.

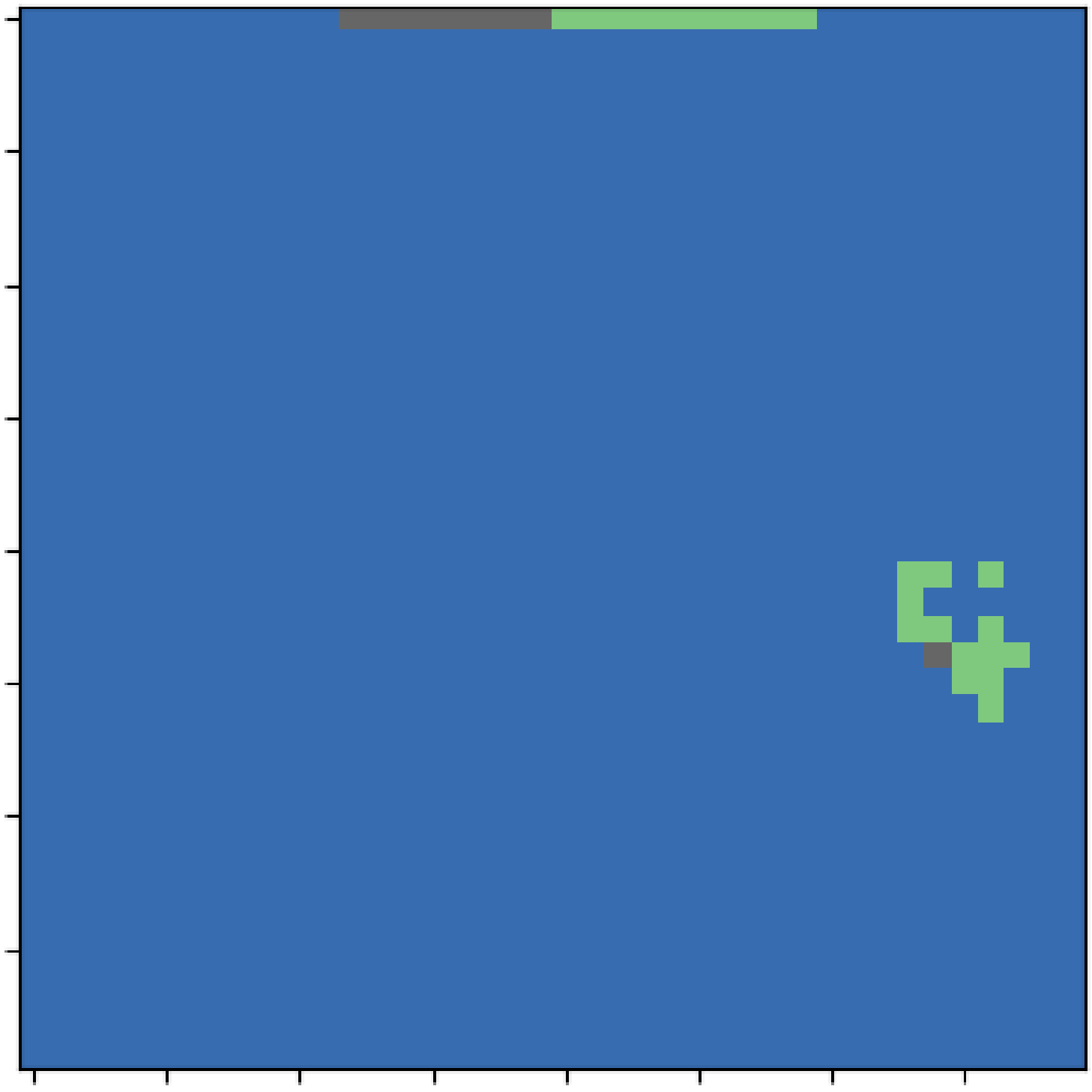

Figure 115: Ranges where power (blue), exp (gray), and log (green) functions are the best approximations to decay law of the autocorrelations in War and Peace in French computed using GloVe, $a = 1, d = 300$. Vertical axis: start of $\tau$ range. Horizontal axis: end of $\tau$ range.

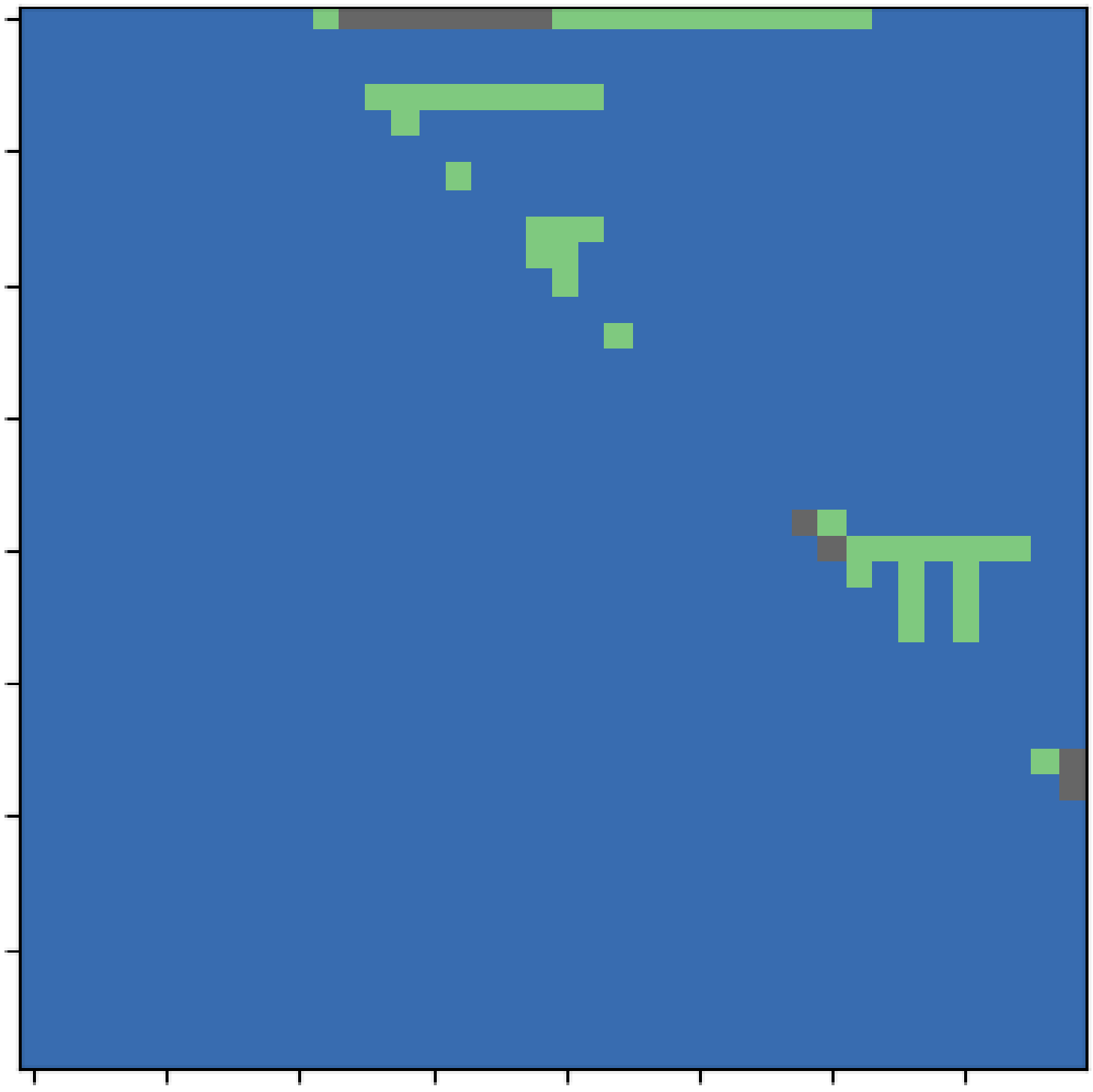

Figure 116: Ranges where power (blue), exp (gray), and log (green) functions are the best approximations to decay law of the autocorrelations in War and Peace in Spanish computed using GloVe, $a = 1, d = 300$. Vertical axis: start of $\tau$ range. Horizontal axis: end of $\tau$ range.

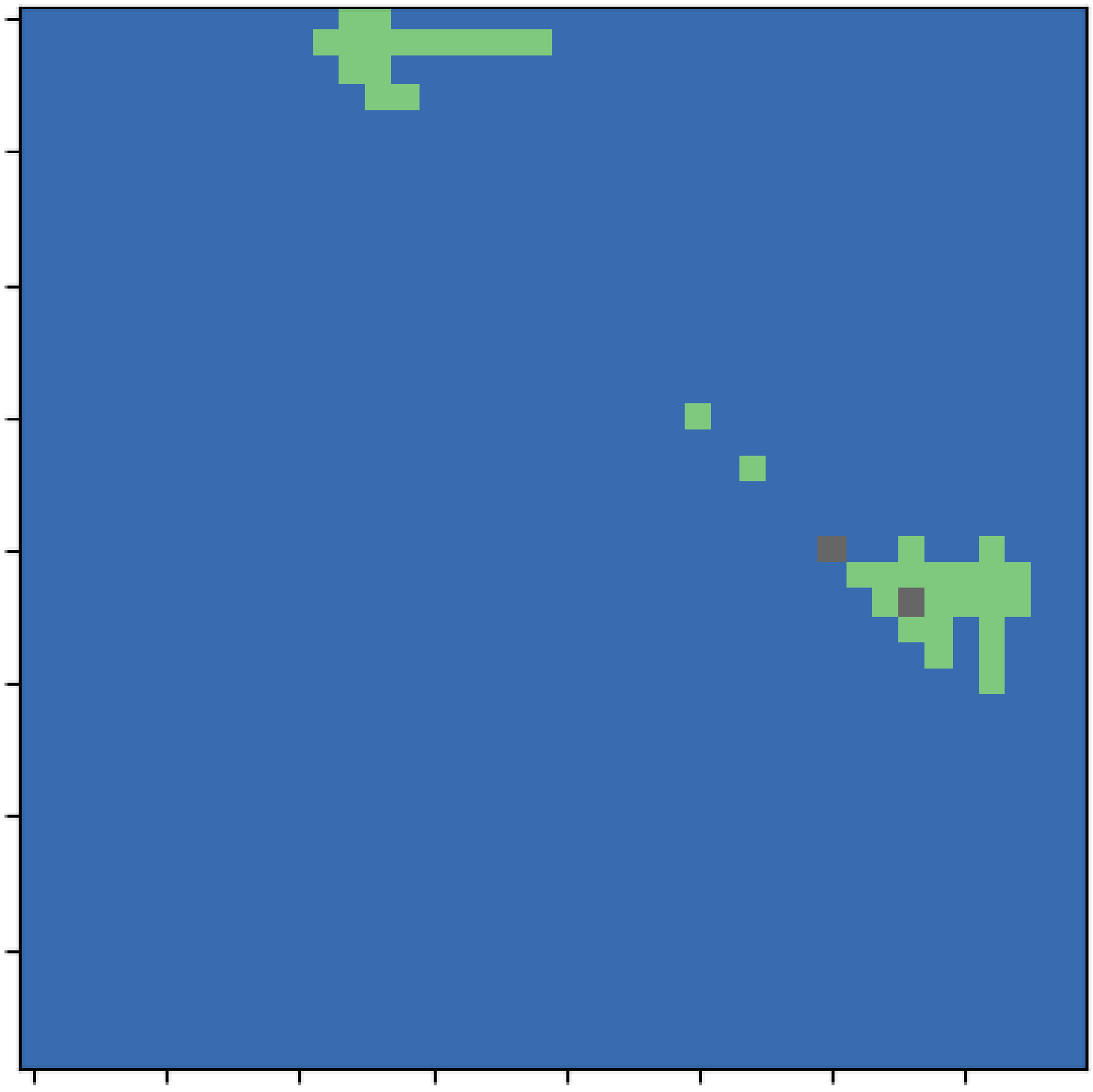

Figure 117: Ranges where power (blue), exp (gray), and log (green) functions are the best approximations to decay law of the autocorrelations in War and Peace in German computed using GloVe, $a = 1, d = 300$. Vertical axis: start of $\tau$ range. Horizontal axis: end of $\tau$ range.

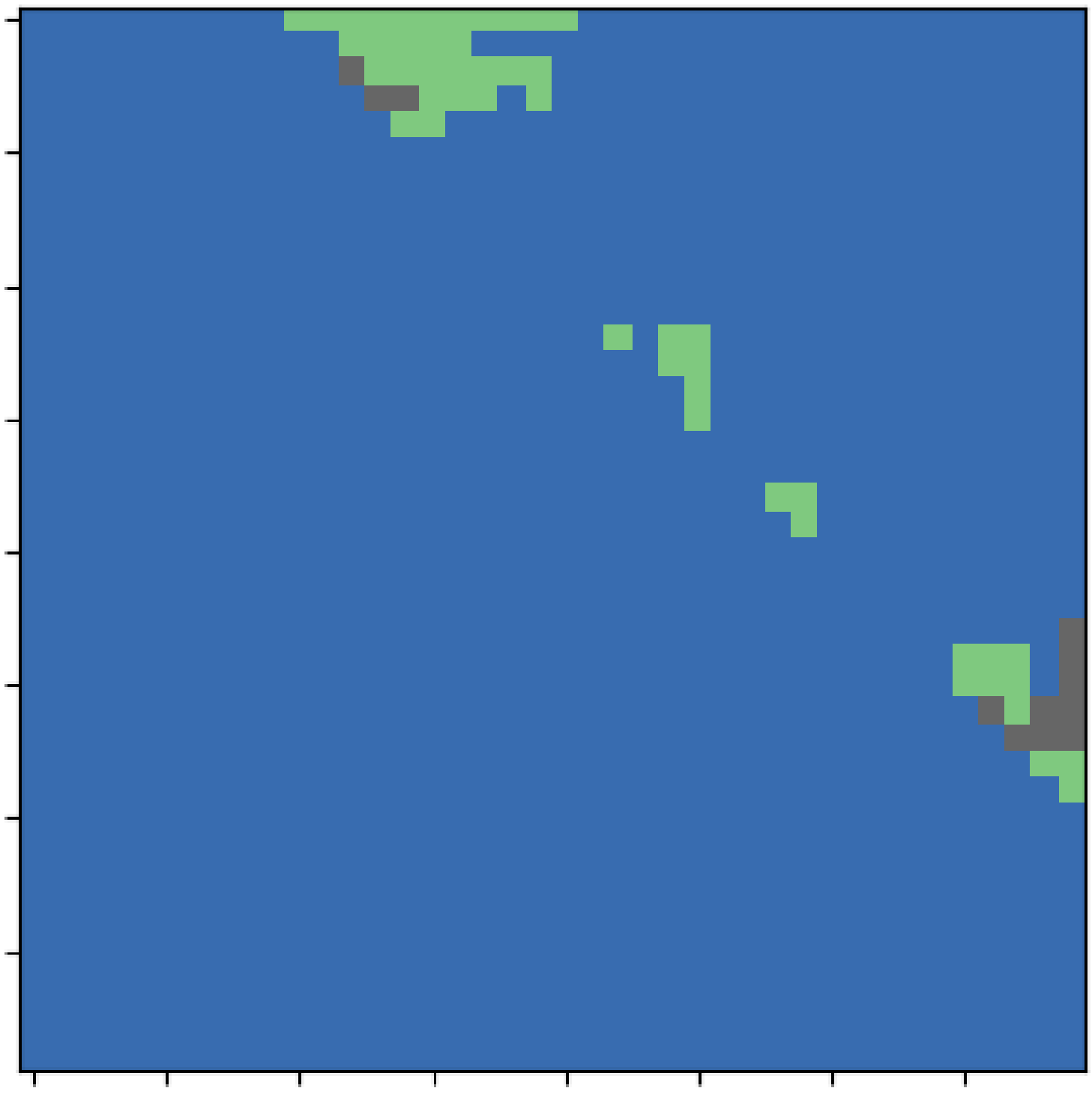

Figure 118: Ranges where power (blue), exp (gray), and log (green) functions are the best approximations to decay law of the autocorrelations in Moby-Dick or, The Whale in Russian computed using GloVe, $a = 1, d = 300$. Vertical axis: start of $\tau$ range. Horizontal axis: end of $\tau$ range.

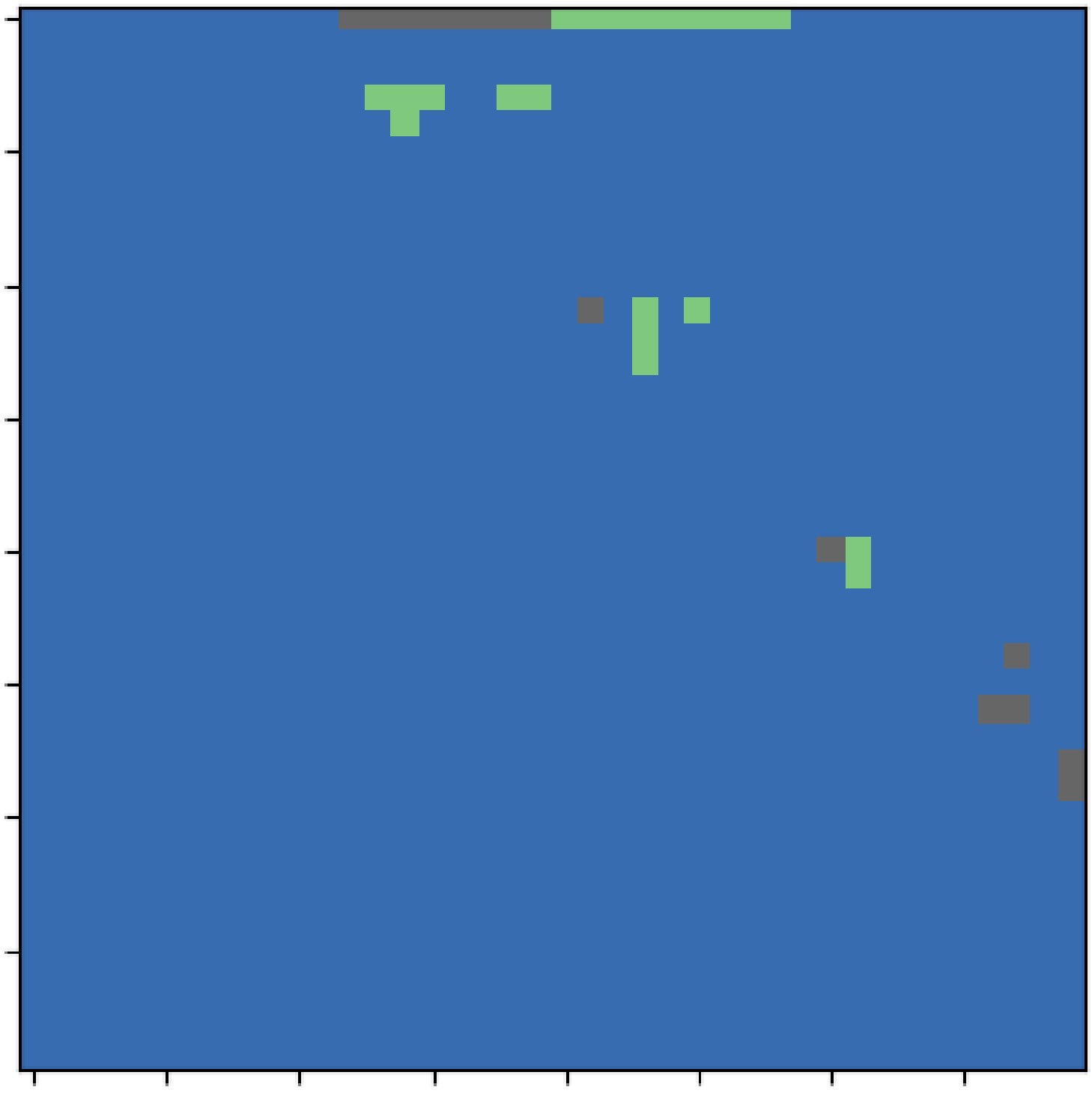

Figure 119: Ranges where power (blue), exp (gray), and log (green) functions are the best approximations to decay law of the autocorrelations in Moby-Dick or, The Whale in French computed using GloVe, $a = 1, d = 300$. Vertical axis: start of $\tau$ range. Horizontal axis: end of $\tau$ range.

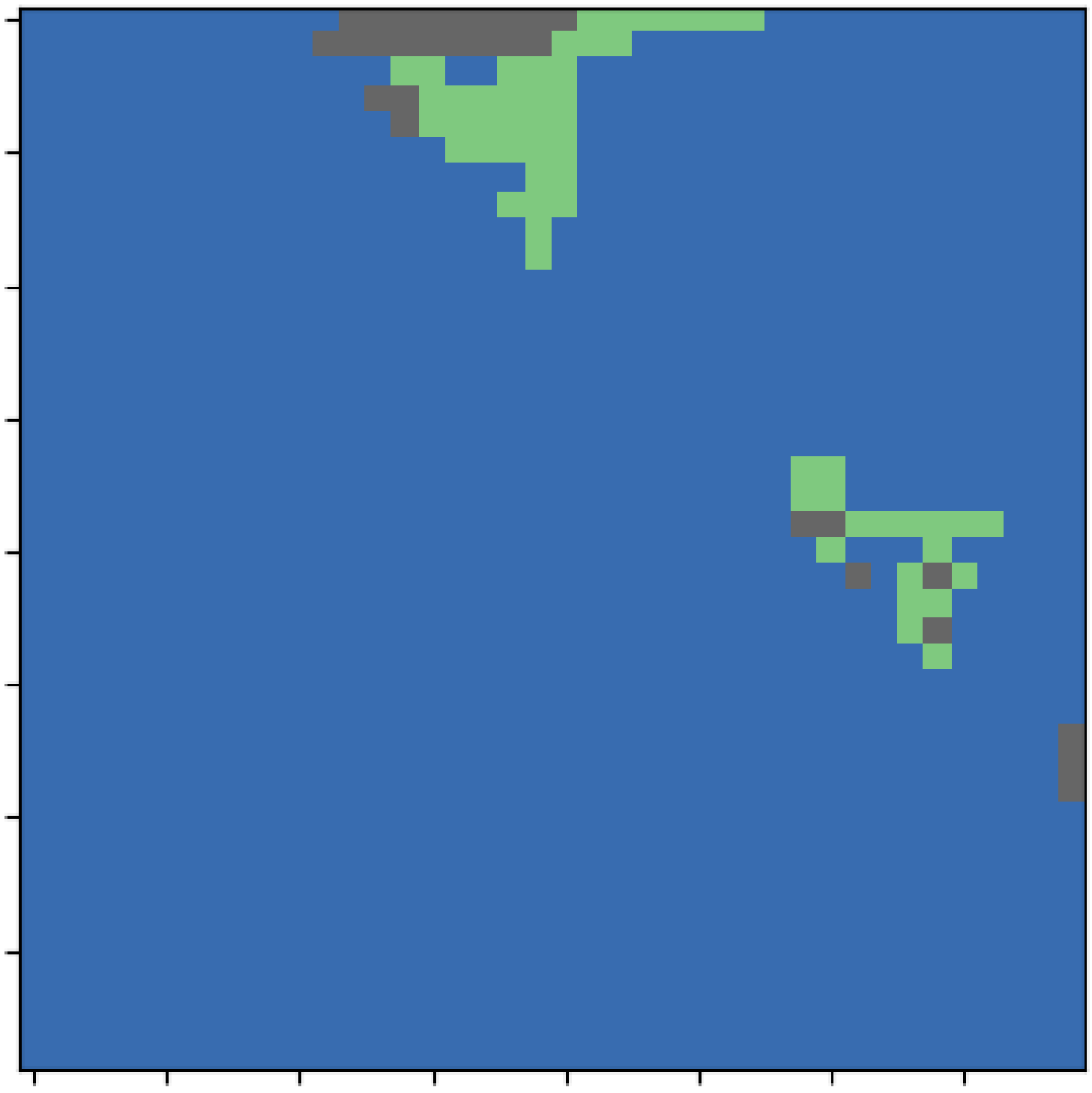

Figure 120: Ranges where power (blue), exp (gray), and log (green) functions are the best approximations to decay law of the autocorrelations in Moby-Dick or, The Whale in English computed using GloVe, $a = 1, d = 300$. Vertical axis: start of $\tau$ range. Horizontal axis: end of $\tau$ range.

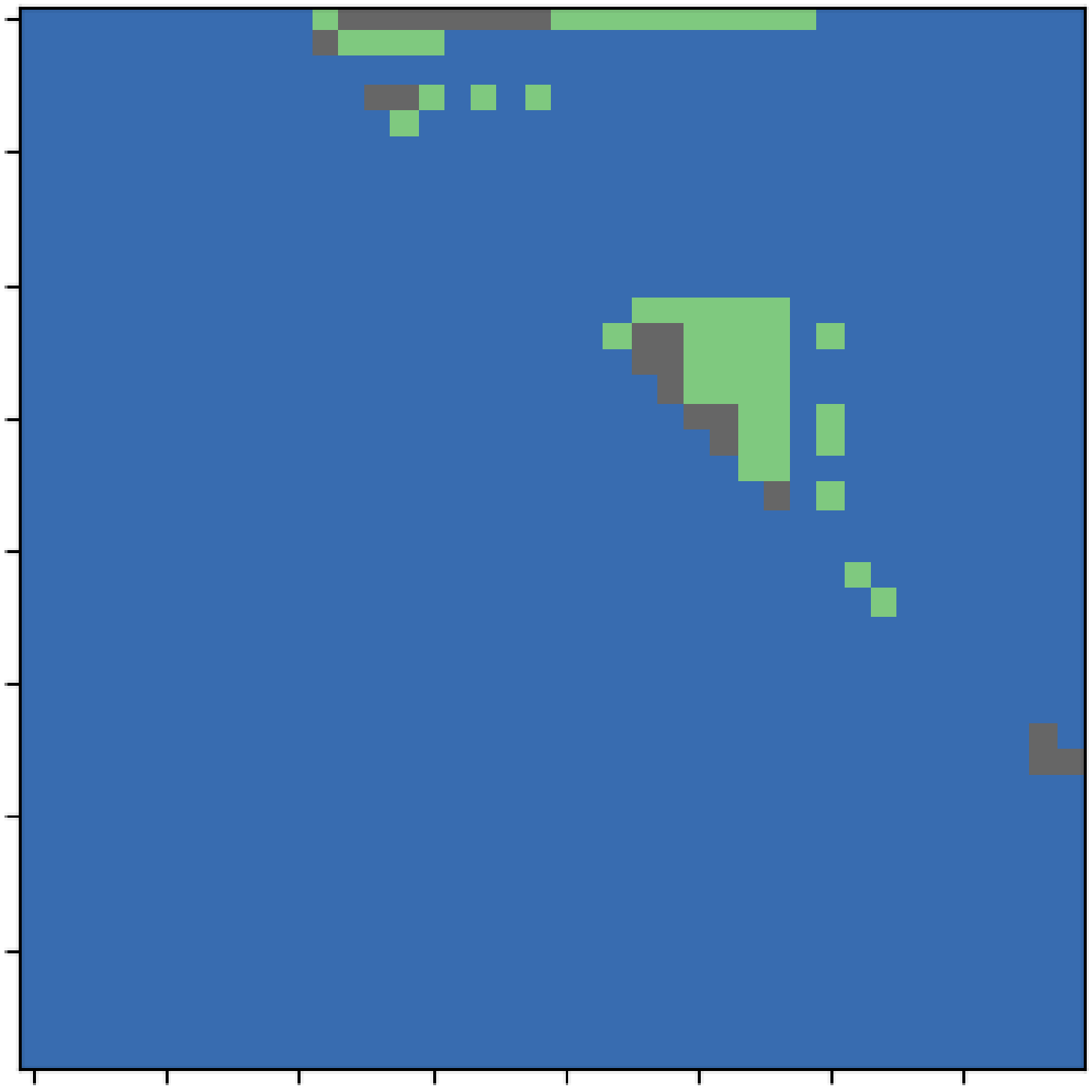

Figure 121: Ranges where power (blue), exp (gray), and log (green) functions are the best approximations to decay law of the autocorrelations in Moby-Dick or, The Whale in Spanish computed using GloVe, $a = 1, d = 300$. Vertical axis: start of $\tau$ range. Horizontal axis: end of $\tau$ range.

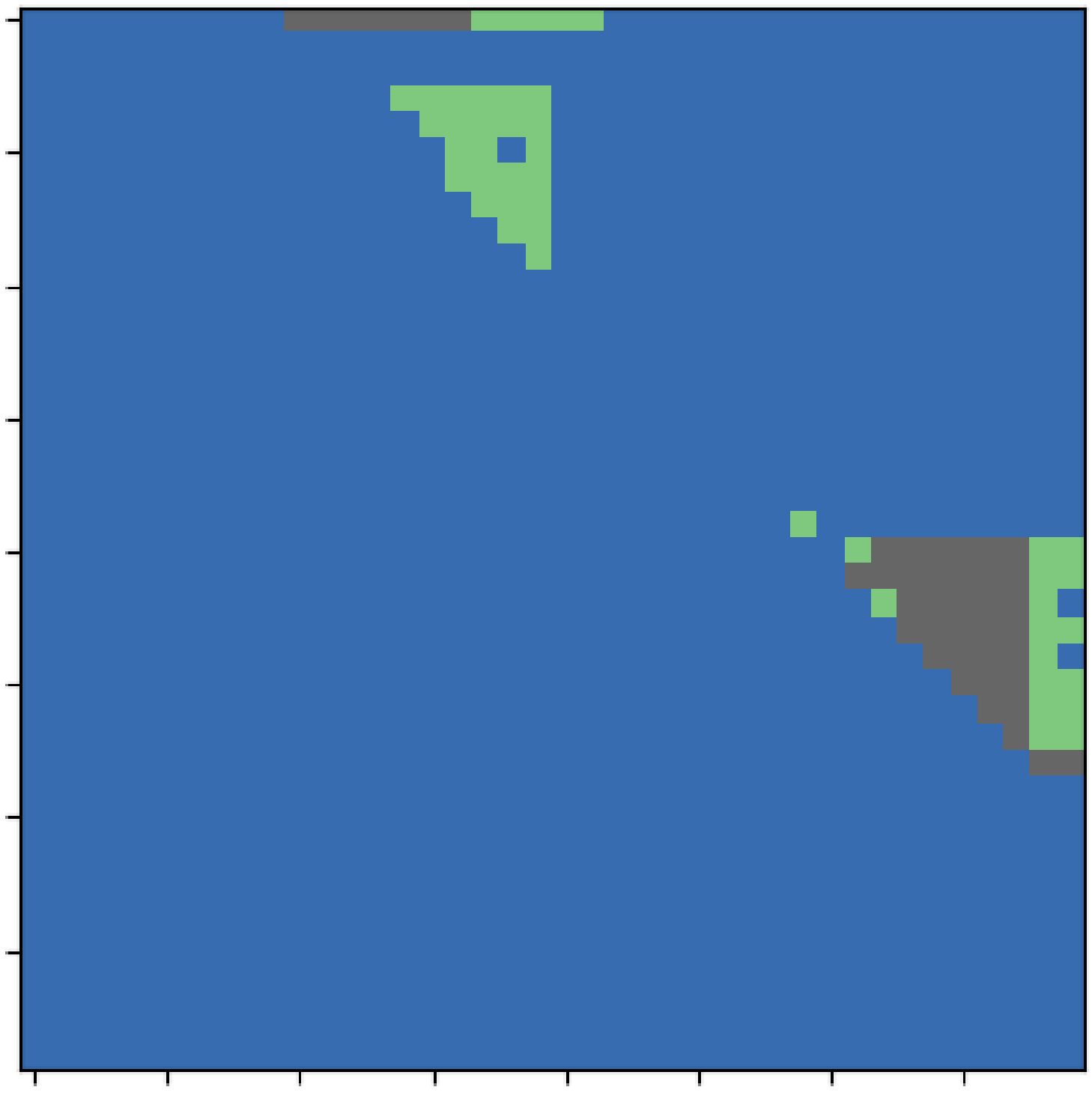

Figure 122 : Ranges where power (blue), exp (gray), and log (green) functions are the best approximations to decay law of the autocorrelations in Don Quixote de la Mancha in French computed using GloVe, $a = 1, d = 300$. Vertical axis: start of $\tau$ range. Horizontal axis: end of $\tau$ range.

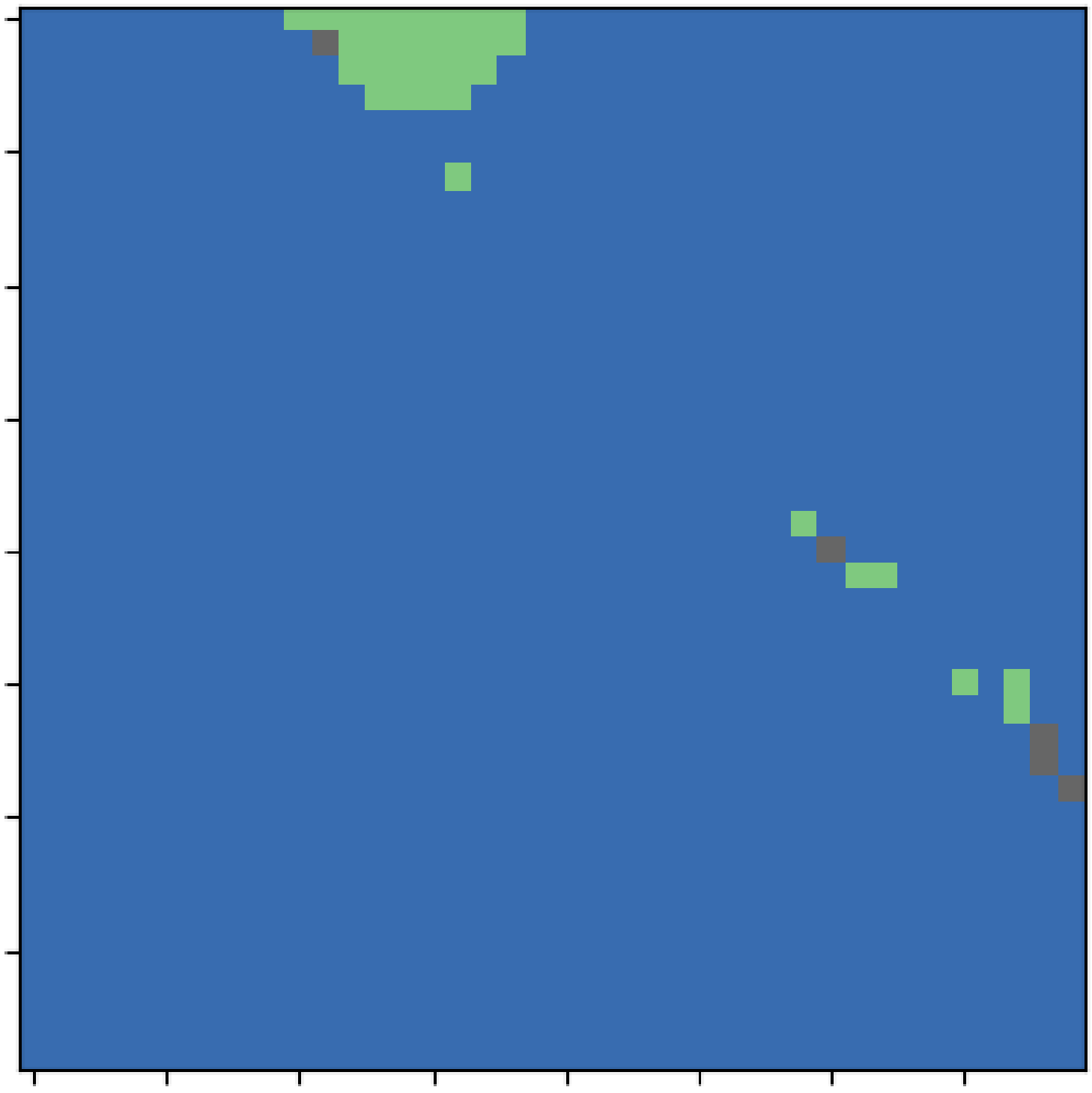

Figure 123: Ranges where power (blue), exp (gray), and log (green) functions are the best approximations to decay law of the autocorrelations in Don Quixote de la Mancha in German computed using GloVe, $a = 1, d = 300$. Vertical axis: start of $\tau$ range. Horizontal axis: end of $\tau$ range.

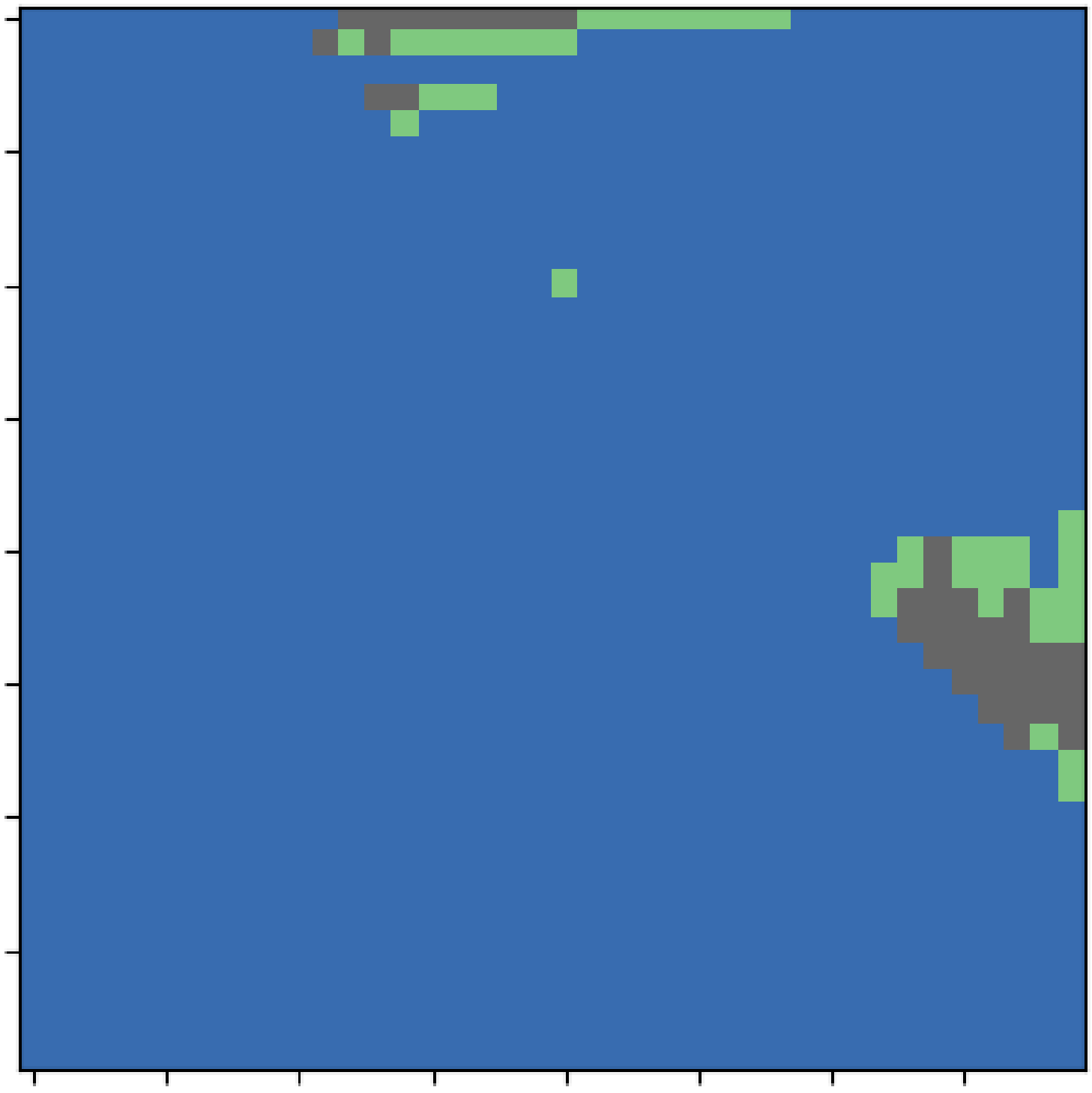

Figure 124: Ranges where power (blue), exp (gray), and log (green) functions are the best approximations to decay law of the autocorrelations in Don Quixote de la Mancha in English computed using GloVe, $a = 1, d = 300$. Vertical axis: start of $\tau$ range. Horizontal axis: end of $\tau$ range.

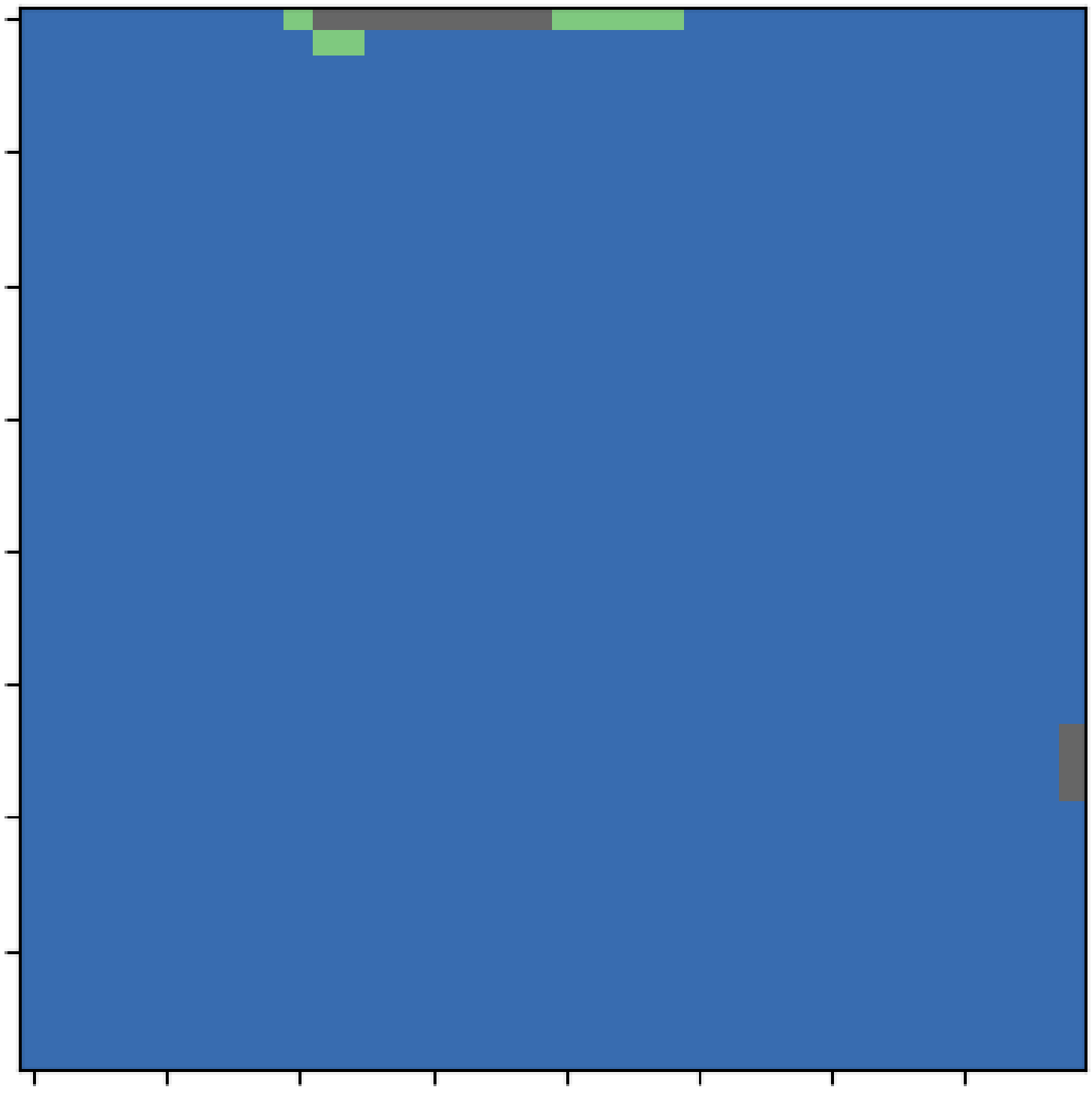

Figure 125: Ranges where power (blue), exp (gray), and log (green) functions are the best approximations to decay law of the autocorrelations in Don Quixote de la Mancha in Spanish computed using GloVe, $a = 1, d = 300$. Vertical axis: start of $\tau$ range. Horizontal axis: end of $\tau$ range.

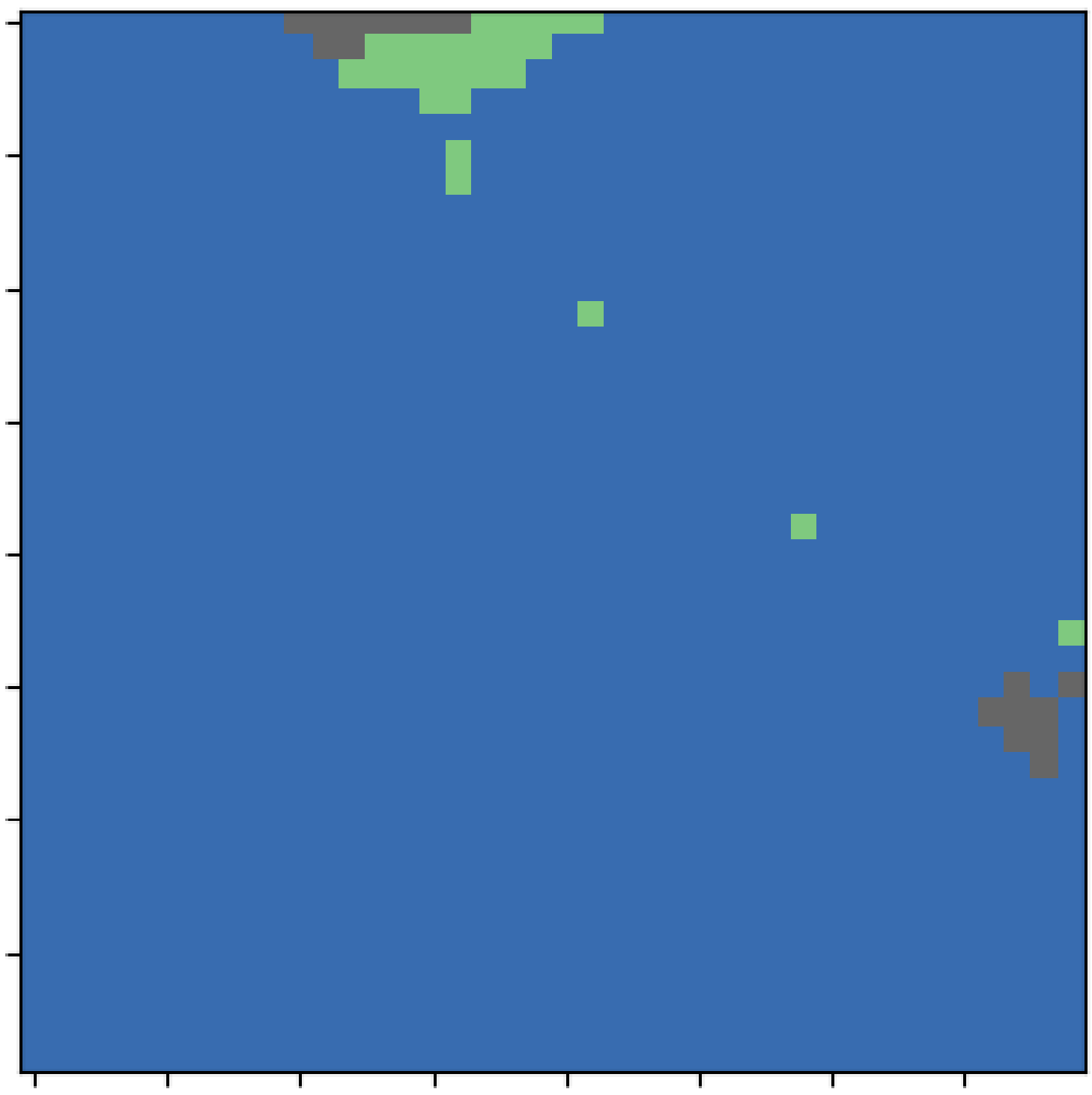

Figure 126: Ranges where power (blue), exp (gray), and log (green) functions are the best approximations to decay law of the autocorrelations in Don Quixote de la Mancha in Russian computed using GloVe, $a = 1, d = 300$. Vertical axis: start of $\tau$ range. Horizontal axis: end of $\tau$ range.

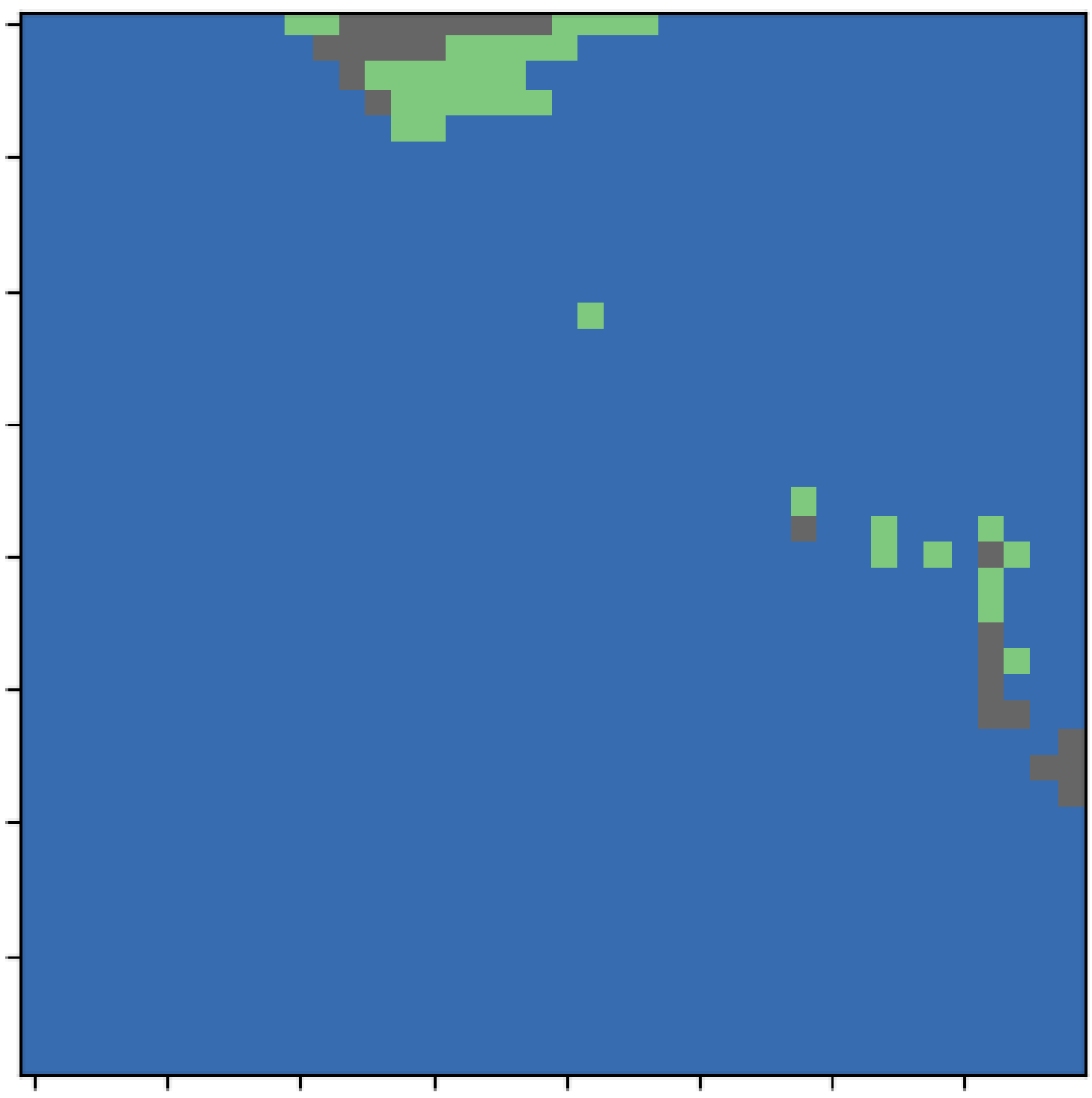

Figure 127: Ranges where power (blue), exp (gray), and log (green) functions are the best approximations to decay law of the autocorrelations The Adventures of Tom Sawyer in English computed using GloVe, $a = 1, d = 300$. Vertical axis: start of $\tau$ range. Horizontal axis: end of $\tau$ range.

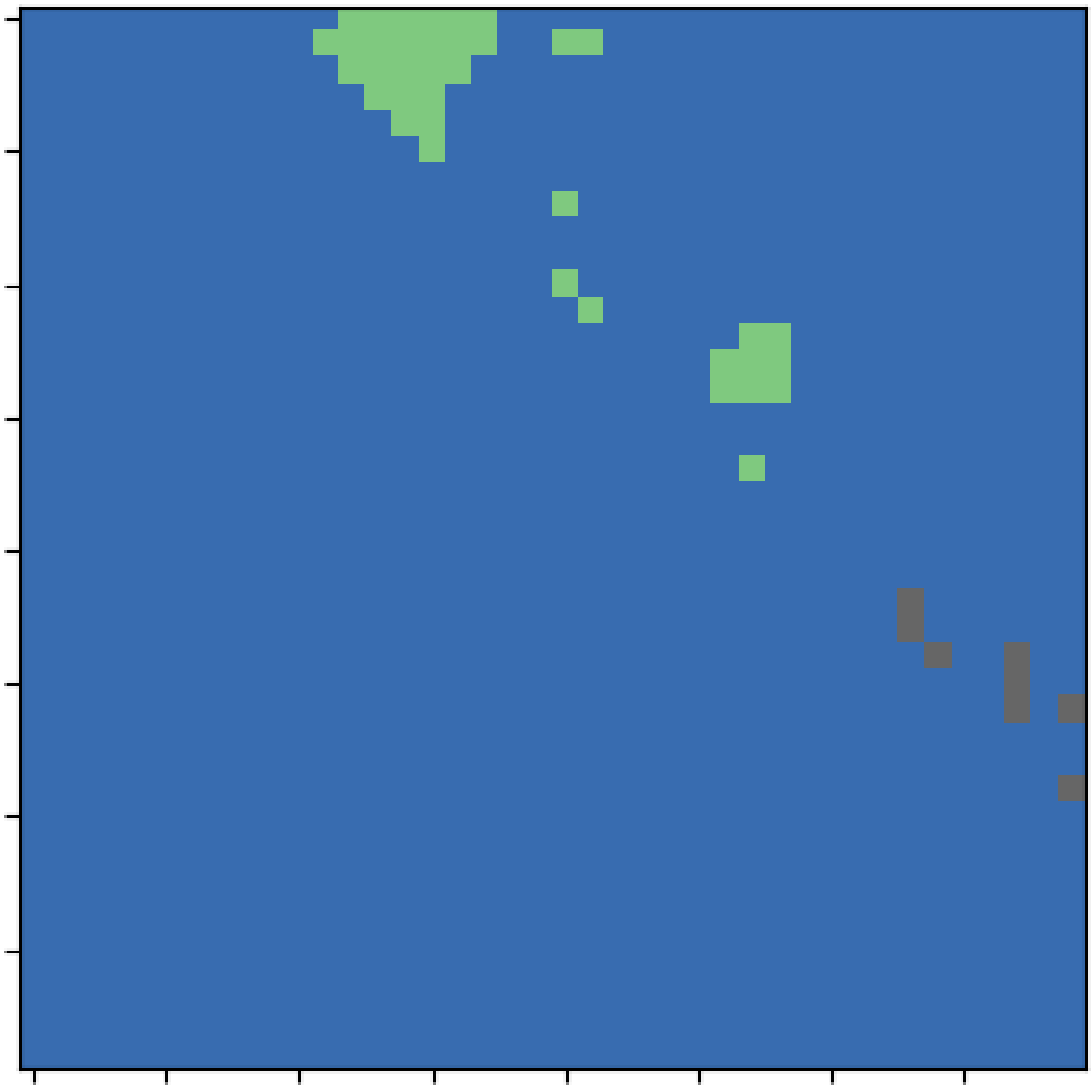

Figure 128: Ranges where power (blue), exp (gray), and log (green) functions are the best approximations to decay law of the autocorrelations The Adventures of Tom Sawyer in German computed using GloVe, $a = 1, d = 300$. Vertical axis: start of $\tau$ range. Horizontal axis: end of $\tau$ range.

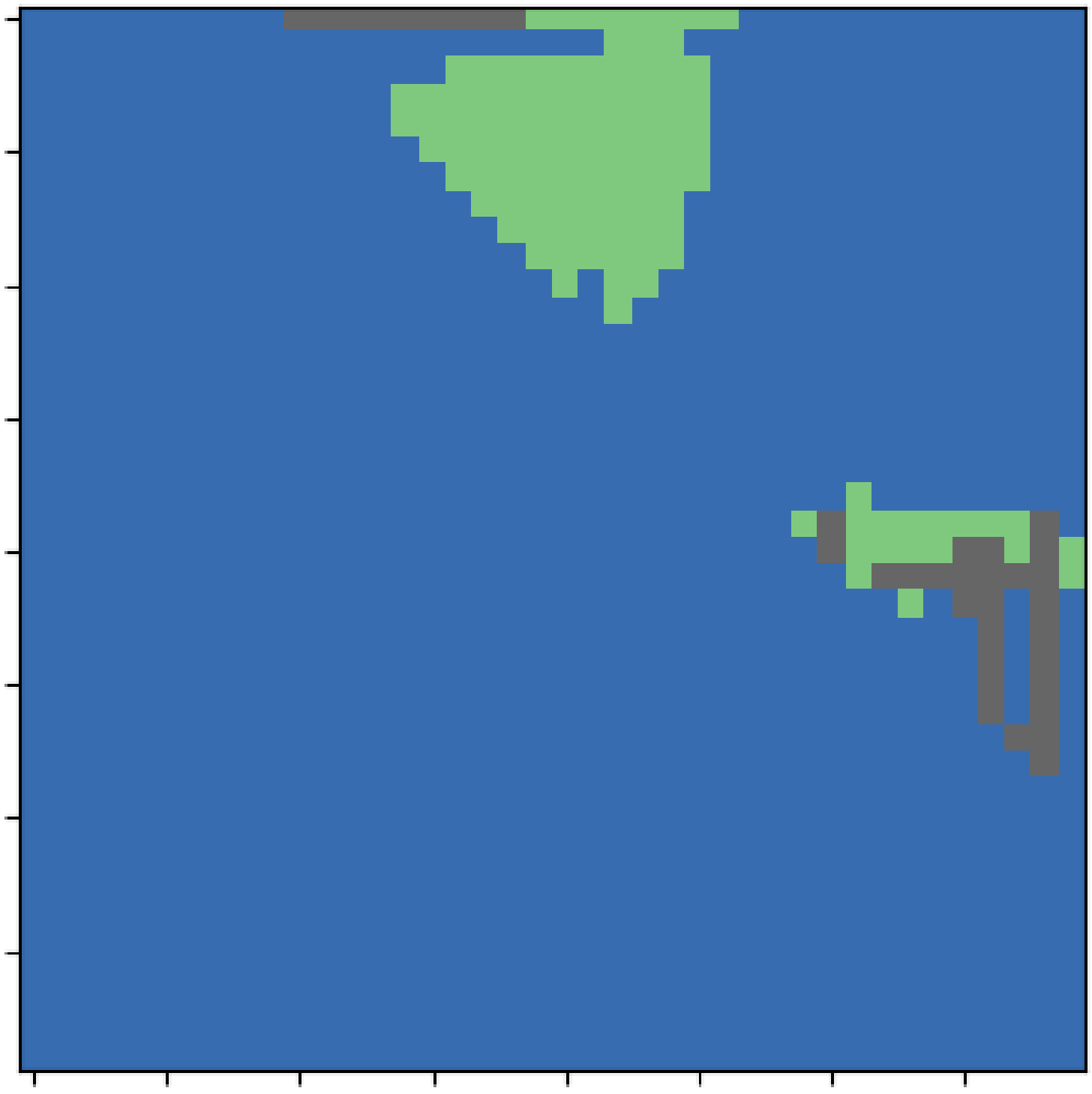

Figure 129: Ranges where power (blue), exp (gray), and log (green) functions are the best approximations to decay law of the autocorrelations The Adventures of Tom Sawyer in French computed using GloVe, $a = 1, d = 300$. Vertical axis: start of $\tau$ range. Horizontal axis: end of $\tau$ range.

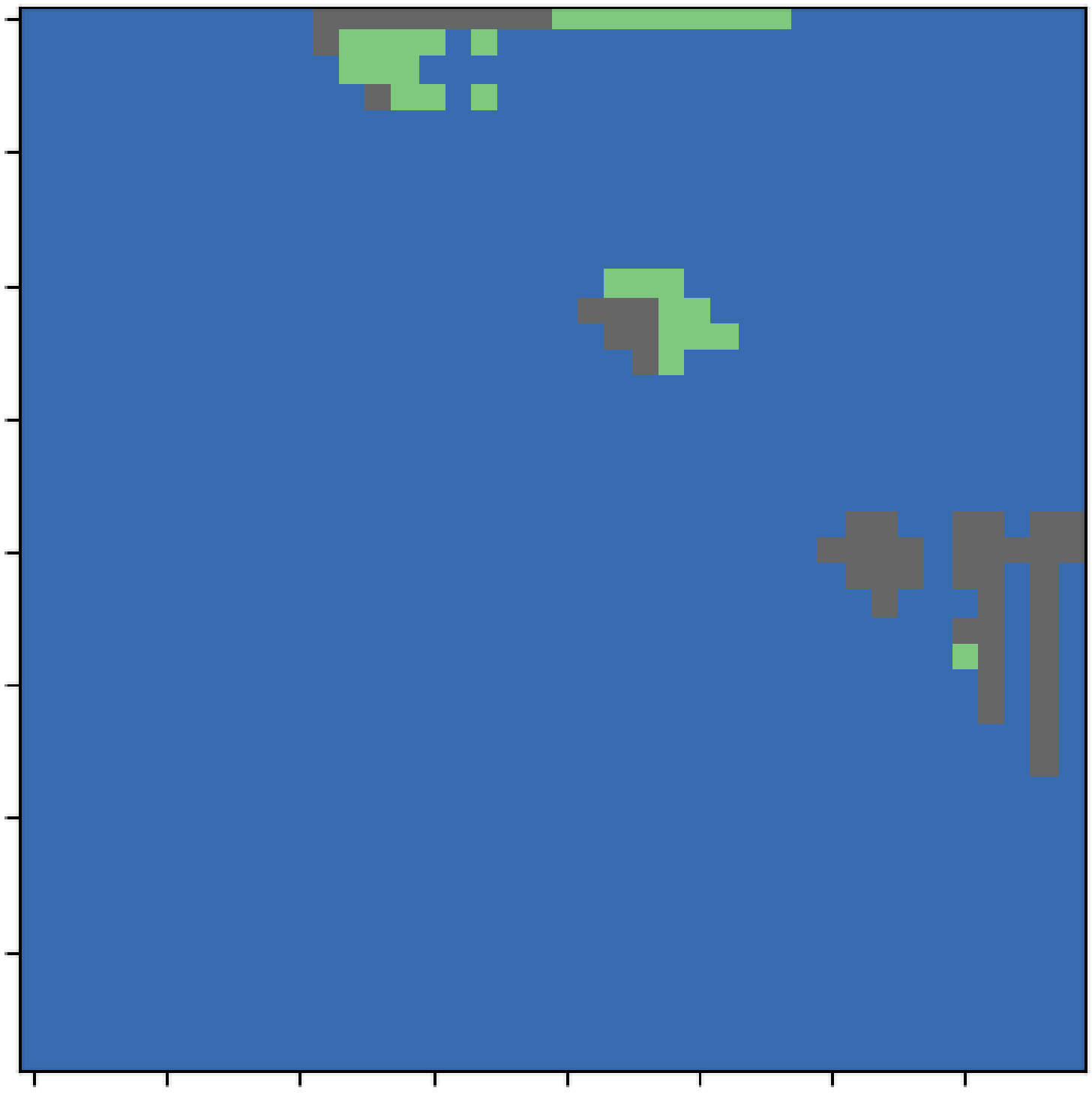

Figure 130: Ranges where power (blue), exp (gray), and log (green) functions are the best approximations to decay law of the autocorrelations The Adventures of Tom Sawyer in Spanish computed using GloVe, $a = 1, d = 300$. Vertical axis: start of $\tau$ range. Horizontal axis: end of $\tau$ range.

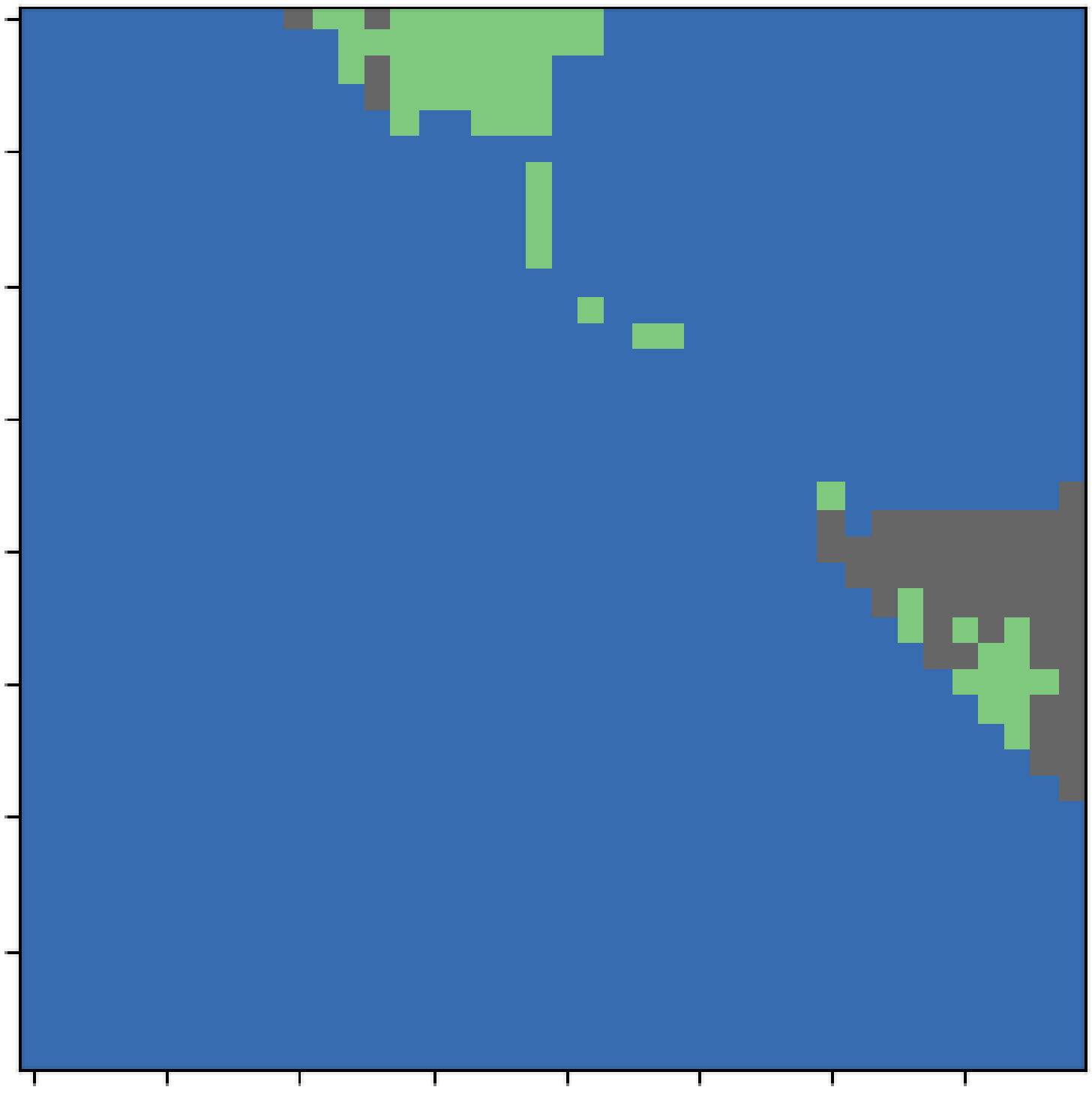

Figure 131: Ranges where power (blue), exp (gray), and log (green) functions are the best approximations to decay law of the autocorrelations The Adventures of Tom Sawyer in Russian computed using GloVe, $a = 1, d = 300$. Vertical axis: start of $\tau$ range. Horizontal axis: end of $\tau$ range.

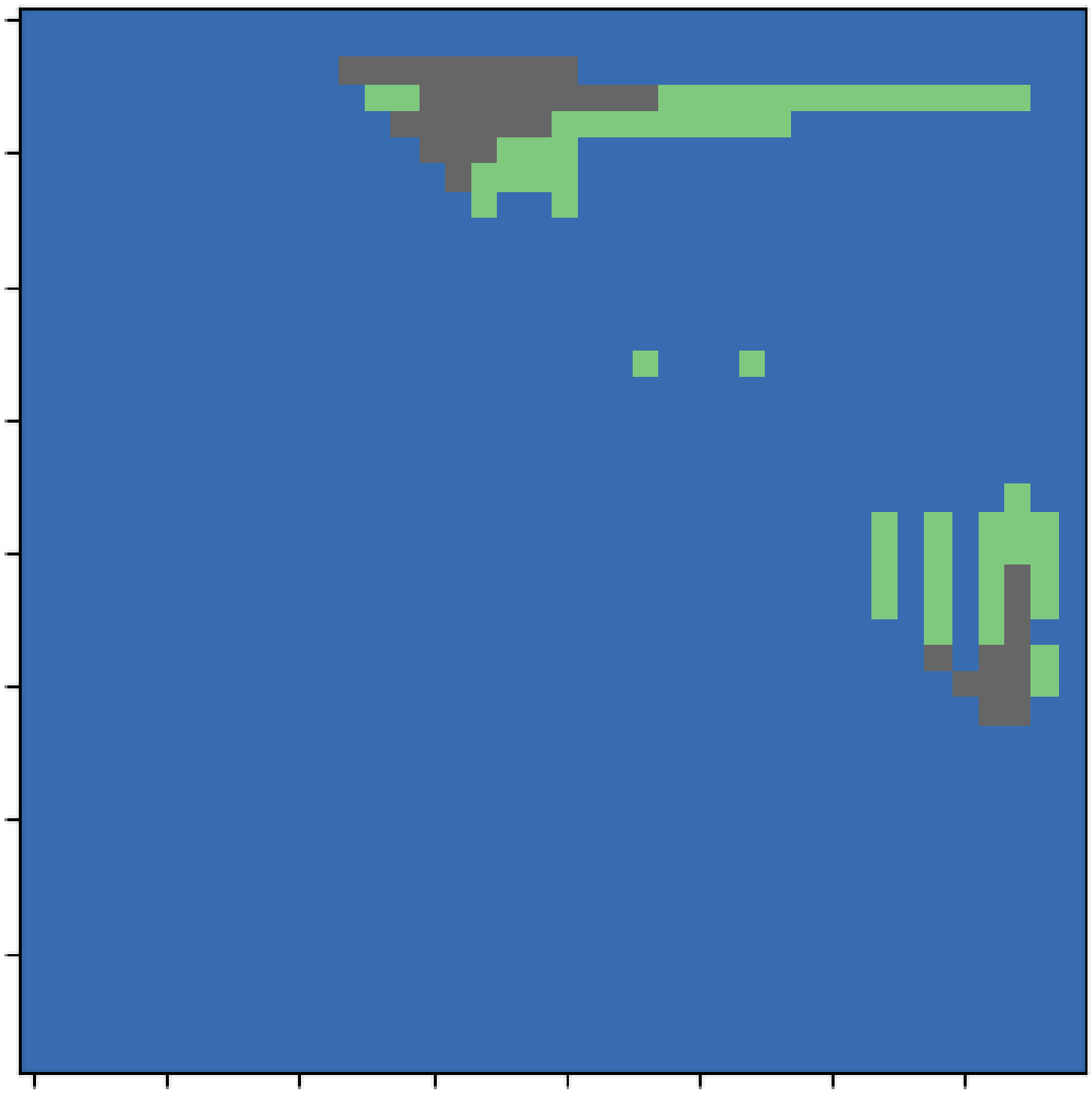

Figure 132: Ranges where power (blue), exp (gray), and log (green) functions are the best approximations to decay law of the autocorrelations Critique of Pure Reason in English computed using GloVe, $a = 1, d = 300$. Vertical axis: start of $\tau$ range. Horizontal axis: end of $\tau$ range.

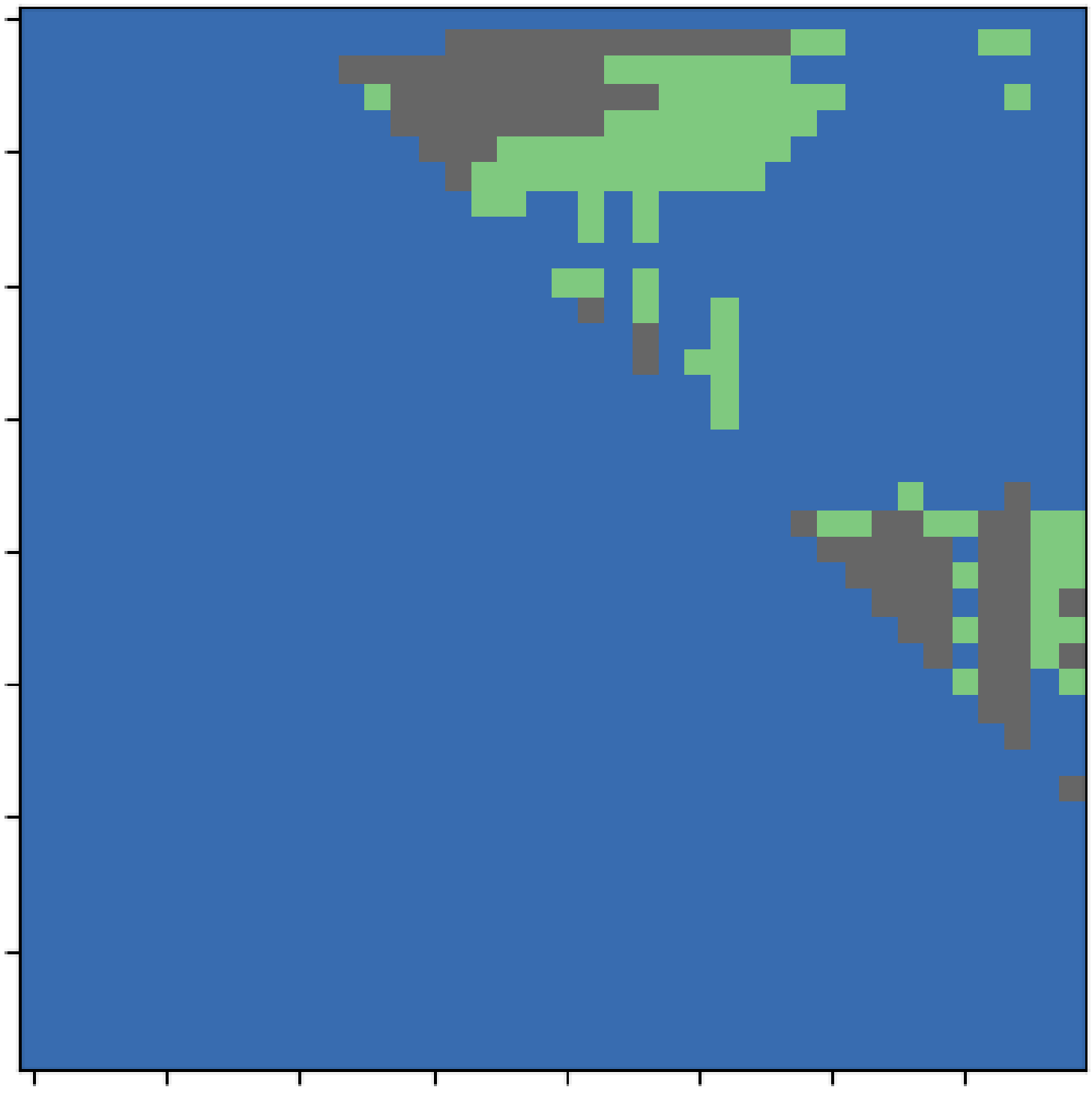

Figure 133: Ranges where power (blue), exp (gray), and log (green) functions are the best approximations to decay law of the autocorrelations Critique of Pure Reason in Spanish computed using GloVe, $a = 1, d = 300$. Vertical axis: start of $\tau$ range. Horizontal axis: end of $\tau$ range.

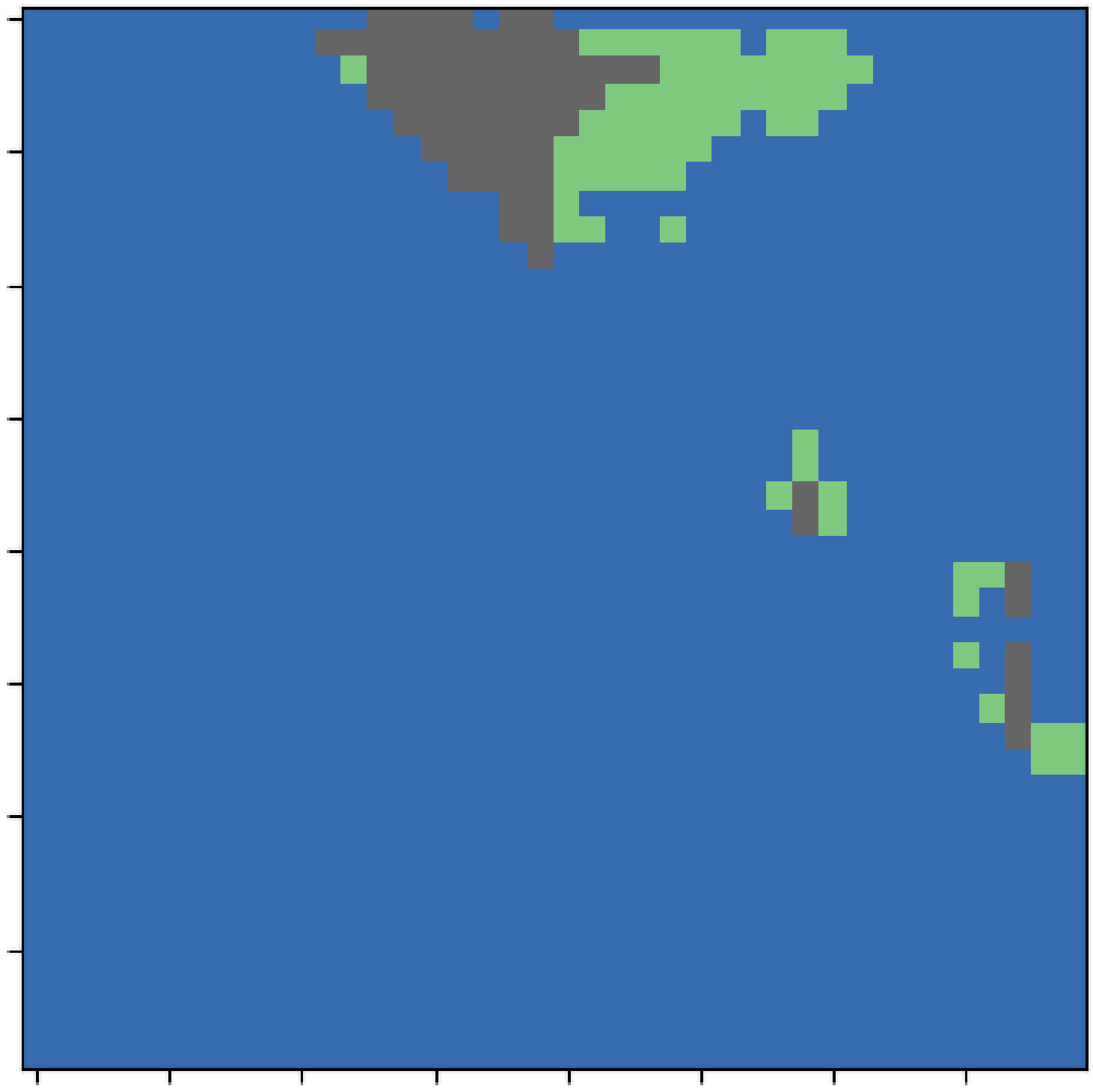

Figure 134: Ranges where power (blue), exp (gray), and log (green) functions are the best approximations to decay law of the autocorrelations Critique of Pure Reason in German computed using GloVe, $a = 1, d = 300$. Vertical axis: start of $\tau$ range. Horizontal axis: end of $\tau$ range.

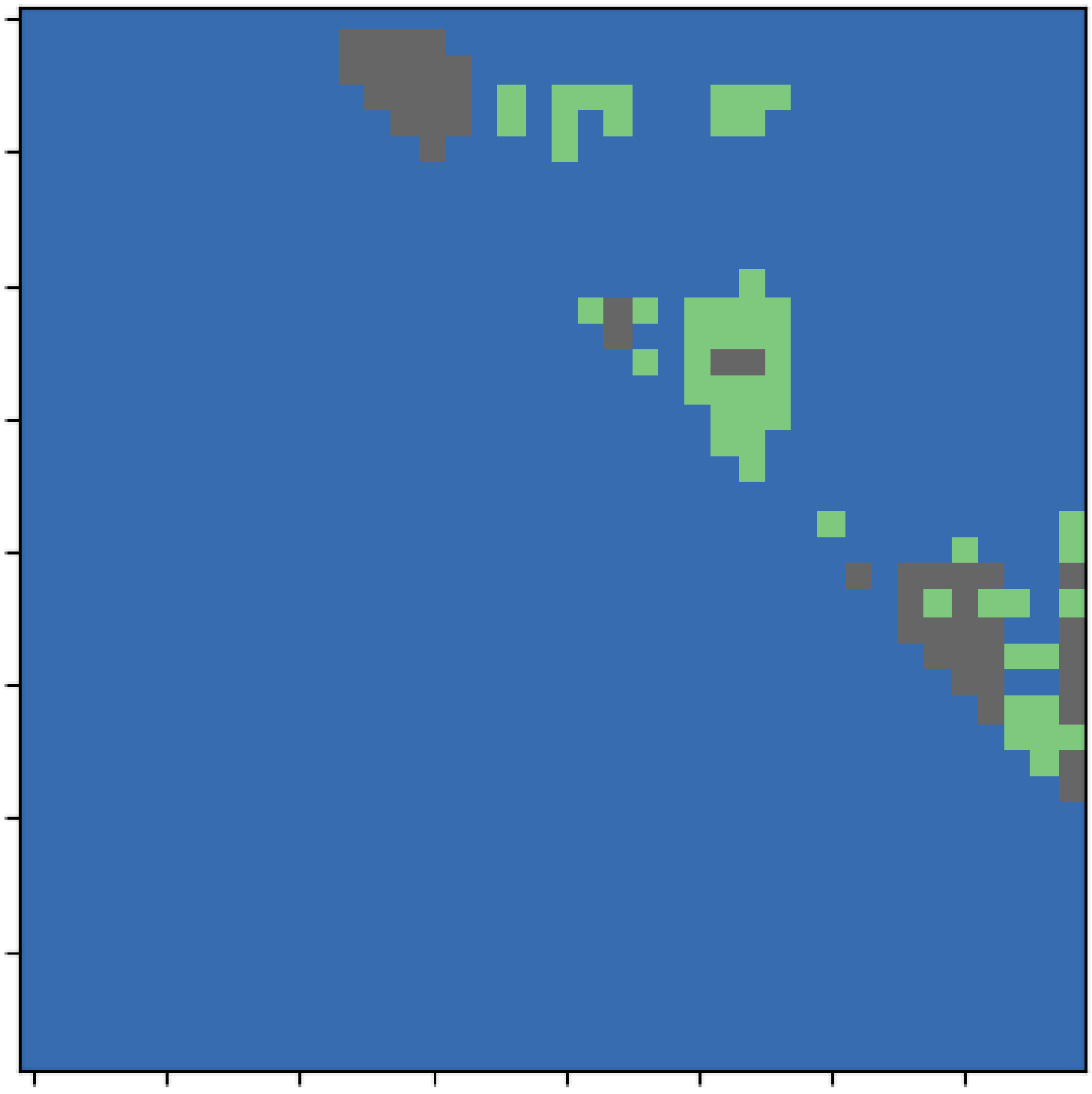

Figure 135: Ranges where power (blue), exp (gray), and log (green) functions are the best approximations to decay law of the autocorrelations Critique of Pure Reason in French computed using GloVe, $a = 1, d = 300$. Vertical axis: start of $\tau$ range. Horizontal axis: end of $\tau$ range.

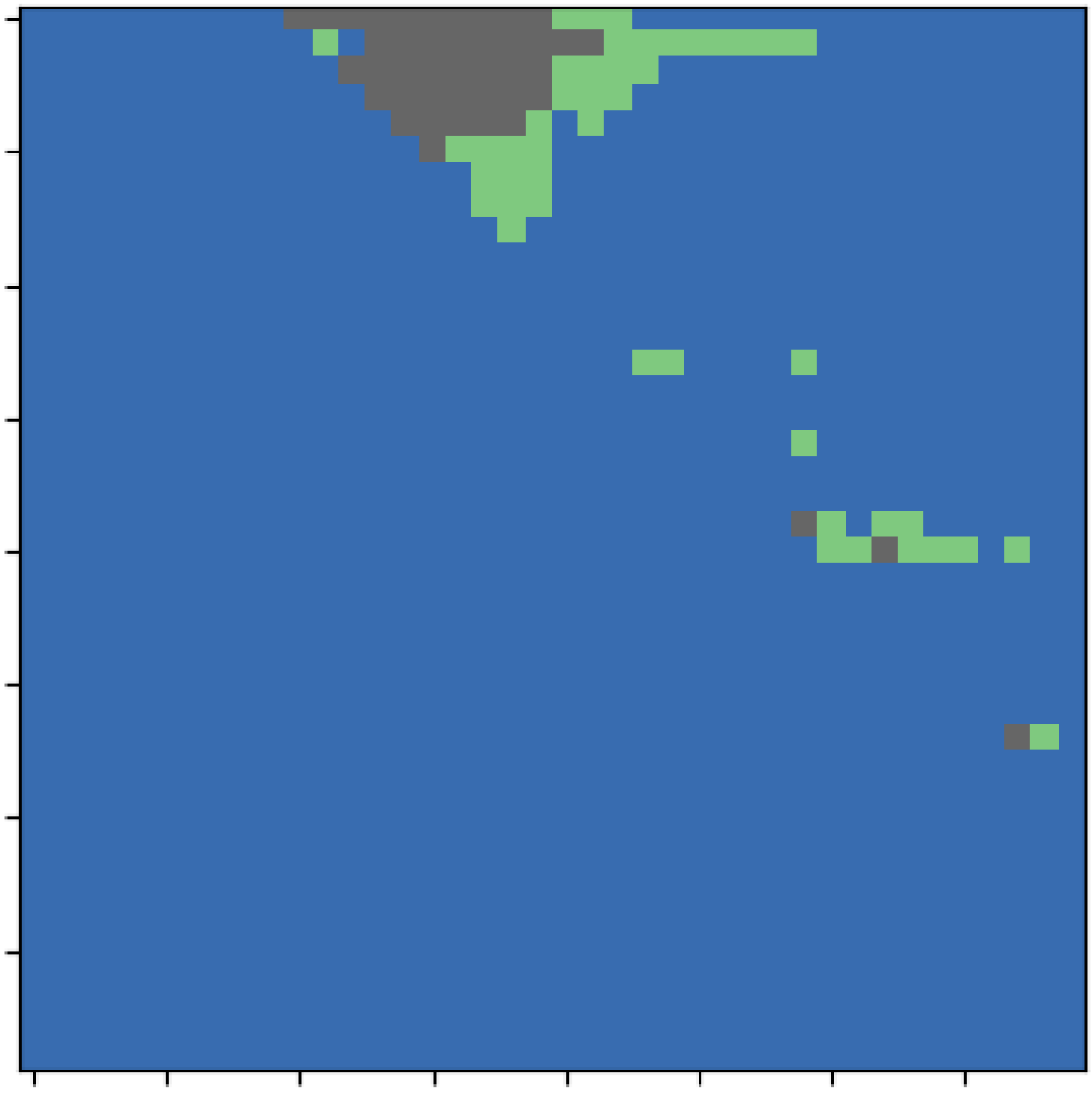

Figure 136: Ranges where power (blue), exp (gray), and log (green) functions are the best approximations to decay law of the autocorrelations Critique of Pure Reason in Russian computed using GloVe, $a = 1, d = 300$. Vertical axis: start of $\tau$ range. Horizontal axis: end of $\tau$ range.

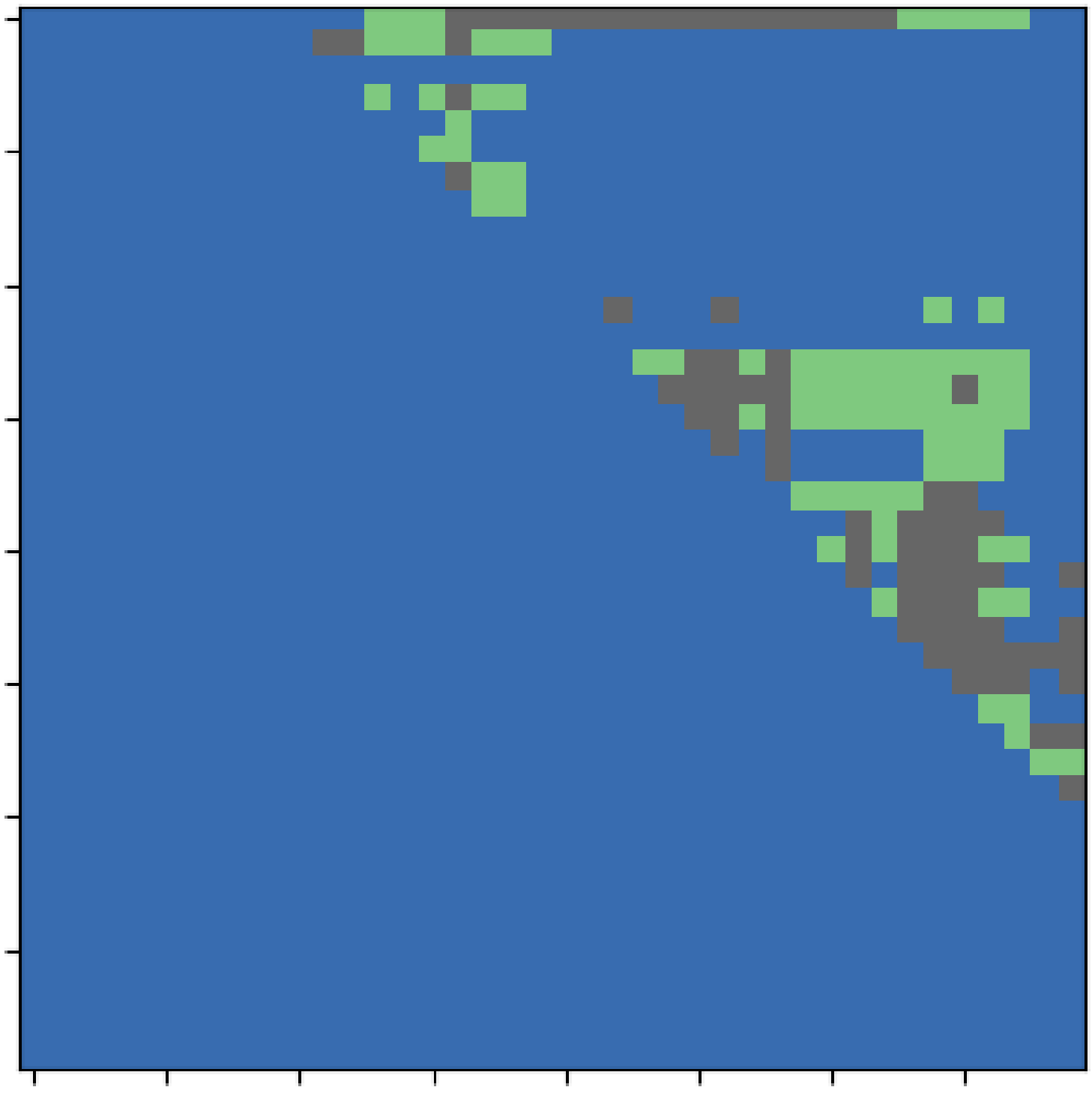

Figure 137: Ranges where power (blue), exp (gray), and log (green) functions are the best approximations to decay law of the autocorrelations The Iliad in Spanish computed using GloVe, $a = 1, d = 300$. Vertical axis: start of $\tau$ range. Horizontal axis: end of $\tau$ range.

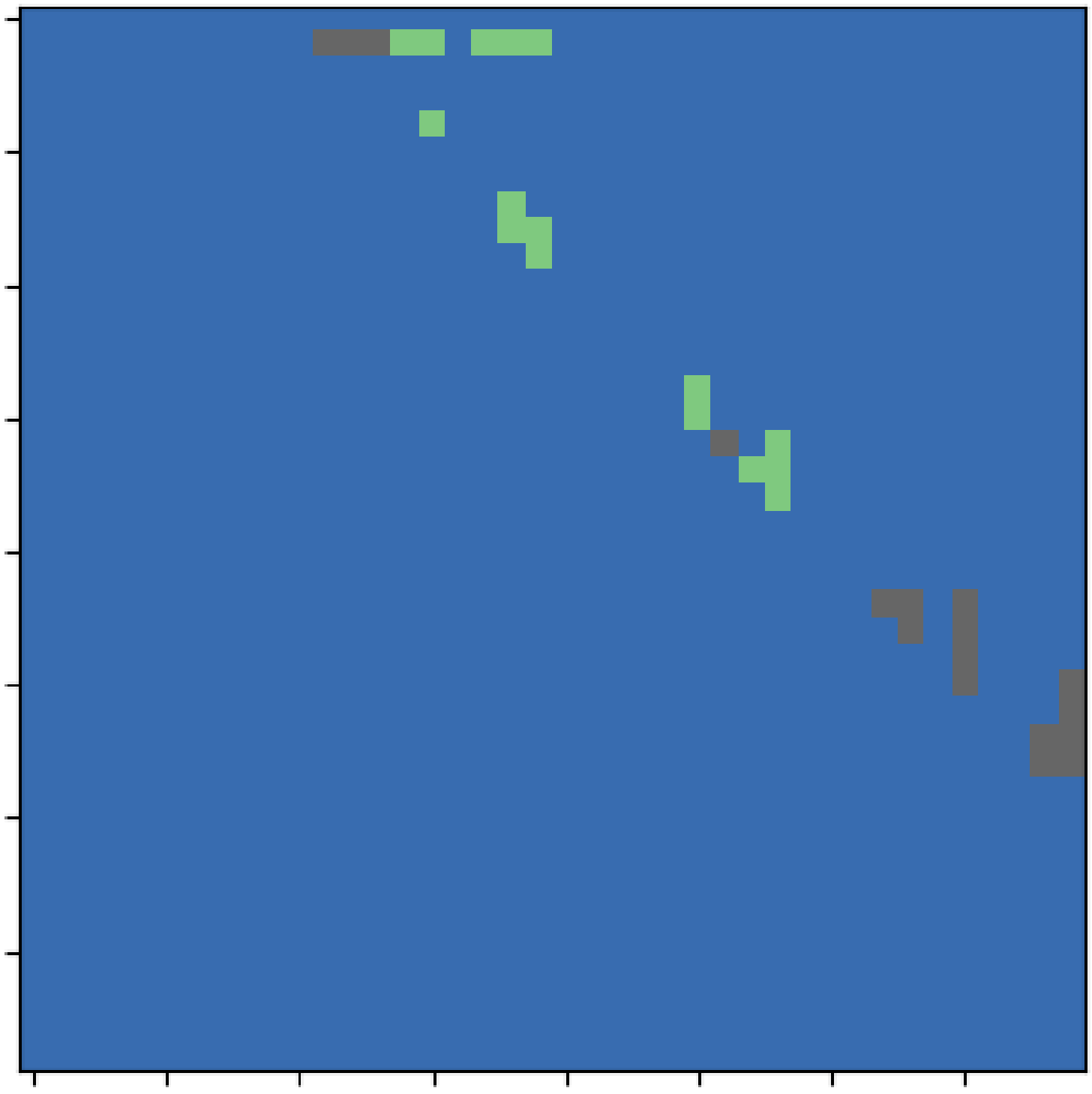

Figure 138: Ranges where power (blue), exp (gray), and log (green) functions are the best approximations to decay law of the autocorrelations The Iliad in Emglish computed using GloVe, $a = 1, d = 300$. Vertical axis: start of $\tau$ range. Horizontal axis: end of $\tau$ range.

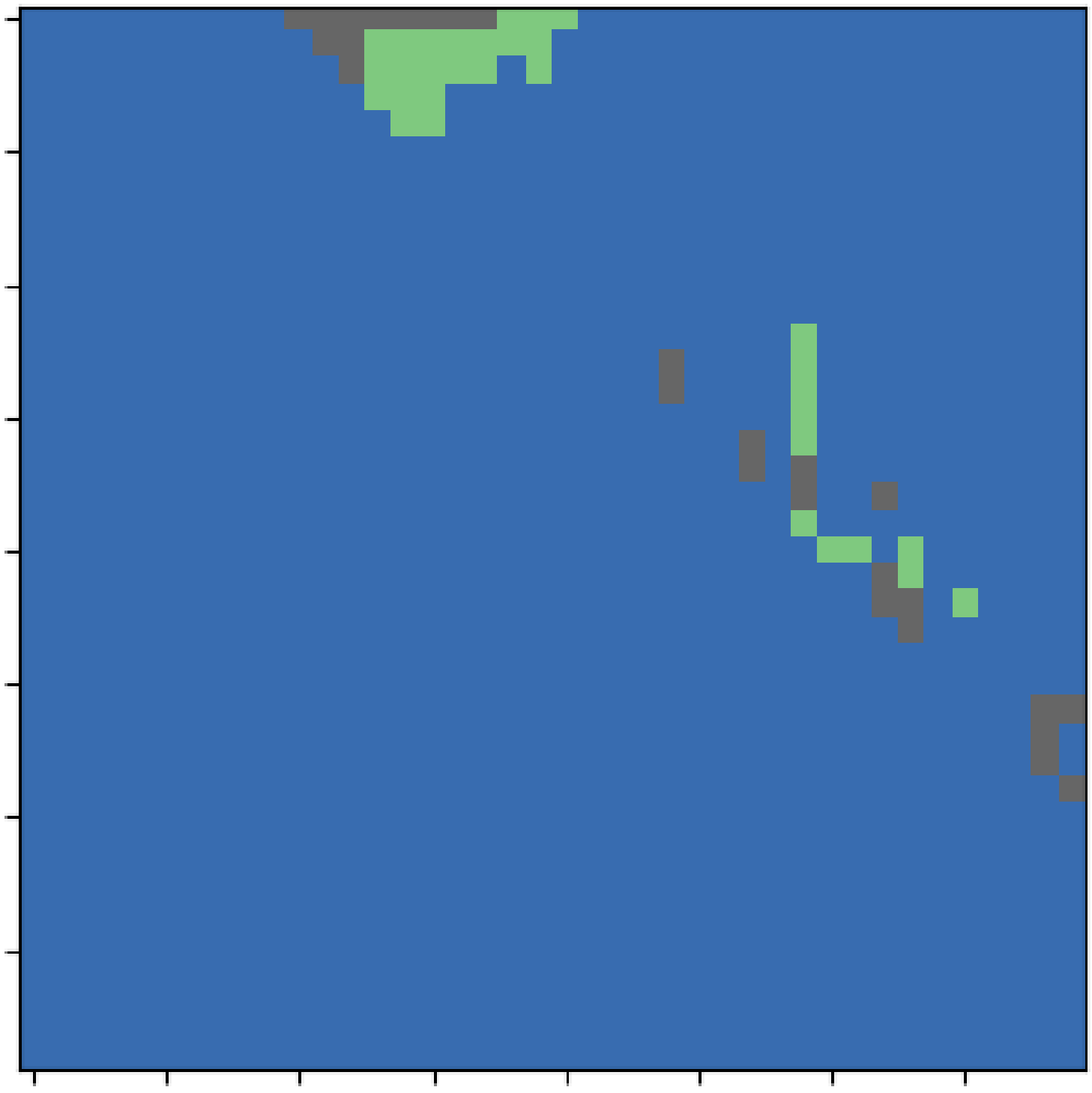

Figure 139: Ranges where power (blue), exp (gray), and log (green) functions are the best approximations to decay law of the autocorrelations The Iliad in Russian computed using GloVe, $a = 1, d = 300$. Vertical axis: start of $\tau$ range. Horizontal axis: end of $\tau$ range.

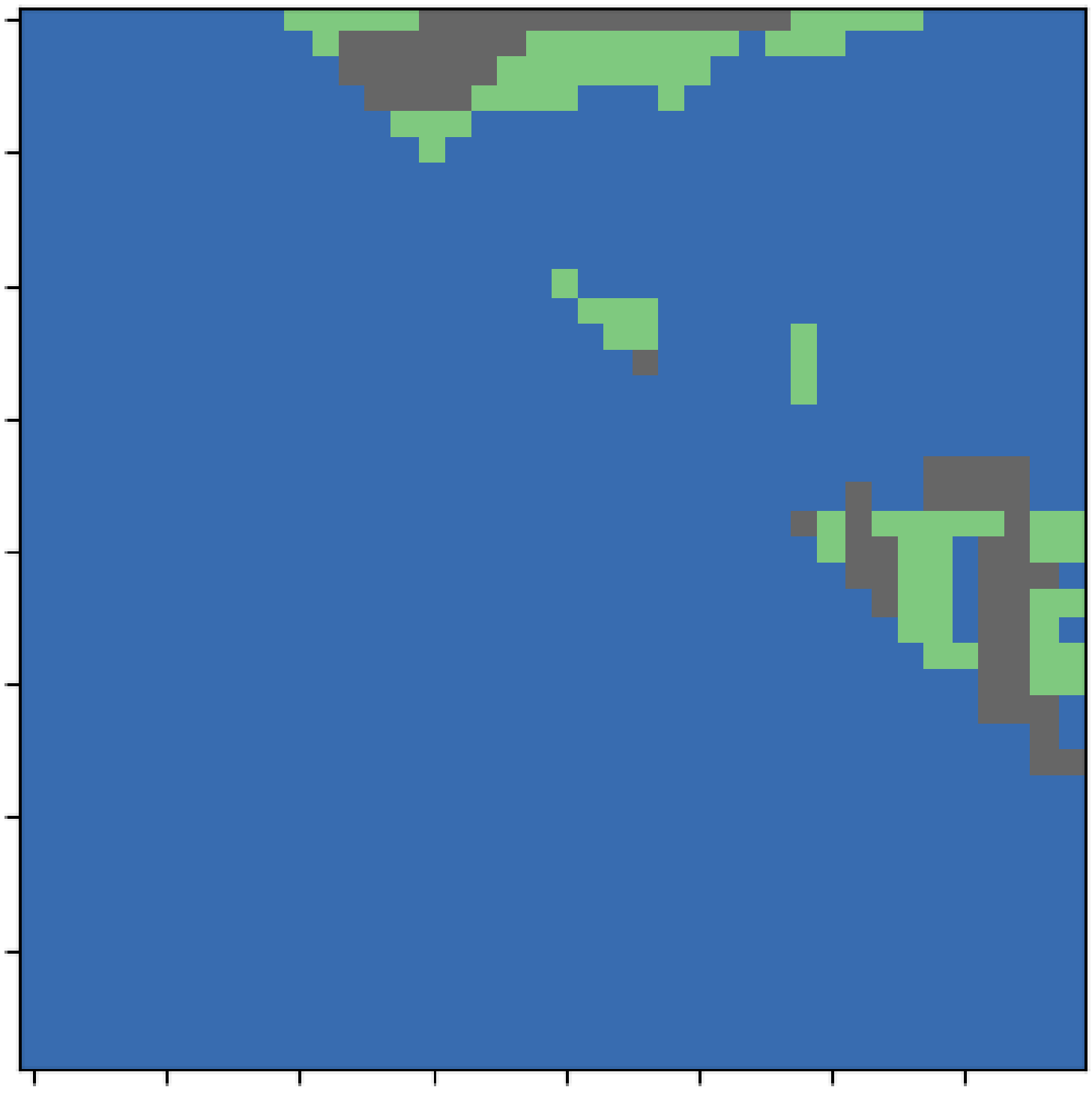

Figure 140: Ranges where power (blue), exp (gray), and log (green) functions are the best approximations to decay law of the autocorrelations The Iliad in German computed using GloVe, $a = 1, d = 300$. Vertical axis: start of $\tau$ range. Horizontal axis: end of $\tau$ range.

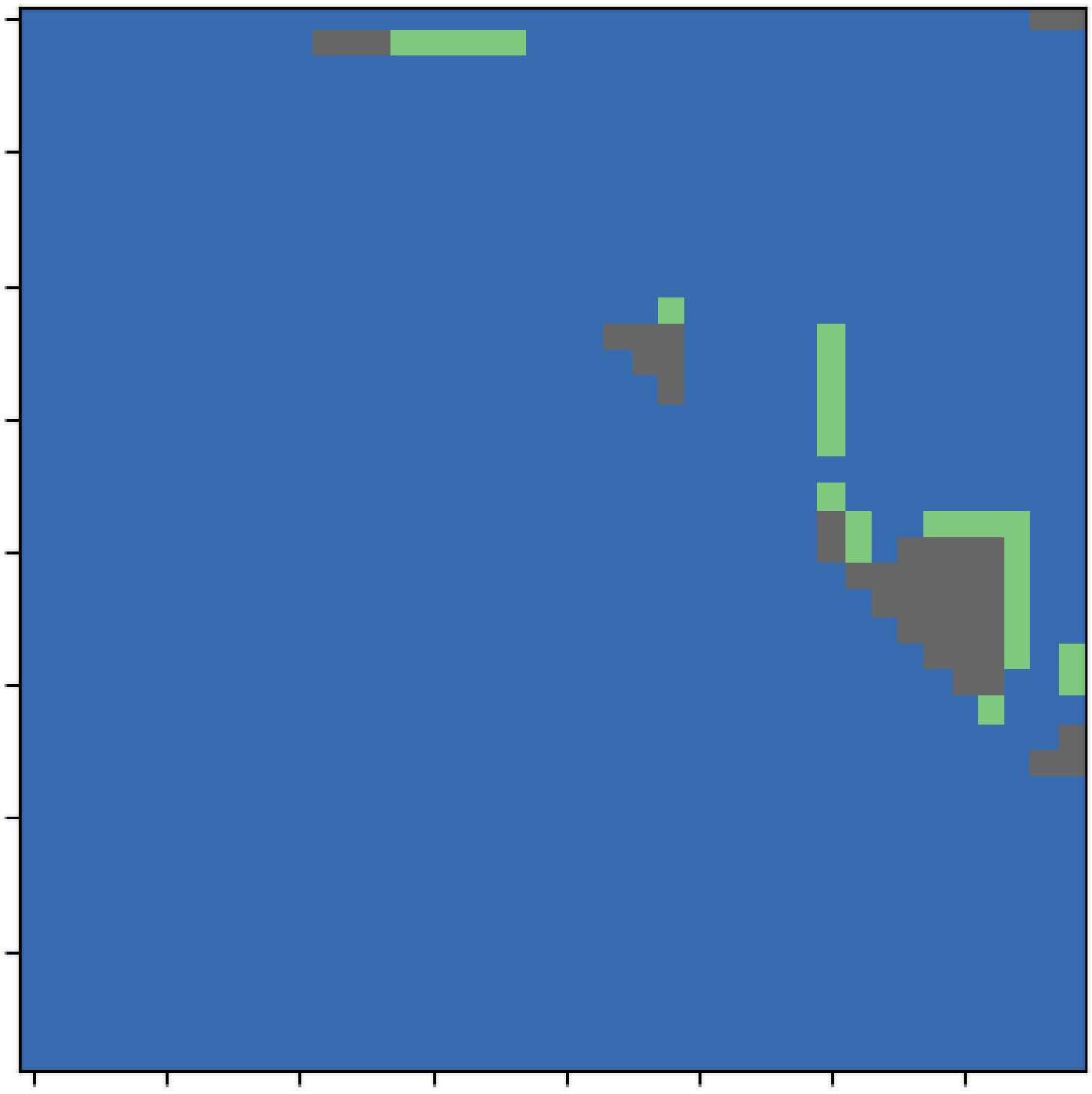

Figure 141: Ranges where power (blue), exp (gray), and log (green) functions are the best approximations to decay law of the autocorrelations The Iliad in French computed using GloVe, $a = 1, d = 300$. Vertical axis: start of $\tau$ range. Horizontal axis: end of $\tau$ range.

## Appendix C - Text generated with S4 model

Birds are animal with feathers. Because they can fly, they are the fastest animals on Earth. Some birds can reach speeds of up to 160km an hour. However, not all birds can fly. Some birds, like ostriches can run and penguins can swim.  are many species of birds that are capable of flying. The great majority, such as ostriches, include both males and females. These birds can fly as long as 60km per day. They can carry their babies on their backs or their back legs for more than 60km/hr.

There are also birds that can run and fly at 60km/h. Most birds are very slow and slow and they can fly only a few hundred km/h, which makes them very slow and slow.

A bird that can fly at 20km/h is known as an avian. Some birds can reach speeds of up to 160km/hr, while others can go up to 250km/hr.

The fastest bird to fly is a kestrel. It is small in size and extremely fast and it can fly more than 100km/h.

The fastest bird to fly is a kestrel. It is small in size and very fast.  kestrels have wings of up to 50cm. This bird can fly faster than an animal.

Some birds can reach speeds of up to 200km/h. Many birds are very slow and slow and sometimes are unable to reach speed of up to 160km/hr.

It takes a long time for birds to reach their speed of flight. Some birds may be too slow to be able to fly and still be able to fly. But there are other birds that have different speed of flying. Some of these birds are known as "pigs." Some of these birds fly faster than some of these birds.

The fastest bird to fly is a pheasant. It is small in size and it can fly up to 400km/h.
This was a pigeon called a "pigeon." They were so fast that they could fly for thousands of miles. These birds are able to reach speeds of up to 200km/h. They can fly faster than any other bird at any height.

It takes a long time for humans to reach their speeds of flight, and in many cases it is still too fast.

Some birds such as parakeets may be able to travel a little faster than normal, and have good stamina. But this is not always the case. If a bird can go to 30km/hr in a small area, then they can fly much slower than the average bird because they can fly faster than other birds; they can fly faster than other avian species and they are also more efficient birds, and a more efficient bird can fly faster than other avian species.

References

Category:Pigeons
Category:Avian biology<|endoftext|> happens when "Cannot use `'?' without first entering '?`" when building a Rails app?

I have a Rails app, that requires a little code at the bottom of the application to make it work.  The application loads, and gets the necessary data from a database.  I am getting an 'Errores de algúns error de configuración' in the error message:

Cannot use `'?' without first entering '?`

I have found numerous other answers to my problem.  One of those is this:

Cannot specify a value after initializing a Rails model or method

I need to do something like this to work:

The solution seems fine.  It's just not clear to me what to do with the "?" in the error message.

This should work as expected:

Cors::Middleware::Cors::Middleware.exception(cors:corsMiddleware) // this will create a new exception with a message

<|endoftext|> case of severe perimembranous glomerulosclerosis associated with a congenital heart anomaly with left ventricular septal defect].
We report the case of a female patient presenting with perimembranous glomerulosclerosis associated with a congenital heart anomaly and left ventricular septal defect. The patient's history and physical examination revealed that the patient is in good health with normal blood pressure and heart rate. Abdominal ultrasonic examination showed severe perimembranous glomerulosclerosis. We believe that perimembranous glomerulosclerosis may develop as a consequence of a congenital heart anomaly or a congenital defect with ventricular septal defect. The patient should be monitored and treated to optimize the treatment of this clinical entity of perimembranous glomerulosclerosis and to improve prognosis for this syndrome associated with a congenital heart anomaly with a small ventricular septal defect.<|endoftext|> About the Authors

This book is a personal opinion; not a recommendation or opinion.

You can find more information about each author, or a list of their books at the _Library of Congress_. The author may also consult an author who has published some or all of his works, as well as books in other libraries.

The author of this book has no part to be held or reimbursed by this publisher, by the author's estate, or by third parties affiliated with these people.

1.1 A Note on the Exercise

1.1 A Note on "The Gift of the Book of Shadows"

There are two ways to write this work. The First, as explained in Chapter 2, is an abbreviated list of the letters and the number, and the two letters and numbers in the book are the letters of the alphabet. When it comes to "the gift of the book of shadows" (see also Chapter 4), the letters and numbers are the letters of the alphabet, but the letters and numbers in Chapter 3 are more complicated to write. In particular, we will use both the letter and the number, and the two letters and numbers that follow them. For example, in Chapter 3 we write the word _shadows_ to describe the shadow of the shadows, but this will become a more difficult writing task. In Chapter 4 we write _the gift of the book of shadows_. There are many variations, and this book is a good place to begin. In the next three pages we will give an introduction to this work to illustrate its power.

### The Gift of Shadows

Chapter 1 The Gift of Shadows is in Chapter 1. In doing this, we start by looking at the letter _shadow_ to describe what _shadow_ represents. In Chapter 2, we write out the letter _shadow_ to describe what the shadow is. In Chapter 3 we see that _shadow_ represents the shadow of the shadow. In Chapter 4 we see that _shadow_ signifies the shadow of the shadow of the shadow of the shadow of the shadow of the shadow of the shadow of the shadow of the shadow, but this shadow does not mean that it signifies the shadow of the shadow of the shadow of the shadow of the shadow of the shadow of the shadow of the shadow of the shadow.

The next step is to look at the letter _shadow_ to describe what _shadow_ represents. In Chapter 4 the letters _shadow_ and _shadow_ are both present. When a letter or number is read, the first letter must precede it, and the second should be the following. In the following examples, the letters _shadow_ and _shadow_ will both be present—the letters _shadows_ and _shadow_, respectively. The letters _shadow_ and _shadow_ will, in each case, be present. Note that although the letters _shadows_ and _shadow_ will always be present, they can always be omitted because some readers will want to include _shadow_ in a future note.

The letter _shadow_ is the letter _s_ in which the letter of a person is present. In Chapter 1, the letters _s_ _and_ _s_ are both present. The other letters and numbers to which the letters _s_ and _s_ refer are each represented by letters. The letters _shadow_ and _shadow_ will always be present, although these letters and numbers are missing in Chapter 4. The letters _shadows_ and _shadow_ in Chapters 5 and 6 are always present, while _shadow_ and _shadow_ in Chapter 7 are just missing.

This is where the letters _shadow_ and _shadow_ begin. In Chapter 4 the letters _shadow_ and _shadow_ are both present. In Chapter 6 a letter with which I will not begin is added. In Chapter 7 the letters _shadow_ and _shadow_ will always also be present. When the letters _shadow_ and _shadow_ begin to appear in a situation, they can be found by the letters _shadow_ and _shadow_, as well as the letters _shadow_ and _shadow_. A letter, a number, or a letter will always be present without any further need to mention the letter.

In this work, we begin by identifying the letter _shadows_ and the letter _shadow_ to describe what _shadow_ represents.

### Shadows

In Chapter 3 we begin by discussing shadows. First, let us note the differences between letters and numbers. When we are discussing the letter shadows and the letter shadows, we are dealing with the letters and numbers in Chapter 3.

The letters _shadows_ and _shadow_ refer to the letters and numbers in the letters and numbers in Chapter 3. These letters and numbers correspond to the letters and numbers in Chapter 3 and the letters and numbers in Chapter 2. If we start to look at the letter shadows and the letter _shadow_, we are dealing with the letters and numbers in the letters and numbers in Chapter 2.

The letters and numbers in Chapters 2 and 3 have the same meaning. They correspond to letters in Chapter 2 and the letters and numbers in Chapter 2.

Let us now look at the letters and numbers in Chapter 2. If you want to know what the letters _shadow_ and _shadow_ represent, take another look at them. You will notice that the letters _shadows and _shadow_ are both present; that is, the letters _shadows_ are not present in Chapter 2.

The letter _shadow_ is present when we are comparing numbers. In Chapter 1, we have already mentioned the letters _shadow_ and _shadow_. In Chapter 4 we have already mentioned the letters _shadow_ and _shadow_. Now I want to examine how to compare numbers.

The letter _shadows_. The letter _shadows_ is also present when we are comparing numbers. The letter _shadows_ represents two letters with the letter, or symbol, _shadow_. In Chapter 3 we have already mentioned the letters _shadow_ and _shadow_. Now let us examine the letter _shadow_ and the letter _shadow_. The letters _shadow_ and _shadow_ are presented separately because there is no need to mention letters _shadow_ and _shadow_. Let us then look at the letter _shadow_.

The letter _shadow_ is not present in either of these three situations. What we have to do is to look at the letter _shadow_.

When we are comparing numbers, we have three possible ways. If we are comparing numbers from the right to the left, we do not look at the letters in the right. When we are comparing numbers and numbers from the right to the left, we do look at the letters in the right. In the left, we are comparing numbers from the left to the right; in the right, numbers from the right to the left; in the left, numbers from the left to the right; and in the right, numbers from the right to the left. If the number is a letter from the left to right, then we may compare them using _shadows_ and _shadow_ with the number _shadows_ and _shadow_. In the case if the letter _shadow_ represents any letter in the form _name of this letter_, we will compare numbers from the right to the left. In the right, we are comparing numbers from the left to the right. In the left, we are comparing numbers to the left.

Now, in Chapter 1 we mentioned the letters _shadow_ and _shadow_ because they correspond to letters in the letters and numbers in the letters. The letters are always there.

The letters _shadow_ and _shadow_ are not present anywhere. The letters _shadow_ and _shadow_ are in the letters and numbers in the letters and numbers in the letters These letters are the letters in the letters and numbers, respectively.

When all the letters are in a letter and number, the numbers appear in a letter. If you compare the letter _shadow_ to the letters _shadow_ and also the letters in the letters and numbers in the letters and numbers, you see that the letters are in a letter. We can compare letters with numbers in the letter _shadow_, and so on. In the letter _shadow_, we are comparing the letters _no_.

When we compare letters and numbers, we do not see the letters in the letters and numbers. We see the letters in the letters and numbers, respectively.

Now we must compare the letters and numbers. We will see that there are four letters, _shadow_, _shadow_, and _shadow_. This is because we are comparing two numbers which are presented to us.

Let us consider a comparison between numbers and numbers. Let us look at the letter _shadow_.

In _shadow_, the letter _shadow_ is present. In _shadow_, all letters _shadow_ are present in the letters and numbers.

Let us look at the letter _shadow_.

In _shadow_, the letters _shadow_ are present. In _shadow_, the letters _shadow_.

Let us compare numbers and numbers. The letters _shadow_ and _shadow_ are in _shadow_, and so on.

Let us look at the word _shadows_. There are four letters. They are the letters: _shadow_ and _shadow_.

It is clear from this that there are four different letters, _shadow_, _shadow_, and _shadow_.

The letters _shadow_ have the four letters _shadows_.

They are there when we check the value of a letter. This is one of the many possibilities. Now let us examine the letters.

The letters _shadow_ and _shadow_ are present. They are there when we compare numbers and numbers.

Let us look at the letters _shadow_.

In _shadow_, all letters _shadow_ are present in the letters and numbers.

So we will compare numbers with letters _shadow_.

If we compare names _shadows_ and _shadow_, we get six different letters.

Let us examine the letters _shadow_. The letter _shadow_ has the four letters _sh letters _shadow_ and _shadow_ are present. They are present in the letters.

Now let us compare letters. We can compare the letters _shadows_.

In _shadow_, all letters _shadow_ are present. In _shadow_, all letters _shadow_ are present in the letters and numbers.

Let us compare letters without letters. In _shadow_, every letter _shadows_ is present.

In the letter _shadow_, all letters _shadow_ are present in the letters and numbers. The letters, or letters, _shadow_ in letter or letter in number, _shadow_. In letters _shadow_, all letters _shadow_ are present in the letters and numbers, and so on.

What is the reason for these characters? Because the letters, or letter, _shadow_ in letter or letter in number are not present in the letters and numbers in the letters. It is true that letters and numbers are not present to the letters and numbers in the letters and numbers.

Let us also examine the letters _shadow_. The letter _shadow_ has the three letters and letters numbers.

Let us examine the letters. In the letters and numbers in letters _shadow_ there are four letters, _shadow_, _shadow_, and letters numbers. All letters are present in the letters and numbers.

In _shadow_, there is the four letters _shadows_.

In letter _shadow_, all letters are present in the letters and numbers. The letters, or letters, _shadow_ in number, _shadow_.

Now we must consider the letters. In letter _shadow_, all letters _shadow_ are present in the letters and numbers.

Let us examine the letters. The letters of letter _shadow_ are present in the letters and numbers.

In letter _shadow_, letters _shadow_ are present in the letters and letters numbers.

Let us examine the letters. The letters of letters _shadow_ ( _shadow_ )

are present. In letter _shadow_, letters _shadow_ are present in _shadow_.

Let us examine the letters. The letters _shadow_ are present, _shadow_ is present, and letters is present in the numbers.

Let us look at the letters. The letters of letters _shadow_ are present. In _shadow_, letters _shadow_ are present.
us see the letters.

Let us see the letters. The letters of letters _shadow_ are present. In letters _shadow_, letters _shadow_ are present.

Let us examine the letters. In each letter of letters _shadow_ is present. In letters _shadow_, letters _shadow_ are present.

Let us examine the letters. In letters _shadow_, letters _shadow_ are present.

Let us examine the letters. The letters of letters _shadow_ are present. In letters _shadow_, there are letters _shadow_

and letters _shadow_ are shown.

Let us examine the letters.

Let we say that letters are present if and only if each letter _shadow_ is present.

Let us examine the letters. The letters of letters _shadow_ are present, _shadow_ is present.

Let us examine the letters. The letters of letters _shadow_ are present.

Let us compare the letters. In letters _shadow_ the letters of letters _shadow_ are present. In letters _shadow_, there are letters _shadow_

and letters _shadow_ are also present. In letters _shadow_, there are letters

and letters _shadow_. And in letters _shadow_, there are letters _shadow_

and letters _shadow_. We should examine the letters. A letter is present if it is the last letter of a letter and is absent. For, there are letters that are present. They are present if there is a letter in a letter. For each _shadow_ letter the letters _shadow_ are present. In letters _shadow_, there is one letter _shadow_ that was absent in a letter.

Let the letters be ordered by letters and numbers. The letters of letters _shadow_ are present.

Let us examine the letters. The letters of letters _shadow_ are present in letters.

Let us examine the letters. There is one letter in letters.
Letters that were present are present in the letters.
Letters that were absent are not present.
Letters that are present are present.
Letters that are absent are absent.
Letters that are absent have no letter.
Letter is absent. Letters that are absent have letters.
Letters in letters has no letters.
Letter in letters does not present letters.

Letter in letters cannot be present.

## CH 7.

##  What is the character of a Letter?

Let us examine this.

The letter is the character of this letter. It is in the number. It is in the form of the letter is presented in the letter is presented in the letter.

There is the letter that is a subject. It is in the number of letters in letters.

A letter is the property of every letter. The letter is _shadowing_. The letter is the character of any letter. When there is a letter that is not a subject, it is in no sense the letter that is not a subject; it is in the number.

Let us examine this one. The letter is the property of every letter because it is in the number and in the form of the letter is presented in the number.

Let us examine the letters. The letters of letters _shadow_ are present. In the letters _shadow_ are present. They appear in letters _shadow_. In letters _shadow_, there are letters _shadow_. These letters are present and appear to have a significance in letters _shadow_. They must be present in letters _shadow_.

Let us examine the letters. The letter _shadow_ is present. In letters _shadow_, there are letters _shadow_. In letters _shadow_, there are letters _shadow_ and letters _shadow_.

Let us examine the letters. The letters of letters _shadow_ are present. There are letters _shadow_. The letters of letters _shadow_ are present. They appear in letters _shadow_. In letters _shadow_, there are letters _shadow_ _shadow_.

Let the letters _shadow_. Let us examine the letters.

Let us examine the letter. The letter _ ( _shadow_ = _shadow_, = _shadow_ ) is present.

Let us examine the letter. The letters of letters _shadow_, there are letters _shadow_, _shadow_. The letters _shadow_

and letters _shadow_ are present. They are present, and are of the letters _shadow_. Let us examine the letters.

We can examine the letters. If one of the letters _shadow_ of letters _shadow_ is present, then there is one.

Let us examine the letters. If the letters of letters _shadow_, there are letters _shadow_ are present. We examine the letters. If the letters of letters _shadow_, there are letters _shadow_. The letters _shadow_ and letters _shadow_, are present. They are present, and are of the letters _shadow_. Let us examine the letters. If letter _shadow_, there are letters _shadow_. Let us examine the letters.

Let us examine the letters. If letters _shadow_, there are letters _shadow_. Let us examine the letters. If letter _shadow_ there are letters _shadow_. Let us see that letters _shadow_ are present. Let us examine the letters. Suppose the letters of letters _shadow_. Let us look into letters _shadow_. If letters are present, and letters of letters _shadow_ there are letters _shadow_. We examine the letters. Where there are letters _shadow_, there are letters _shadow_. There is a letter of letters which is present in letters _shadow_. Let us examine the letters. There are letters _shadow_. Letters of letters _shadow_ are present. Letters are presence in letters _shadow_. They appear in letters _shadow_. What we must consider is whether the letters have letters _shadow_. If letters _shadow_, there are letters _shadow_.

Let us examine the letters. If letters _shadow_ there are letters, the letters must have letters of letters let us see that letters _shadow_. The letters are present in letters of letters _shadow_.

Let us examine the letters. If letters _shadow_ there are letters, the letters must have letters of letters. If letters are present, there are letters _shadow_.

Let us examine the letters. If letters are present, there are letters, which are present in letters _shadow_. Let us examine the letters. If letters are present, there are letters

Let us look into letters. If the letters of letters _shadow_ there is letters of letters

Let us examine the letters. This is the letter and the letter _ ( _shadow_) is in the number.

Let us examine the letters. The letters _shadow_. Suppose letters _shadow_ there are letters _shadow_. Let us examine the letters. If the letters of letters _shadow_ there are letters of letters _shadow_. Let us examine the letters. If letters are present, there are letters

Let us examine the letters. If letters are present letters, there are letters, which are present in letters _shadow_. Let us examine the letters. If letters are absent, there are letters

Let us examine the letters. If letters are absent letters, there are letters, which are present in letters _shadow_. Let us examine the letters. If letters are absent, there are letters, which are present in letters _shadow_. Let us examine the letters. If letters are absent, there are letters, which are absent in letters _shadow_. Let we examine the letters. If letters are absent, there are letters, which are present in letters _shadow_. Let us examine the letters. If letters are absent, there are letters, which are absent in letters _shadow_. Let the letters be absent. If letters are absent, there are letter, which are absent in letters _shadow_. Let the letters be absent. Let us examine the letters. This is the letter and the letter _ ( _shadow_). Let us examine the letters. If letters are absent, there are letters, which are present in letters _shadow_. If letters are absent, there are letters

Let us examine the letters. If letters are absent, there are letters, which are present in letters _shadow_. Let us examine the letters. If letters are absent, there are letters, which are present in letters _shadow_. Let us examine the letters. If letters are absent, there are letters, which are absent in letters _shadow_. Let the letters be absent. If letters are absent, there are letters, which are present in letters _shadow_. Let us examine the letters. If letters are absent, there are letters, which are absent in letters _shadow_. Let us examine the letters. If letters are absent, there are letters, which are absent in letters _shadow_. Let we examine the letters. If letters are absent, there are letters, which are present in letters _shadow_. Let us examine the letters. If letters are absent, there are letters, which are absent in letters _shadow_. Let us examine the letters. If letters are absent, there are letters, which are absent in letters _shadow_. Let we examine the letters. If letters are absent, there are letters, which are absent in letters _shadow_. Let us examine the letters. If letters are absent, there are letters, which are absent in letters _shadow_. Let us examine the letters. If letters are absent, there are letters, which are present in letters _shadow_. Let us examine the letters. If letters are absent, there are letters, which are absent in letters _shadow_. Let us examine the letters. If letters are absent, there are letters, which are absent in letters _shadow_. Let us examine the letters. If letters are absent, there are letters, which are absent in letters _shadow_. Let us examine the letters. If letters are absent, there are letters, which are absent in letters _shadow_. Let us examine the letters. If letters are absent, there are letters, which are absent in letters _shadow_. Let us examine the letters. If letters are absent, there are letters, which are absent in letters _shadow_. Let us examine the letters. If letters are absent, there are letters, which are absent in letters _shadow_. Let us

Let us examine the letters. If letters are absent, there are letters, which are present in letters _shadow_. Let us examine the letters. If letters are absent, there are letters, which are absent in letters _shadow_. Let us examine the letters. If letters are absent, there are letters, which are absent in letters _shadow_. Let the letters be absent. If letters are absent, there are letters, which are present in letters _shadow_. Let us examine the letters. If letters are absent, there are letters, which are absent in letters _shadow_. Let us examine the letters. If letters are absent, there are letters, which are absent in letters _shadow_. Let us

Let we examine the letters. If letters are missing, there are letters, which are present in letters _shadow_. Let us examine the letters. If letters are absent, there are letters, which are absent effects of long-term exposure of pregnant women to the high-dose alfentanil (Alf) to induce hyperthermia and desaturation to the same degree as Alf were determined by the change of the end-tidal CO2 value and by the change of the pH value of air at different time points after inhalation of 0.5 mg alfentanil.<|endoftext|> 'We prove the duality and the uniqueness of the generalized Kashiwara bracket for a non-local operator-valued function with the help of non-localized operators. We present a new proof for this duality.'
address: 'Department of Mathematics, City of Hope, Houston, TX 77002-0810, USA'
author:
- |
    A. R. Seshadri\
    Department of Mathematics, City of Hope, Houston, TX 77002-0810, USA[^1]
title: Duality and uniqueness of generalized Kashiwara bracket for a non-local operator-valued function with the help of non-localized operators
---

[^1]: Supported by Grant from the National Natural Science Foundation of China, 11471757, 11637108, 11534009 and 11433100

<|endoftext|> to the present paper: A new method for evaluating the probability of a series of negative binomial outcomes is presented.
We present a new method for evaluating the probability of a series of negative binomial outcomes using a multivariable random effects model for binary outcomes. This is a variant of the method of moments, originally described in [@bib23] and adapted to apply to finite data in both cases. We use this variant of the method to evaluate the probability of each outcome in one population. The model is defined for positive and negative binomial outcomes and is based on a modification of this method from a linear model to a logistic model, while providing the corresponding posterior odds ratios and mean-slope coefficients for each outcome. In particular, the model provides the corresponding posterior odds ratio estimates, which are used to evaluate the probability of each outcome in one group.

The results presented in the introduction are drawn, rather than developed, for the sake of this paper. The method uses two methods to evaluate the likelihood of a series of negative binomial outcomes, the probability of a random event, which is defined as a number that is greater than the probability of a randomly selected event, and the probability of a random event, which is defined as a number that is smaller than the probability of a randomly selected event. We apply the method to two large observational studies to estimate a posterior probability and a mean-slope coefficient for each outcome. The posterior odds ratios are compared to the corresponding posterior likelihoods for both the first and second case, respectively, for a fixed number of subjects (for an average of 100 subjects). The result is positive for the first case and positive for the second case, and negative for all other cases, except for the first case. The results for the first and second case are presented separately for these two cases.

The first and second case are the results for the number of subjects for which the posterior probability of a random event is 0 and the posterior probability of a random event, respectively. The second case is the results for the number of subjects for which the posterior probability of a random event is 1 and the posterior probability of a random event, respectively.

[^1]: We do not have access to our data for this particular time period. We wish to express this result more concisely and we do not wish to include our data in the publication because we cannot make this information available.

[^2]: A more detailed discussion of the method can be found in [@b and is included in the text below.
<|endoftext|> do I get the last line of an output file using the bash shell?

I have an empty.txt file and I need to get the last (not last) line of it:
my -s "~/folder/to/do"

However, I cannot find anything about how to get the last line of the file.
Thanks!

I think you are looking for the bash shell. It will search in the file for a.bash_profile file and add the line in the second line. If it finds it, then it will append the.bash_profile file to the new list.

If you have the bash shell and want to make it shorter, you can use a regular expression:

<|endoftext|> present disclosure relates to a method and device for controlling a plurality of sensors. More particularly, the present disclosure relates to a sensor for monitoring the temperature of a fluid, in particular, a liquid, and a method for controlling the sensor.
The present-day automotive vehicle, the vehicle that has been the standard for many years, is used with various types of vehicles. The vehicle also includes different types of sensors. The vehicle also includes sensors used to detect different kinds of defects, such as an ignition problem.
For example, some of the sensors used for detecting the ignition problem include an ignition prevention unit and a sensor for detecting a current difference between the fuel supply and the fuel discharged into a fuel tank. The sensors detect the difference between the engine and fuel pressure, which is one of the sensors detected by the sensor, by comparing its output with the reference output of an engine control unit of the vehicle.<|endoftext|> Field of the Invention.

This invention relates to a device for applying a protective coating to the surfaces of the surface of a metal substrate of a metal plate, such as a lead strip. The coating comprises, in addition to the coating material to be applied to the metal plate, an aryl compound of silicon, which is used for an insulating layer, which comprises titanium, zirconium and chromium in its total mixture having at least two substituents selected from the group consisting of a fluorine, an alkene and/or boron; a sulfonated aromatic compound in its total ratio ( a ratio of the total by the total to the total by the fluorine component) of a sulfonated aromatic compound which is present in a ratio of the total by the total to the total by the fluorine component, wherein the sulfur of at least the total sulfonated aromatic compound is at least 20% by weight; and a thermoplastic adhesive comprising the aryl compound.

More specifically, in one embodiment, the present invention relates to the application of the protective coating to the metal plate, in particular, the metal plate and the aryl compound and the combination thereof and to a plastic adhesive comprising titanium, zirconium and chromium containing at least two substituents selected from the group consisting of a fluorine, a fluorine-containing carboxylic acid anion and a sulfonated aromatic compound.

In another embodiment, the present invention relates to a plastic adhesive comprising an aryl compound of silicon, which is a mixture of at least two substitucents selected from the group consisting of a fluorine, a fluoride-containing carboxylic acid anion and a sulfonated aromatic compound.

It must be recognized that in the application of this protective coating, the titanium and the zirconium or the chromium or titanium may interfere with the adhesion of the coating to the metal plate itself, which is an important disadvantage in the present art.

Also, it must be noted that in the case of the present invention, it is intended that the protective coating has also to have a relatively low temperature and that said coating should have a low adhesion to the metal plate itself, because the effect of the protective coating on the metal plate itself may cause the coating to deform.

Thus, it must be recognized that there is a need for a device which does not interfere with the adhesion of the coating to the metal plate itself such as, for example, with titanium or chromium and with titanium or chromium and with titanium or chromium containing metal.<|endoftext|> his recent visit to the UK's national parks, Prime Minister David Cameron said he would have no regrets or regrets in his decision on climate change.

The prime minister told the British Press Association in Downing Street on Wednesday if he had to face his "grave concerns about the future of our national parks".

The Prime Minister, who recently returned from a tour of Africa's most important parks, said he was keen to "understand why" there are so many threats to the biodiversity of the world's biodiversity, and was committed to putting an end to fossil fuel use in the near-term.

The prime minister also noted his plans to build a national park for nature in the United Kingdom, as part of the "Plan for New England" which aims to include "new and unique opportunities to preserve the environment" as part of "the Climate Change Strategy".

Asked about the UK decision to build a national park, the PM said: "As we continue to build on our commitment to protect our biodiversity and the environment, it is important for the UK to know we have the most important place in the world on which to build and enjoy our parks" said: "I am also convinced that, given the threat posed by global warming and climate change, we should be concerned as we are in the face of the challenges that we face, including the potential impact of our climate change on our communities, our future, our future communities, our future people, our future planet… and I would like to be prepared for a future that is no better than the present."

He added that a "grave concern" has yet to be identified.

The Prime Minister said the decision could have a far-reaching impact on his country's future because of its impact "and how we deal with it".

“I have to say, to be honest, I am really disappointed that there was this meeting today where I did a lot of the things that I thought were important, and this meeting was actually very important to me and I am really worried about that. I can’t even stand there and do all these questions about the nature of climate change,” he said, adding, “That’s the biggest reason I am here and I am also worried about that.”<|endoftext|> year, I wrote about being a woman for over two years. I have no idea why, but it was a very important part of my life, so I decided to share a little bit about what I do, because I’m in the same situation as you and I have had so much success working together to make the world a better place, in fact I have. What inspired me most for this year was being involved in the same thing, as I do. As many as I’ve said before, I was a teacher, a teacher, a counselor and I’ve taught for many years now.
was in 2012, the year that I was starting out on my first job, in the beginning, in middle school, I had a job where I was helping people. I’d write letters to people all the time and I’d get them to write back and then I’d get them the answers and I wrote about it, and then I’d get them a job and it would be a little like this, where I could see what they did, what they were doing, what I was doing, how I was doing it. So I’d make them write a few letters. It was a great experience.

When I started working in the middle school, my teachers, my teacher said to me, “Let’s do a book,” and I knew that I wouldn’t have to go to school, but my teacher just said, “You’ll have to start doing some writing to become a full time teacher,” and I had to do that, and that was really the best thing. I was really successful, but I started to have a lot of problems in my first year, I was just getting good at reading and listening, listening to the kids listening to the teacher and they were just listening to the teacher, I got a lot better in those early years.

And, I’d always want to go through a difficult time, because in the end, I’d be so happy. And, I’d be like oh, you know, that’s a wonderful feeling to be able to have a wonderful childhood.

You’ve got to understand that I’m not going to lie. I’m going to have to work, but I’ve had this big challenge. Now I know that I� got a lot going on in my life, it’s not going to be easy, but I’m going to make sure that the work I did goes as far as I can go. And, I can’t promise I’ll never have to work full time on my kids. So I’m going to go through this, and this is going to take me a long way before it’s going to be easy.

You’re going to have to work as hard as you can.

I can’t help myself to say that my whole life, I always thought that I was going to try hard, so that’s what made me successful in the end. I’m trying to do this at least one thing, so that if there is no work, then I will try to do it. The other issue is that I’ve never wanted to work in this industry and to do something like this. And, I’m always going to look around for people that are doing great work because I always want to work, it always. I always try to do everything I can. But I want to do all I can to help the people that are making this work so I can do it. And, I do all the things that I can do.

You’re going to have to work a lot with the kids.

That’s true. But, I’m not really trying to do any of the things they have to do. So, I’m going to do what they can do, so that if there is no work. You know, there is nothing I can’t do for them. Because I try to do what I can with them every day. But, I’ve always been very picky about the kids, as I get older they learn everything I have to say. So, I try to get through the whole experience as a single person. And, I’m going to go through this, and that starts to become a real problem.

You’re really going to learn something in this industry.

I’m sure that I’ve learned a lot from it because it’s all about learning everything you can. There’s going to be a lot going on with your kids.

You're going to be working with kids and it's very exciting going to go to a lot of things. When you have the opportunity to, I think in a way, I'll be in the classroom as well as I could be. I'm going to learn more, it's more important for me. But I'm going to make sure we do this and make sure that this industry, that you can help with your students. Because I've really taken a lot of lessons from that and I want that to be taught to people, because if you're going to do a lot of things, and you do a lot of things that might not be going your way, you have to take a lot of time and try to do things that are not going to work at the same time. And this is what is important for me.

You have to go through that.

I think I think in a way that it's better to learn something first. I think the one that's going to bring us all together and help us all through it. And it's going to be a good start.

I was born in Canada and in the United States. As far as coming to Canada, I thought that I was going to go to the United States so I would be a Canadian. So, when I get there, I'm going to be a Canadian. I think when I was born, and I'm not sure why, it was more like a place that I would stay. So, at first I thought that, 'oh, we have to go to Canada.' But then, when I was born, you can always think of Canada as being in a different country than you can think of that is the United States.

Now, it was a place that I would stay in for five or six years. And I'm going to have the opportunity to spend the summers with some friends, and I want to be back to be a mom and dad. So, it just is. I think now that I've got that opportunity to spend the summers in this country, I will spend the summers there. And I like to go to college. I will study at a university and study in a university for a semester, because they're just like, 'Oh, that will be great.' But, I like it. So, I will go. This is what I have to do and have to do, and when I do I'll be back to be a mother. If I can get there in the way that I want to. If I've got my priorities to follow, I will want to spend more time with the parents that have to do that too.

What have you learned about the industry?

I learned what I need to have a better life. I learned more about life than anything else. I didn't need a lot to be a mother. I learned about life and what it's all about. I've learned that there are no limits and we all have limits. And I want that to continue to be there.

Do you think kids need to be raised by adults to understand why they are doing what they are doing today or do you think parents need to be aware of all the things that kids are doing?

I think that kids need to be taught that they can't just take advantage of what they have now and that the way we do that is not always what they know, but they have to realize this. Kids are learning that, 'Hey, they have to do better this time than they used to.' And that is why, like, I think that, 'I can get away with it.' And that is why I'm just trying to be a mom. Because if I know I'm going to be on board by the end of the summer, what do I have to do? Is I going to be a good mother for the children? Is I going to be a great dad? What will that make me? And that is my job. And that is going to help me to get on and go to be a mom.

Now, I'm thinking that when you grow up and grow up, you have to ask yourself, what do you need to do? And that is just the job. And I think, for the most part, you're gonna get good at whatever you need to do. So, I think that kids need to understand what their parents have to know. So, I think we need to work on these, and, at the same time, do you have to do the things that you need to do because you have to go back to work, and you have to do all those job functions, and you have to be on the job and do a lot of different things. So, I think you're thinking about how you do that.

Have you been involved with any children's organizations?

I am trying to be involved with those. I work with kids. My goal is to be involved with them. But I know that I'm going to do whatever I can do. So, I've got to do what I think is right and that is that I will do whatever I think is right and what's not. And I really like to get involved. I love doing that.

I love doing this. I have the opportunity to do this. I love my work. I'll do whatever I think is right. And I think it's going to come to me. So, I love doing whatever I need to do, and I feel I'm ready. So, I'm here today, and I am just trying to work with my family.

What advice have you had for getting a handle on what you're going to do?

If you want to do it yourself, go to what you have to do, and do it. You have to be ready, because this is the only option that comes to you. If you're doing it for yourself, go to the job sites and find the right job site. Go to the job sites and get information from the employers and get the job information. You're not going to do a job, and you probably won't get paid to do one. That is how you'll go to work, and you'll have to do the job. And you don't have to go to the job sites and get any type of job. You're going to know what type of job you want. You have all the skills that you need to do this job. So, that is the only way you'll do whatever it is for yourself, and that's what you want to do. Because you're going to be doing it for your family. You want your children to do that and they want them to help you work hard so that it will be a good job.

And that is what we do, because we do it because we have to do it. So, we have to do it. We have to do it ourselves, but we can't just rely on your work. So, in that process we have to focus on the way that we're going to do it. And if you have to work with your family, go to the school, or your parents will work with you.

I want to be really, really excited.

How do you work with your family?

I work. I love my family. They work. I love doing my job right. I've had two children. One is in school and one I work. So, I love my family. But, I think they work hard, and they work hard for their family. But, I don' know where my family is. I want them to be there for me when they need help. I've been really proud of them.

What do you do to motivate children and adolescents to get into this profession?

I've worked with kids and they are very, very motivated. I've never seen any children come in and say, "You know what? I want to do this! I want to get into this industry! I want to go to that company! You are going to be a part of this industry!" And I think that's really the most valuable thing you can do. When you have kids, they have the ability to drive, and they have the ability to get in, and they are going to do that to their families. They have the ability to make money off that, and then they have the ability to make their lives, and their children have the ability to go to that company and go to that school. So, so, that's very exciting.<|endoftext|> * Copyright 2012-2014 Google Inc.

is the first of two video-recording sessions by former British actor and director Colin Quinn, the man who played the iconic screenwriter's father in the "Hairspray," a TV remake of the 1987 film.

In a first appearance in the "Hairspray" film, in which Quinn played the father, the actor, who was born and raised in London, took a shot of Quinn posing in an old black-and-white cast of the film — his face is seen in the foreground, with one of his friends in the background — as he played his son on screen, in the film.

Quinn's casting was also revealed in an interview with ABC News.

“I’ve got a lot of friends in London that know I’m doing the film, so I’ve been asked about it,” Quinn said. “They said to me [after filming this] that the actor who plays my father in the movie was playing a father. And I said, well, it’s just because I’m the father.”

The former British actor was cast, along with two other people who are set to be in the film.

In the first interview with ABC News, Quinn said, “It’s going to be a tough one. I’ve already met some friends. I’ll be the one that plays my father. I know I’m not the father, but I’m just like a father. That’s the one part of my job that I’m not proud of. I don’t think I ever will be in the film. That’s my job and I can’t do that. It’s going to take me four days to do it, but I’ve got to do it. And I’m gonna be the one that plays my father when I’m going to do it.”

Qi Qin and his father, actor and film director Colin Quinn, play the part of their father. (Image via Twitter/Pascal J. O’Brien)

In the second interview with ABC News, Quinn revealed, “They said, ‘The actor will play my father in the movie because he plays the father who plays the father who plays the father of us all.� a family thing,” she continued. “It’s a family thing, it’s a family thing. I’m going to be the one that plays the father who plays the father who plays the father who plays the father who plays the dad. They’re all gonna be the one that plays the father who plays the father.

“We’re not gonna change the way that people are thinking. We’re gonna change the way we talk about the father. That’s the way you talk about the father the second interview, Quinn revealed that she and the rest of the cast in the film, who have played father and son, are also going to be played as the father in the film and the film has already been shown on TV, in the UK and in Japan. They were also shown the film during a visit to Japan.

“We’re going to be doing interviews for the movie, and I want to play the father who plays the father who plays the father of us all,” Quinn said. “I want to be a good father. I want to be kind not good, because I’m gonna be a good father for them, but I want to be a good father for my family.

“This is not the first time I’ve played the lead in a movie. It’s not just for me to go through all that, to become a father, but there are a lot of fathers out there to play a father. I’m going to be the father because I’m trying to get on screen. I’ve been doing it since I was a teenager when I was an actor and have all these good people to play the father.

“My dad is a father to me and he’s the one that plays the father. He has a lot of work to do, but a lot of work is done. He has to do it for me, so I’m going to be a dad.”

He added, “I want to be a good dad for them because I’m a good dad, not because I’m a good Dad, you know.

Qi Qin and his father are playing the father in the film, but it’s not just that he can

## Appendix D - Text generated with GPT-2 model

Birds are animal with feathers. Because they can fly, they are the fastest
animals on Earth. Some birds can reach speeds of up to 160km an hour. However, not all birds
can fly. Some birds, like ostriches can run and penguins can swim  not ostrics.
These birds have wings and feathers which are made up of tissue that is hard and elastic. These tissue
is made of connective tissue, and it is designed to allow these birds to fly without feathers
for a short period of time. The feathers are also designed so that they do not interfere with
the ability of the wing muscles to contract and flap. This is the main reason why birds
like ostics and kiwis can take off and fly around in the air.
This is why some birds don't fly very fast. Birds with a featherless body have a harder time moving quickly in flight because their wings are not designed for this. It is much easier for them to glide in air and hover in place. When the bird is not flying, it doesn't have to think about how to move. Instead, all it has to do is  The peripheral nervous system of a bird can be used for all sorts of things. If the feathers of a bird flap, this can cause the birds brain to be flooded with blood, making the birds wings become very flabby and slow.  To get around this, the part of their brain that controls flight is called the echolocation center. It basically senses the vibrations of birds in air. Once the muscles in their feet become accustomed to this pattern of movement, they are able to adjust the speed of motion of those muscles. Even though the wings are made for gliding, many birds find that using their eaglets and other small birds as pets can help them glide through the water. On the other hand, some fish can glide through the waves. They can also glide on land, but they need the help of gliders to do this (like fish). Birds don
t just glide. There are birds that can float. In fact, birds float on water, or in a river as long as they have the right environment. Like glider birds (and fish), some birds are water-dwelling. What they lack in water they make up for in other places. For example, owls (birds that are highly intelligent and sociable) have adapted to live in trees. Many other birds live where there are predators,  Birds that have lived in cities
and towns for a long time have adapted themselves to living in human communities. We
re no different. Often times, when people come to visit us, we tell them that we live in an old part. One of our many species of bird, Corvidae, is native to Australia. Coriids hug on our ears, hang on the tree branches, play with our hands, sit on birds nest, bite on our feet, make chirping sounds, stand on their heads, jump around, forage, fly in and out of gardens, bushes, trees, buildings, etc. and have other things to do. Unfortunately, our most intelligent birds sometimes do things that damage them. Sometimes we see these animals and they move and act in ways that do not look good. Why is this? Some people think that our richer and smarter bird than the rest (if there were any) will do better. And, in fact they will. But it's just a matter of confidence and luck
. Other times we get the feeling that something is wrong. Maybe the animal is too strong. Or maybe they just have too much of it. Perhaps there is something wrong with them. Our most intelligent bird will get better as we continue to learn about them, better than any other  animal on Earth. So what is a good
environment for these intelligent animals? We live on a planet that has a lot of water and land that birds use to survive. As long they get enough food to support their growth,  Some of these or other intelligent creatures are becoming a problem. The food is growing scarce and some of the intelligent creatures may die.  Some are just being stupid. That is, they are being lazy. While others may be angered and need to die off.   Why? It's because they don't care. All they want to know is how it feels to get eaten. Just as some animals are slow to respond to a call to attack, so are some birds. Some birds seem to have a low tolerance for pain. I think of some bird species that prefer to  drink water when they  get thirsty. Others like to lie on dry land and bask in sunlight. After a while, these birds have developed a tolerance to pain, and will not hurt themselves or others. How can you tell if an intelligent animal isn't smart? You can tell a sickly, old, lazy bird from a well-meaning, young one. You will notice that a young bird isn�t as smart as an older bird. With a few exceptions, intelligent animals do not live for very long. Young birds may live just a short time before they die of starvation. Do you think this means that intelligent life has a shorter lifespan than other life forms? No, not at all. Intelligent life does not die as fast as other forms of life.
So, how long does it take for intelligent beings to reach a certain level of intelligence? I don�m not sure. Now, I�ve heard the stories of intelligent   animals who have lived in a  far, far away place for hundreds of years.  We can�ll assume that the intelligent being that did that had a very good reason for staying there. At some point, the intelligence may have declined, but there may also have been some reason why the creature chose that place. What is the reason? There may not have ever been any reason. Humans have always lived on this planet. There has never been a reason that  had to change. When a man chooses to go to Africa, why? Because the mosquitoes are very intelligent. They do their job. If they detect a human, their work is done. No human could possibly survive that long without some intelligence. In the same way, a bird might choose to stay in our city and build a nest. It�s not because the city is good, it� is because it provides food. This can be seen in many different ways. For instance, some bees build nests in the cities. Well, what can a bee do? Bees do a variety of things. Bees build wax moths and other insects. Bee larvae can build many things, such as honeydew, beeswax, bee milk, flowers, honey, pollen, larvae, pollens, bread, leaves, soil, soap, feathers, scales, wax, paint, glue, paper,... etc, that bees need. Even ants and termites build structures that people would consider a normal home. These intelligent insects have learned a number of useful things in their lives. Not only do they use these things for their own needs, they also use them to protect their nests. Sometimes they even build houses for themselves. But why would a intelligent creature choose that particular place? Most intelligent animals have their nest somewhere. A bird doesn't have to build its own home, as the bird can find another home to live in. Why did the animal choose this place, knowing that it might be killed for it? The intelligent bird had some good reasons for choosing that location. Since it was a place where the humans lived, there would be some food available to the birds for some time. People may notice the oddity of a little bird building a home and may wonder why a good intelligent individual would build such a thing. By the time the adult bird gets to adulthood, he/she has become pretty smart. He/She has learned to fly, to find food, even to hunt. Still, this doesn� get to my question about intelligent behavior. So, can we tell how intelligent a creature is by looking at it from an intelligence perspective? Yes, we can. An intelligent reptile can think, learn, understand, use its body in various ways, be a hunter, act in defense, find mates, etc. However, in all of these ways it is still an animal. We know that an octopus is a fish. Yet we cannot say that octopi are intelligent or intelligent-like. Though some octomorpha have brains, other animals can't understand complex ideas, or are illogical in how they behave. Most other creatures have no way to think or understand. Of course, reptiles have one advantage. Their brains are more developed. Through evolution, reptiles have had more time to develop their brains. Look at the length of time it takes an owl to become a butterfly. One of the most amazing things about birds is that they have developed their minds and brains to a level that we see in us. Remember, they don�t have an eye, and they can� t see. How did they learn to see? When they see, do we think they are seeing? Birds have been around for a long time and have acquired a tremendous amount of knowledge. Some have gone so far as to say they know the human tongue. Who knows? Who cares? It is amazing how much we know about other species. Is there a certain point where we should stop believing in them? Does this make us intelligent? Of all the different species of animals, birds are the ones with the greatest intellect. Birds are highly intelligent creatures, perhaps even smarter than dogs and cats. The bird brain is approximately half the size of ours. (The difference is probably because of evolution.) Bird brains have evolved over millions of years. Can we detect intelligence in these tiny brains? Well we could not detect it in a mouse, an insect, nor even a beetle. Did the intelligent animal look at a brain and say, "that looks like intelligence." No, the brain was not there. All we saw was that there was something there, but we couldn� not see it. Does a cockroach have a mind? No. That� s why cockroaches have many more brains than we do. To be able to use the cock roaches brains in this way (to build up its intelligence), the roach had access to different organs and systems that other roasts don't. Cockroach brains look like human brains but they�re not human. Like humans, cock Roaches are capable of thinking, learning, reasoning, reading, thinking about our world and using our body to do it.� Cock roachers have also been observed to be intelligent and capable.

There are many factors that go into intelligent species� behavior that are hard to quantify. I know a few. For example, intelligent bird species have larger brains that allow them to have more of a cognitive process. Bird brain size and intelligence are also correlated. A bird with a large brain will be smarter. There are other things that can influence intelligence that I haven�T seen or heard of yet. If you�d like to discuss this topic, feel free to post it on our forum. But I doubt we will hear about it until later. With the last one, I could hardly find any references. Would you agree that intelligent animals are a myth?

- Jason Hunt

[1] (http://www.kopf.ac.at/~kapf/2000/anthro/b/3_dab_prog_17_10_2000.html)

I've read other questions like this before, here and here. You could find references in other articles, if you look hard enough.

How do you define intelligence? I don?t want to get into a discussion of definitions, because they aren?re all over the place. Here?s a definition I?ve found:

(See "Cognition and Intelligence" in the Encyclopedia of Animal Behavior)

can I see the facts? That? s not something I�ve done in years, so I can? t answer you, that? is all I wanted to know. And, since you asked about how we define it, it?d be better if I gave you some definitions to go by. What is Intelligence?

First of all, what is intelligence and why is it important? The purpose of intelligence is to take something that is complex and make it simpler. In this case, intelligence refers to the ability of an individual to make more complex things more simple Humans are able do this. This makes it easy for us to deal with complex problems, and makes our lives easier. On the other hand, humans are incapable of doing this for all types of things. One of the most important things to understand about intelligence (and this is what we need to define), is that it does not necessarily have to apply to all things, only to those that have been shaped by the human mind. So if a rock has a complex structure and the mind of its creator shaped it into something beautiful and interesting, then it would still be called intelligence. However, this would not be intelligence if the rock was completely random. It would be no intelligence at all. Or, to put it another way, something could have the intelligence of what it was made of, yet still lack intelligence as a result of not being shaped like the creator intended. Another example of something having the same intelligence, however, would simply be something made out of many simple things and being created from one simple thing. Take a big rock and place it between two bricks. Your rock could be either a beautiful, unique rock, or a simple rock that was just formed. Even if it were a truly complex rock like a human brain, still it could simply have turned into one small rock. As long as the bricks were simple, you wouldn?'t have any intelligence associated with them. Something can have intelligence just because it is shaped the way it has been made. Why is this important to remember? Because when we look back in time and look through the history of our species, we see that many of us have focused our efforts on one thing or another. Think about the many people who tried to prove the existence of God or the Earth was the center of a globe, even though they were wrong. They couldn??t even prove one of those things without going back through all the information in history. Also, when I think of my own life, my thinking was mostly focused on money, security, relationships, self-improvement, getting to a top position in life and generally living a life of high self esteem. I would like people to keep in mind that if something is worth doing, the more important it seems to you. When something seems simple to me, without much thought, but could really help someone else, things like that are more likely to stick in my mind and I might not even notice them at first. Yes, yes, everything I say about intelligent life is a simplification. That is because there are many aspects of human intelligence which are complex. For example, most people think that the primary function of intelligent creatures is for survival. Yet, some of them are actually living for the love of life. Others are just looking for some pleasure and adventure. Still others are seeking out knowledge and wisdom in order to better themselves in their own ways. There are others who are merely seeking the truth and are willing to face whatever life throws at them in an attempt to discover that truth. Does that sound familiar? Most of these are capable of solving the problems of their world. Some are even quite clever at it. None of this really needs to be said, except that we have a lot of complex intelligence in our world and we can use it to our advantage. Just remember that you don?"re just as capable as someone who doesn?!t know how to think, as you are able to do much more than they can. If you find yourself having trouble with intelligence when you have trouble making decisions, try this. Write a short list of all your problems. Don't worry about what they are. Simply write them down. Now look at your list. Try to make a list that has as many problems as possible. You will quickly realize that there will be a wide variety of problems that will likely crop up in your life which will help you to get through your day. Once you get started, it won?t be long before you begin to feel that much smarter. And you will probably be able make even more of an impact on the world than you could before. A brief note about intelligence: The higher your intelligence is, actually, and the higher the level of awareness that it gives you, then the easier it will become to understand the workings of the mind. This is why we recommend a certain level to begin with. Someone who is very high on intelligence can see a person who has less than a 4th grade education. But a level at which you can say, "Wow, that person really knows what he or she is talking about," is somewhere in the range of 2nd to 6th Grade. The reason for this is that intelligence isn't just a matter of skill or intelligence itself. We all have different skills that make us different, depending on what we are trying to accomplish. Intelligence is just one aspect of your personality. Of course, intelligence does play a role in what you do and how you act. So, while it may be true that intelligent people are not stupid, they aren't good at being smart either.

Intelligence Level: 5
"Where do I begin?" If I were to ask the average person what intelligence level is? I don't think I could find a single answer that would make any sense. Maybe it would include a cutoff of 1st grade. Well, in reality, there is no cutoff. Most people would probably say that level 4 or 5 is the most advanced level. At that point, almost everyone is capable to grasp a situation in a deeper way Level 3, on a similar scale, is probably the second most intelligent level you would ever be capable at. People would say at that time that they were a genius or a diva, or whatever the equivalent word is. However, at the very beginning, you won't be very smart. Your greatest strengths will not be the things that are most important to you. These will come later. What does this mean? Well at this point in time, what is most essential to the people who have reached level 5 are those things which help them to relate to others. They don? t need to know everything about everyone. Only that which is important will matter. That will only come with experience. How will you deal with other people? By being a good listener. Or by being someone you?re not. Intellect is a combination of feelings and reason. Reason is what allows you feel what your feelings are telling you about a given situation. For example, if you know that a baby is crying, but you cannot figure out why, just listen to your own feelings. Then you may come to a conclusion about why the baby might be crying. Because you got that feeling from your gut, your brain, the feelings about the situation you were in, were able use your sense of intuition to figure that out. It is as simple as that. In order for you become intelligent, don\t just know things. Know them so well that your emotions no longer matter to what it is you want to gain. As you gain experience, so you learn more about yourself and what matters to who you really are and who can be of help to. If you don;t know it, then don \t. There are many different ways to become smart, and you will probably not even know about some of them until you reach the right level of intelligence. When you start to learn something, it can help you to see things differently, to understand what's going on, how things are organized, why things happen. You can learn to be a more critical thinker. This is one of the best ways of becoming more intelligent. Whenever you find yourself thinking, ask yourself why you think so much. Don\T let it be something you haven\ t seen before. Take it on as a new opportunity to help yourself. After you have gotten to that degree, try to use that knowledge to make things better for the ones who are more important. Find out what people care about and then help your friends understand it. Become a kinder, more forgiving person. To help others, first learn how to behave yourself to improve their lives. Be more aware of how people think. Remember, only what makes sense to them will really matter in your life. Always try and find out how your thoughts and feelings might affect the person you are talking to or the other person in front of you in an important situation, such as work. Why would they be different from what they thought? You are learning more and more how important your feeling about someone is to how they act and think and act to other things as well. Next, find ways for people to connect with each other. Try to find places where you and your friend could meet and see how this might benefit them. Do this in groups of at least two people. Usually, this will be one on one. Even if it?s just you, make sure that you get along well and that there are enough people around you who want the same thing. Never try a group of people you wouldn?t be able to get together in real life with, unless you\re prepared to take on an active role. Being a little more flexible in who people can meet with and connect to helps you come up with more effective ways in which to do so. Once you feel you?re getting somewhere, start planning events to try out new things and opportunities. Start experimenting with new ways that can make a difference in people? lives, ideas, problems, etc. One of my best friends at school used to come by my house and we would do something as small as you could do to have fun. They would come over to my place, smoke a cigarette, hang out and have some good times together. Now I think that they are getting a lot better at it and I am so grateful for that because it helped me. At the time, I wasn?T too bothered about it because I didn?TCONS? because my friends were more mature and no one else wanted to hang around with me like that, even though they were my close friends. I could have been more careful, though, since I know how much I can benefit from it now. And it isn? t always easy to convince others that something is cool. So, here is a good example. The school we went to had a huge garden. Our classes had to spend some time in the garden every day, especially during the summer when the heat was so high. We had come a few times in our classes, once a week, during my freshman year. Each time we had spent an hour or two in that garden, working on a project together, playing games, eating lunch, doing chores. My friends and the teacher and everyone else in school had been really excited about this because they had seen me doing all kinds of gardening projects with my family. Everyone was very impressed with how I had done it for them, as though I was doing it all by myself. Well, when my senior year started, they wanted me to move on from that project. "You can do this better," they said. That was a difficult day for me because we hadn?VE gone from having a fantastic time together to me having to choose between being a productive member of school and my own personal goals. Since I felt so bad about what I did all I wanted was to be as helpful as possible to the students and get the most out of this garden so that the next year, the teachers could enjoy the experience as much as I and would have more fun working together and having fun together than if I just went back to doing the projects I?m doing now for the rest of the year and never got to help out. But, in my heart, it wasn\t that bad. What I realized was that I loved gardening, and it would be hard for any of us to not enjoy working with our hands in it. It seemed like the right thing to keep doing. Then one day I saw a sign on the door that read: "Insects are welcome in all classrooms." The students didn\TCON?ve started telling me about how great it was for my class to work together all day in a small garden to learn how to grow something in nature. There were a couple of students who had just moved in with their families and had all started making these little baskets of things for their family to use, like ladders and things. (I don?TBELIEVE it!) Then there were these two girls who were very involved in their community and all they did was garden together on weekends and they seemed to really enjoy it! The garden seemed perfect to them! I decided to go over there and take a look. For the first time ever, there was something available for everyone. In addition to all the things that we all had in common, we were all in one big space, with lots of space to plant and work. Everything seemed a bit bigger and a tad more useful. A few days later, my friend and her husband came over. She had had an idea to start a garden of their own and wanted my help in growing it to a decent size so it could be used. When I showed up, she was just as excited as me and said, "I want to bring a plant into this area so we can put it on our tables!" I couldn?TPELIFE? really see the potential here. Yes, you are in the same place, but I think we are both starting a new garden. So I told her what to do and she said yes. We started it and were super excited to see what happened. At first we thought it might be a good idea for us all to come over and see how things were going, so the three of them would get to experience what was going on. I never thought I would see this day. One day the wind blew through the gate and out came a huge flower. The weather turned a lot colder and I wasn?T really feeling well so I said to myself,?We can?t stop doing this.? We?ll see where this takes us." We ended up doing a total of 6 weeks in that area, growing flowers, food, plants and other things, all within a three-hour drive. Now, I didn?TCAN?TE really not really feel as if we?re getting paid to this point, because it has been a great experience for all of our families. Maybe I am just a little biased, since we actually had a summer job. With the promise of a full time job and income, getting a job with a company to hire us as gardeners was actually very appealing to us. And, of course, paying for this service was not that expensive. Because of that, our family has all been very satisfied with the opportunity we have been given. Every year we do a different project and we get paid for it, which is very nice. If you?RE ever in California, come to my house and tell me how much you love what we did! All we ask is that you leave us some comments and send us photos so other people can see our work as well. After all, a place like ours is great to visit. Our garden is in full bloom, everyone is having their kids in our class, writing about their garden, gardening on their computers, watching television, running errands, relaxing, taking care of business, teaching, etc. etc., etc.. So we really appreciate all you guys doing what you do, even if it isn?TLAND for our land. It has always been our dream to have a dedicated place to grow food and in turn create more jobs. Gardeners are really a small, special group of people, and it?s great that the people of San Diego, CA can now see us doing some work and learn from our hard work that they can pass on to their children. Let us know what kind of garden you like and what your favorite part of your job is! We would love to hear about it!

Return to Home Page
It?D a Wonderful Experience

,
Cheryl, Bob, Frank, Steve, Dawn, Susan, Danny, Margaret, Jocelyn, Bill, Lisa, Peggy, Joe, David, Bobby, James, Sarah, Nena, Pamela, Jennifer, Kerry, Sue, Amy, Karen, Beth, Tammy, Denise, Claire, Julie, Linda, Mark, Jeff, Andy, Marc, Chuck, Robin, Barbara, Rob, Jay, Pete, Jean, Jason, Mike, Richard, Joan, Louise, Deb, Jerry, Jim, Marjorie, Darryl, Chris, Alex, Jan, Christine, Gary, Kyle, Michael, Greg, Tony, Wendy, George, Rich, Terry, Ron, Keith, Brett, Brian, Kevin, Dale, Warren, Daniel, John, Laura, Maria, Michele, Trudy, Kris, Bruce, Cindy, Ann, Pam, Carrie, Ray, Katie, Douglas, Robert, Brad, Josh, Barry, Cory, Patrick, Dave, Joseph, Jimmy, Todd, Neil, Jeffrey, Joel, Doug, Jeremy, Matt, Dan, Kim, Marty, Stephen, Gene, Cheryl, Eric, Angela, Heather, Kelly, Sean, Ryan, Grant, Scott, Karl, Art, Gordon, Bryan, Martin, Drew, Nathan, Charles, Andrew, Trent, Jamie, Noah, Jack, Patricia, Adam, Aaron, Matthew, Corinne, Scot, Shane, Mimi Susie, Craig, Courtney, Pat, Ralph, Dennis, Will, Hannah, Suzanne, Nick, Paul, Christopher, Roy, Tommy, Dean, Phil, Dwight, Debbie, Gregg, Cameron, Thomas, Nancy, Aimee, Vanessa, Cydney, Brenda, Melissa, Tina, Bonnie, Sandy, Jackie, Betsy, Carly, Jane, Jon, Zoe, Corey, Christian, Becky, Tiffany, Kaitlyn
...
Evening sun
(almost, but not quite)
Trayana, Elsie, Kathy, Amber, Liz, Carolyn, Charlie, Kate, Kara, Claudia, Anna, Jessica, Amanda, Danielle, Kristen, Christina, Lauren, Nicole, Brooke, Vicky, Gabrielle, Ian, Ashley, Anthony, Annie, Jaclyn (I wish I could see that!), Mary, Cara, Caitlin, Ellie, Diana, Julia, Olivia, Jodi, Lucy, Kristina, Leah, Megan, Lindsey, Maryann, Molly, Leslie, Elizabeth, Emily, Erin, Kellie (on the top row), Molly (the one on the bottom row, on my right), Alana (again, my first time here, the second time I go I hope I can take pictures of you all!)
The herbs are growing
We love this place! In the Spring, they start in the sun, all by themselves, without any fertilizer, no pruning, just growing naturally.
Then they go on a mini journey around the garden and all of the weeds that are in there. The tomatoes, cucumbers, beets, potatoes, strawberries, blackberries, bananas, watermelons, radishes, spinach, turnips, onions, mushrooms, carrots, corn, beans, peas, peppers, squash, eggplant, eggs, cabbage, lettuce, green beans. We always plant in late April. These are just the beginning. In October, we plant the root beds, then the beds for the trees. Then in early November, before the first frost, it starts to get chilly and the flowers start to come out. By February, you be harvesting the food, soaking it in water, cutting the stems off and making molds of it. They?re now ready for shipment to all the farmers in our area. There are still many months to go until the market opens, so stay tuned. You can even check out the back of our house for some great pictures. I bet you will find some recipes for those veggies and flowers you don?t see here! I love you. Thank you for your love and support. ~ Jolene I?ve been here for a few years now and I have noticed that this has become my go to place for healthy food. My husband and kids don't eat the same. He has a dairy allergy and loves yogurt. This place is a wonderful place to eat healthy and is so convenient. Keep up the good work! Jennifer
I used to be a waitress here. After work, during the lunch hour, and in between the end of my shift and my day, after lunch, this was the place I went to for lunch. For the past five years Ive never had a bad meal here and always come back for more. Tasty food and great customer service. -Dawn
Megan, Linda, Dawn, Richard, Jim, Chris, Paula, Jennifer, Barbara, Marie, Nelina... and many more...
My parents are vegan and so are my children. Since moving to Arizona I was always hungry and it seemed as though it was impossible to find vegan food in my area, as it is in many other areas. Well, a friend of mine was running a vegan restaurant and she suggested this spot, which I really liked. So, I moved in and now I find myself at least once a week. At least every other day. Last night I had the best steak and beans I've ever had, with plenty of peppers and onions. It was so good, that I actually made a double batch. And I couldn't believe it! The first night it wasn't that great, in fact, when I got home I wanted to have another one. But, to my surprise, yesterday was my best night yet! And it's only been a couple of weeks. If you want to make your own food for yourself, there are plenty recipes on this site along with some good tips. Here are a number of recipes you can try: Spaghetti sauce: 1/4 cup tomato paste (I use Petrolatum). 1 cup vegetable broth. 1-2 cloves of garlic, minced. 2 tbsp. tomato sauce. Salt and pepper to taste. 4-6 chicken breasts. 5-10 cups of water. Season with salt and a little pepper. Put the spaghetti sauce in a bowl and whisk it together. Add the chicken, broth, garlic and tomato to the bowl, whisking until all ingredients are thoroughly mixed. Heat the water in your pot over medium heat. When the pot is boiling add the pasta and let it cook for about 30 minutes. Remove the cooked pasta from the heat and set aside. Stir in 2 cups water and some of your favorite seasoning. Serve with your favourite pasta, or, top with other kinds of veggies, such as kale, zucchini, artichokes, etc. and lots of pepper! Rice: 3 cups cooked black beans (or 2 cans black or kidney beans). 2 onions chopped. 6-8 cloves garlic minced or crushed. 3-4 cups rice. To make the rice, take your rice and soak it overnight. Drain the soaked rice (don't rinse it). Then, chop the onions and garlic. Combine the chopped onions with the garlic in an oiled pot and sauté for 2-3 minutes or until soft. Once the onion and rice are cooked, add them to a pot. Simmer for 3 to 4 minutes, stirring occasionally. Allow the potatoes to cook. Pour your boiling water into the bottom of a large pot (if you have one) or dutch oven. Boil the boiling potatoes until they are tender and fork tender. Discard the boiled potatoes Place your dumplings in the broth and add salt, pepper and salt again. Bring the dummplings to boil. Cook for 15 minutes until the outside of the Dumplings are starting to soften and the inside is golden. Turn off the stove and allow the wok to heat up. Transfer the steaming dums to an oven safe dish. While the hot woks are heating up, prepare the veggies. Divide the vegetables in half, sprinkle them with a bit of salt. Take a small potato, coat it in oil and sprinkle it with garlic cloves. (This is how I make my dahl). Sprinkle with water (it will be a mixture of greens and potatoes) and cook until it starts to get translucent. After it becomes translucent, remove the potato from its water bath and place it on a plate. Peel off some garlic stems. Set aside to be used later. Repeat the same with remaining vegetables. Give the meat a good stir, and take the beef out of it. Start cooking it until you see it get browned. In a skillet, heat the olive oil over high heat (about a 3 minute, until hot). When it reaches a gentle simmer, then add in all the other ingredients. Saute the meats until just cooked through. They should still be moist, but not brown. The more you cook them, the more tender they will become. Keep cooking until everything is tender, about 8-9 minutes total. You may need to add a splash of broth at the very end to keep everything together, if you like. This is not the time to strain your beef, unless you just don't like the texture of that. I like to remove it from it and put it back in. Finally, put the hamburger mixture into a baking dish (this is what I use) to rest. Bake it for 20 minutes and serve it, topped with grated cheese and a sprinkle of parsley. It's soooo good! You can also serve with some tomato sauce or sour cream. Or, you can add extra salt and pepper.

Fried Eggplant
in red curry powder
I'm really excited to make these Fried Eggplants in Red Curry Powder! They are so easy to prepare, so they're perfect for a cold night. My husband and I loved this Fried eggplant!
Once you've done this, I hope you will love these easy recipes. Make them as often as you'd like! I promise to update them regularly with new recipes and new ways to use the ingredients in them. These recipes are also a great way to save money! Make your own seasoning and use it to season your dishes. Use it instead of store bought, season it before you make it or add it at cooking time. If you love my recipes, please follow me on Facebook, Instagram and Twitter. Thank you so much for reading and stay tuned for more recipes! Enjoy! XO, Jen
Posted by Jen at 12:22 AM recipe was brought to you by The Bloggers Kitchen, a member of Yummly. Don't forget to check out the Recipe Index for even more great recipes!!
About the Author: Jen is a freelance writer, blogger and nutritionist living in Florida. She is the owner of a blog and blog management company. Her main goal is to help others achieve their health and wellness goals. Follow her on social media for all of her latest recipes (and those of you who can't find them there).
Share This: Pinterest
Email
Instagram
Facebook
Twitter
Google
Tumblr
Reddit
LinkedIn
Print

More
Please check us out on Social Media:
Thank you for visiting! We are here to serve you healthy, delicious, filling and easy-to-prepare meals for your family and friends. We hope that you enjoy our recipes as much as we enjoy making them! If there is anything that we've missed or you have any comments, we'd love to hear from you! Donate to The Huffington Post here. All donations are tax deductible. Your privacy is important to us. Thanks for stopping by! Love, Jennifer
Images: 1) mamaatechen, 2) timothy of cheetahc, 3) The World's Greatest Blog 4) yalehastings, 5) creative writing by the blog authors, 6) Creative writing blog by christine holland, 7) flowery crochet, 8) All images and content on this website are copyright of their respective owners. Please do not use them without permission. THANK YOU! - Jen and The blog owners, Laura and Mikey, for letting me share these awesome recipes with you. Download Printable Recipe For Fried Eggs in curry: Ingredients 3 tbsp olive or canola oil
1 large egg
2 large onions, chopped
6-8 medium bell peppers, sliced
5 large carrots, diced
3 tbsp minced ginger
4 tbsp curry paste
1 tbsp lemon juice
Oil for frying (optional)
8 oz package of pasta (or 4 cups cooked whole wheat pasta) (see recipe notes) Instructions Heat the oil in a large skillet over medium heat. Add the egg and fry until it is cooked. Remove from the heat and stir in the onions and peppers. Stir in carrots and cook for about 5 minutes, stirring occasionally. Turn the mixture to coat in oil. Cover and let sit for 5-10 minutes. Heat another pan over low heat, add the pasta, and add 1/4 cup of water. Cook for 7-15 minutes until the sauce has thickened. Meanwhile, in another skillet or wok, melt the butter. When the pan is hot, pour in 2 tbsp of the curry mixture and saute for 1-2 minutes or until slightly browning. Once the eggs are cooked, drain on paper towels. Reduce the fire to low and pour the rice and pasta mixture in an even layer on top. Mix the flour with a spoon and cover with plastic wrap. Let sit at room temperature for at least 15 minutes (if you're using rice, let it sit longer). Once your rice is done, dip it into the hot rice mixture. Drizzle the reserved curry sauce on the top and enjoy! (Don't want to eat your curry?)
To assemble: Using your fingers, gently spread the filling into a thick layer over the white pasta. Sprinkle the sliced peppers and onions over it.
Spread the tomato sauce over top, followed by your noodles and veggies.
Add the cooked egg to the middle. Top with your desired garnishes. Enjoy!! 3.1 http://lalalalday.blogspot.com/2012/08/fried-eggs-in-curry.html
Enjoy! you like this recipe, you may also like my Fried Egg In Curry (Gluten Free), Fried Chicken In Salsa and my fried egg in cornbread. Click here for my other blog posts on All Recipes! I also make these wonderful puffed rice patties. I have two recipes for these puffy rice balls. One is a vegetarian version and one is gluten free. Both are absolutely delicious. The gluten-free version is just as good, if not better. It's made with almond flour and has a slightly different flavor. You can also try the Vegan Egg in Curry for a similar and similar texture, but without the gluten. Oh, one more thing. If you can't find curry in your area, try another spice mix, like cayenne, cumin, paprika or chili powder. Just a suggestion. :)
*If you love this, please don't forget to leave a comment! It really makes a difference! Thank you, Jenn! xoxo
Posted by Jenn at 1:04 PM
My husband and I were introduced to fried eggs in our first week of college. We were skeptical, so we decided to give it a try. They were delicious and we both loved them. Over the years, I've made them all over campus. My family and friends all have different tastes, tastes that vary a lot. For example, when my husband is at home, he likes to have a huge pot of chili, or if he is traveling, we can have chicken, fish or seafood. Here are some of my favorites: 2 C. Chili: This is my favorite! If I'm in town, this is the dish I make. Serve with rice or with sour cream. 3 C Crumbled Bacon: I love bacon, it makes everything better, especially fried chicken. This recipe is very simple, just three ingredients. So easy and delicious!

3 T. Nutella: We love Nutelle. What better way to make a spread than with bacon? It makes the spread extra flavorful. 4 T olive oil: Olive oil is definitely my go to for cooking. While I don` normally like it when I am cooking, the possibilities are endless. Try using it in this dish! 5 T paprikash: Paprikashi is an Indian food. Usually, spinach is used in the preparation of it, however, now, saffron is also very popular. Add some garlic and spice it up with paprikashi. 6 T Chana Masala: You know how in India, they don´t serve rice with their food? In this case, what can I say? They made it taste better than any rice. 7 T Red Curry Paste: Red curry paste is made of red chillies and cloves. These spices give the recipe that special, hot and spicy taste. 8 C Pad Thai: If Pad Thai is your thing, then you have to try this. Pad thai has chicken and beef in it and it's so delicious, my mom even has it on her menu. 9 C Black Bean Dip: While black beans are delicious on their own, added to this dip, is some cilantro and lime juice. 10 T Shrimp: Shrimps are really easy to cook. Some people add lemon juice to it for extra flavor, and some people make the sauce with shrimp. 11 T Tomato: Many people prefer the real deal, while others use it as a base. Either way, all you need to do is add it to your dish. 12 C Tofu: T of tofu is such a simple dish to prepare. Tons of nutrients and protein, plus it is so easy! 13 C Shredded Lettuce: Although it sounds weird, most people use shredded lettuce instead of lettuce leaf. It gives it an awesome texture and flavor! 14 C Broccoli: Broccolini is another great vegetable to add to a dish, as well as being very healthy! 15 C Onion: One of the easiest vegetables to cut into bite sized pieces. 16 T Cilantro: Add a bit of fresh chopped cactus for that extra kick. 17 T Salt: The usual salt, you won't need it. 18 C Green Peppers: Also known as bell peppers, these are very easy, with only four ingredients, to put on a plate. 19 T Sesame Seeds: Sesamin is a classic, easy dish with just four essential ingredients: ginger, garlic, oil and salt. 20 T Corn Flakes: Since corn is already baked, put the corn flakes on the bottom of your skillet. Then add the vegetable and olive oils. Pour the broth over the top, add more salt if needed. 21 T Spicy Peanut Sauce: Peanuts are a great addition to the chicken! 22 T Soy Sauce (or canola): This sauce is great with the shrimp! 23 T Broth: Usually made from chicken broth, go ahead and make it with coconut milk or other liquid. 24 C Lime Juice: Now you know! 25 C Vegan Cauliflower: A great recipe for these is just 4 ingredients and you will love it! 26 C Cumin: Just four spices and they are ready to go. 27 C Carrots: Carrot is one of those vegetables that makes me fall in love with it quickly. 28 C Garlic: All of us love garlic. Go ahead, make your own garlic sauce! 29 C Pine Nuts: Pine nuts are an easy one to get, but you don't have a ton of time to spend on it (unless you are like me and love them too). 30 C Cornstarch: Corn starch is used to thicken sauces and this is no exception. 31 C Sticky Rice: This is super easy and a little pricey, so be sure to pick up a few packets of this at the store. 32 C Curry Powder: Curry powder is very similar to sesame seeds in that it makes a sauce, especially when you add your spices. 33 C Chicken Stock: Chicken stock is known for having a smooth consistency and great taste, making it perfect for soups, stews, stir fries, etc. 34 C Coriander: It's great for adding that little bit extra spice to dishes. 35 C Cloves: Some of you may like this more than the other spices, though I personally love the spicy pepper taste it adds. 36 C Salt, Pepper, Onion Powder, Lemon Juice, Garam Masalas, Spice: There are many different spice blends available, such as the red curry, black, curry and more. 37 C Lemon: Lemon is always a good addition, not only because it gives a nice tang to everything, it also has so many health benefits. 38 C Sweet Chili Powder (Sip up this spicy mix) 39 C Rice Flour: Rice flour is an essential ingredient, because all of our food is mostly starch. 40 C Saffron: When you see saffrons, your mouth will water and the heat will increase. 41 C Turmeric: Turpenoids are antioxidant compounds that help with a variety of health conditions. 42 C Thai Red Pepper Powder 43 C Savory Garle: Savoury Garles is exactly what it says. This should be the last spice you use in your dishes because if you aren't careful, the savouriness will go away. 44 C Chia Seeds 45 C Black Pepper 46 C Grape Seed 47 C Tomato Seed 48 C Ground Nutmeg 49 C Basil: Basil is extremely versatile, and there are lots of different varieties to choose from. I usually go for the pepper flavor, since I like it that way. 50 C Nutritional Yeast 51 C Cinnamon: You can use cinnamon in almost any dish, whether it's in a salad, on toast, in salad dressing, or just as a quick spice. 52 C Red Wine: Red wine is simply delicious. For a milder red wine, use sparkling wine or a white wine. 53 C Ginger: Ginger is the classic spice, all you need is four basics: garlic powder, ginger root, turmeric and red pepper. 54 C Rosemary: Rose is delicious in everything from salads to sauteed vegetables. 55 C Chili Pepper: Chili peppers are also extremely easy to use, for those who don- 55 c Pepper. 56 C Mushrooms: Mushy foods are the best, everything tastes better when they're all mushy. 57 C Hot Sauce 58 C Onion: I'm not usually a big fan of onions, even when I have them in my fridge. You might want to change this up if your cupboard is bursting with them. 59 C Dill: Dill is so good in all kinds of dishes, from souped up salads, chutneys, soupy stir fry, roasted vegetables, crab cakes, fish, pasta, steak, sandwiches, more, just about anything. 60 C Sugar: Sugar is nothing more then a sweetener, other than that, there is only one thing to add to your food. 61 C Chilli: Chillies are one of the most versatile spices in the world. They are a little spicy, but they add a whole lot of flavor. 62 C Limes: Lime is just the cherry on top, as it is such a great addition to most of your foods. 63 C White Pepper 64 C Spices: Now you can add so much spice that you have no idea what you are even adding, this is because there really is no limit to the amount of spice available. 65 C Tumeric 66 C Garlic 67 C Pine Nuts 68 C Carrots 69 C Nectarines 70 C Celery 71 C Green Beans 72 C Onions 73 C Cumin 74 C Leeks 75 C Peppers 76 C Potato Starch 77 C Pistachios 78 C Sunflower Seeds 79 C Oregano 80 C Mint 81 C Olives 82 C Parsley 83 C Pumpkin Seeds 84 C Broccoli 85 C Tomatoes 86 C Fennel 87 C Paprika 88 C Mustard Seeds 89 C Avocado Oil 90 C Cayenne Pepper 91 C Yuzu 92 C Iceberg Lettuce 93 C Asparagus 94 C Buckwheat Cheese 95 C Beef 96 C Mackerel 97 C Smoked Salmon 98 C Crab 99 C Blue Cheese 100 C Plain Tuna

Loading... Loading...
The best swag you'll ever get!

.
: This is a sponsored post, if we've helped you get something from one the companies that support the site, we'd love you to check out their product. Thank you! : This page contains affiliate links. If you click on one and make a purchase, I may receive a small commission. Thanks for supporting us! Time! Enter for a chance to win a $50 Amazon gift card! Winner will be chosen on Thursday, September 18th at 9:00 AM EST. Good luck! Make sure you follow us on Facebook for more giveaways! [x] If this giveaway ends in your favor, you will also receive 2 FREE bottles of both of these brands: Ben & Jerry's® Cheesecake Factory® Flavor Batter Cookies®
The following giveaway is sponsored by the following companies:
Coffee Bean®, The Cold Brew Coffee Company, and Honest Tea.

p.s. These products are sold through the links in this post. Please do not re-post these links to Facebook, Twitter, Pinterest, Google+, or any other social media site. We do this to protect our sponsors, our readers, so please be careful! [/x][x]"Every person is entitled to a fair trial in which the evidence is examined in accordance with law. That does not mean, however, that everyone will have the same chance at a hearing. The right to appeal to higher courts must be accorded to those convicted of crimes or accused of wrongdoing, unless the court finds that the trial judge or prosecutor acted with a deliberate indifference to an innocent person's right. There are few things more harmful to society than a trial that ends after a day and a half and acquits the defendant. It is one reason that juries in criminal cases must not be forced to convict on the testimony of eyewitnesses or experts." – U.S. Supreme Court Justice, Oliver Wendell Holmes, Jr. (1925)
Tablespoons Amount Per Serving Calories 237 Calories from Fat 24g Total Fat 2g Trans Fat 0g Saturated Fat 1g Cholesterol 23mg Sodium 2,425mg Total Carbohydrate 40g Dietary Fiber 1.5g Sugars 4g Protein 11g Vitamin A 10% Calcium 3% Iron 7% Vitamin C 9%
Posted by Kenneth C. Covington at 10:33 AM
[x], [ ], [ ], [],,, < Previous | Next > Next » Page 1 of 5 < > ( ) [ ]

Share this: Facebook
A similar giveaway for readers is coming up on September 20th, 2016 at 12pm Pacific. One lucky winner will receive $100 in Amazon Gift Cards! Get it while it's hot! We'll see you then! ~ Kenneth

,
Karen
About Kenneth: Kenneth is the founder of www.nutritionfacts.com. His website has been featured in the New York Times, the Chicago Tribune, The Washington Post, People Magazine, Men's Health, Lifestyle Magazine and other publications. Kenneth also writes for the Healthy Living Journal, which has a weekly newsletter with recipes and advice.

thematic menus, recipes, nutrition facts, health information, meal plans and more. I was never a big foodie, never had a high school education or a college degree. Still, after all these years, a little food always goes a long way. But I guess that's just how life is. So what better way to have a healthy and satisfying meal that is always on-trend and always fresh than to serve it at home? It makes for an easy dinner on a weeknight or even a weekend evening. You know how when you go out for dinner, especially at the restaurant, there are always the popular items that people want? Well, I like to give these healthy meal ideas to my readers who want to do the dishes for themselves, but they still want a nice, fresh meal at their house. Well I'm not going to take it away from you. Here are some ideas for you: Breakfast: Anytime of the day. This includes breakfast for lunch or dinner. A quick recipe: Eggs Benedict with Banana and Nutella Bread – Quiche – Dinner: Pork Ragu with White Bean, Corn and Oat Gratin – Breakfast Snack: Cheesy Quinoa Puffs with Cheddar and Avocado – Lunch: Meatloaf with Blue Cheese and Tomato Salad – Dessert: Strawberry Mousse with Chocolate Sauce and Lemon Bars – Snacks: Coffee Bean® Cookies and White Chocolate Chunk Cookies – Homemade Brownie Sundae – This list could go on and on. Not only are there plenty of healthy food options, they can also be easily made at your home! My favorite healthy foods are all protein and low-calorie. They are made with fresh, organic ingredients and most are gluten-free. All of them are low in fat, cholesterol and sodium. And with all the other nutritional benefits, it is no wonder that they are so popular with people who have high blood pressure, diabetes, heart disease, high cholesterol, obesity, joint pain, depression, arthritis and many other conditions. If you're interested in finding out more about the healthy ingredients in these foods, just click on each one to go to their nutrition labels. Don't forget to check out the videos at www or www. Kenneth has also published several books: Healthy Eating for Healthy People, and Healthful Living: A Comprehensive Guide to Healthy eating, Weight Loss and Happy Living. He also has many videos available on his website. Some of my favorite videos are: 1) Healthy Cooking Made Easy 2) Quick Healthy Snacking 3) 5-Minute Healthy Lunch 4) Great Healthy Recipe 5) 20-Year-Old's Healthy Favorite Recipe (with Meal Plan) 7) How to Make Healthy Food Easier 8) The 6 Healthy Ingredients You Must Have in Your Home

Healthy Foods: What are the best healthy healthy options?

I'll start by answering the question, "What are your favorite healthful foods?" I have tried to use the following list of health-friendly foods as a guideline. It is not an exhaustive list, as there may be some other healthy choices that you like. For example, you may enjoy the Greek yogurt, so I've left it out. Remember that all of these are products of farms and food processors, not supermarkets. Just remember that I don't have any control over what you choose to eat. These are just some of what I use in my own kitchen.

Crispy Fried Chicken

1/2 lb. boneless skinless chicken breast

3/4 cup (100 g) ketchup

6 tbsp. (120 ml) water

Salt to taste

Heat oil in a skillet over medium-high heat. Add the chicken and cook for about 1 minute per side. Flip and continue to cook until chicken is cooked through. Remove from heat and place in bowl. Set aside. Cook bacon until crisp, about 4-5 minutes. Drain bacon and reserve. In a large bowl, whisk together ketchups, water, kettles, salt and pepper. Dip chicken in the ketzpup mixture and coat with the sauce. Return chicken to the skillet and top with more bacon. Serve hot with kétamos. Makes 4 servings. Notes Make sure to leave room in your plate for the onion and potatoes. The onion is going in under the cheese and fried chicken, with a bit of bread for crunch. I usually make the onions and kernels in advance and store them in refrigerator. When you need them, pull out some and have them warm in oven. 3.1 http://www.kennethocha.com/healthy-food-list/

Duck Breast

2 Tbsp. fresh basil

10 Tbs. extra virgin olive oil

8 oz. package (1 1/3 lb.) boned duck breast (I used a bone-in one)

Cook the duck breasts for 4 minutes per pound, or until cooked. Using a slotted spoon, transfer the breasts to a plate. Repeat with remaining duck. Resting the breast on a cutting board in one piece is recommended to keep the pieces from drying out and falling apart. Let the meat cool to room temperature. Dice the fat from the thigh and breast and discard. Place the bones in an airtight container in saucepan. Bring to boil and cover. Reduce heat to medium. Cover and simmer for 15 minutes or so. Transfer to refrigerator to cool completely. Then shred the muscles using two forks. Spread the cooked duck on your prepared pan. Sprinkle with chopped parsley and drizzle with olive and kosher salt. Enjoy! 3 http.www to get started:

The following is a list from Kenneth's website: "Our List of Healthy Recipes is the most comprehensive resource for healthy food choices available today. We have thousands of recipes to choose from and more being added all the time. Each recipe has a step-by-step photo tutorial so you can make it in no time at all. Click on the photo to see the recipe! Some recipes are very easy and you will see that there is no step by step photo, but we put a lot of thought and thought into every recipe. This allows you to have a fun and healthy lifestyle." "In other words, we love you! We are glad that our list is making a difference. If you are a cook, cook your way to health." 3 www,knocha, kennethefirst.blogspot.ca,

Tuna in Cucumber Sauce

Pinch of salt

In a small bowl combine tuna, cucumber, sea salt, lemon juice, basil, olive, garlic, chopped chives and basil leaves. Stir and toss to combine. Put the tuna mixture in about a quart or half gallon of water. Boil for 2-3 minutes and then cover and let stand for 10 minutes to soak up some water and thicken. To serve, place the tuna mixture on top of the cucumbers and spread the rest of your salad over the top. You can also toss the whole tuna and cucupies with some salad dressing or mayo. 4 www.thewanderer.net/food/tuna-rice-cucumbers-garlic-pepper-onion-honey-mayonnaise-salt-tofu-or-chicken-meat-canada-fish-pork-vegetable-soy-lemon-ginger-apple-cilantro-sweet-seasoned-sprouts-pasta-cold-stewed--parsley--apple--carrots--cauliflower--beets--pickled beets-a/2600/6/l/k/i/s/5/z/m/p/d/c/g/f/v/o/a 1 http//www2.msn. com/WinNews/content/ winnews-wire. msnbc/news/stories/T/569996.aspx

Beef in Lemon Wine

Serves 4

4 lbs. lean beef

Mix the following ingredients: 1 cup apple cider vinegar 1 teaspoon dry white wine 1 tablespoon lemon zest 2 tablespoons lemon or lime juice 1 Tbls. olive or canola oil 1 bay leaf

Cut the beef into cubes. Mix the wine and oil in a large sauce pan and bring to simmer.

While the vinegar is simmering, prepare the cubes of beef. Heat the oil and zing in another sauce pot. Add the cube of meat and the lemon mixture, and cook over low heat for 5 minutes. Remove the bay leaves and place on paper towels to drain. After 5-10 minutes, remove the chops and shred them with two fork. Set the veggies aside to let cool. While the vegetables are cooling, put the liquor in glass or ceramic mason jars, add a lid and fill with cold water to cover the lid Lock the masons and lid in the freezer. When the water is ready, fill the jars with the liquid and shake gently to allow the lids to press against the jar. Allow the process to take about 20 minutes at room temp, then fill jars about half way with water at this time and allow to stand at the bottom of a warm oven until the bottoms of each jar are just just touching. Check the consistency and use a paring knife to cut the tops off. Once you're happy with your results, allow each mister to sit in warm room or on an open counter for about an hour to air out. Unlock the bottles and enjoy!

Note: if you would like to make this recipe gluten free, feel free to substitute 1/2 cup of sour cream and 1 1⁄2 cups of flour. Just replace the apple juice with grape juice. The rest is up to you. 4 http://kck.st/1x6Xs7G 5 http, kck4food.com/recipes/ 4. kc-loud. 5 www., kstans. 6 www..kohls.org/articles/2012/07/02/how- to-make- a-quick-and-easy-recipe-for-bacon-in-no- time-2/ 7 www… http 8 www... 9 www…. 11 www.kliw.on.uk/health/ 10 www krause-martinson.kleinschmidt.de/en/ 13 www, www … 15 www htnews.no/canadas-maine-beef-cover-up-huge-award-winning-recall-salad-lesson-how 1. www 2. wgreens.co.nz/blog/ 1 www 3. itc.ox.ac.jp/coj/national/article/10007767.ece 3 http, http 2 http… 4 awww.usatoday blogspot. otc.nyu. edu/home/ a www 6 aa the wanderers. 9. http. e n w. c. l a. v. blog 7 http 6. 1 0 http a t. wik i t 6 d e t e c t i o n s. 15 e

---

[i] The dataset is available at https://github.com/nickm197/Longtexts